\documentclass[10pt,twocolumn,letterpaper]{article}

\usepackage{cvpr}              
\usepackage[utf8]{inputenc}
\usepackage[font=small]{caption}
\usepackage{amsmath}
\usepackage{amsthm}
\usepackage{booktabs}
\usepackage{algorithm}
\usepackage{algorithmic}
\usepackage[switch]{lineno}
\usepackage{adjustbox}
\usepackage{multirow} 
\usepackage{makecell}
\usepackage{color}
\usepackage{makecell}
\usepackage{indentfirst}
\usepackage{amssymb}
\usepackage{array}
\usepackage{textcomp}
\usepackage{stfloats}
\usepackage{url}
\usepackage{verbatim}
\usepackage{graphicx}
\usepackage{float}
\usepackage{times}
\usepackage{soul}
\usepackage{microtype}
\usepackage{epsfig}
\usepackage{enumitem}
\usepackage{tabularx}
\usepackage{placeins}
\usepackage{colortbl}
\usepackage{pifont}
\usepackage[hang,flushmargin]{footmisc}

\definecolor{cvprblue}{rgb}{0.21,0.49,0.74}
\usepackage[pagebackref,breaklinks,colorlinks,allcolors=cvprblue]{hyperref}

\usepackage[capitalize]{cleveref}
\crefname{section}{Sec.}{Secs.}
\crefname{table}{Table}{Tables}
\crefname{figure}{Fig.}{Figs.}

\def\paperID{10658} 
\def\confName{CVPR}
\def\confYear{2025}

\title{HVI: A New Color Space for Low-light Image Enhancement}

\author{Qingsen Yan$^{1}$\thanks{These authors contributed equally to this work.}
~~~~~Yixu Feng$^{1*}$~~~~~Cheng Zhang$^{1*}$\\
Guansong Pang$^2$~~~~~
Kangbiao Shi$^1$~~~~~
Peng Wu$^1$~~~~~
Wei Dong$^3$~~~~~
Jinqiu Sun$^1$~~~~~
Yanning Zhang$^1$\thanks{Corresponding author.}\\
$^1$Northwestern Polytechnical University~~~~~
$^2$Singapore Management University\\
$^3$Xi’an University of Architecture and Technology\\
}

\begin{document}
\maketitle

\begin{figure*}[htp]
    \centering
    \includegraphics[width=0.96\linewidth]{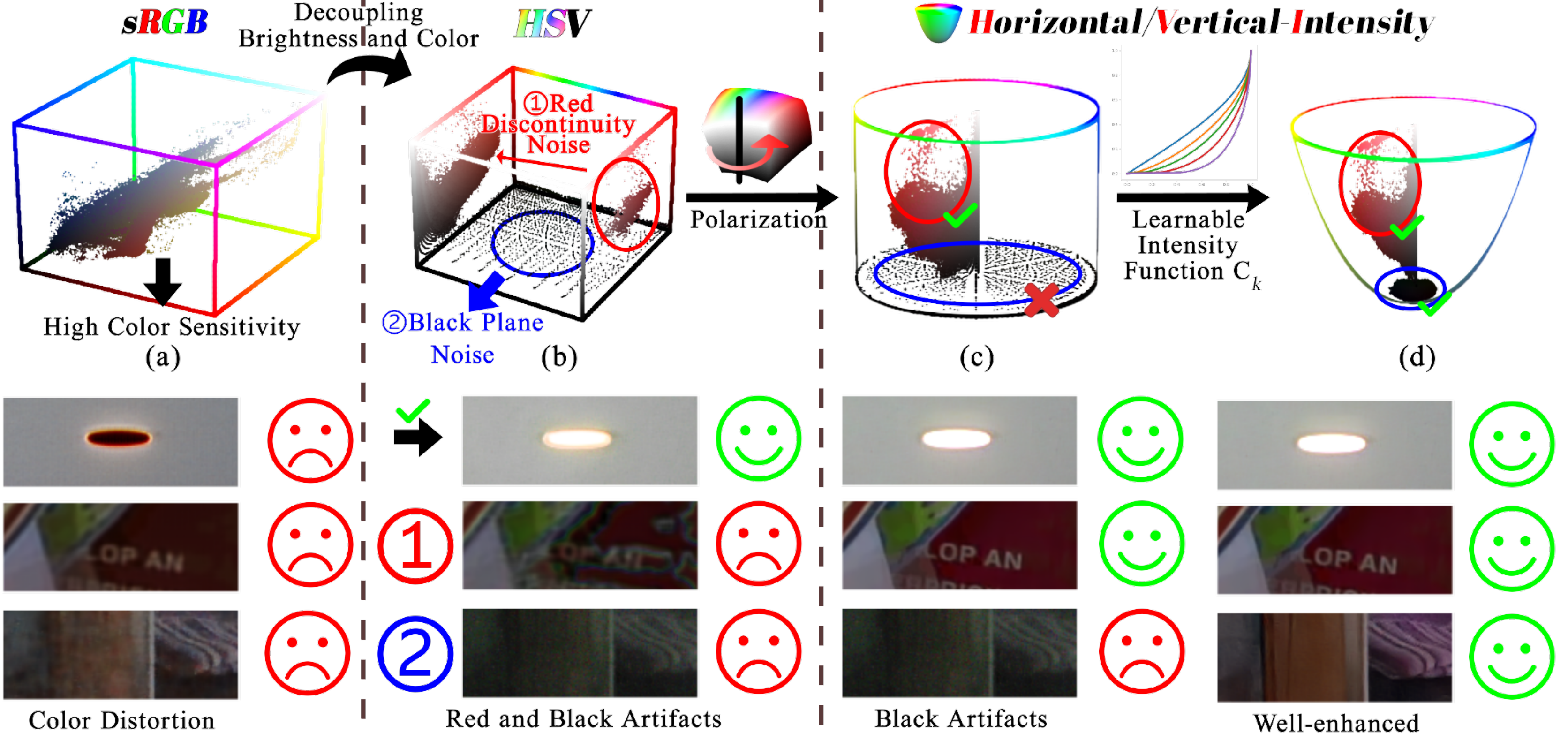}
    \vspace{-0.2cm}
    \caption{The top row illustrates the process of transforming images from the sRGB color space, via HSV, to the HVI color space. The bottom row presents the corresponding test results. The sRGB color space is known for its high color sensitivity, often causing color distortions in test images. By decoupling brightness and color to obtain the HSV color space, the illumination enhancement appears normalized. However, this transformation introduces varying levels of red discontinuities and black regions, which subsequently cause artifacts in the enhanced images. Introducing polarization to HSV ensures continuity in the red regions. Introducing a learnable intensity function $\mathbf{C}_k$, on the other hand, helps collapse the black regions, yielding the HVI color space with optimal image enhancement.}
    \vspace{-0.2cm}
    \label{fig:1}
\end{figure*}

\begin{abstract}
Low-Light Image Enhancement (LLIE) is a crucial computer vision task that aims to restore detailed visual information from corrupted low-light images.
Many existing LLIE methods are based on standard RGB (sRGB) space, which often produce color bias and brightness artifacts due to inherent high color sensitivity in sRGB.
While converting the images using Hue, Saturation and Value (HSV) color space helps resolve the brightness issue, it introduces significant red and black noise artifacts.
To address this issue, we propose a new color space for LLIE, namely Horizontal/Vertical-Intensity (\textbf{HVI}), defined by polarized HS maps and learnable intensity. The former enforces small distances for red coordinates to remove the red artifacts, while the latter compresses the low-light regions to remove the black artifacts. To fully leverage the chromatic and intensity information, 
a novel Color and Intensity Decoupling Network (\textbf{CIDNet}) is further introduced 
to learn accurate photometric mapping function under different lighting conditions in the HVI space. 
Comprehensive results from benchmark and ablation experiments show that the proposed HVI color space with CIDNet outperforms the state-of-the-art methods on 10 datasets. The code is available at \href{https://github.com/Fediory/HVI-CIDNet}{https://github.com/Fediory/HVI-CIDNet}.
\end{abstract}

\vspace{-0.45cm}
\section{Introduction}
\label{sec:intro}

Under low-light conditions, imaging sensors often capture weak light signals with severe noise, resulting in poor visual quality for low-light images.
Obtaining high-quality images from such degraded images necessitates Low-Light Image Enhancement (LLIE) that aims at improving the image brightness while simultaneously reducing the impact of noise and color bias \cite{2022LLE}.

The majority of existing LLIE approaches \cite{KinD, EnGAN,Zero-DCE,Han_ECCV24_GLARE,10543170,GSAD} focus on finding an appropriate image brightness, which is typically done by employing deep neural networks to learn a mapping relationship between low-light images and normal-light images within the standard RGB (sRGB) space.
However, the image brightness exhibits a strong coupling with the color from the three sRGB channels, \textit{a.k.a.} high color sensitivity in \cite{gevers2012color,lee2024rethinking}, causing an obvious color distortion of the restored image in these LLIE methods \cite{RetinexFormer,EnGAN}, as shown in Fig. \ref{fig:1} (a).

Inspired by Kubelka-Munk theory \cite{gevers2012color}, recent methods \cite{li2020low,zhou2023low,zhang2021better} have sought to transform images from the sRGB color space to the Hue, Saturation and Value (HSV) color space. These methods help achieve the brightness enhancement more accurately, but they amplify local \textbf{color space noise} \cite{gevers2012color}, introducing severe artifacts in the results. 
As illustrated in Fig. \ref{fig:1} (b), the transformation from sRGB to the HSV disrupts the continuity of the red (\textcolor{red}{\ding{172} Red Discontinuity Noise}) and black (\textcolor{blue}{\ding{173} Black Plane Noise}) color, resulting in increased Euclidean distances for similar color and the introduction of artifacts in the final images (see the zoomed in images of \ding{172} and \ding{173}). 
These two types of noise can cause severe artifacts in the enhancement of red-dominated or extremely dark images.

To address these issues,
we introduce a new color space named \textit{\textbf{H}orizontal/\textbf{V}ertical-\textbf{I}ntensity (\textbf{HVI})}, designed specifically for the LLIE task.
The key intuition 
is that minimizing color space noise, by reducing Euclidean distances in similar colors.
To this end, we polarize in the Hue and Saturation (HS) plane to enforce smaller distances for similar red point coordinates, which eliminates the red discontinuity noise in the primary HSV space (see Fig. \ref{fig:1} (c)). 
For the black plane noise issue, we introduce a trainable darkness density parameter $k$ and its corresponding adaptive intensity collapse function $C_k$, which compresses the radius of low-light regions to zero, with the flexibility to gradually expand to the value of one as the intensity increases, as illustrated in Fig. \ref{fig:1} (d). This helps remove the black noise artifacts while maintaining the primary image appearance.

We further propose an LLIE network, named \textit{\textbf{C}olor and \textbf{I}ntensity \textbf{D}ecoupling Network (\textbf{CID}Net)}, to leverage the chromatic and intensity information for optimizing the color and intensity in the HVI space. After transforming the image into the HVI space, CIDNet leverages two network branches, namely HV-branch and intensity-branch, to respectively model decoupled color and brightness information for
learning accurate photometric mapping under different lighting conditions and restoring more natural colors. 



Our contributions can be summarized as follows:
\begin{itemize}
    \item We introduce a new HVI color space for the LLIE task, which is uniquely defined by polarized HS and trainable intensity. This offers an effective tool that eliminates the color space noise arising from the HSV space, largely enhancing the brightness of the low-light images.
    \item We further propose a novel LLIE network, CIDNet, to concurrently model the intensity and chromatic of low-light images in the HVI space. Despite being lightweight and computationally-efficient with relatively small parameters (1.88M) and computational loads (7.57GFLOPs), it enables the learning of accurate photometric mapping under different lighting conditions.
    \item Comprehensive results from quantitative and qualitative experiments show that our approach outperforms various types of state-of-the-art (SOTA) methods on different metrics across 10 datasets. 
\end{itemize}

\section{Related Work}
\label{sec:related}

\subsection{Low-Light Image Enhancement}

\textbf{Single-stage Methods.} Single-stage deep learning approaches \cite{LLNet,KinD,EnGAN,RetinexFormer,10244055,9809998} has been widely used in LLIE. 
Existing methods propose distinct solutions to address the aforementioned issues.
For instance, RetinexNet \cite{RetinexNet} enhances images by decoupling illumination and reflectance based on Retinex theory. 
Bread \cite{Bread} decouples the entanglement of noise and color
distortion by using YCbCr color space. Furthermore, they designed a color adaption network to tackle the color distortion issue left in light-enhanced images. 
Still, RetinexNet and Bread can show 
inaccurate control in terms of brightness and biased
color in black areas.

\textbf{Diffusion-based Methods.} With the advancement of Denoising Diffusion Probabilistic Models (DDPMs) \cite{Ho2020Denosing}, diffusion-based generative models have achieved remarkable results in the LLIE task. It has shown the capability to generate more accurate and appropriate images. However, they still exhibit issues such as local overexposure or color shift. Recent LLIE diffusion methods have attempted to address these challenges by incorporating global supervised brightness correction or employing local color correctors \cite{zhou2023pyramid,wu2023reco,hou2024global,wang2024zero,jiang2023low,feng2024difflight}.
Other methods such as Diff-Retinex \cite{yi2023diff} rethink the retinex theory with a diffusion-based model in the LLIE task, aiming to decomposed an image to illumination and reflectance from sRGB color space. 
However, these diffusion models often fail to fully decouple brightness and color information.

\subsection{Color Space}

\label{section:CS}

\textbf{RGB.} 
Currently, the most commonly used is the sRGB color space. For the same principle as visual recognition by the human eye, sRGB is widely used in digital imaging devices \cite{POYNTON2003187}. Nevertheless, image brightness and color exhibit a strong interdependence with the three channels in sRGB \cite{gevers2012color}.
A slight disturbance in the color space will cause an obvious variation in both the brightness and color of the generated image. Thus, sRGB is not a desired color space for low-light enhancement.

\textbf{HSV and YCbCr.} The HSV color space represents points in an RGB color model with a cylindrical coordinate system \cite{Foley1982FundamentalsOI}. Indeed, it does decouple brightness and color of the image from RGB channels. However, the red color discontinuity and black plane noise pose significant challenges when we enhance the images in HSV color space, resulting in the emergence of various obvious artifacts. To circumvent issues related to HSV, some methods \cite{Bread,brateanu2024lytnet} also transform sRGB images to the YCbCr color space, which has an illumination axis (Y) and reflect-color-plane (CbCr). Although it solved the hue dimension discontinuity problem of HSV, 
the Y axis is still coupled with the CbCr plane partially, leading to severe color shifts.

\section{HVI Color Space}
The HVI space is built upon the HSV color space, which is proposed to address the color space noise issues arising from the HSV space. The key intuition in HVI is that the restored images should have good perceptual quality for respective colors, \ie, similar colors have small Euclidean distances. Below we introduce the HVI transformation in detail, where the HSV color space is first applied to decouple the brightness and color information of input images, which could cause color space noise (\eg, red discontinuity and black plane). We then introduce our proposed polarized HS operations and learnable intensity collapse function in HVI to effectively address these issues.

\subsection{Color Space Noises in HSV}

\textbf{Intensity Map.} In the task of LLIE, one crucial aspect is accurately estimating the illumination intensity map of the scene from a sRGB input image.
Previous methods \cite{RetinexNet, PairLIE, KinD} largely rely on the Retinex theory \cite{land1977retinex}, using deep learning to directly generate the corresponding normal-light map. 
While this approach aligns with statistical principles, it often struggles to fit physical laws and human perception \cite{gevers2012color}, resulting in limited generalizability \cite{wang2024zero}. 
Therefore, we instead refer to the Max-RGB theory \cite{land1977retinex} to estimate the intensity map, \textit{a.k.a.}, Value in HSV, rather than using neural networks to generate it.
According to Max-RGB, for each individual pixel $x$, we can estimate the intensity map of an image $\mathbf{I}_{max} \in \mathbb{R}^{\mathrm{H\times W}} $ as follows:
\begin{equation}
\mathbf{I}_{max}(x) =\max_{\mathbf{c}\in \{R,G,B\}} (\mathbf{I_{c}}(x)).
\label{eq:1}
\end{equation}

The intensity map then goes through the sRGB-HSV transformation that can lead to various red and black noises. We introduce these separately as follows.

\textbf{Hue/Saturation Plane.}
Real-world low-light images often contain significant noise, making its identification and removal a key challenge in LLIE.
Recent studies \cite{RetinexNet,yi2023diff} indicate that the noise in low-light images is a primary cause to shifts in Hue and Saturation, \textit{a.k.a.} a general case of Reintex theory \cite{wang2024zero}, while having minimal impact on light intensity.
Therefore, decoupling the sRGB color space, known for its high color sensitivity, can be advantageous for the LLIE task. 
By leveraging pixel-based photometric invariance \cite{gevers2012color} and dichromatic reflection modeling \cite{shafer1985using}, sRGB can be decoupled into illuminance and chromatic components, yielding the HSV (Hue/Saturation-Value) color space. 
In this representation, the Value map ($\mathbf{V}$) component corresponds to light intensity map ($\mathbf{V}=\mathbf{I}_{max}$), while the HS plane forms a chromaticity plane independent of illuminance constraints. Specifically, the transformation of sRGB image to Saturation map ($\mathbf{S}$) is defined as follows
\begin{equation}
\mathbf{s}  = \begin{cases}
0, &  \mathbf{I}_{max}   = \mathbf{0} \\
\frac{\Delta}{\mathbf{I}_{max} }, &  \mathbf{I}_{max}  \ne \mathbf{0} \\
\end{cases} \\
,
\end{equation}
where $\Delta=\mathbf{I}_{max}-min(\mathbf{I}_c)$ and $\mathbf{s}$ is any pixel in $\mathbf{S}$. The Hue map ($\mathbf{H}$) is formulated as
\begin{equation}
\mathbf{h}= \begin{cases}
0, &\text{if } \mathbf{s} = 0 \\
\frac{\mathbf{I_{G}} - \mathbf{I_{B}}}{\Delta} \mod{6},  &\text{if } \mathbf{I}_{max}  = \mathbf{I_{R}} \\
2+\frac{\mathbf{I_{B}} - \mathbf{I_{R}}}{\Delta},  &\text{if } \mathbf{I}_{max}   = \mathbf{I_{G}} \\
4+\frac{\mathbf{I_{R}} - \mathbf{I_{G}}}{\Delta},  &\text{if } \mathbf{I}_{max}   = \mathbf{I_{B}}
\end{cases}
,
\label{eq:hue}
\end{equation}
where $\mathbf{h}$ is any pixel in $\mathbf{H}$. 

\textbf{Color Space Noises.} 
Converting an sRGB image to HSV space effectively decouples brightness from color, enabling more accurate color denoising and more natural illuminance recovery. 
However, this transformation also amplifies noise in the red and dark regions \cite{gevers2012color}, which are critical for LLIE tasks. 
As illustrated in Fig. \ref{fig:1} (b), enhancing the image brightness within the HSV color space yields a more balanced brightness level. 
However, excessive noise in the red discontinuities and the black plane introduces significant artifacts, particularly in red and dark areas of output image, which greatly degrade the perceptual quality. 

To address this issue, we propose the HVI color space as follows, which effectively preserves the decoupling of brightness and color while minimizing these artifacts.

\subsection{Horizontal/Vertical Plane with Polarized HS and Collapsible Intensity}
\label{sec:HVI}
Our primary approach to addressing the color space noise issue is to ensure that more similar colors exhibit smaller Euclidean distances. 
Along the Hue axis, the color red appears identically at both $\mathbf{h}=0$ and $\mathbf{h}=6$, due to the modular arithmetic of Hue-axis, which splits the same color across two ends of the spectrum. 
In particular,
to address the red discontinuity issue, we apply polarization to the Hue axis ($\mathbf{h}$) at each pixel in $\mathbf{H}$, obtaining orthogonal $h= \cos (\frac{\pi \mathbf{h}}{3})$ and $v= \sin (\frac{\pi \mathbf{h}}{3})$.
When the Hue axis is polarized, it forms an angle within the orthogonalized $h-v$ plane, with $\mathbf{s}$ representing the distance from the origin.

For the black plane noise issue, we aim to collapse regions of low light intensity while preserving those with higher intensity. However, the optimal extent of collapse varies across different datasets and networks. 
Therefore, it is important to make this region adaptively collapsible through a learning process. 
To achieve this, we introduce an adaptive intensity collapse function $\mathbf{C}_k$ as follows
\begin{equation}
\mathbf{C}_k(x)=\sqrt[k]{\sin (\frac{ \pi \mathbf{I}_{max}(x) }{2} )+\mathcal{\varepsilon} },
\label{eq:2}
\end{equation}
where $k\in \mathbb{Q^+}$ is a trainable parameter to control the dark color point density, and a small  $\mathcal{\varepsilon}=1\times10^{-8}$ is used to avoid gradient explosion. 
Essentially, $\mathbf{C}_k$ serves a radius mapping function, with smaller $\mathbf{C}_k$ corresponding to smaller radius or lower intensity values. Thus, black points are clustered together as $\mathbf{C}_k$ decreases. We then formalize the Horizontal ($\mathbf{\hat{H}}$) map and Vertical ($\mathbf{\hat{V}}$) map as
\begin{equation}
\begin{split}
    \mathbf{\hat{H}} &= \mathbf{C}_k \odot  \mathbf{S}  \odot H,\\
    \mathbf{\hat{V}} &= \mathbf{C}_k \odot  \mathbf{S}  \odot V,
\end{split}
\label{eq:7}
\end{equation}
where $h \in H$, $v \in V$, and $\odot$ denotes the element-wise multiplication. $\mathbf{\hat{H}}$, $\mathbf{\hat{V}}$, and $\mathbf{I}_{max}$ can be concatenated to form an HVI image.

Thanks to these operations, the HVI space builds a strong color space that maintains the advantages of HSV while removing the red and black noises. More importantly, since HVI is trainable due to $k$ and $\mathbf{C}_k$, specifically designed neural networks, as discussed in the following, can be created to optimize LLIE upon HVI under various lighting conditions.
\section{Color and Intensity Decoupling Network}

\begin{figure*}[!t]
    \centering
    \includegraphics[width=1\linewidth]{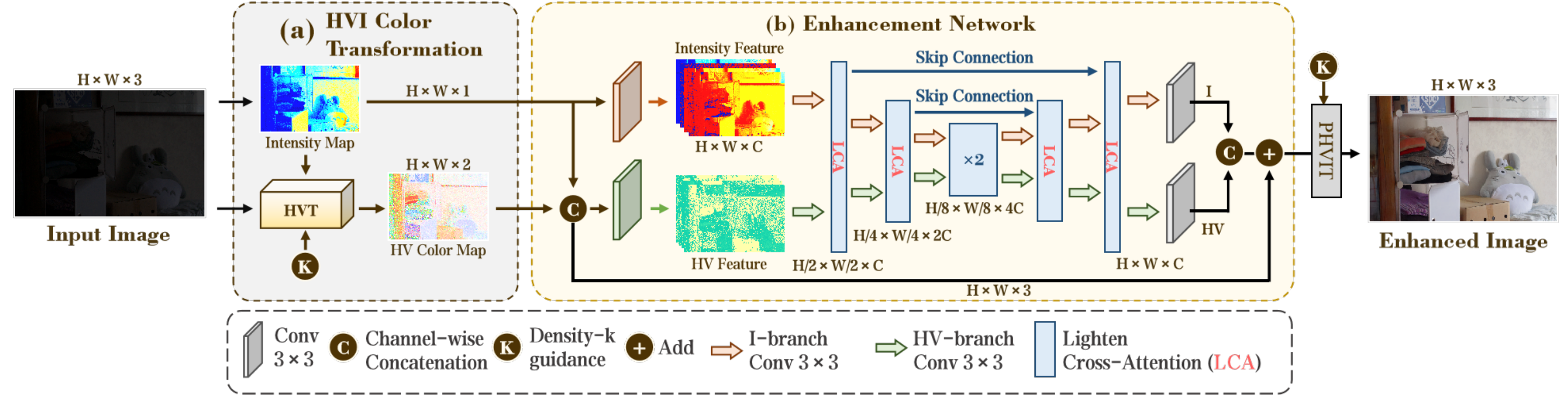}
    \caption{The overview of the proposed CIDNet. \textbf{(a)} HVI Color Transformation (HVIT) takes an sRGB image as input and generates HV color map and intensity map as outputs. \textbf{(b)} Enhancement Network performs the main processing, utilizing a dual-branch UNet architecture with six Lighten Cross-Attention (\textcolor[RGB]{255,95,95}{LCA}) blocks. Lastly, we apply Perceptual-inverse HVI Transform (PHVIT) to take a light-up HVI map as input and transform it into an sRGB-enhanced image.}
    \label{fig:CIDNet-pipeline}
\end{figure*}

To more effectively utilize chromatic and brightness information in the HVI space, we introduce a novel dual-branch LLIE network, named Color and Intensity Decoupling Network (CIDNet), to separately model the HV-plane and I-axis information in the HVI space, as shown in Fig. \ref{fig:CIDNet-pipeline}. CIDNet employs HV-branch to suppress the noise and chromaticity in the dark regions and utilizes I-branch to estimate the illuminance of the whole images.

The overall framework of CIDNet can be divided into three consecutive main steps. There is an HVI transformation applied before the dual-branch enhancement network. After the enhancement, CIDNet performs perceptual-inverse HVI transformation to map the image to the sRGB space. Below we introduce each step in detail.

\subsection{HVI Transformation}
As described in Sec. \ref{sec:HVI}, the HVI transformation decomposes the sRGB image into two components: an intensity map containing scene illuminance information and an HV color map containing scene color and structure information. 
Specifically, we first calculate the intensity map using Eq. \ref{eq:1}, which is $\mathbf{I_{I}}=\mathbf{I}_{max}$. Subsequently, we utilize the intensity map and the original image to generate HV color map using Eq. \ref{eq:7}.
Furthermore, a trainable density-$k$ is employed to adjust the color point density of the low-intensity color plane, as shown in Fig. \ref{fig:CIDNet-pipeline}(a).


\subsection{Dual-branch Enhancement Network}
\label{sec:dual}
As illustrated in Fig. \ref{fig:CIDNet-pipeline} (b), the dual-branch network is built upon the UNet architecture, involving an encoder and an decoder with respective three Lighten Cross-Attention (LCA) modules, and multiple skip connections. Two key designs in the network are the dual-branch structure and the LCA module.
We explain the key intuition behind these designs as follows.

The LLIE task can be decomposed into two sub-tasks: noise removal in low-light regions and brightness enhancement. 
By converting the image to the HVI color space, where luminance and color are decoupled, we can apply brightness mapping to the Intensity map and denoising to the HV color map. 
Since these two sub-tasks follow distinct statistical patterns \cite{2022LLE}, inspired by Retinex-based methods \cite{RetinexNet,URetinexNet,PairLIE}, we use separate branches, the I-branch and HV-branch, to address each sub-task individually.
Additionally, the input to the HV branch is formed by concatenating the Intensity map with the HV color map, as we observed that severely low-light images contain a small amount of noise in the luminance component as well.

The cross-attention between the two branches is used, rather than using self-attention individually to each branch. One main reason is that the illumination intensity is inversely proportional to image noise intensity. 
A low-light image may also contain high-illumination regions requiring only minimal denoising and enhancement. 
Therefore, using the intensity features to guide the HV-branch in denoising can reduce global color shifts and achieve more effective noise suppression.
For another reason, the noisy intensity information, after being denoised in the HV-branch, is transferred to the I-branch through cross-attention, resulting in smoother enhancement outcomes.

\subsection{Perceptual-inverse HVI Transformation}
To convert HVI back to the HSV color space, we perform a Perceptual-inverse HVI Transformation (PHVIT), which is a surjective mapping while allowing for the independent adjustment of the image’s saturation and brightness.

The PHVIT sets $\hat{h}$ and $\hat{v}$ as an intermediate variable as $\hat{h}=\frac{ \mathbf{\hat{H}}}{\mathbf{C}_k+\mathcal{\varepsilon}}$,$\hat{v}=\frac{ \mathbf{\hat{V}}}{\mathbf{C}_k+\mathcal{\varepsilon}}$, where $\mathcal{\varepsilon}=1\times10^{-8}$ is used to avoid gradient explosion. 
Then, we convert $\hat{h}$ and $\hat{v}$ to HSV color space.
The Hue ($\mathbf{H}$), Saturation ($\mathbf{S}$) and Value ($\mathbf{V}$) map can be estimated as
\begin{equation}
\begin{split}
    \mathbf{H} &= \arctan(\frac{\hat{v}}{\hat{h}})\mod 1,\\
    \mathbf{S} &= \alpha_{S}\sqrt{\hat{h}^{2}+\hat{v}^{2}},\\
    \mathbf{V} &= \alpha_{I}\hat{\mathbf{I}}_{\mathbf{I}},
\end{split}
\end{equation}
where $\alpha_{S},\alpha_{I}$ are the customizing linear parameters to change the image color saturation and brightness. Finally, we will obtain the sRGB image with HSV image \cite{Foley1982FundamentalsOI}.

\subsection{Loss Function}
To provide comprehensive supervision for training CIDNet, we guide the enhancement from two key perspectives, including the GroundTruth in the sRGB space and the HVI map in the HVI space. Specifically, given the enhanced HVI map $\mathbf{\hat{I}_{HVI}}$ and the restored sRGB image $\mathbf{\hat{I}}$ output from CIDNet, we aim to minimize their difference to the primary sRGB GroundTruth $\mathbf{I}$ and its corresponding HVI map $\mathbf{I_{HVI}}$:
\begin{equation}
L= \lambda\cdot l(\mathbf{\hat{I}_{HVI}},\mathbf{I_{HVI}}) + l(\mathbf{\hat{I}},\mathbf{I}),
\end{equation}
where $\lambda$ is a weighting hyperparameter to balance the losses in the two different color spaces. This helps achieve not only more closely with the probabilistic distribution of sRGB in the HVI space, especially the red and black ones, due to the optimized $k$ and $\mathbf{C}_k$, but also inheritance of pixel-level structure detail in the sRGB space. 
\section{Experiments}

\subsection{Datasets and Settings}
We employ seven commonly-used LLIE benchmark datasets for evaluation, including LOLv1 \cite{RetinexNet}, LOLv2 \cite{LOLv2}, DICM \cite{DICM}, LIME \cite{LIME}, MEF \cite{MEF}, NPE \cite{NPE}, and VV \cite{VV}. 
We also conduct further experiments on two extreme datasets, SICE \cite{SICE} (containing mix and grad test sets \cite{SICE-Mix}) and SID (Sony-Total-Dark) \cite{SID}. 

\textbf{LOL.} The LOL dataset has v1 \cite{RetinexNet} and v2 \cite{LOLv2} versions. Compared to LOL-v1 that contains both real and synthetic data, LOL-v2 is divided into real and synthetic subsets. 
For LOLv1 and LOLv2-Real, we crop the training images into $400\times400$ patches and train CIDNet for 1,500 epochs with a batch size of 8. For LOLv2-Synthetic, we set the batch size to 1 and train CIDNet for 500 epochs without cropping.

\textbf{SICE.} The original SICE dataset \cite{SICE} contains 589 low-light and overexposed images, with the training, validation, and test sets divided into three groups according to 7:1:2. CIDNet is optimized using $160\times160$ cropped images from the training set for 1,000 epochs with a batch size of 10 and tested on the datasets SICE-Mix and SICE-Grad \cite{SICE-Mix}. 

\textbf{Sony-Total-Dark.} This dataset is a customized version of a SID subset \cite{SID}.
To make this dataset more challenging, we convert the raw format images to sRGB images with \textit{no gamma correction}, resulting in images with extreme darkness. We crop the training images into $256\times256$ patches and train CIDNet for 1,000 epochs with a batch size of 4.

\textbf{Experiment Settings.} We implement our CIDNet by PyTorch. The model is trained with the Adam \cite{Adam} optimizer (\textit{$\beta_{1}$} = 0.9 and \textit{$\beta_{2}$} = 0.999) by using a single NIVIDA 2080Ti or 3090 GPU. The learning rate is initially set to $1 \times 10^{-4}$ and then steadily decreased to $1 \times 10^{-7}$ by the cosine annealing scheme \cite{sgdr} during the training process. 

\textbf{Evaluation Metrics. }For the \textit{paired} datasets, we adopt the Peak Signal-to-Noise Ratio (PSNR) and Structural Similarity (SSIM) \cite{SSIM} as the distortion metrics. 
To comprehensively evaluate the perceptual quality of the restored images, we report Learned Perceptual Image Patch Similarity (LPIPS) \cite{LPIPS} with AlexNet \cite{Alex} as the reference.
For the \textit{unpaired} datasets, we evaluate single recovered images using BRISQUE \cite{BRISQUE} and NIQE \cite{NIQE} perceptually.

\begin{table*}
    \centering
    \renewcommand{\arraystretch}{1.}
    \caption{Quantitative results of PSNR/SSIM$\uparrow$ and LPIPS$\downarrow$ on the LOL (v1 and v2) datasets. Due to the limited number of test set in LOLv1, we use GT mean method during testing to minimize errors. This approach will be explained in supplementary. The FLOPs is tested on a single $256\times256$ image. The best performance is in \textcolor{red}{red} color and the second best is in \textcolor{cyan}{cyan} color.}
\vspace{-0.2cm}
    \resizebox{\textwidth}{!}{
    \begin{tabular}{c|c|cc|ccc|ccc|ccc}
        \Xhline{1.5pt}
        \multirow{2}{*}{\textbf{Methods}}&\multirow{2}{*}{\textbf{Color Model}}&\multicolumn{2}{c|}{\textbf{Complexity}}&\multicolumn{3}{c|}{\textbf{LOLv1}} & \multicolumn{3}{c|}{\textbf{LOLv2-Real}} & \multicolumn{3}{c}{\textbf{LOLv2-Synthetic}}\\
        
        ~&~&Params/M&FLOPs/G&PSNR$\uparrow$&SSIM$\uparrow$&LPIPS$\downarrow$&PSNR$\uparrow$&SSIM$\uparrow$&LPIPS$\downarrow$&PSNR$\uparrow$&SSIM$\uparrow$&LPIPS$\downarrow$\\
        \hline
        RetinexNet \cite{RetinexNet}& Retinex& 0.84& 584.47 &	18.915 &	0.427 & 0.470&	16.097 &	0.401 	&0.543 &	17.137 &	0.762 &	0.255
\\
        KinD \cite{KinD}& Retinex & 8.02& 34.99 &23.018& 	0.843& 0.156&	17.544& 	0.669 &	0.375	&	18.320 	&0.796 &0.252
\\
         ZeroDCE \cite{Zero-DCE}& RGB&0.075&4.83& 	21.880 &	0.640&0.335& 16.059 &	0.580 & 0.313&	17.712& 	0.815 &	0.169
\\

         

         RUAS \cite{RUAS}& Retinex & 0.003& 0.83 &	18.654& 	0.518 &	0.270&15.326 &	0.488 &	0.310&13.765 	&0.638 	&0.305
\\
        LLFlow \cite{LLFlow}& RGB&17.42 & 358.4 &24.998 &	0.871&	0.117&17.433 &	0.831 & 0.176&	24.807 &	0.919 &	0.067
\\
         EnlightenGAN \cite{EnGAN}& RGB& 114.35 & 61.01 &  20.003 & 0.691&0.317 &18.230 & 0.617 &0.309 & 16.570& 0.734&0.220
\\

         SNR-Aware \cite{SNR-Aware}& SNR+RGB& 4.01 & 26.35 &26.716 	&0.851 	&0.152 &21.480 &	\textcolor{cyan}{0.849} &0.163	&	24.140 &	0.928 & 0.056
\\
        Bread \cite{Bread}& YCbCr&2.02 & 19.85 &25.299 	&0.847 	&0.155 &20.830 &	0.847 &0.174	&	17.630 &	0.919 & 0.091
\\
        PairLIE \cite{PairLIE}& Retinex& 0.33 & 20.81 &	23.526 &	0.755& 0.248&	19.885 &	0.778 & 0.317&	19.074	&0.794&	0.230
\\
        LLFormer \cite{LLFormer}& RGB& 24.55&22.52&25.758&0.823&0.167&20.056&0.792& 0.211&24.038&0.909&0.066
\\
         RetinexFormer \cite{RetinexFormer}& Retinex& 1.53 & 15.85 & 	27.140 & 	0.850  & 0.129&	\textcolor{cyan}{22.794}  &	0.840  & 0.171&	\color{cyan}{25.670}  &	\textcolor{cyan}{0.930}  &	0.059
\\
        GSAD \cite{hou2024global}& RGB& 17.36 & 442.02 & 	\color{cyan}{27.605}& 	\color{cyan}{0.876}  & \color{cyan}{0.092}&	20.153  &	0.846  & \color{cyan}{0.113}&	24.472 &	0.929  &	\color{cyan}{0.051}
\\
       QuadPrior \cite{wang2024zero}& Kubelka-Munk& 1252.75 & 1103.20 & 	22.849& 	0.800  & 0.201&	20.592  &	0.811  & 0.202&	16.108 &	0.758 &	0.114
\\
        \textbf{CIDNet(Ours)}& HVI& 1.88 & 7.57 & 	\color{red}{28.201} &	\color{red}{0.889} &\color{red}{0.079}	&\color{red}{24.111}& 	\color{red}{0.871} &\color{red}{0.108}  &	\textcolor{red}{25.705} &	\color{red}{0.942} & \color{red}{0.045}
\\
         \Xhline{1.5pt}
    \end{tabular}
    }
    \label{tab:table-LOL}
\end{table*}

\begin{figure*}[!t]
\centering
\begin{minipage}[t]{0.12\linewidth}
    \centering
    \vspace{3pt}
    \centerline{\includegraphics[width=\textwidth]{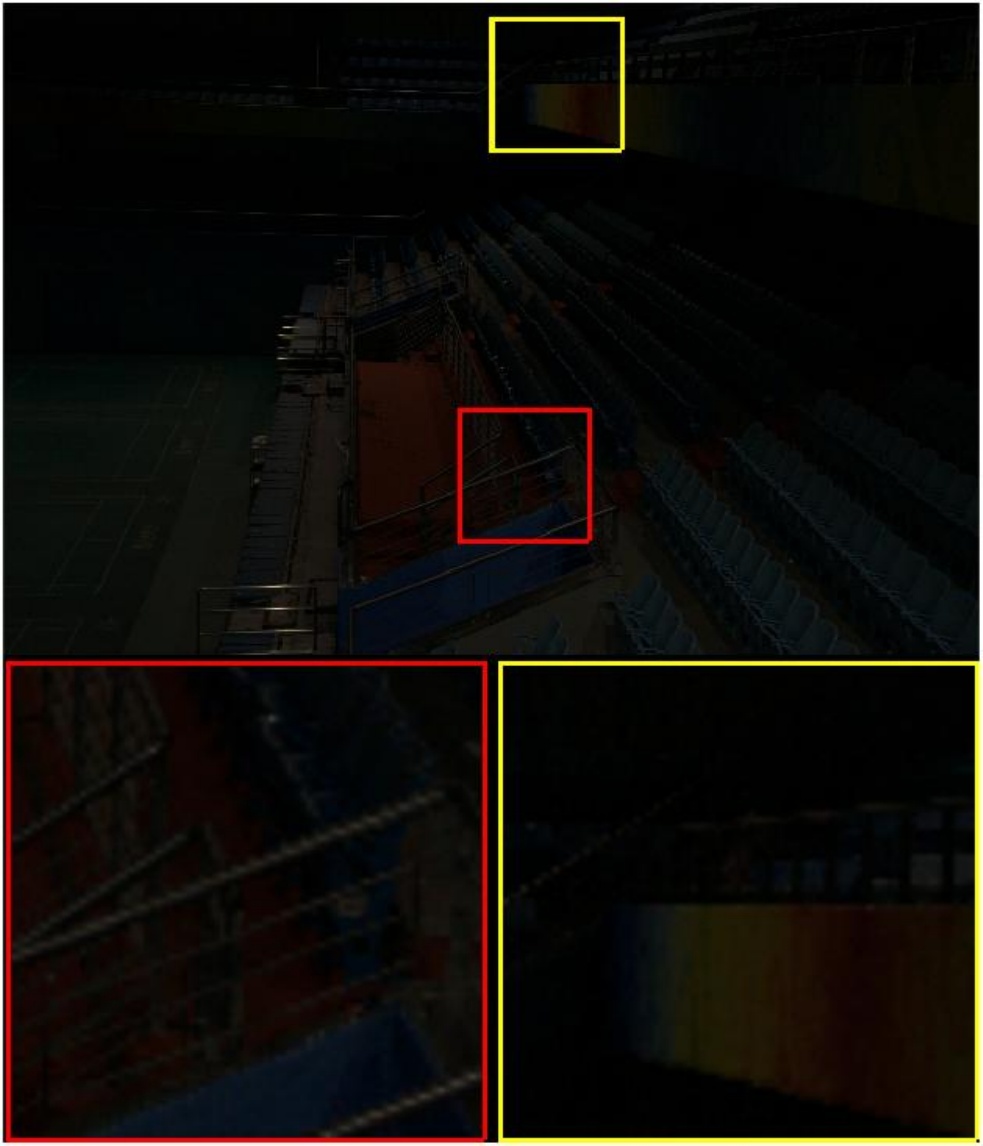}}
    \centerline{\includegraphics[width=\textwidth]{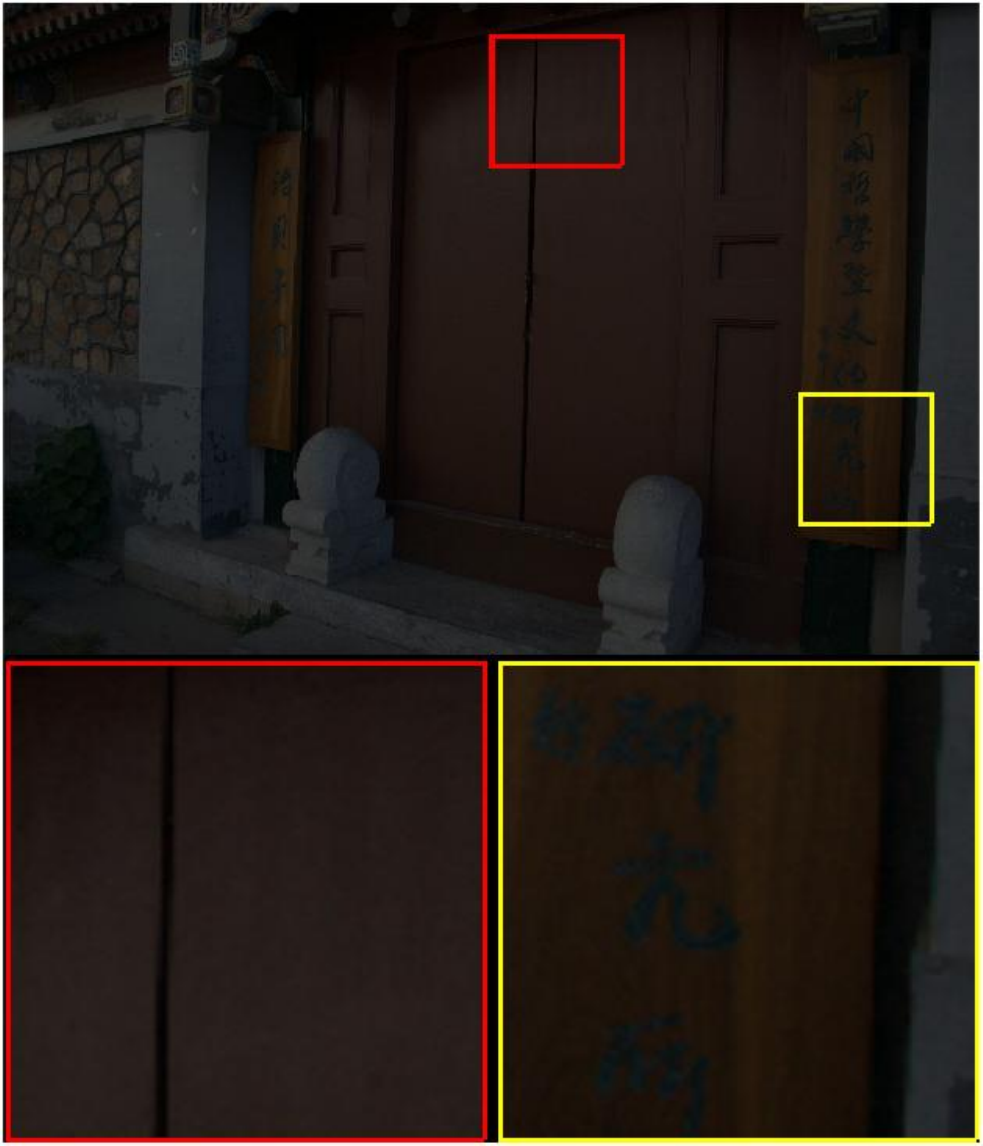}}
    \centerline{\small Input}
\end{minipage}
\begin{minipage}[t]{0.12\linewidth}
    \centering
    \vspace{3pt}
    \centerline{\includegraphics[width=\textwidth]{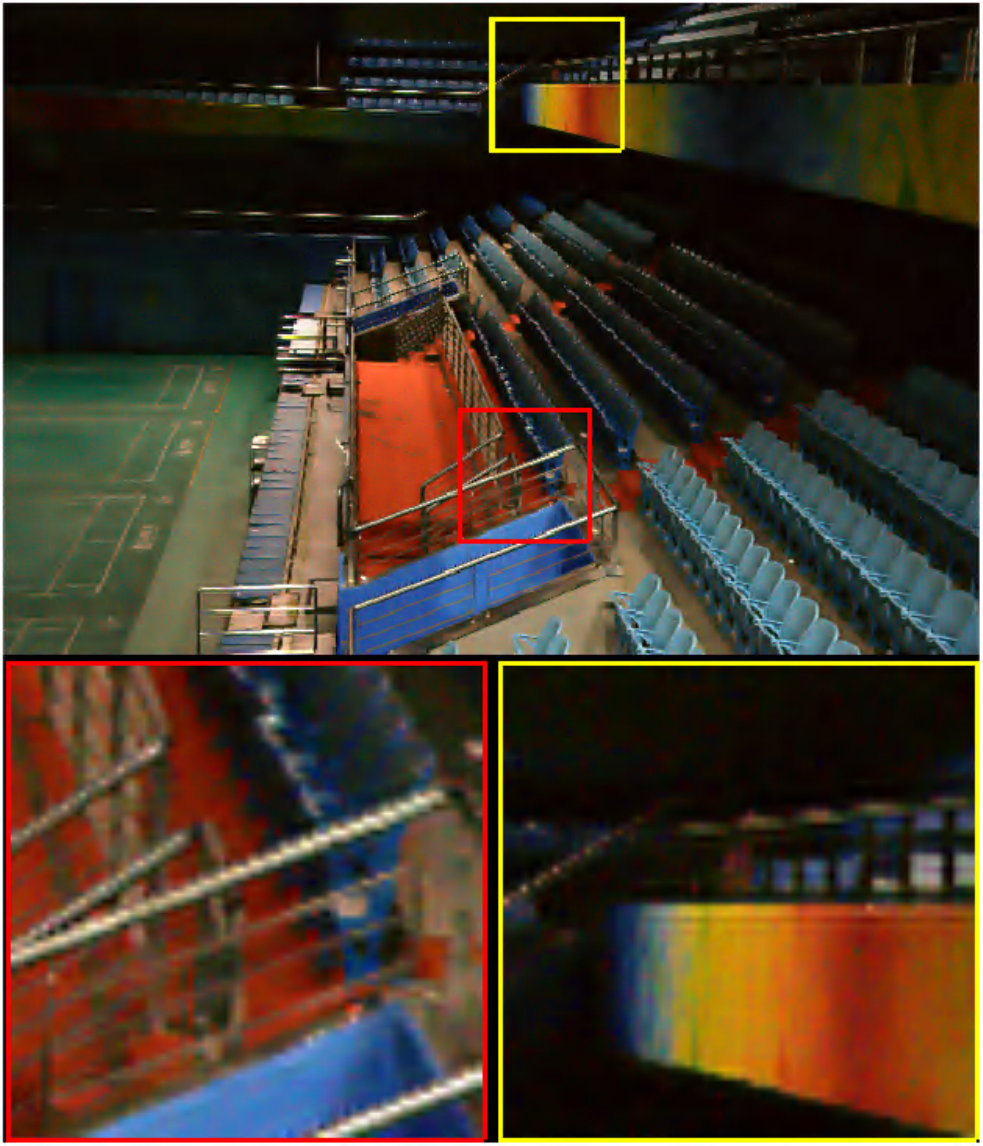}}
    \centerline{\includegraphics[width=\textwidth]{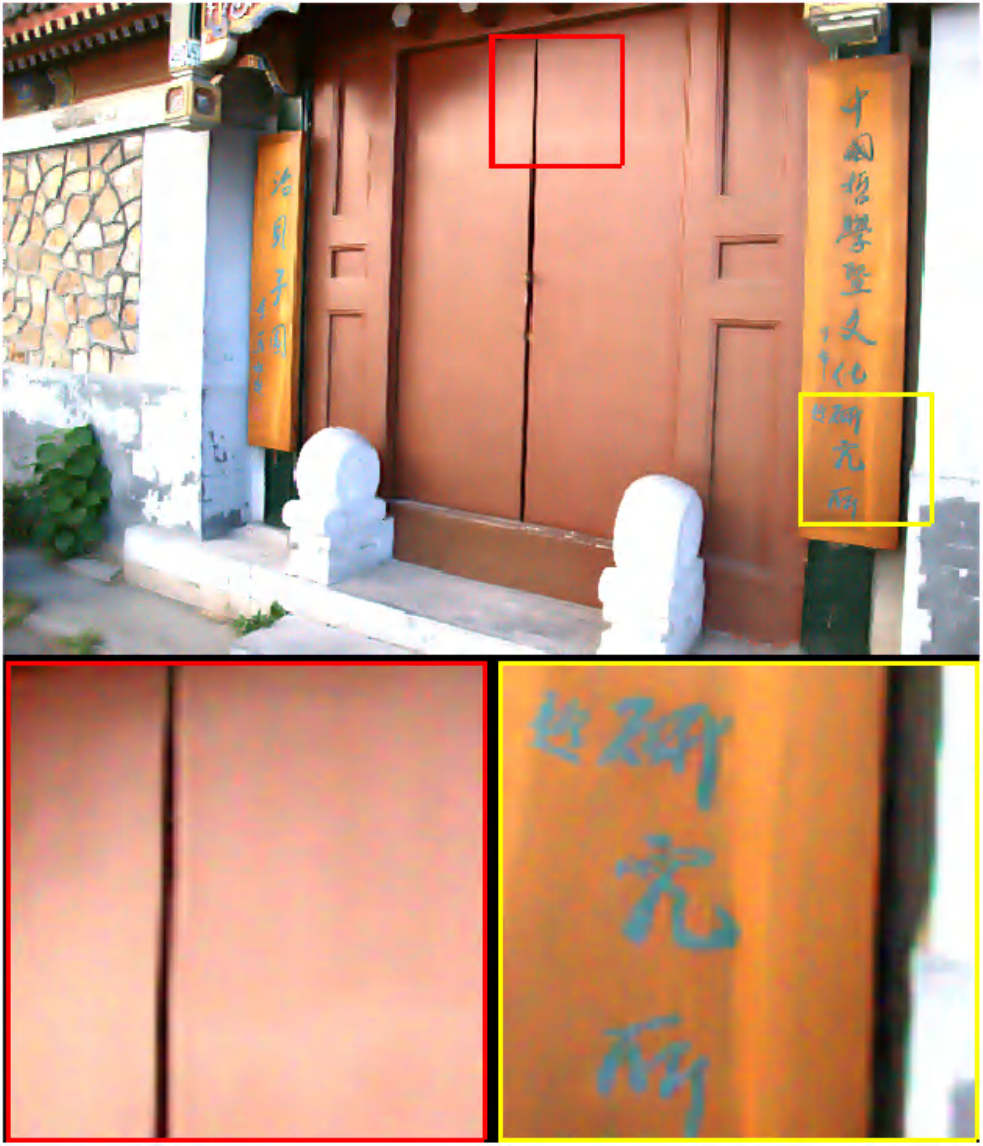}}
    \centerline{\small RUAS \cite{RUAS}}
\end{minipage}
\begin{minipage}[t]{0.12\linewidth}
    \centering
    \vspace{3pt}
    \centerline{\includegraphics[width=\textwidth]{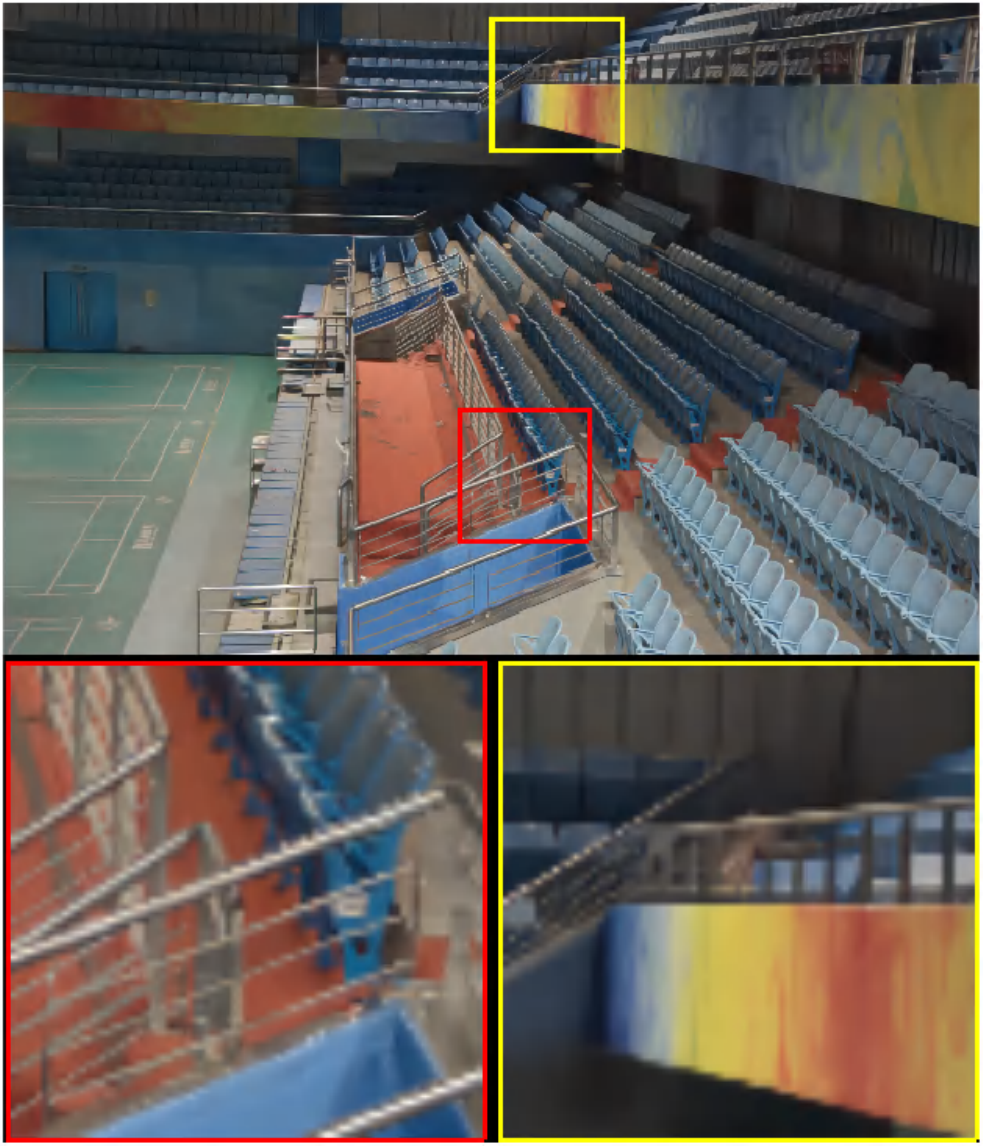}}
    \centerline{\includegraphics[width=\textwidth]{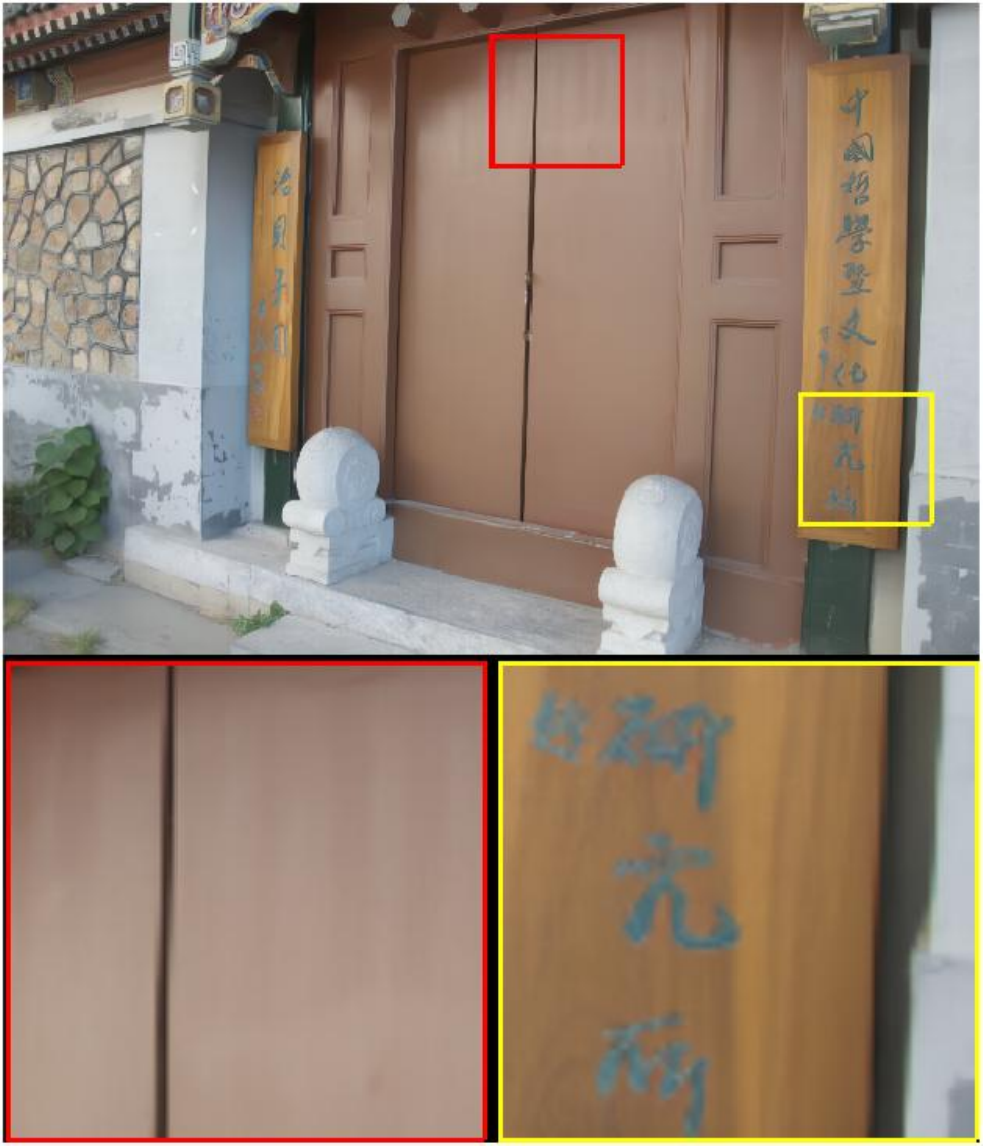}}
    \centerline{\small LLFlow \cite{LLFlow}}
\end{minipage}
\begin{minipage}[t]{0.12\linewidth}
    \centering
    \vspace{3pt}
    \centerline{\includegraphics[width=\textwidth]{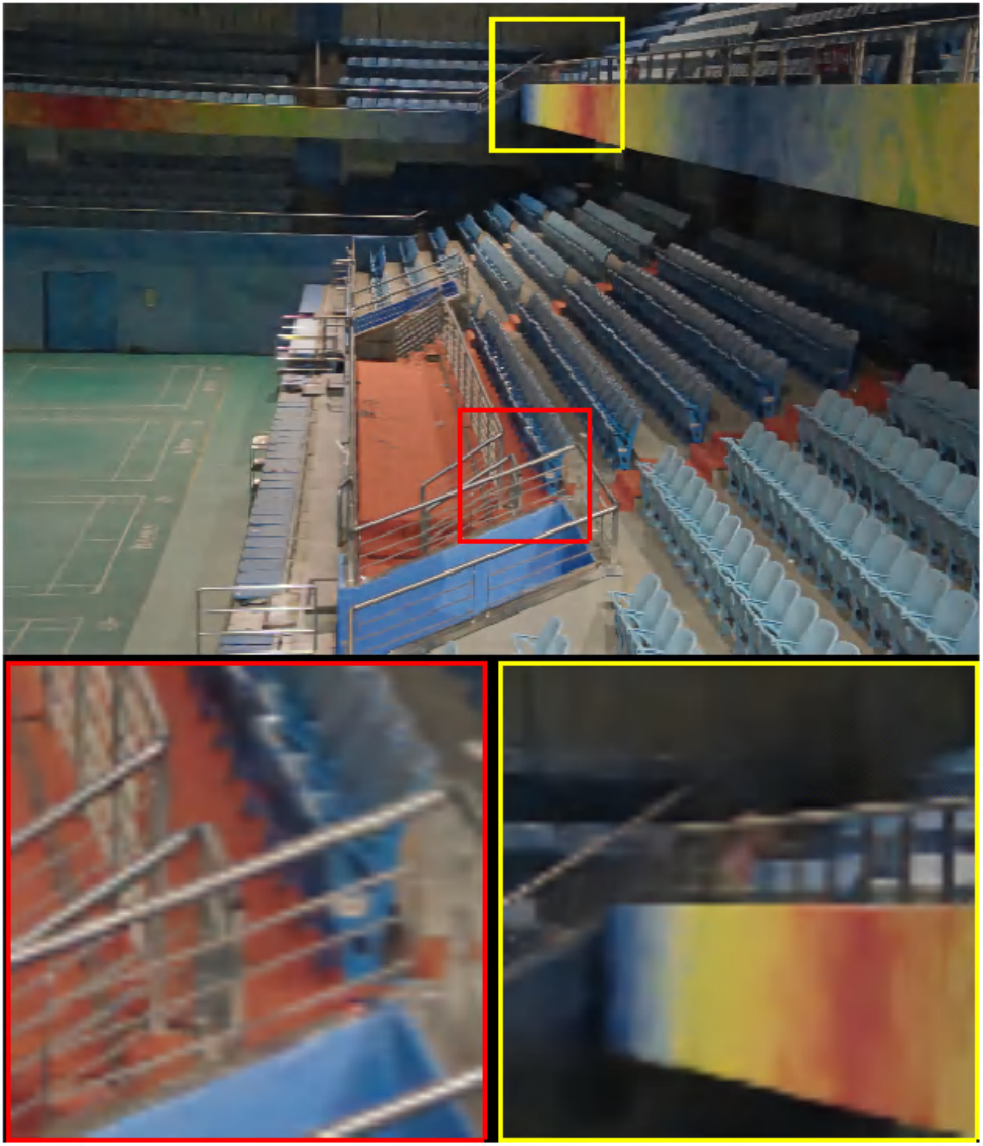}}
    \centerline{\includegraphics[width=\textwidth]{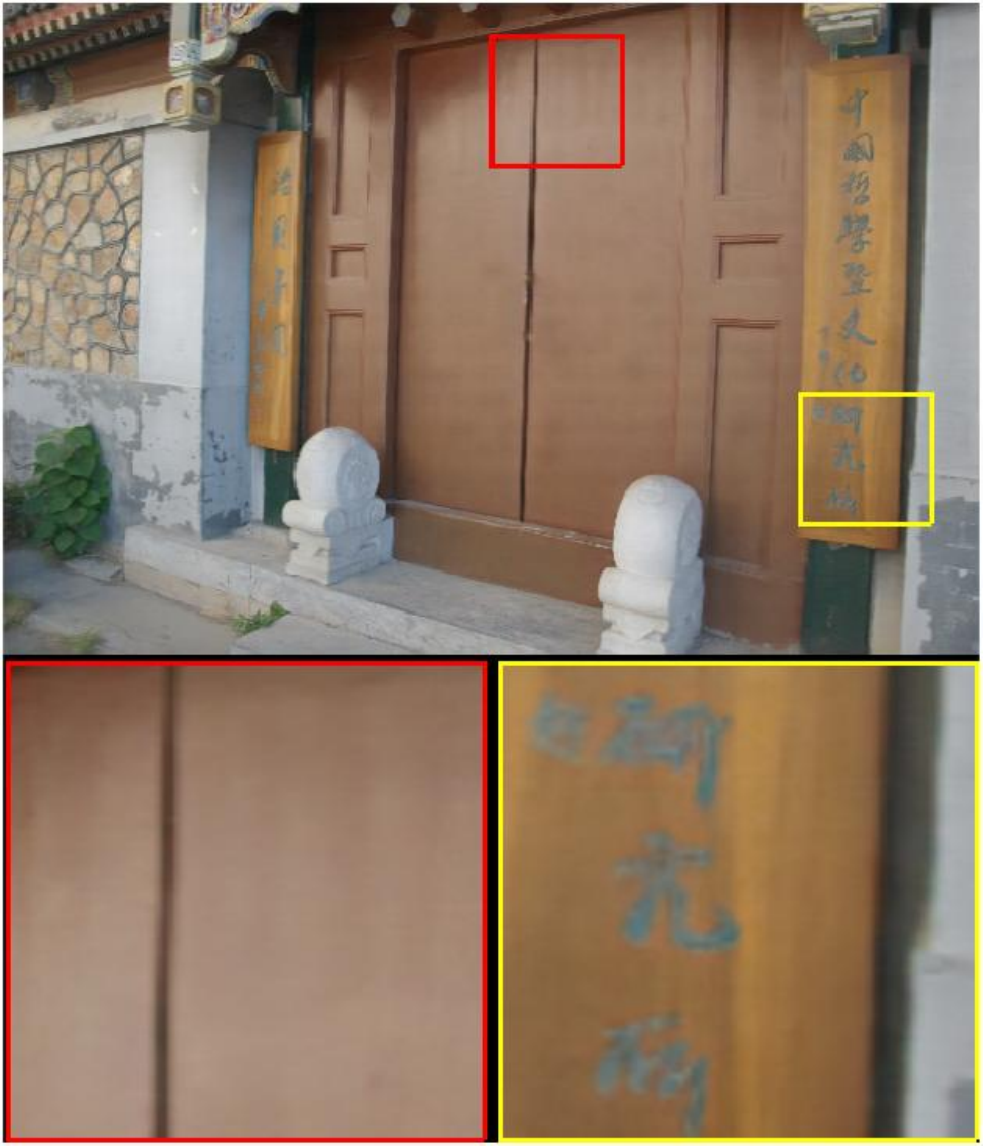}}
    \centerline{\small SNRNet \cite{SNR-Aware}}
\end{minipage}
\begin{minipage}[t]{0.12\linewidth}
    \centering
    \vspace{3pt}
    \centerline{\includegraphics[width=\textwidth]{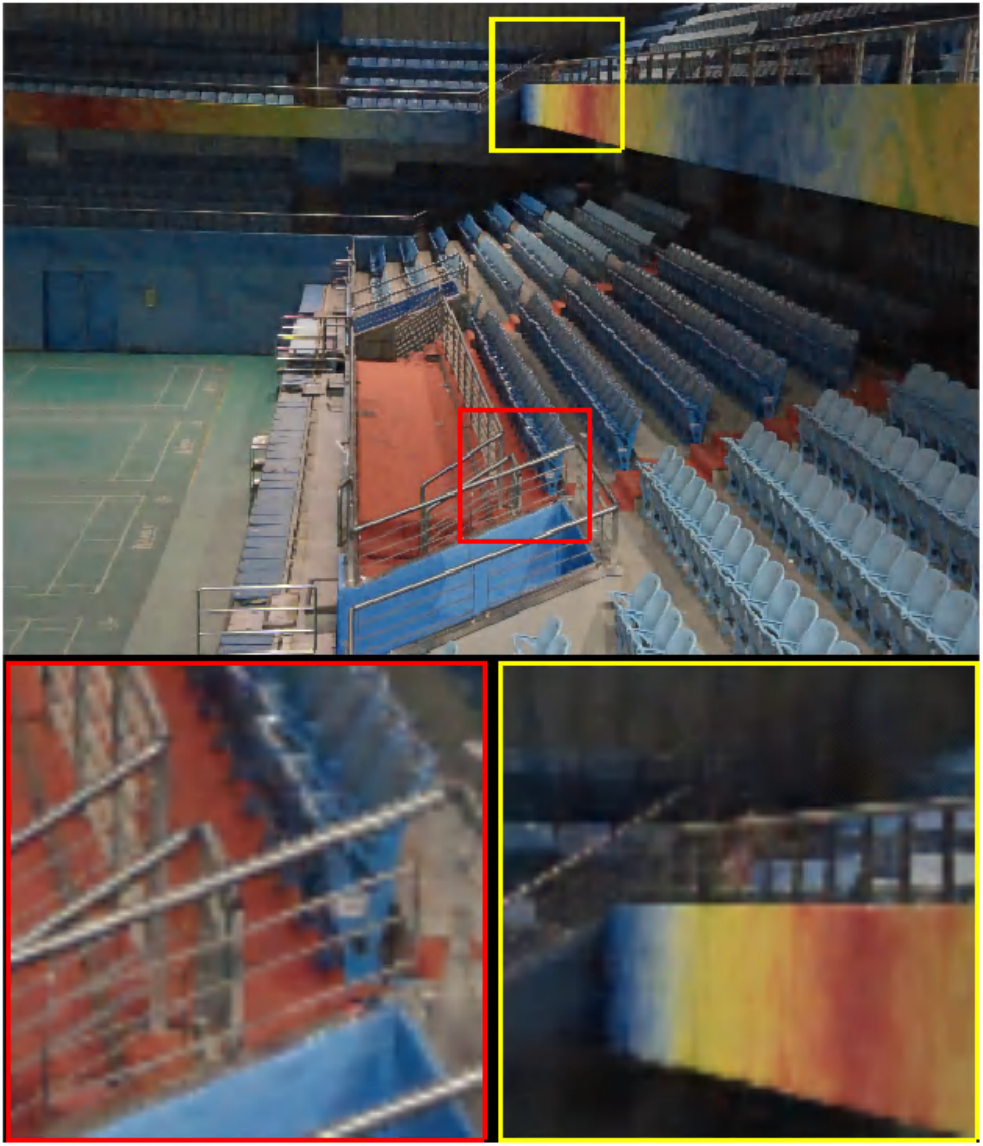}}
    \centerline{\includegraphics[width=\textwidth]{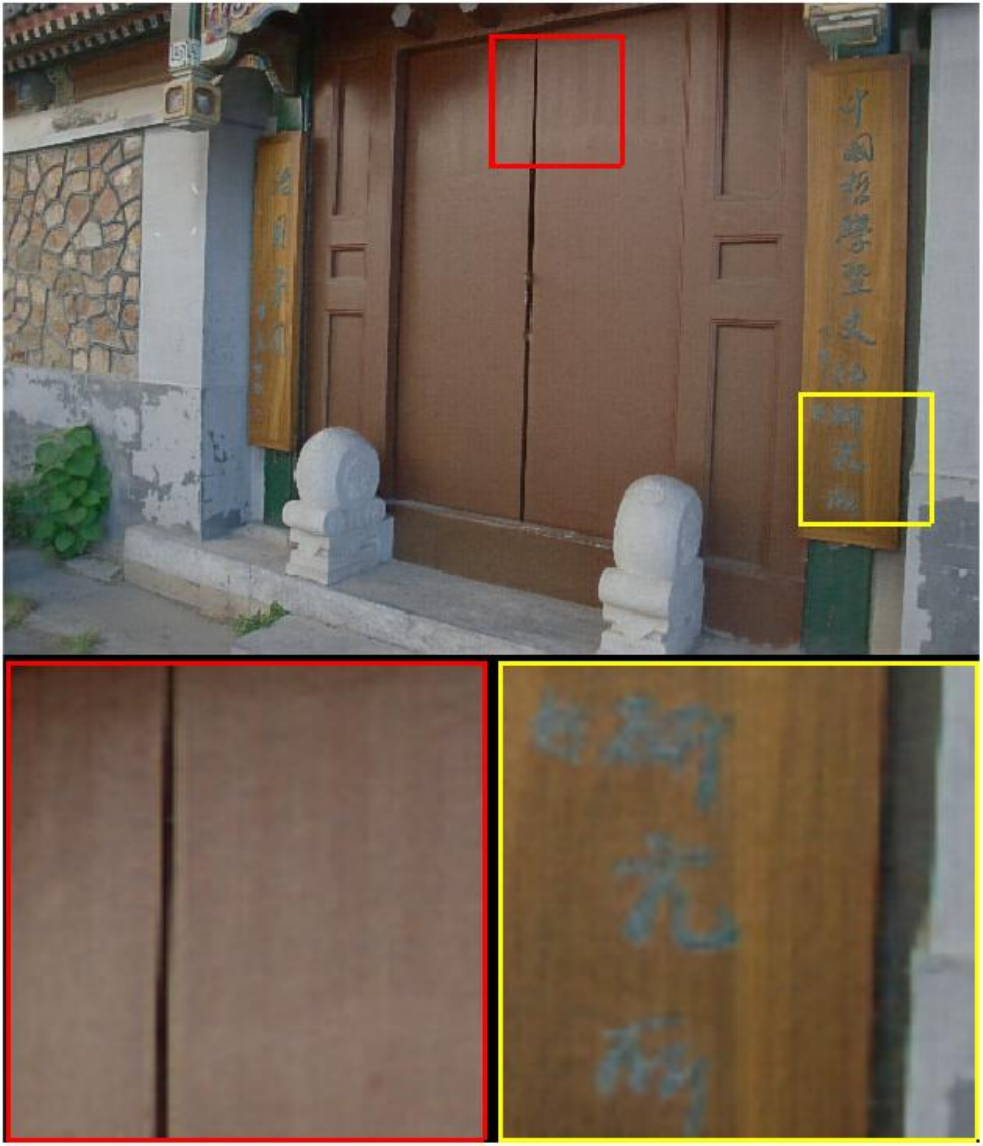}}
    \centerline{\small RetFormer \cite{RetinexFormer}}
\end{minipage}
\begin{minipage}[t]{0.12\linewidth}
    \centering
    \vspace{3pt}
    \centerline{\includegraphics[width=\textwidth]{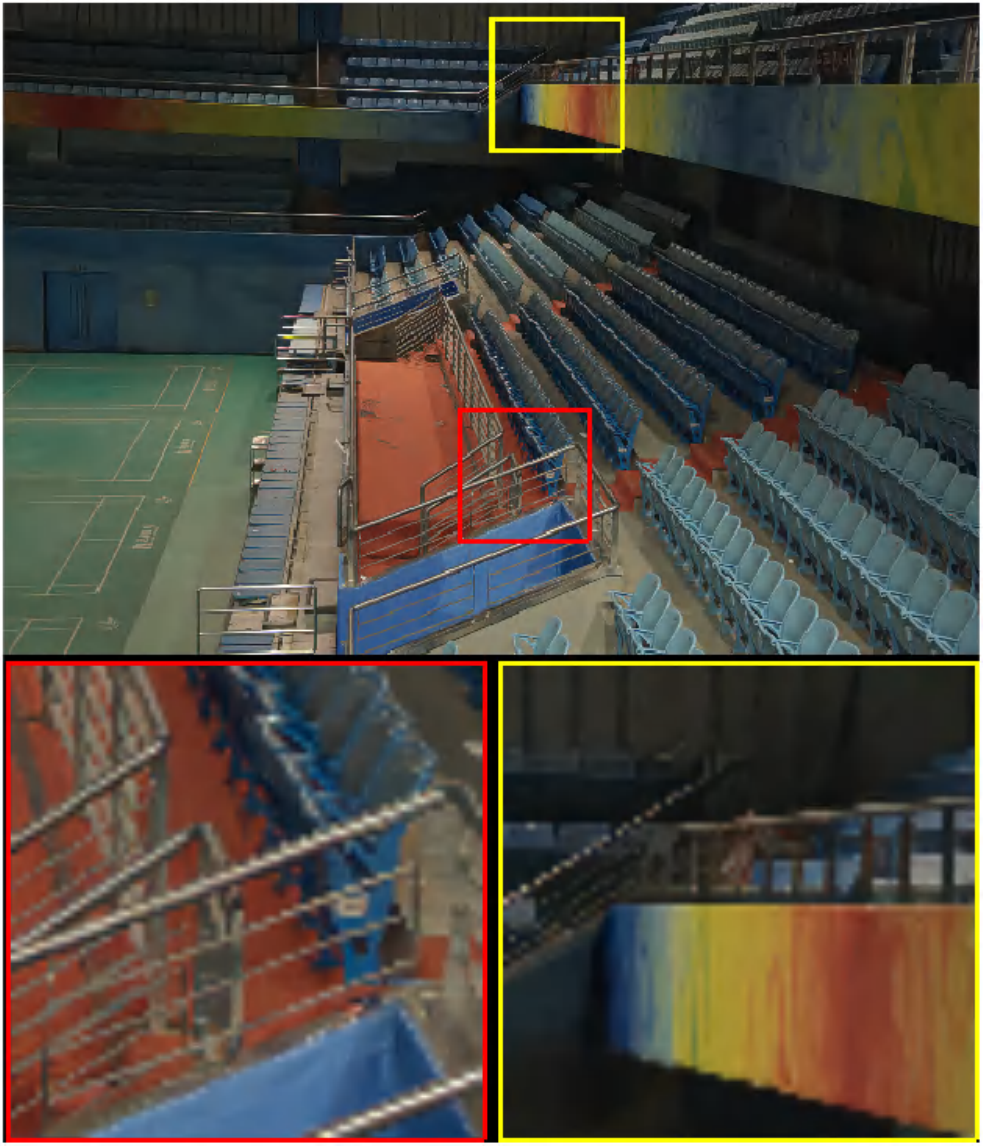}}
    \centerline{\includegraphics[width=\textwidth]{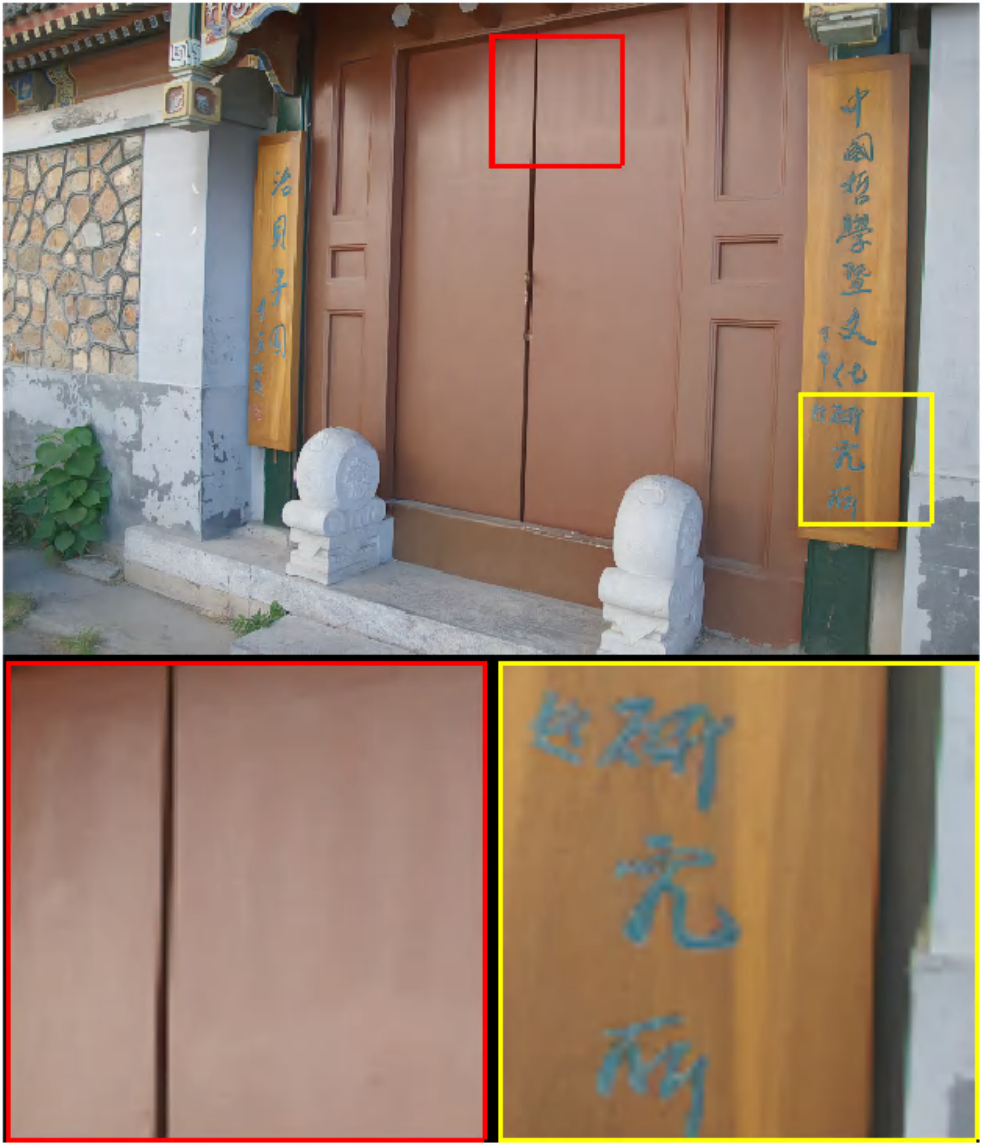}}
    \centerline{\small GSAD \cite{GSAD}}
\end{minipage}
\begin{minipage}[t]{0.12\linewidth}
    \centering
    \vspace{3pt}
    \centerline{\includegraphics[width=\textwidth]{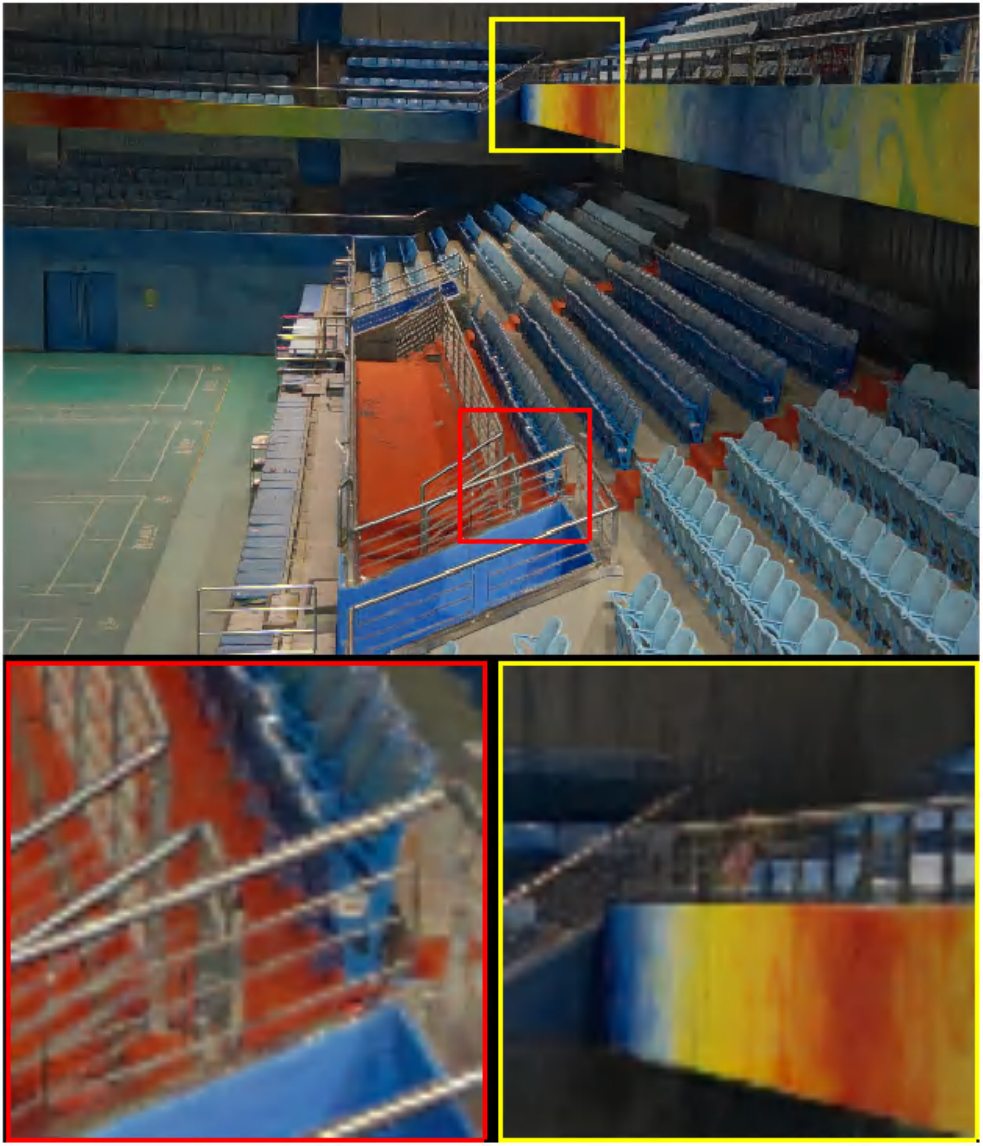}}
    \centerline{\includegraphics[width=\textwidth]{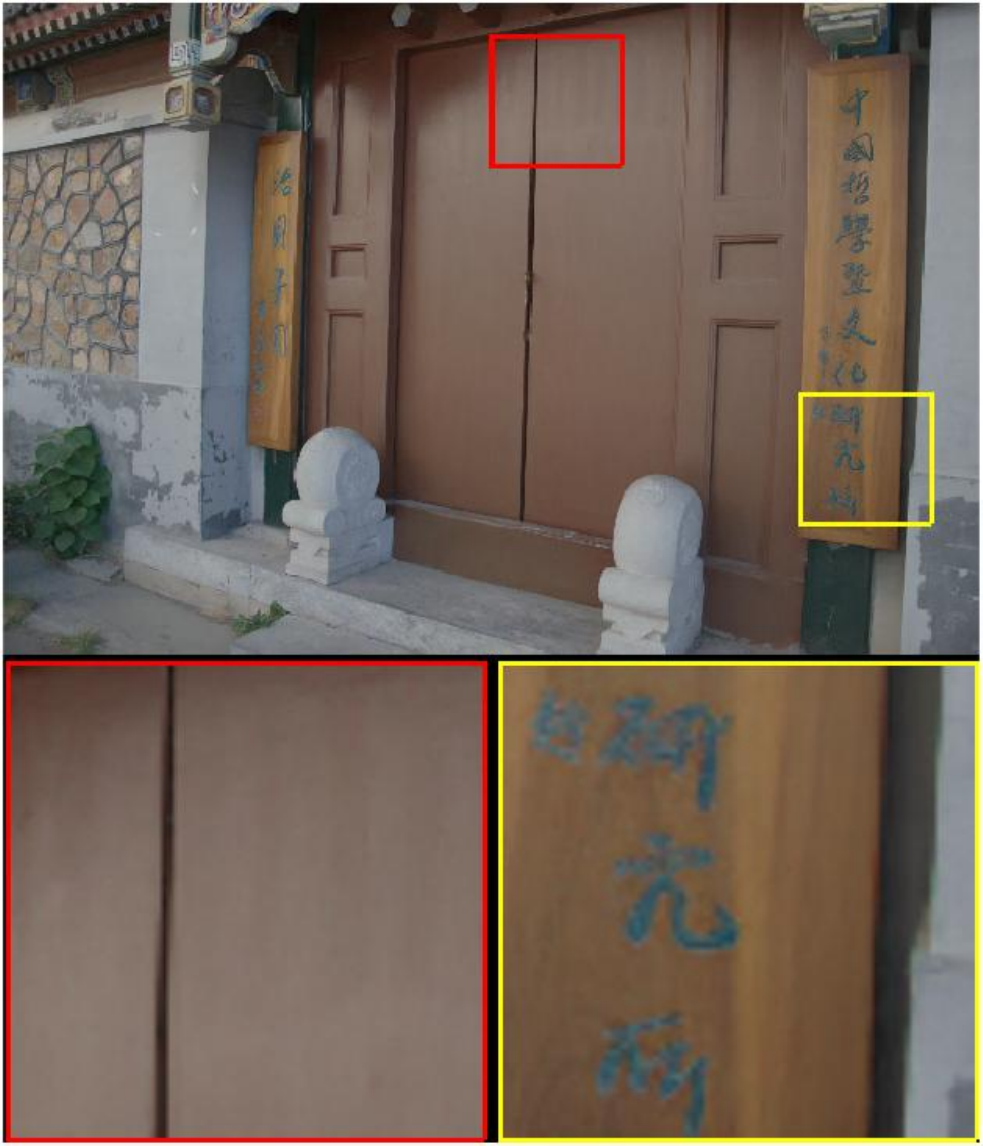}}
    \centerline{\small CIDNet(Ours)}
\end{minipage}
\begin{minipage}[t]{0.12\linewidth}
    \centering
    \vspace{3pt}
    \centerline{\includegraphics[width=\textwidth]{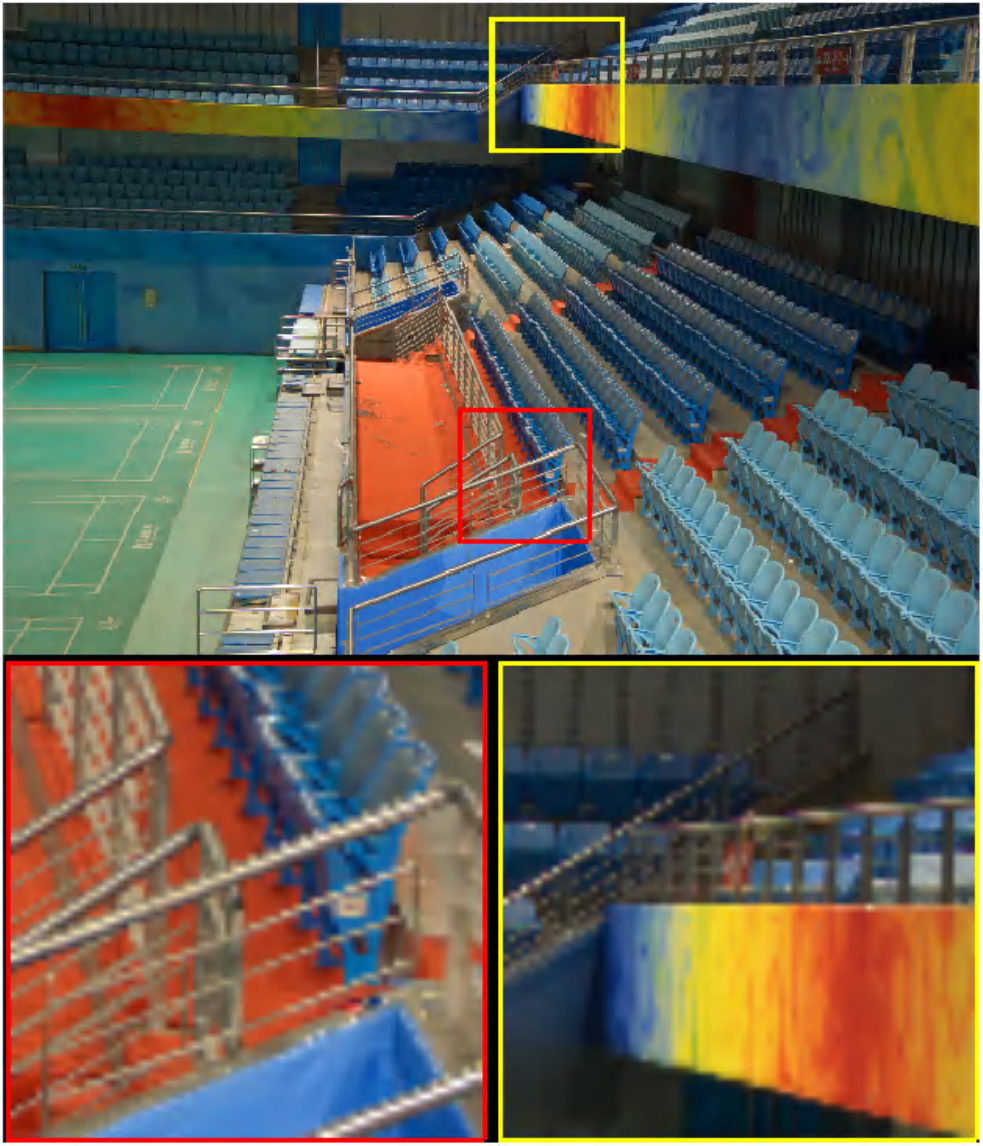}}
    \centerline{\includegraphics[width=\textwidth]{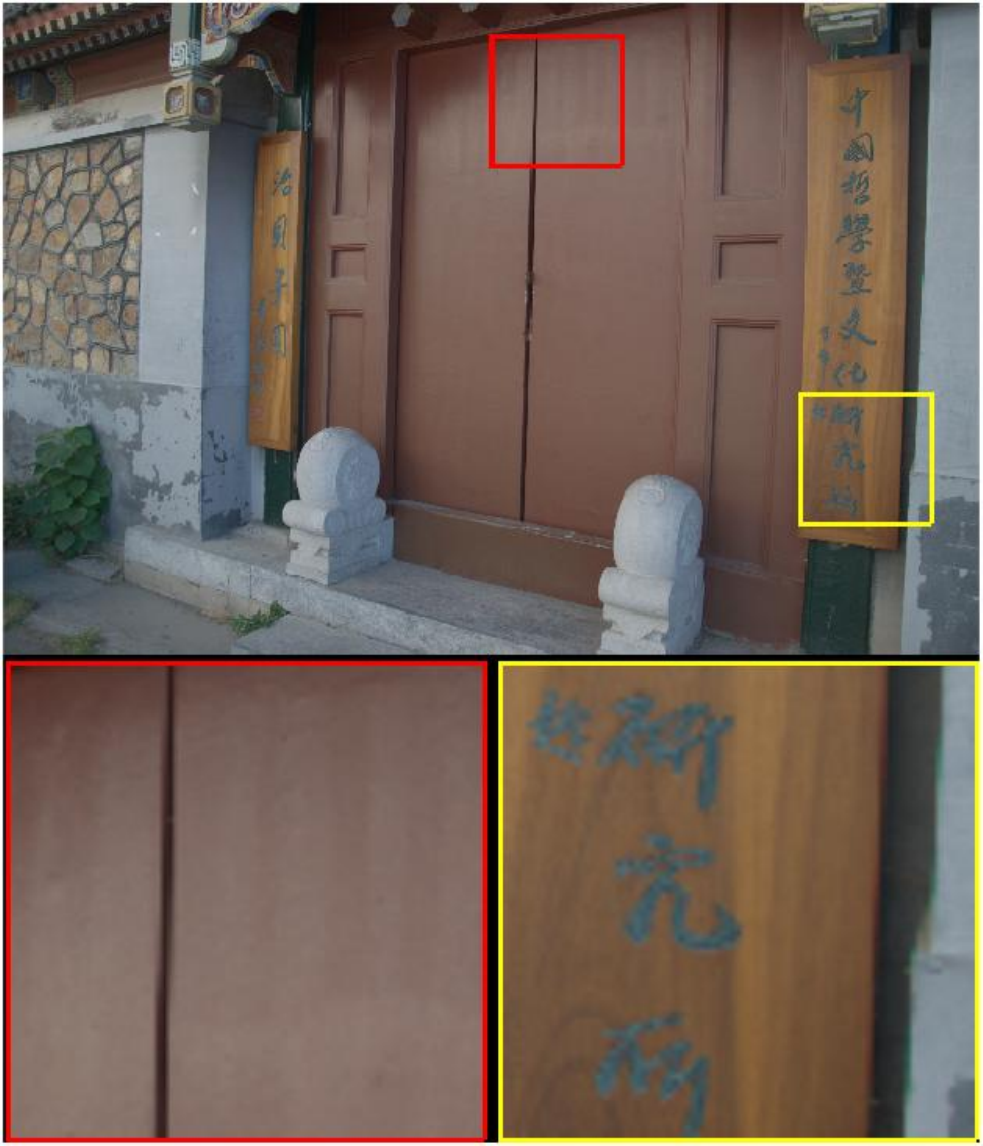}}
    \centerline{\small GroundTruth}
\end{minipage}
\vspace{-0.2cm}
 \caption{Visual comparison of the enhanced images yielded by different SOTA methods on LOLv1 (top row) and LOLv2 (bottom row).}
 \label{fig:LOL}
\end{figure*}

 \begin{table}
      \centering
        \renewcommand{\arraystretch}{1.}
        \caption{Quantitative result on SICE, Sont-Total-Dark, and the five unpaired datasets (DICM \cite{DICM}, LIME \cite{LIME}, MEF \cite{MEF}, NPE \cite{NPE}, and VV \cite{VV}). The top-ranking score is in \textcolor{red}{red}.}
\vspace{-1.8mm}
        \resizebox{\linewidth}{!}{
        \begin{tabular}{c|cc|cc|cc}
        \Xhline{1.5pt}
        \multirow{2}{*}{\textbf{Methods}}&
         \multicolumn{2}{c|}{\textbf{SICE}}& 
         \multicolumn{2}{c|}{\textbf{Sony-Total-Dark}} & 
         \multicolumn{2}{c}{\textbf{Unpaired}}\\

        ~&
            PSNR$\uparrow$&	SSIM$\uparrow$& 
            PSNR$\uparrow$&	SSIM$\uparrow$&
            BRIS$\downarrow$& NIQE$\downarrow$\\
            
            \hline

            RetinexNet \cite{RetinexNet}&
            12.424& 0.613&
            15.695& 0.395& 
            \color{red}{23.286}& 4.558
\\
            ZeroDCE \cite{Zero-DCE}&
            12.452& 0.639& 
            14.087&	0.090& 
            26.343&	4.763
\\
            URetinexNet \cite{URetinexNet}&
            10.899& 0.605& 
            15.519& 0.323&
            26.359& 3.829
\\
            RUAS \cite{RUAS}&
            8.656& 	0.494& 
            12.622& 0.081&  
            36.372& 4.800 
\\
            LLFlow \cite{LLFlow}&
            12.737&	0.617& 
            16.226&	0.367& 
            28.087&	4.221 
\\

             \textbf{CIDNet (Ours)}&
            \color{red}{13.435}&	\color{red}{0.642}&
            \color{red}{22.904}&	\color{red}{0.676}&
            23.521&	\color{red}{3.523}\\
            \Xhline{1.5pt}
        \end{tabular}
        }
        \label{tab:SID}
\end{table}

\subsection{Main Results}
\textbf{Results on LOL Datasets.} 
In Tab. \ref{tab:table-LOL}, it can be found that our method is optimal on all metrics for both LOLv1 and LOLv2 datasets with 1.88M parameters and 7.57 GFLOPs. 
We outperform the best RGB-based method GSAD (diffusion)
in terms of all PSNR, SSIM, and LPIPS metrics, while utilizing only 10.8\% parameters of GSAD.
Compared to RetinexFormer, a SOTA method based on Retinex theory, CIDNet delivers higher image quality while reducing computational cost by 8.28 GFLOPs.
Subjective results are illustrated in Fig. \ref{fig:LOL}, which demonstrates that our method not only more accurately recovers multi-color regions compared to GroundTruth but also achieves stable brightness enhancement, thanks to the HVI color space. 
More visualization comparison can be found in supplementary materials.

\textbf{Results on SICE and Sony-total-Dark.} 
To validate the performance on large-scale datasets, we evaluate CIDNet on SICE (including Mix and Grad) and SID-Total-Dark. 
The results are presented in Table \ref{tab:SID}, where it is clear that CIDNet is the best performer in both PSNR and SSIM metrics on the two datasets. 
Notably, on Sony-Total-Dark, our model surpasses the second-best method by 6.678 dB in PSNR. 
This improvement is due to the extreme darkness of the dataset images, which substantially increases the difficulty of distinguishing details from noise. 
However, CIDNet leverages the intensity collapse function $\mathbf{C}_k$ to effectively maintain an optimal signal-to-noise ratio during training, enabling better detail recovery.

\begin{figure*}[!t]
    \centering
    \includegraphics[width=1\linewidth]{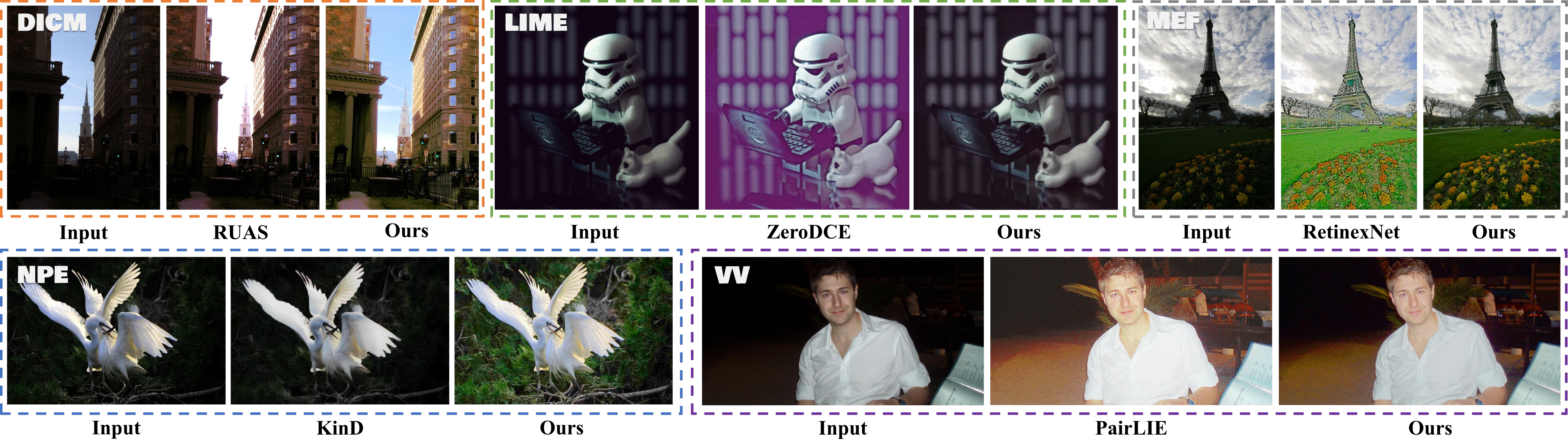}
    \vspace{-0.55cm}
    \caption{Visual comparison on the five unpaired datasets. Follow RetinexFormer \cite{RetinexFormer}, we select one image in each dataset to compare our method with the other methods. More visual comparison can be found in the supplementary materials.}
    \label{fig:unpaired}
\end{figure*}

\textbf{Results on Unpaired Datasets.} 
We evaluate the effectiveness of models trained on LOLv1 or LOLv2-Syn using various methods, and report their performance using BRISQUE and NIQE metrics in Tab. \ref{tab:SID}. Our method exhibits a substantial improvement in the NIQE metric compared to other approaches.
As shown in Fig. \ref{fig:unpaired}, while CIDNet does not outperform RetinexNet in the BRISQUE metric in Tab. \ref{tab:SID}, its recovered perceptual results are closer to realistic appearances than RetinexNet. 
This may be attributed to the fact that building upon the HSV space, the HVI color space is derived from real-world perceptual models \cite{gevers2012color}.

\begin{table*}[!t]
    \centering
    \renewcommand{\arraystretch}{1.}
    \caption{Results of applying HVI transformation as a plug-in to various LLIE methods
    on LOLv2-Real. 
    \textcolor{red}{Values} in brackets represent the absolute improved performance gain. The best PSNR/SSIM$\uparrow$, LPIPS$\downarrow$, and inference time$\downarrow$ are in \textbf{bolded}.} 
\vspace{-1.8mm}
    \resizebox{\linewidth}{!}{
    \begin{tabular}{cccccccc}
    \Xhline{1.5pt}
         \cellcolor{gray!10}Methods& 
         \cellcolor{gray!10}FourLLIE \cite{wang2023fourllie}&
         \cellcolor{gray!10}LEDNet \cite{LEDNet}&
         \cellcolor{gray!10}SNR-Aware \cite{SNR-Aware}& 
         \cellcolor{gray!10}LLFormer \cite{LLFormer}& 
         \cellcolor{gray!10}GSAD \cite{GSAD}& 
         \cellcolor{gray!10}DiffLight \cite{feng2024difflight}&
         \cellcolor{gray!10}\textbf{CIDNet}\\
    \Xhline{1.5pt}
         PSNR$\uparrow$&
         22.730(\textcolor{red}{+0.381})&
         23.394(\textcolor{red}{+3.456})&
         22.251(\textcolor{red}{+0.771})&
         22.671(\textcolor{red}{+2.615})&		
         23.715(\textcolor{red}{+3.562})&	
         23.969(\textcolor{red}{+1.364})&
         \textbf{24.111}
\\
         SSIM$\uparrow$&
         0.856(\textcolor{red}{+0.009})&
         0.837(\textcolor{red}{+0.010})&
         0.840(-0.009)&	
         0.852(\textcolor{red}{+0.060})&		
         \textbf{0.876}(\textcolor{red}{+0.030})&	
         0.859(\textcolor{red}{+0.003})&
         0.871

\\
        LPIPS$\downarrow$&
        0.125(+0.011)&
        0.115(\textcolor{red}{-0.005})&	
        0.117(\textcolor{red}{-0.054})&	
        0.117(\textcolor{red}{-0.094})&	
        \textbf{0.103}(\textcolor{red}{-0.010})&	
        0.109(\textcolor{red}{-0.012})&
        0.108
\\
        GPU Time/s$\downarrow$&
        0.075&
        0.054& 	
        0.070& 
        0.139& 	
        0.315& 
        0.578&
        \textbf{0.053}
\\
        Model Type&
        CNN&
        CNN&
        Transformer&
        Transformer&		
        Diffusion&
        CNN+Diffusion&
        Transformer
\\
    \Xhline{1.5pt}
    \end{tabular}
    }
    \label{tab:HVI}
\end{table*}

\begin{table}[!t]
\centering
\captionof{table}{Model ablation. Dual denotes the dual-branch network. }
\label{tab:ablation}
\vspace{-1.8mm}
\renewcommand{\arraystretch}{1.}
\resizebox{\linewidth}{!}{
    \begin{tabular}{ll|ccc}
    \Xhline{1.5pt}
         \multicolumn{2}{l|}{\cellcolor{gray!10}\textbf{Metrics}}&	\cellcolor{gray!10}PSNR$\uparrow$&	\cellcolor{gray!10}SSIM$\uparrow$&\cellcolor{gray!10}LPIPS$\downarrow$
\\
    \Xhline{1.5pt}
         \multirow{4}{*}{\textbf{Color Space}}&	sRGB	&20.062& 	0.825& 0.137
\\
         ~&HSV		&21.349 & 0.801	& 0.167
\\
         ~&HVI (w/ Polarization Only)& 21.558 &0.821 & 0.149
\\
         ~&HVI (w/ $\mathbf{C}_k$ Only) &  21.536  & 0.825 & 0.179
\\
\midrule
         \multirow{3}{*}{\textbf{Structure}}&UNet Baseline \cite{petit2021unet}&  19.306  &  0.778 & 	0.222
\\
         ~&SelfAttn \cite{Restormer}  &22.313&	0.835& 0.126
\\
         ~&Dual+SelfAttn \cite{Restormer}&23.159& 	0.856&   0.116
\\
\midrule
        \multirow{2}{*}{\textbf{Loss}}&HVI Only&23.221& 0.854&0.132
\\
        ~&sRGB Only&23.319& 0.857& 0.123
\\
\midrule
        \multicolumn{2}{l|}{\textbf{Full Model (HVI-CIDNet)}}&24.111& 0.871&0.108
\\

    \Xhline{1.5pt}
    \end{tabular}
    }
\end{table}

\textbf{Generalizing HVI to Other LLIE Models.}
To verify the effectiveness of the HVI color space, we further evaluate its performance when it is used with different LLIE models. In particular, HVIT, together with its inverse mapping PHVIT, is used as a plug-and-play module into six SOTA methods that use sRGB images as input and are independent of specific color space characteristics. 
The results are reported in Tab. \ref{tab:HVI}. It is clear that transforming to the HVI color space improves PSNR, SSIM, and LPIPS metrics across various methods compared to the results in the sRGB color space. 
Notably, the GSAD method demonstrates the most significant improvement, with a PSNR increase of 3.562 dB.
This demonstrates not only the generalizability of HVI to various sRGB-based methods but also its general effectiveness as a color space for the LLIE task. 

For the inference time results in Tab. \ref{tab:HVI}, it is evident that the diffusion-based methods require longer GPU time but achieve better enhancement results. 
In contrast, CIDNet shows the most efficient inference while achieving the highest PSNR and the second-best SSIM and LPIPS scores. 
This highlights the strong ability of CIDNet in balancing the efficiency and effectiveness within the HVI color space.
    
\subsection{Ablation Study}
\begin{figure*}[htp]
\centering
\begin{minipage}[t]{0.136\linewidth}
    \centering
        \vspace{1pt}
        \centerline{\includegraphics[width=\textwidth]{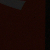}}
        \centerline{\includegraphics[width=\textwidth]{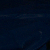}}
        \centerline{\includegraphics[width=\textwidth]{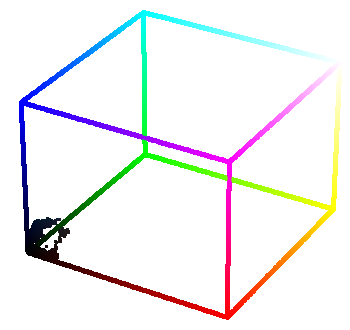}}
    \centerline{\small (a) Input}
\end{minipage}
\hfill
\begin{minipage}[t]{0.136\linewidth}
    \centering
        \vspace{1pt}
        \centerline{\includegraphics[width=\textwidth]{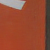}}
        \centerline{\includegraphics[width=\textwidth]{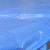}}
        \centerline{\includegraphics[width=\textwidth]{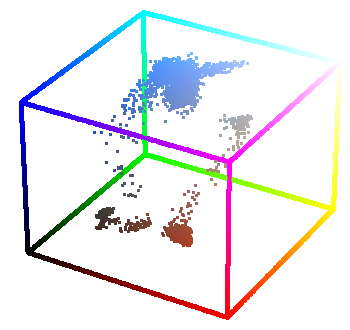}}
    \centerline{\small(b) sRGB}
\end{minipage}
\hfill
\begin{minipage}[t]{0.136\linewidth}
    \centering
        \vspace{1pt}
        \centerline{\includegraphics[width=\textwidth]{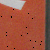}}
        \centerline{\includegraphics[width=\textwidth]{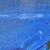}}
        \centerline{\includegraphics[width=\textwidth]{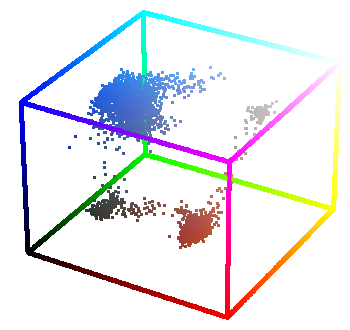}}
    \centerline{\small(c) HSV}
\end{minipage}
\hfill
\begin{minipage}[t]{0.136\linewidth}
    \centering
        \vspace{1pt}
        \centerline{\includegraphics[width=\textwidth]{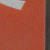}}
        \centerline{\includegraphics[width=\textwidth]{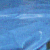}}
        \centerline{\includegraphics[width=\textwidth]{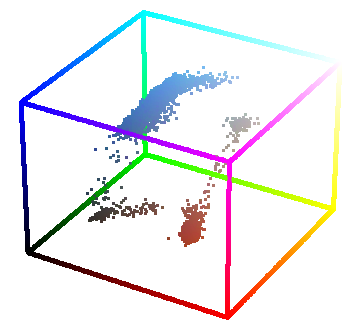}}
    \centerline{\small(d) w/ Polarization}
\end{minipage}
\hfill
\begin{minipage}[t]{0.136\linewidth}
    \centering
        \vspace{1pt}
        \centerline{\includegraphics[width=\textwidth]{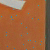}}
        \centerline{\includegraphics[width=\textwidth]{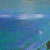}}
        \centerline{\includegraphics[width=\textwidth]{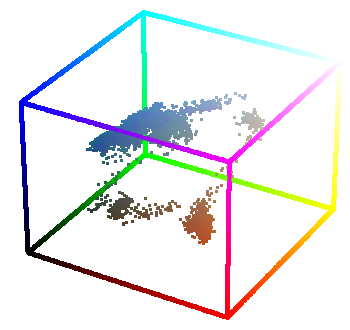}}
    \centerline{\small(e) w/ $\mathbf{C}_k$}
\end{minipage}
\hfill
\begin{minipage}[t]{0.136\linewidth}
    \centering
        \vspace{1pt}
        \centerline{\includegraphics[width=\textwidth]{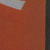}}
        \centerline{\includegraphics[width=\textwidth]{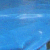}}
        \centerline{\includegraphics[width=\textwidth]{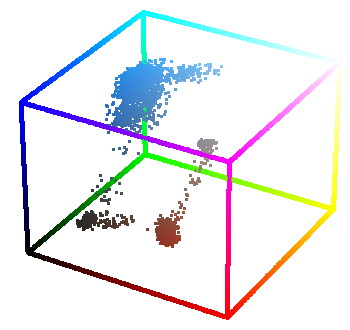}}
    \centerline{\small(f) HVI}
\end{minipage}
\hfill
\begin{minipage}[t]{0.136\linewidth}
    \centering
        \vspace{1pt}
        \centerline{\includegraphics[width=\textwidth]{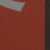}}
        \centerline{\includegraphics[width=\textwidth]{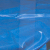}}
        \centerline{\includegraphics[width=\textwidth]{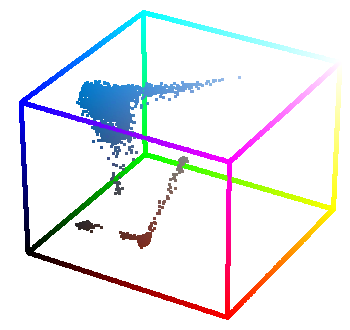}}
    \centerline{\small(g) GroundTruth}
\end{minipage}
\vspace{-0.18cm}
\caption{Top and middle rows are ablation results on LOLv2-Real for five different color spaces used by CIDNet. The bottom row provides a visual comparison by mapping the pixel values of the results to sRGB.
Note that due to the dual-branch and the cross-attention mechanism are specifically designed for HVI, we only use UNet \cite{petit2021unet} with self attentions \cite{Restormer} for a fair comparison. }
\label{fig:hvi2}
\end{figure*}

\begin{figure}
\centering
\begin{minipage}[t]{0.237\linewidth}
    \centering
        \vspace{1pt}
        \centerline{\includegraphics[width=\textwidth]{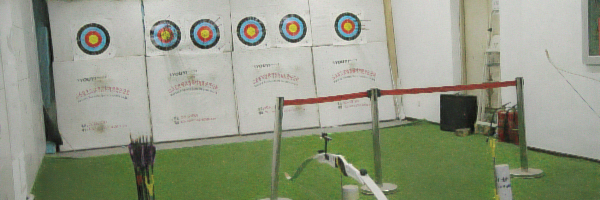}}
        \centerline{\includegraphics[width=\textwidth]{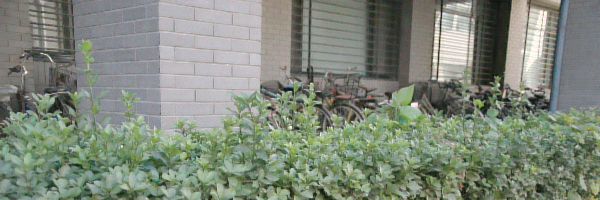}}
        \centerline{\includegraphics[width=\textwidth]{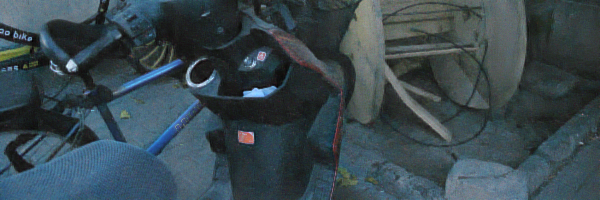}}
    \centerline{\small SelfAttn}
\end{minipage}
\hfill
\begin{minipage}[t]{0.237\linewidth}
    \centering
        \vspace{1pt}
        \centerline{\includegraphics[width=\textwidth]{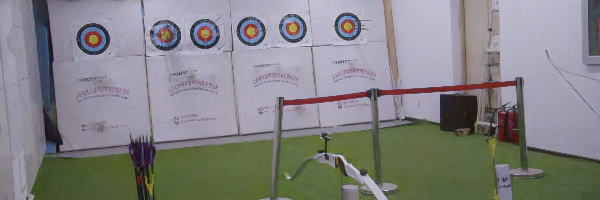}}
        \centerline{\includegraphics[width=\textwidth]{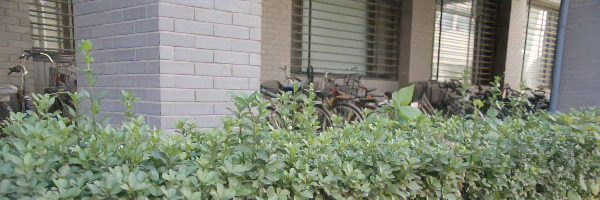}}
        \centerline{\includegraphics[width=\textwidth]{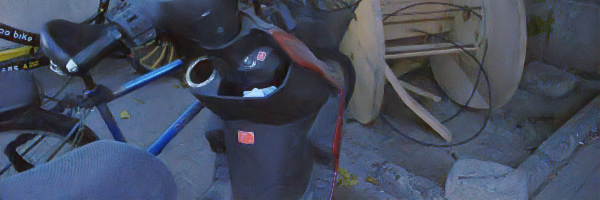}}
    \centerline{\small Dual + Self}
\end{minipage}
\hfill
\begin{minipage}[t]{0.237\linewidth}
    \centering
        \vspace{1pt}
        \centerline{\includegraphics[width=\textwidth]{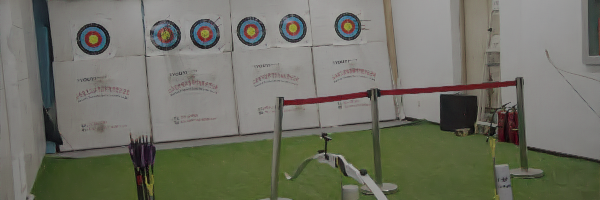}}
        \centerline{\includegraphics[width=\textwidth]{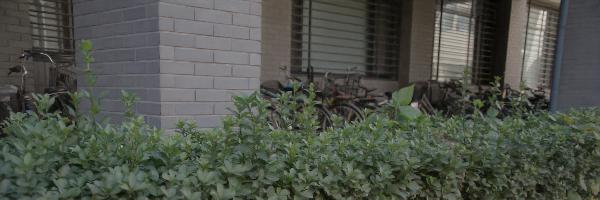}}
        \centerline{\includegraphics[width=\textwidth]{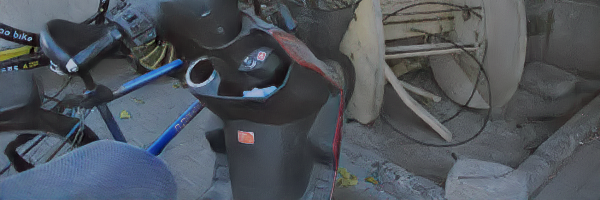}}
    \centerline{\small Dual + Cross}
\end{minipage}
\hfill
\begin{minipage}[t]{0.237\linewidth}
    \centering
        \vspace{1pt}
        \centerline{\includegraphics[width=\textwidth]{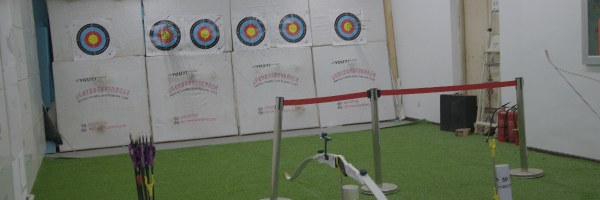}}
        \centerline{\includegraphics[width=\textwidth]{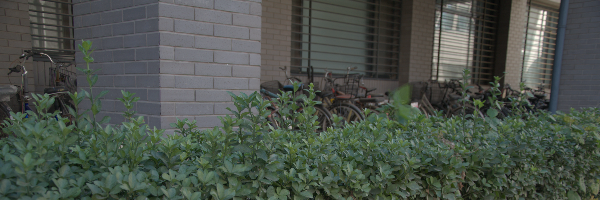}}
        \centerline{\includegraphics[width=\textwidth]{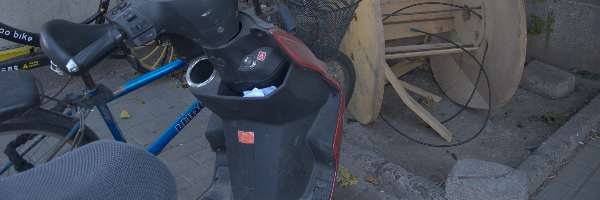}}
    \centerline{\small GroundTruth}
\end{minipage} 
\vspace{-0.18cm}
\caption{Results of using different structures on LOLv2-Real.}
\vspace{-0.15cm}
\label{fig:struct}
\end{figure}

We validate our HVI color space and the key modules in CIDNet using both quantitative (Tab. \ref{tab:ablation}) and qualitative results (Figs. \ref{fig:hvi2} and \ref{fig:struct}). The experiments are all performed on LOLv2-Real for fast convergence and stable performance.

\textbf{HVI Color Space.} 
It can be seen in Tab. \ref{tab:ablation} that the enhancement in the sRGB color space leads to chromatic aberration and luminance bias, as demonstrated by the difference between Figs. \ref{fig:hvi2}(b) and (g). 
Compared to sRGB, using HSV yields images aligned more closely with the GroundTruth in both luminance and color due to its effectiveness on decoupling brightness from color, which can be observed the enhancement from Fig. \ref{fig:hvi2}(b) to Fig. \ref{fig:hvi2}(c).  
This results in improved PSNR and LPIPS in Tab. \ref{tab:ablation}. However, it significantly introduces more noise due to the red discontinuity in HSV, leading to the noisy black spots in the red regions in Fig. \ref{fig:hvi2}(c) and the degraded SSIM in Tab. \ref{tab:ablation}. 
Using the polarization or the $\mathbf{C}_k$ solely in the HVI space can lead to similar image quality in PSNR and SSIM compared to those in the HSV color space. Qualitatively,
as shown in Fig. \ref{fig:hvi2}(d), using polarization only helps cluster similar red tones, avoiding the red discontinuity.
Relying solely on the intensity function $\mathbf{C}_k$ helps adjust the brightness, but leads to confusion between red and other colors.
Consequently, dot-like artifacts appear not only in the red regions but also color shifts in other areas, as shown in Fig. \ref{fig:hvi2}(e).
These issues are all effectively mitigated when the polarization and the $\mathbf{C}_k$ function are applied together, as shown by consistent improvement of the Full Model in all three metrics in Tab. \ref{tab:ablation} and the better image enhancement in Fig. \ref{fig:hvi2}(f).

\textbf{Dual-branch Network Structure.} 
In Tab. \ref{tab:ablation}, adding self-attention to the baseline noticeably improves all three metrics, indicating that transformer-based models hold potential for application in the HVI color space. 
We then modified the architecture from a single-branch to a dual-branch structure without cross-attention, resulting in a PSNR increase of 0.846 dB, while SSIM and LPIPS showed minimal change. 
Further incorporating the cross-attention into the I-branch and HV-branch (Full Model) obtains the best color restoration, light enhancement, and optimal metric performance, as shown in Tab. \ref{tab:ablation}. This can also be observed in Fig. \ref{fig:struct}.

\textbf{Loss Function.} 
As shown in Tab. \ref{tab:ablation}, 
compared to using both HVI and sRGB losses, relying solely on the HVI loss lacks pixel-level spatial consistency constraints, leading to a loss of structural detail in the image and thus lower performance across the three metrics, especially in the LPIPS metric. 
On the other hand, using only sRGB loss is focused on pixel-space enhancement, neglecting the low-light probability distribution in the HVI color space, resulting in undesired color imbalance. 

\section{Conclusion}

In this work, we introduce the HVI color space and the CIDNet approach to address the color bias and brightness artifact issues that occur to current sRGB-based LLIE approaches. 
By encapsulating polarized HS maps and a learnable intensity component, HVI shows strong robustness to both issues. 
To further enhance LLIE, CIDNet is designed to model decoupled chromatic and intensity information in the HVI space for achieving precise photometric adjustments under varying lighting conditions. 
Experimental results on 10 datasets demonstrate that the HVI color space, combined with CIDNet, outperforms SOTA LLIE methods, establishing it as a robust solution for low-light enhancement.
{
    \small
    \bibliographystyle{ieeenat_fullname}
    \bibliography{trans}
}

\clearpage
\setcounter{page}{1}
\maketitlesupplementary

\section{Supplementary Introduction}
In our appendix, we first provide additional details about the HVI color space, along with an extended version to address crossing-datasets challenge in Low-light Image Enhancement task. 
Next, we present the detailed structure of the LCA module in CIDNet and conduct ablation studies on its submodules. 
Following this, we conduct additional experiments on the HVI-CIDNet described in the main text to validate the advantages of the HVI color space and CIDNet. 
Finally, we analyze the limitations of our approach and offer a discussion on potential improvements.

\section{Details and Extensions of HVI Color Space}
In Sec. \ref{sec:HVI}, we introduced the HVI color space, which addresses the red discontinuity and black plane noise issues in HSV by applying polar coordinate transformation and incorporating the $k$ parameter to construct the $\mathbf{C}_k$ intensity compression formula, collapsing the low-light plane into a compact region. 
In this section, we provide a visual analysis of the effectiveness of $k$ and further discuss the extensibility of HVI, aiming to adapt it for solving additional the crossing-dataset challenge.

\begin{figure}
    \centering
    \includegraphics[width=\linewidth]{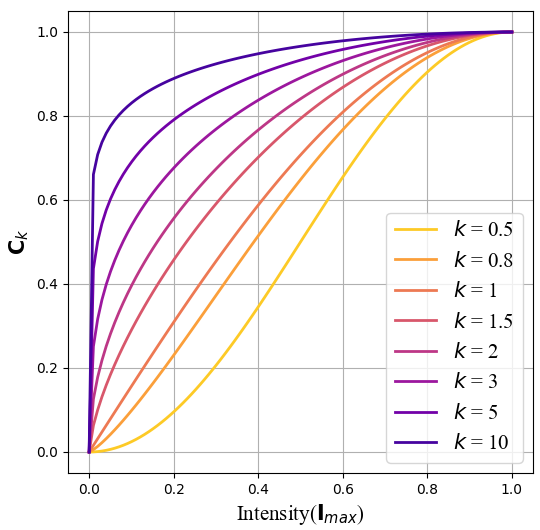}
    \caption{Different Density-$k$s. The independent variable is Intensity ($\textbf{I}_{max}$) and the dependent variable is $\textbf{C}_k$.}
    \label{fig:k}
\end{figure}

\subsection{Visualization and Further Discussion of Parameter $k$}
\begin{figure}[t]
	\begin{minipage}{0.24\linewidth}
		\vspace{3pt}  
		\centerline{\includegraphics[width=\textwidth]{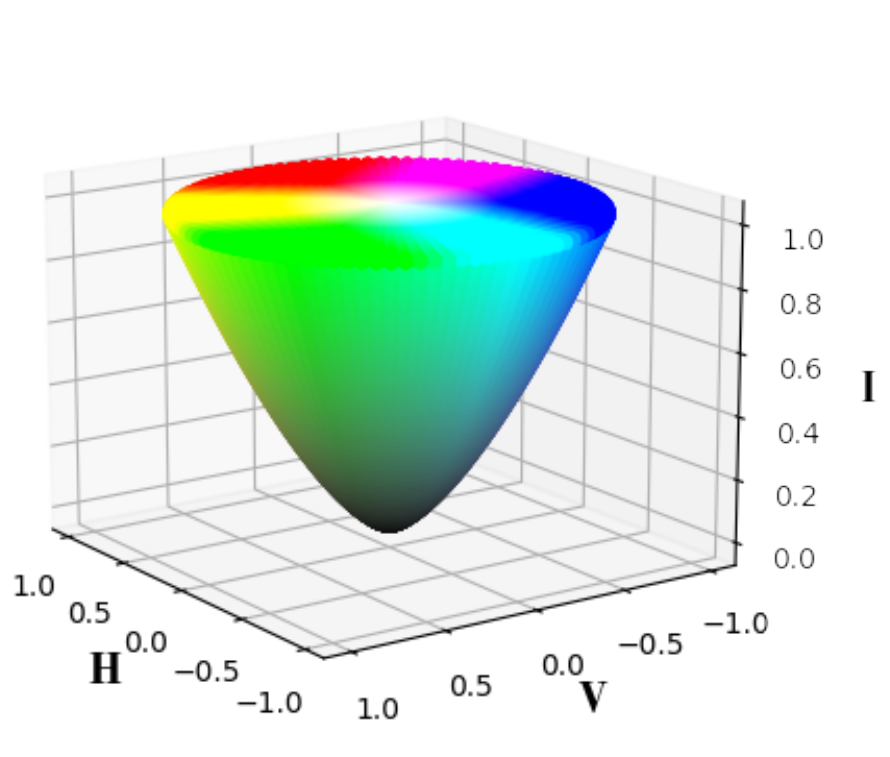}}
		\centerline{$k=0.5$}
	\end{minipage}
	\begin{minipage}{0.24\linewidth}
		\vspace{3pt}
		\centerline{\includegraphics[width=\textwidth]{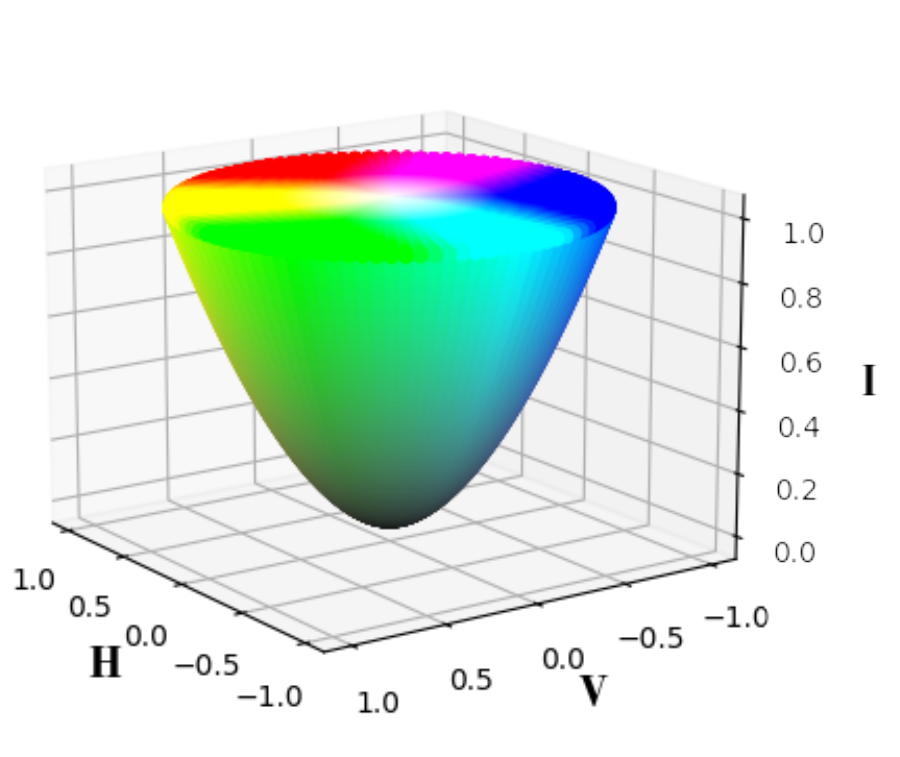}}
	   \vspace{1pt}
		\centerline{$k=0.8$}
	\end{minipage}
	\begin{minipage}{0.24\linewidth}
		\vspace{3pt}
		\centerline{\includegraphics[width=\textwidth]{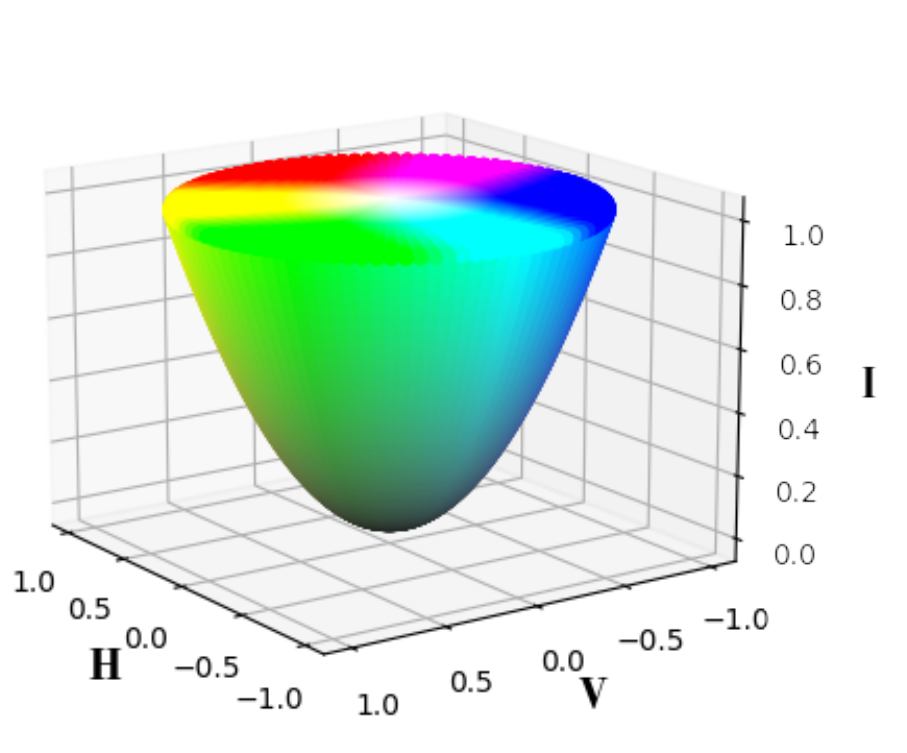}}
		\centerline{$k=1$}
	\end{minipage}
        \begin{minipage}{0.24\linewidth}
		\vspace{3pt}
		\centerline{\includegraphics[width=\textwidth]{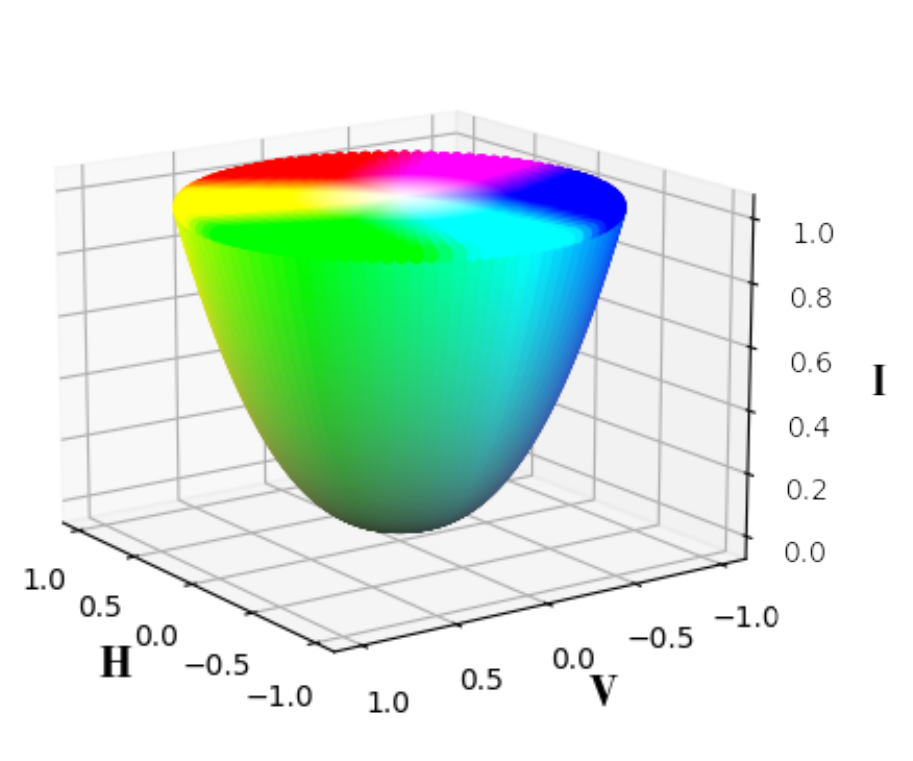}}
		\centerline{$k=1.5$}
	\end{minipage}
	\begin{minipage}{0.24\linewidth}
		\vspace{3pt}
		\centerline{\includegraphics[width=\textwidth]{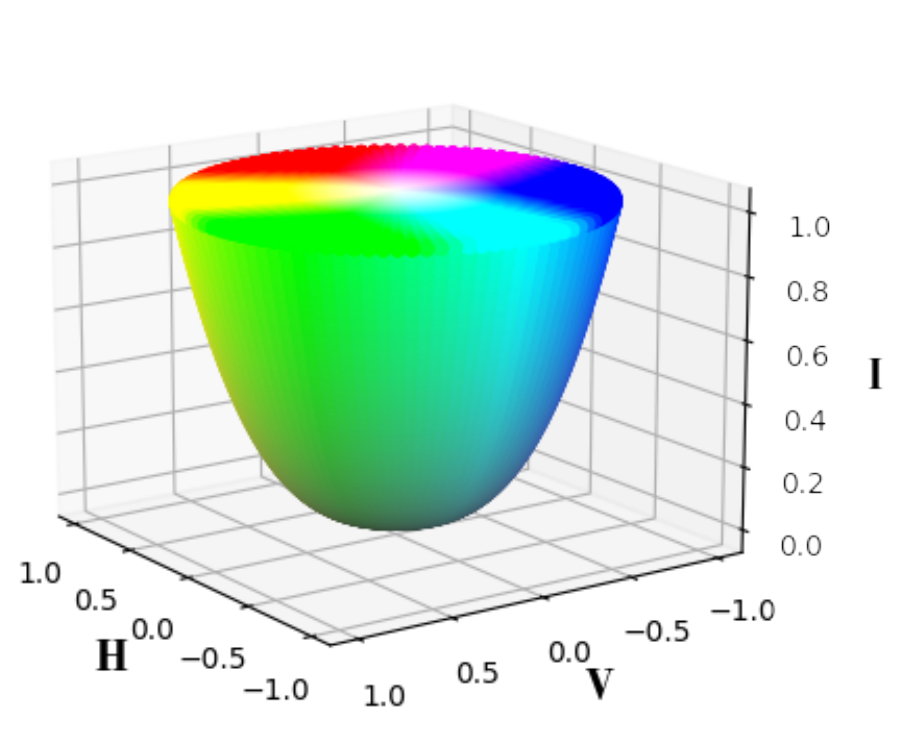}}
		\centerline{$k=2$}
	\end{minipage}
	\begin{minipage}{0.24\linewidth}
		\vspace{3pt}
		\centerline{\includegraphics[width=\textwidth]{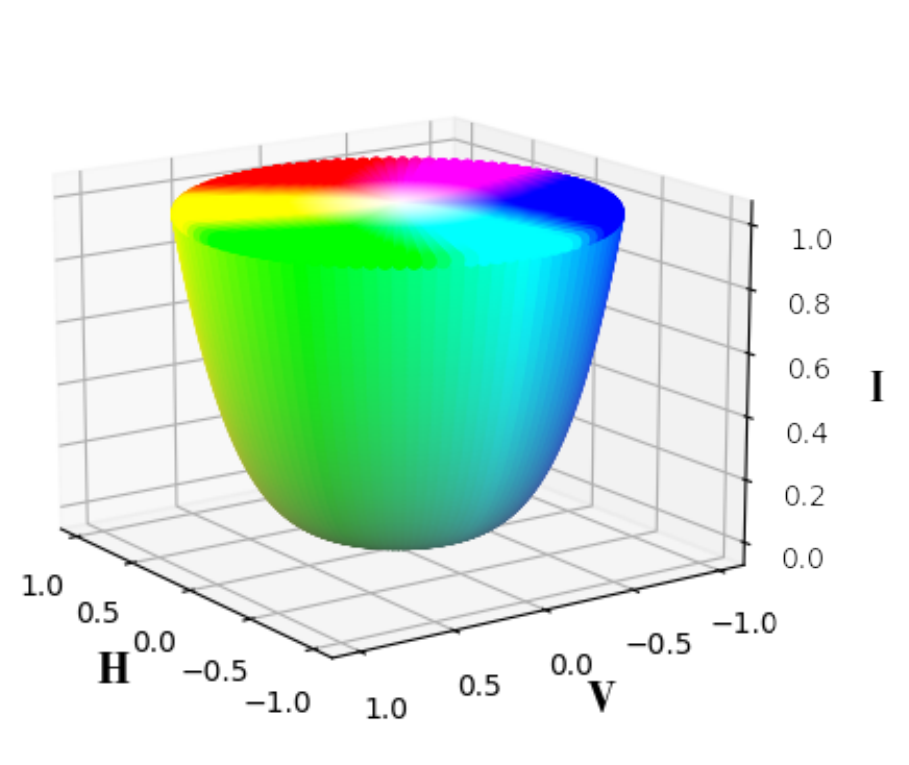}}
		\centerline{$k=3$}
	\end{minipage}
        \begin{minipage}{0.24\linewidth}
		\vspace{3pt}
		\centerline{\includegraphics[width=\textwidth]{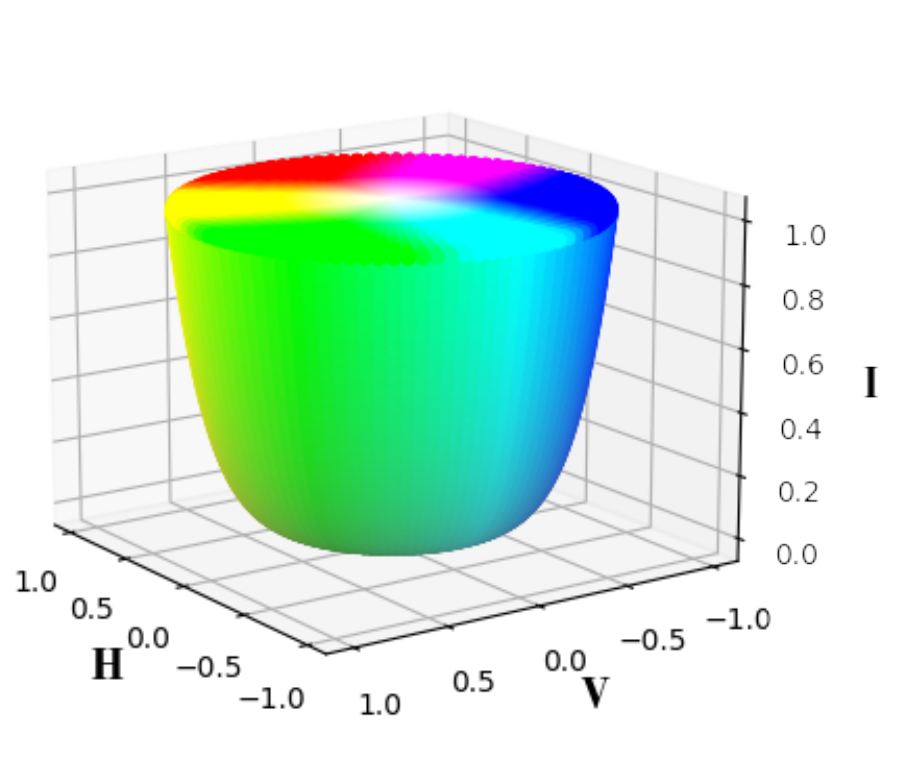}}
		\centerline{$k=5$}
	\end{minipage}
        \begin{minipage}{0.24\linewidth}
		\vspace{3pt}
		\centerline{\includegraphics[width=\textwidth]{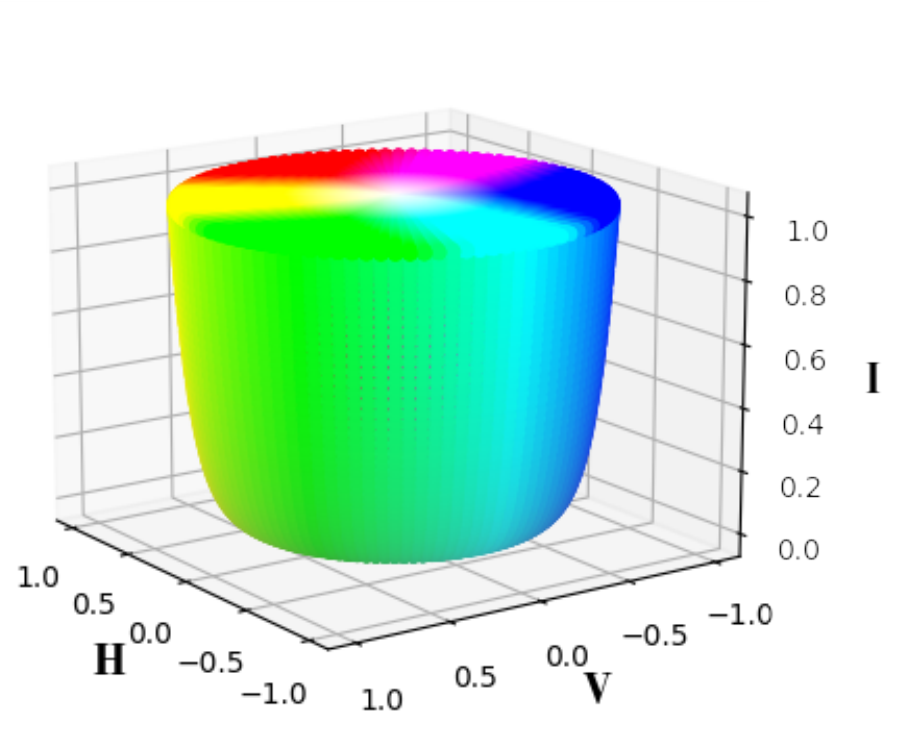}}
		\centerline{$k=10$}
	\end{minipage}
	\caption{Visually comparison of HVI color spaces with different density-$k$.  }
	\label{fig4}
\end{figure}

We visualise Eq. \ref{eq:2} for the generation of $\textbf{C}_k$ and the result is illustrated in Fig. \ref{fig:k}. It can be seen that $\textbf{C}_k$ is essentially a remapping function positively correlated with Intensity from zero to one. 
The purpose of the parameter k is to adjust the gradient of $\textbf{C}_k$ over Intensity.
As shown in Fig. \ref{fig:k}, larger $k$ values exhibit steeper gradients near zero and more gradual slopes as they approach one, which can think of $k$ as a specific low-light characteristic to adjust the density of black color plane points.
We show in Fig. \ref{fig4} the results of a more intuitive visualization of HVI, based on different values of $k$.
As $k$ increases, we can notice that the low-light "chassis" of the HVI space gradually becomes rounded from sharp, and finally widens to approximate a cylinder.
This is because $k$ affects the gradient of $\textbf{C}_k$, and it represents the radius of the HV plane at different Intensities.

\begin{figure}[t]
	\begin{minipage}{0.24\linewidth}
		\vspace{3pt}
		\centerline{\includegraphics[width=\textwidth]{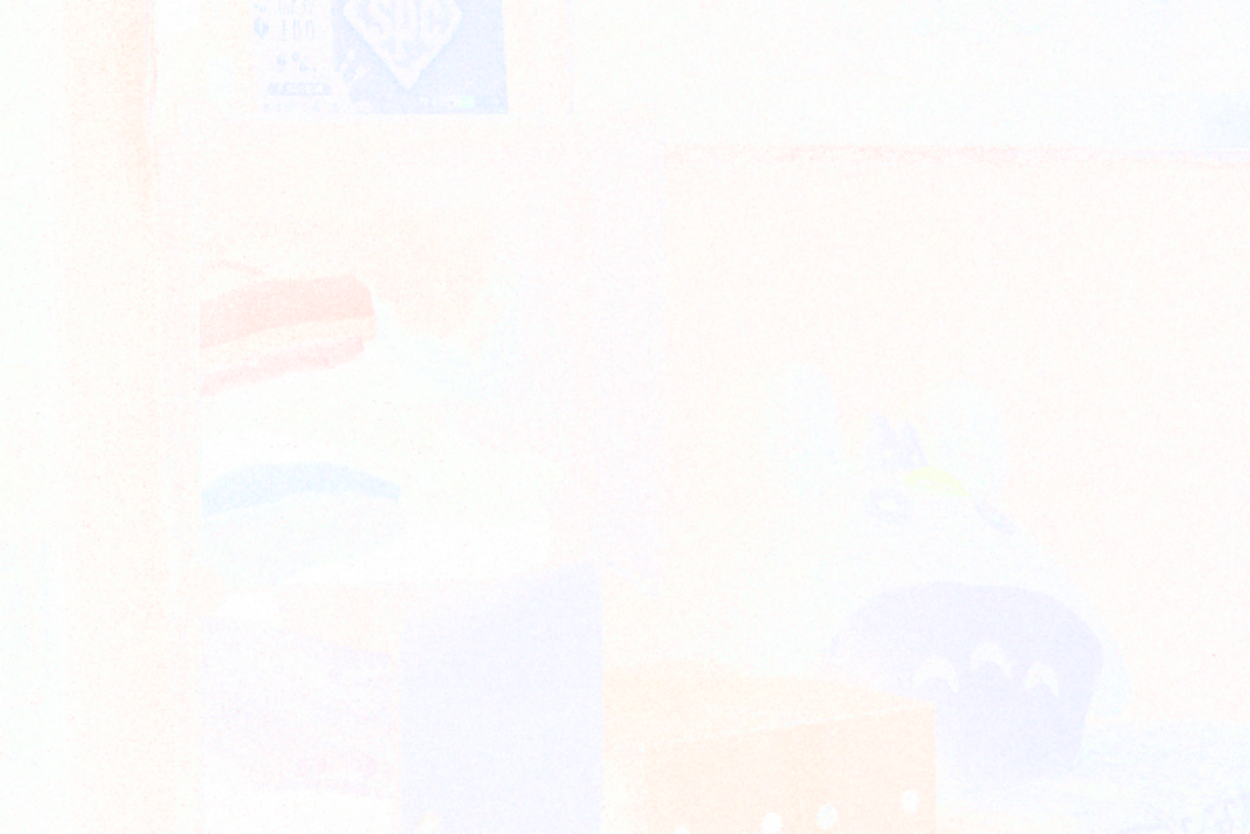}}
		\centerline{$k=1$}
	\end{minipage}
	\begin{minipage}{0.24\linewidth}
		\vspace{3pt}
		\centerline{\includegraphics[width=\textwidth]{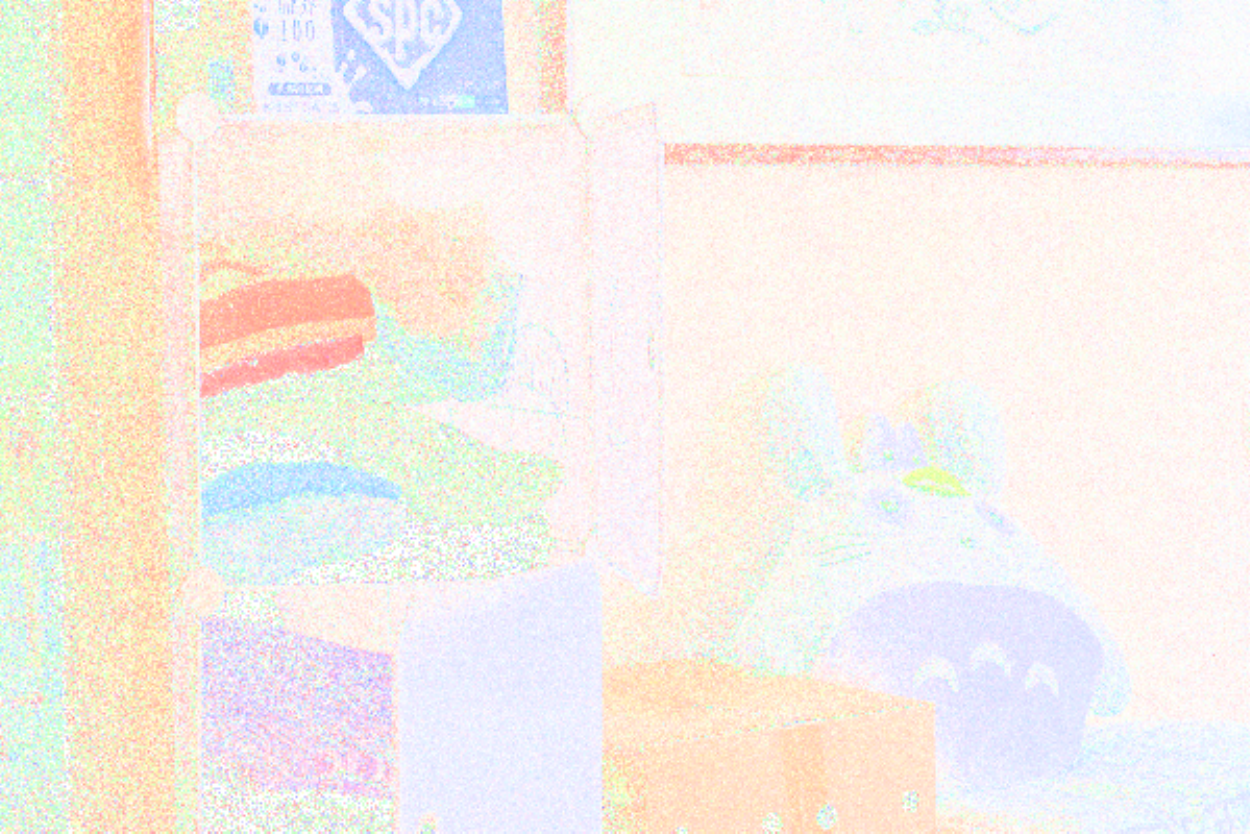}}
	   \vspace{1pt}
		\centerline{$k=3$}
	\end{minipage}
	\begin{minipage}{0.24\linewidth}
		\vspace{3pt}
		\centerline{\includegraphics[width=\textwidth]{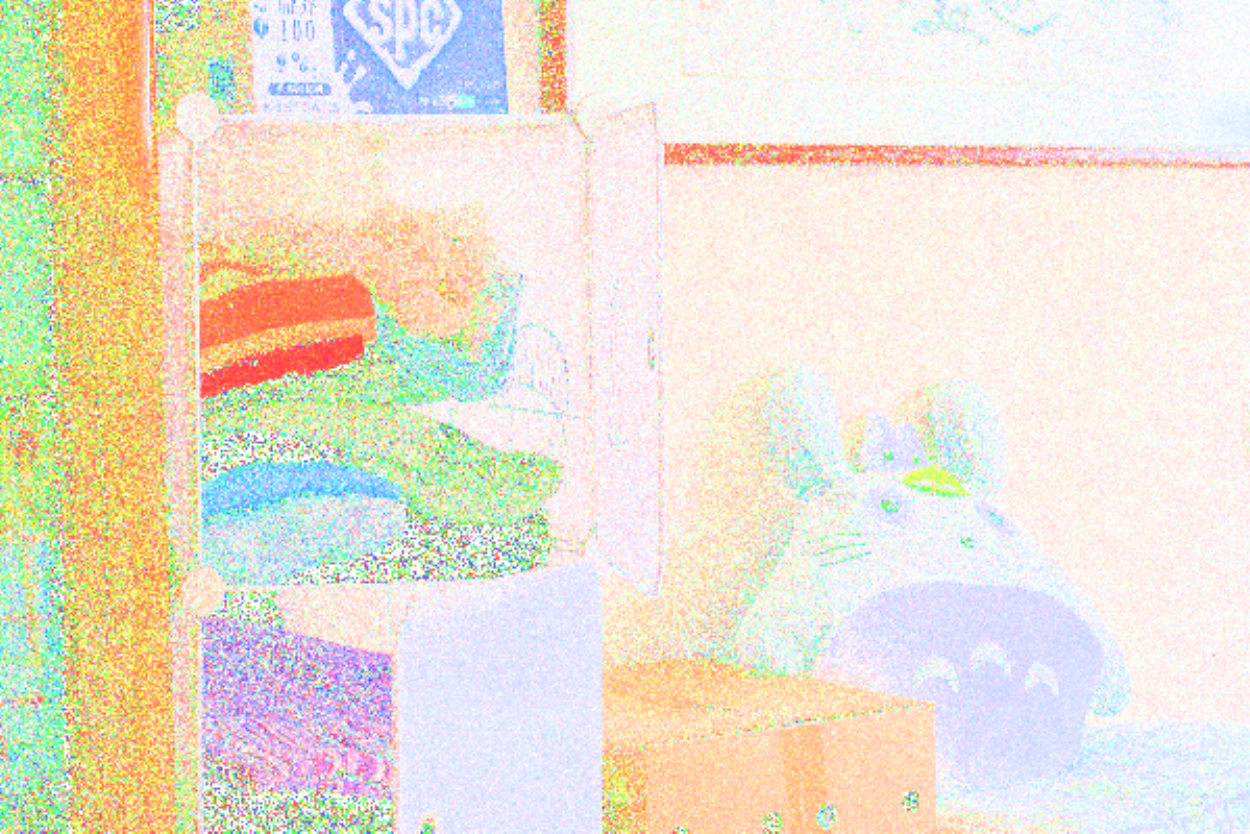}}
		\centerline{$k=10$}
	\end{minipage}
        \begin{minipage}{0.24\linewidth}
		\vspace{3pt}
		\centerline{\includegraphics[width=\textwidth]{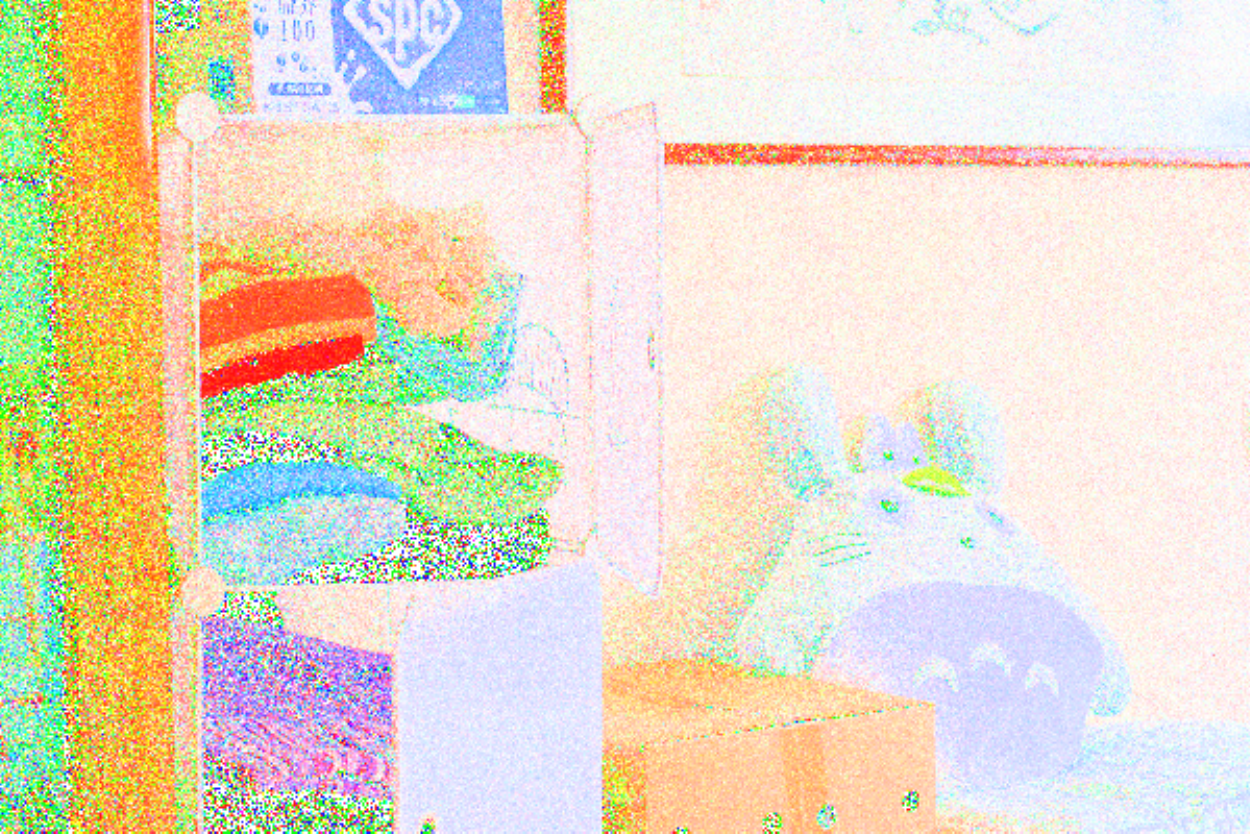}}
		\centerline{$k=50$}
	\end{minipage}
 \caption{HV-map comparison of density-$k$s in low-light images. It can be seen that as $k$ increases, the noise in the image is amplified, while the conflict between detail and noise becomes increasingly obvious. }
 \label{fig5}
\end{figure}

However, this leads to a new question about $\textbf{C}_k$, i.e., what is the key of setting $k$ in Eq. \ref{eq:2} if the low-light plane collapse is also satisfied by setting $\textbf{C}_k = \textbf{I}_{max}$?

The density-$k$ is set to be a hyper-parameter that learns the noise and detail conflicts between different datasets, which the motivation is in Fig. \ref{fig5}. 
As $k$ increases, we observe that the conflict between noise and detail becomes more pronounced, and also leads to greater color deviation with more intense saturation and contrast. 
We argue that the parameter $k$ can be set as a hyper-parameter for regulating the signal-to-noise ratio at low-light color space area, to be added to the subsequent training with the neural network.
Meanwhile, different values of $k$ can be trained using different networks and different datasets. 
From this, we obtain a universal collapse formula as
\begin{equation}
    \mathbf{C}_k(x)=\sqrt[k]{F(\mathbf{I}_{max}(x) )+\mathcal{\varepsilon} },
\label{eq:col}
\end{equation}
where $F(\cdot)$ is a function that passing through $(0,0)$ and $(1,1)$ and consecutive between $[0,1]$. $\mathcal{\varepsilon}$ is a small quantity that prevents the gradient of a power function from exploding when back-propagated, and $x$ denotes any pixel in the image.

When we design Eq. \ref{eq:col}, in addition to the method mentioned in the main text as
\begin{equation}
    F(\mathbf{I}_{max}(x)) = \sin{(\frac{\pi\mathbf{I}_{max}(x)}{2})},
    \label{eq:f1}
\end{equation}
we also considered two alternative approaches,
\begin{equation}
    F(\mathbf{I}_{max}(x)) = \mathbf{I}_{max}(x),
    \label{eq:f2}
\end{equation}
\begin{equation}
    and~F(\mathbf{I}_{max}(x)) = \log_2 (\mathbf{I}_{max}(x)+1),
    \label{eq:f3}
\end{equation}
which are both satisfied the $F(\cdot)$ function definition.
The reason we ultimately choose to use the sinusoidal ($\sin$) formula is because of its characteristic to minimize the risk of gradient explosion ($k<1$ or $\mathbf{I}_{max}(x) \rightarrow 0$ in Eq. \ref{eq:f2}) and gradient vanishing ($k\rightarrow0$ in Eq. \ref{eq:f3}). 
Therefore, our design can greatly improve the success rate during training in order to make modifications to other network structures more efficiently, saving time and cost.


\begin{figure}[t]
    \centering
    \includegraphics[width=\linewidth]{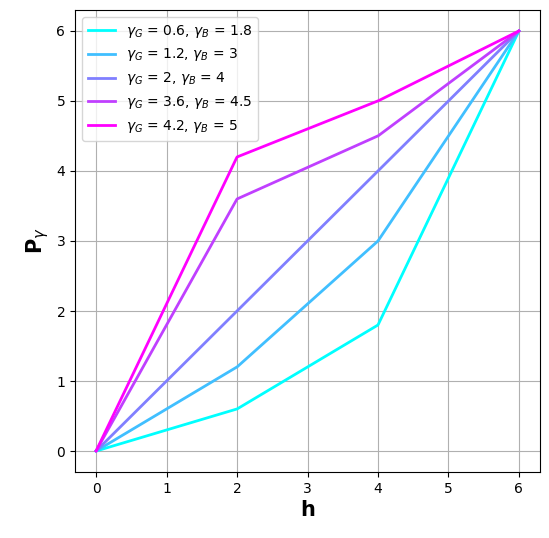}
    \caption{Different $\gamma_G$ and $\gamma_B$s in Eq. \ref{eq:5}. The independent variable is Hue as $\textbf{h}$ and the dependent variable is $\mathbf{P}_\gamma$.}
    \label{fig:gamma}
\end{figure}

\subsection{Further Develop of improving generalization between different datasets}
\textbf{Motivation.} In the field of LLIE, model generalization remains a challenging issue \cite{2022LLE}. 
Models trained on one specific dataset often perform poorly when applied to another dataset \cite{wang2024zero}. 
This dilemma arises primarily from two factors. 
\begin{enumerate}
    \item The varying sensitivity of different types of cameras to the red, green, and blue channel when capturing light intensity. 
    \item The differences in shooting environments across datasets, leading to variations in noise characteristics and brightness mapping statistics.
\end{enumerate}

Previous methods have acknowledged these issues and largely adopted unsupervised or zero-shot approaches to address them \cite{EnGAN,Zero-DCE,yang2023difflle,wang2024zero,PairLIE}. 
While these methods exhibit stronger generalization compared to some supervised approaches, their performance often falls short on the training and testing sets of the same dataset \cite{wang2024zero,PairLIE}. 
To tackle this problem, we explore applying a linear mapping to the Hue transformation $\mathbf{P}_\gamma$ and a saturation mapping function $T(x)$ in the HVI color space to adapt to the varying sensitivity of different cameras to the RGB channels and different low-light scenes.

\begin{figure}
	\begin{minipage}{0.48\linewidth}
		\vspace{3pt}
		\centerline{\includegraphics[width=0.8\textwidth]{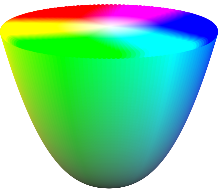}}
		\centerline{\includegraphics[width=\textwidth]{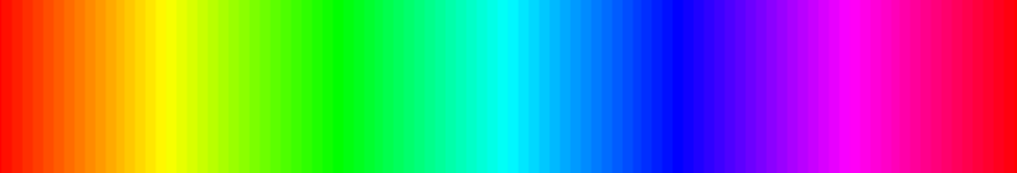}}
		\centerline{(a) $\gamma_G=2,\gamma_B=4$}
	\end{minipage}
\hfill
	\begin{minipage}{0.48\linewidth}
		\vspace{3pt}
		\centerline{\includegraphics[width=0.8\textwidth]{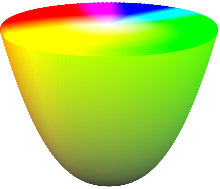}}
		\centerline{\includegraphics[width=\textwidth]{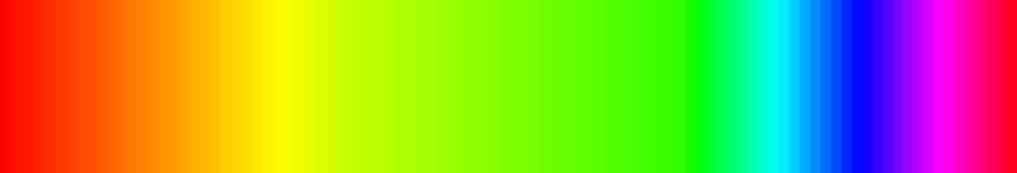}}
		\centerline{(b) $\gamma_G=4.2,\gamma_B=4.8$}
	\end{minipage}

	\begin{minipage}{0.48\linewidth}
		\vspace{3pt}
		\centerline{\includegraphics[width=0.8\textwidth]{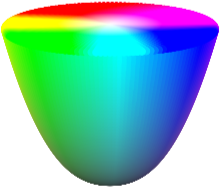}}
		\centerline{\includegraphics[width=\textwidth]{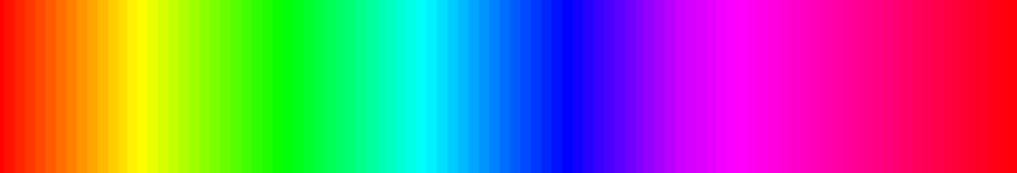}}
		\centerline{(c) $\gamma_G=1.2,\gamma_B=3.6$}
	\end{minipage}
\hfill
    \begin{minipage}{0.48\linewidth}
		\vspace{3pt}
		\centerline{\includegraphics[width=0.8\textwidth]{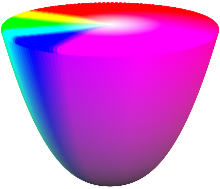}}
		\centerline{\includegraphics[width=\textwidth]{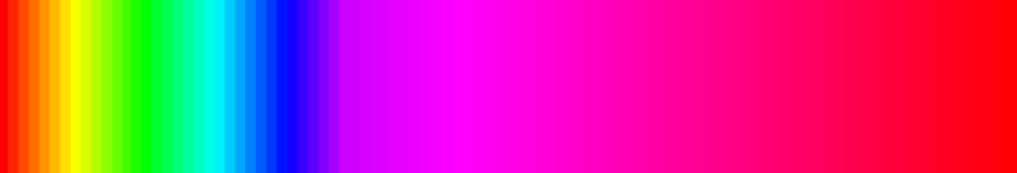}}
		\centerline{(d) $\gamma_G=0.6,\gamma_B=1.2$}
	\end{minipage}
 \caption{A visual presentation for different $\gamma_G$ and $\gamma_B$ values. }
 \label{fig:gamma2}
\end{figure}

\textbf{Methodology.} For Problem 1, we consider that the sensitivity of the camera on RGB can be equated to the relationship between the positions of the three RGB colours on the Hue axis of HSV. If one wants to have higher generalisation on other cameras, one just needs to fine-tune their positions.
Therefore, we set $\mathbf{P}_\gamma$ to be an linear map of Hue in HSV color space as
\begin{equation}
\mathbf{P}_\gamma= \begin{cases}
\frac{1}{2}\gamma_G\mathbf{h},  &\text{if } 0\le \mathbf{h} <2\\
\frac{1}{2}(\gamma_B-\gamma_G)(\mathbf{h}-2)+\gamma_G,  &\text{if } 2\le\mathbf{h} <4 \\
\frac{1}{2}(6-\gamma_B)(\mathbf{h}-6)+6,  &\text{if } 4\le \mathbf{h} \le6
\end{cases}
,
\label{eq:5}
\end{equation}
where $\gamma_G,\gamma_B\in(0,6)$, $\mathbf{h}\in\left[ 0,6\right]$ denotes the hue value mentioned in \ref{eq:hue}. Note that the polarization function should change to $h= \cos (2\pi \mathbf{P}_\gamma)$ and $v= \sin (2\pi \mathbf{P}_\gamma)$ from $\mathbf{h}$ to $\mathbf{P}_\gamma$. 
As Fig. \ref{fig:gamma}, Hue is mapped to $\mathbf{P}_\gamma$ by a trainable segmented linear function with $\gamma_G$ and $\gamma_B$, and each folded line represents a specific make or model of camera. 
We can manually adjust these two parameters in order to allow the HVI space to switch from one camera image distribution, to another. 
The visualization of these two gamma parameters on the color Hue is shown in Fig. \ref{fig:gamma2}. 
The original Hue axis is set by $\gamma_G=2,\gamma_B=4$ as Fig. \ref{fig:gamma2}(a). 
When $\gamma_G=4.2,\gamma_B=4.8$ (see Fig. \ref{fig:gamma2}(b)), green elongated by stretching. 
When $\gamma_G=0.6,\gamma_B=1.2$ (see Fig. \ref{fig:gamma2}(d)), red part is highlighted. 
However, there are limitations. 
It's that we can't be sure of its parameters for an unknown camera, so we can't adjust it to the appropriate $\gamma_G$ and $\gamma_B$ value.
This also limits the generalization of these two parameters, and requires us to conduct in-depth research in the future.

\begin{table*}[btp]
    \centering
    \renewcommand{\arraystretch}{1.2}
    \caption{Robustness testing experiments. All methods is trained on the LOLv1 and tested in the LOLv2-Syn dataset. The best result is in\textcolor{red}{~red} color.}
    \resizebox{\linewidth}{!}{
    \begin{tabular}{c|ccc|cccc|c}
    \Xhline{1.5pt}
         \cellcolor{gray!10}Methods&
         \cellcolor{gray!10}LLFlow \cite{LLFlow}&
         \cellcolor{gray!10}RetinexFormer \cite{RetinexFormer}&	 
         \cellcolor{gray!10}GSAD \cite{GSAD}&	 
         \cellcolor{gray!10}RUAS \cite{RUAS}&	 
         \cellcolor{gray!10}PairLIE \cite{PairLIE}&	 
         \cellcolor{gray!10}EnlightenGAN \cite{EnGAN}&	 
         \cellcolor{gray!10}ZeroDCE \cite{Zero-DCE}&	 
         \cellcolor{gray!10}\textbf{CIDNet}\\
    \Xhline{1.5pt}
         PSNR$\uparrow$&	
         17.119& 	
         16.570& 	
         15.854& 	
         15.326&
         19.074& 	
         16.183& 	
         17.712& 	
         \color{red}{19.457}
\\
         SSIM$\uparrow$&	
         0.812& 	
         0.769& 	
         0.748& 	
         0.488&
         0.797& 	
         0.734& 	
         0.815& 	
         \color{red}{0.817} 
\\
        LPIPS$\downarrow$&	
        0.224& 	
        0.252& 	
        0.243& 	
        0.458&
        0.230&	
        0.220& 	
        \color{red}{0.169}&
        0.193
\\
\midrule
        Train Mode&
        Supervised&
        Supervised&
        Supervised&
        Unsupervised&
        Unsupervised&
        Unsupervised&
        Zero-shot&
        Supervised
\\
        Main Type&
        Flow&
        Transformer&
        Diffusion&
        Unrolling&
        CNN&
        GAN&
        CNN&
        HVI + Transformer
\\
    \Xhline{1.5pt}
    \end{tabular}
    }
    \label{tab:v1-on-v2}
\end{table*}

\begin{figure}
    \centering
    \includegraphics[width=1\linewidth]{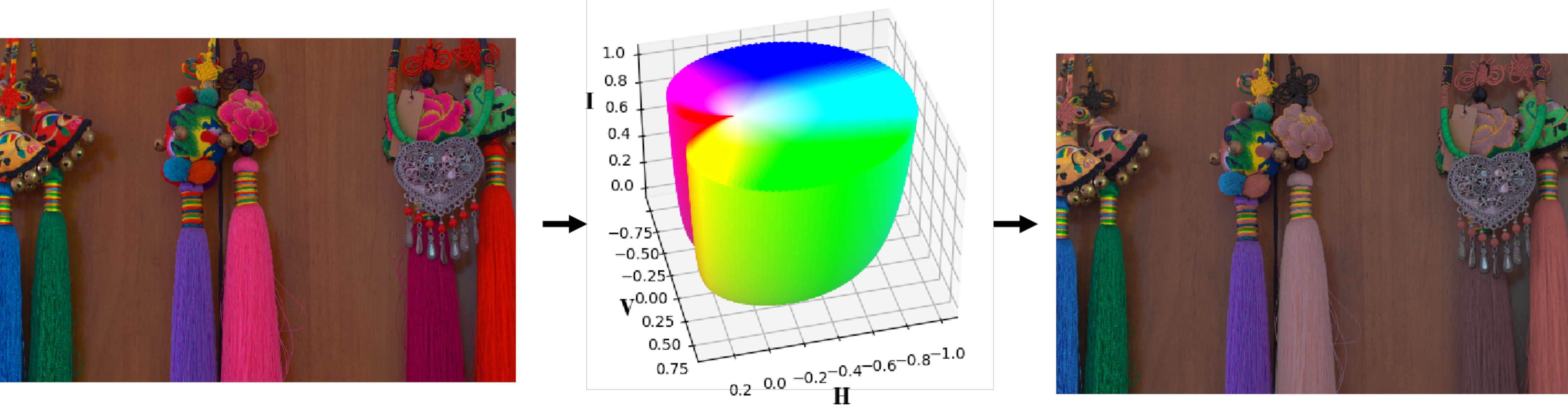}
    \caption{An image transform to HVI space with $T(x)=-4x(x-1)$, which filtered the red related colors. We note that these colors are not disappeared, but rather the colors are hidden in HVI space, preventing the networks from extracting the relevant features.}
    \label{fig:red}
\end{figure}

For the second problem, we try to solve it by setting a mapping equation $T(\cdot)$ on saturation-axis ($\mathbf{s}$). 
We consider that different scenes simply have different saturation on different hues.
Therefore, we introduce the Functional-Saturation-$T$ to establish a saturation mapping relationship between different scenes, which enhances the model's generalization capability.
Specifically, it can be utilized as
\begin{equation}
    \mathbf{D}_T = T(\frac{\mathbf{P}_\gamma}{6}),
    \label{eq:6}
\end{equation}
where $T(\cdot)$ satisfies $T(0)=T(1)$ and $T(\mathbf{P}_\gamma)\ge0$. 
The $T$ equation can be a custom function, a trainable formula, or a neural network, designed to fit the saturation correspondence between different scenes.
For instance, HVI with $T(x)=-4x(x-1)$ is a color space that filter the red related colors to adapt the red-free outdoor scenes as Fig. \ref{fig:red}. 
If we set $T(x)=t|x-0.5|$ where $t$ is a trainable parameter, then $T(x)$ is a trainable filter to adaptively find the colors that need to be filtered out.
If $T$ is a well-trained neural network to adjust different saturation scenes, the generalization of HVI color space could achieve better.
In our main text, we choose $T(x)=1$ as a uniformly sensitive color space for each color, which has a color disc in HV-plane as Fig. \ref{fig:gamma2}(a).

\textbf{Experiments on generalization ability.} 
To verify the generalization ability of our model, we take the model trained on LOLv1 \cite{RetinexNet} and tested it on LOLv2-Syn \cite{LOLv2} as Table \ref{tab:v1-on-v2}. 
Compared with three supervised learning state-of-the-art method, LLFlow \cite{LLFlow}, RetinexFormer \cite{RetinexFormer}, and GSAD \cite{GSAD}, our model comprehensively outperforms in three metrics, thanks to the $T(\cdot)$ function and $\mathbf{P}_\gamma$ mapping. 
And compared with the unsupervised methods RUAS \cite{RUAS}, PairLIE \cite{PairLIE} and EnlightenGAN \cite{EnGAN}, our approach significantly outperforms the once dominant unsupervised models in terms of generalisation ability.
Despite ZeroDCE being one of the most generalizable zero-shot methods, our CIDNet outperforms it in terms of PSNR and SSIM. Nevertheless, our LPIPS score is noticeably worse, indicating that the generated images do not subjectively compare well with the ground truth. This discrepancy may arise from our failure to select more suitable $T$ equation and $\gamma_G$ and $\gamma_B$ parameter, which, although yielding higher metrics, result in less realistic image quality.

\begin{table}
    \centering
    \caption{Ablation study on different types of HVI color space, which is trained on the LOLv1 and tested in the LOLv2-Syn dataset.}
    \renewcommand{\arraystretch}{1.2}
    \resizebox{\linewidth}{!}{
    \begin{tabular}{ccccc}
    \Xhline{1.5pt}
         \cellcolor{gray!10}~& 
         \cellcolor{gray!10}\makecell{w/o Eq. \ref{eq:5} and Eq. \ref{eq:6}}&
         \cellcolor{gray!10}\makecell{w/ Eq. \ref{eq:5} Only}&
         \cellcolor{gray!10}\makecell{w/ Eq. \ref{eq:6} Only}&
         \cellcolor{gray!10}\makecell{Full}\\
    \Xhline{1.5pt}
         PSNR$\uparrow$&17.545&18.112&18.458&19.457\\
         SSIM$\uparrow$&0.794&0.811&0.807&0.817\\
         LPIPS$\downarrow$&0.232&0.219&0.200&0.193\\
    \Xhline{1.5pt}
    \end{tabular}
    }
    
    \label{tab:rob}
\end{table}

\textbf{Ablation Study.} We conduct a further ablation study on CIDNet with and without Eq. \ref{eq:5} and Eq. \ref{eq:6}, respectively, to test the crossing dataset generalization ability of our proposed method.
As shown in Tab. \ref{tab:rob}, With the addition of Eq. \ref{eq:5} and Eq. \ref{eq:6}, all three metrics, PSNR, SSIM, and LPIPS, show significant improvement, which demonstrate the effective of $T(\cdot)$ function and $\mathbf{P}_\gamma$ mapping, by adjusting the relationship between the RGB channels between different cameras and the different scenes that were captured.

\section{Architecture Details of CIDNet}
In this section, we first provide a detailed explanation of the Lighten Cross-Attention (\textcolor[RGB]{255,95,95}{LCA}) module mentioned in Sec. \ref{sec:dual}. 
Next, we discuss the physical-based significance underlying the Intensity Enhance Layer (IEL) and Color Denoise Layer (CDL). 
Finally, we conduct ablation studies on the various sub-modules within the LCA to evaluate their individual contributions.

\subsection{Lighten Cross-Attention}
\begin{figure}
    \centering
    \includegraphics[width=1\linewidth]{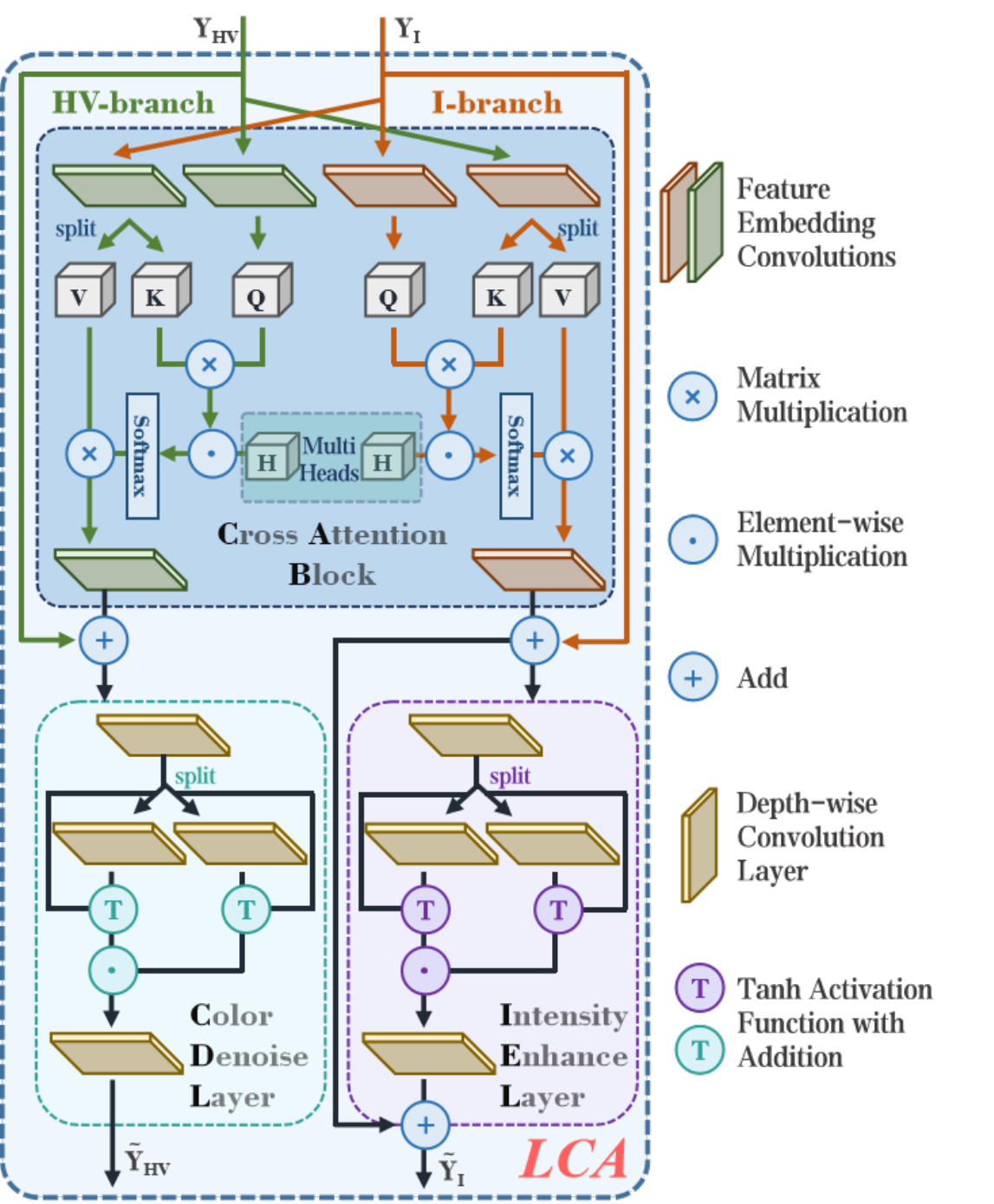}
    \caption{The dual-branch Lighten Cross-Attention (\textcolor[RGB]{255,95,95}{LCA}) block (\emph{i.e.}, I-branch and HV-branch). The LCA incorporates a Cross Attention Block (CAB), an Intensity Enhance Layer (IEL), and a Color Denoise Layer (CDL). The feature embedding convolution layers contains a $1\times1$ depth-wise convolution and a $3 \times 3$ group convolution.}
    \label{fig:CIDNet-structure}
\end{figure}

To enhance the interaction between the structures of images contained in the brightness and color branches, we propose the Lighten Cross-Attention (LCA) module to learn the complementary information of HV-branch and intensity-branch.
As shown in Fig. \ref{fig:CIDNet-structure}, the HV-branch and I-branch in LCA handle HV features and intensity features, respectively. 
To learn the complementary potential between HV features and intensity features during the processing, we propose cross attention block (\textbf{CAB}) to facilitate mutual guidance between HV features and intensity features.
To force the CAB to learn the information from the opposite branch (\emph{i.e.}, HV branch only use the information of I-brach to refine itself), we utilize the one branch as the query and leverage another branch as key and value in the CAB.

As shown in \ref{fig:CIDNet-structure}, 
Specifically, the LCA module consists of a Cross Attention Block (\textbf{CAB}), an Intensity Enhance Layer (\textbf{IEL}) for the I-way, and a Color Denoise Layer (\textbf{CDL}) for the HV-way. 
The CAB exhibits a symmetrical structure between the I-way and HV-way. We use the I-branch as an example to describe the details.
$\mathbf{Y_{I}}\in \mathbb{R}^{\hat{H}\times\hat{W}\times\hat{C}}$ denotes the inputs of I-branch, our CAB first derives query ($\mathbf{Q}$)  by  $\mathbf{Q}=W^{(Q)}\mathbf{Y_{I}}$. Meanwhile, the CAB splits key ($\mathbf{K}$) and value ($\mathbf{V}$) by $\mathbf{K}=W^{(K)}\mathbf{Y_{I}}$ and $\mathbf{V}=W^{(V)}\mathbf{Y_{I}}$. $W^{(Q)}$, $W^{(K)}$ and $W^{(V)}$ represents the feature embedding convolution layers. We formulate as 
\begin{equation}
\begin{split}
\mathbf{\hat{Y}_I}=W(\mathbf{V} \otimes \mathrm{Softmax}\left( \mathbf{Q}\otimes \mathbf{K}/\alpha_H\right) + \mathbf{Y_I}),
\end{split}
\label{eq:11}
\end{equation}
where $\alpha_H$ is the multi-head factor \cite{dosovitskiy2021image} and $W(\cdot)$ denotes the feature embedding convolutions.

Next, following Retinex theory, intensity enhance layer (IEL) decomposes the tensor $\mathbf{\hat{Y}_{I}}$ as $\mathbf{Y}_I=W^{(I)}\mathbf{\hat{Y}_{I}}$ and $\mathbf{Y}_R=W^{(R)}\mathbf{\hat{Y}_{I}}$. The IEL is defined as 
\begin{equation}
\begin{split}
\mathbf{\tilde{Y}_{I}} = W_{s}(&(\tanh{(W_{s}\mathbf{Y}_I)} + \mathbf{Y}_I) \\ \odot&(\tanh{(W_{s}\mathbf{Y}_R)} + \mathbf{Y}_R)),
\end{split}
\label{eq:12}
\end{equation}
where $\odot$ represents the element-wise multiplication and $W_{s}$ denotes the depth-wise convolution layers. Finally, the output of IEL adds the residuals to simplify the training process.

\subsection{Physical Significance of CDL And IEL}
We first explain the physical significance of the Color Denoise Layer as \textcolor[RGB]{46,162,152}{CDL} block. 
Each light ray that passes through our retina can be broken down into any number of different wavelengths of light with corresponding light saturations \cite{BORN1980133,fairman1997cie} as
\begin{equation}
\begin{split}
\mathbf{P} = \Sigma (S \odot W)
\end{split}
\label{eq:p1}
\end{equation}
where $\mathbf{P}$ is a map or image, $S,W$ represent the \textcolor[RGB]{84,130,53}{Saturation} and \textcolor[RGB]{46,117,182}{Wavelength}. A normal-light feature map $\hat{\mathbf{P}}$ can be generated by a low-light feature map $\mathbf{P}$ as
\begin{equation}
\begin{split}
\hat{\mathbf{P}} &= \mathbf{P} + \Delta\mathbf{P}\\
&=\Sigma ((S+ \Delta S) \odot (W+ \Delta W))
\end{split}
\label{eq:p2}
\end{equation}
where $\Delta S$ and $\Delta W$ formulate the color bias and feature noise of HV-features.

\begin{figure*}[htp]
    \centering
    \includegraphics[width=1\linewidth]{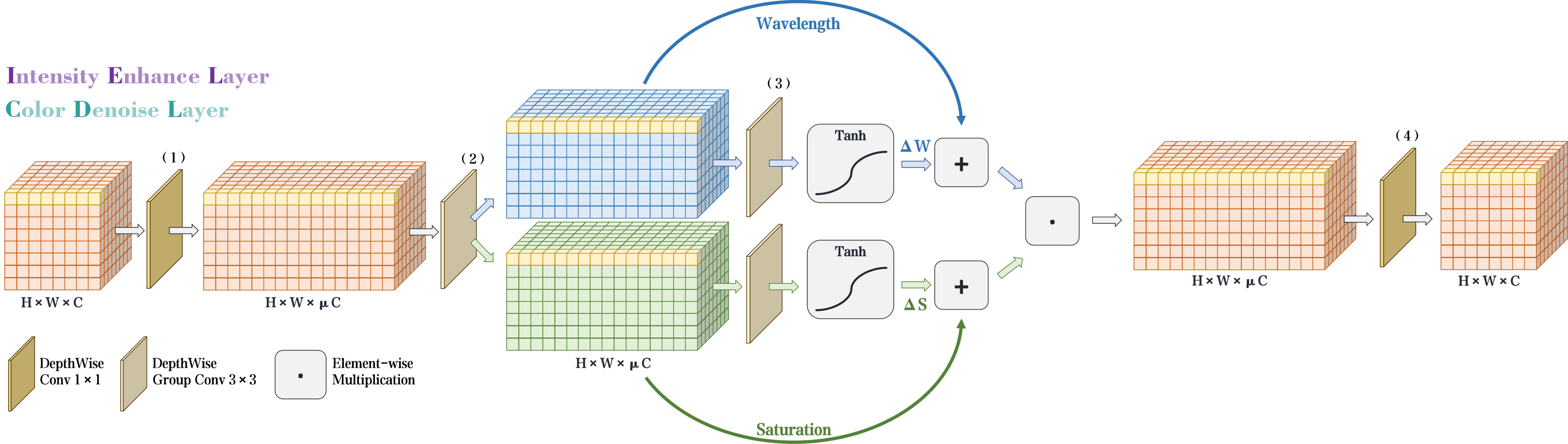}
    \caption{Structural diagrams of \textcolor[RGB]{112,48,160}{IEL} and \textcolor[RGB]{46,162,152}{CDL} that follow two different theories but have a consistent structure. \textbf{(1)} Pixel-by-pixel photometric decomposition. \textbf{(2)} Decoupling of each photometric component into \textcolor[RGB]{46,117,182}{Wavelength} and \textcolor[RGB]{84,130,53}{Saturation} corresponding to that wavelength. \textbf{(3)} Find the $\Delta$ of \textcolor[RGB]{46,117,182}{Wavelength} and \textcolor[RGB]{84,130,53}{Saturation} to solve the color bias and noise problem. \textbf{(4)} Re-stack the decomposed photometric channel-wise components to their original size to complete the enhancement task.}
    \label{fig:CDL}
\end{figure*}

Inputted a low-light tensor $\mathbf{P}$ with $H\times W \times C$, any small $1\times1\times C$ tensor on channel-wise can be considered as a photometric decomposition feature at that pixel. 
Our Color Denoise Layer (\textcolor[RGB]{46,162,152}{CDL}) first use a depth-wise $\texttt{Conv}1\times1$ to preliminary photodecomposition to a $H\times W \times \mu C$ tensor as Fig. \ref{fig:CDL}(1). 
Next, it's decomposed into \textcolor[RGB]{46,117,182}{Wavelength} and \textcolor[RGB]{84,130,53}{Saturation} by a depth-wise group $\texttt{Conv}3\times3$ layer as Fig. \ref{fig:CDL}(2). $\Delta W$ can be formulated as
\begin{equation}
\Delta W = \tanh{(\texttt{DWConv}_{3\times3}(W))}
\label{eq:1s}
\end{equation}
where $\texttt{DWConv}$ represent the Depth-wise grouped $\texttt{Conv}3\times3$ layer as Fig. \ref{fig:CDL} (3), and $\Delta S$ is excatly the same as $\Delta W $. 
The reason we use $\tanh$ as an activation function is that its dependent variable domain $\in(-1, 1)$. The depth-wise tensor $\hat{\mathbf{P'}}$ generated by $\hat{\mathbf{P'}}=(S+ \Delta S) \odot (W+ \Delta W)$ as Eq. \ref{eq:2}. 
Finally, $\hat{\mathbf{P'}}$ is multiplied by a Point-wise $\texttt{Conv}1\times1$ into a new light-up post-tensor as Fig. \ref{fig:CDL} (4).

\begin{figure*}[htp]
\centering 
\subfloat[\centering Input $\text{PSNR}\uparrow/\text{SSIM}\uparrow$]{\includegraphics[width=0.145\textwidth]{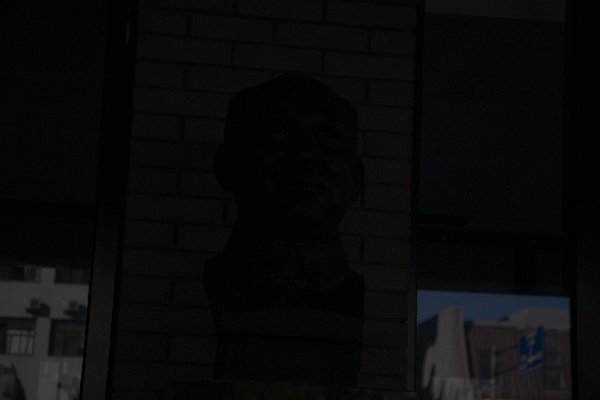}}
\quad
\subfloat[\centering w/o CAB $14.15/0.843$]{\includegraphics[width=0.145\textwidth]{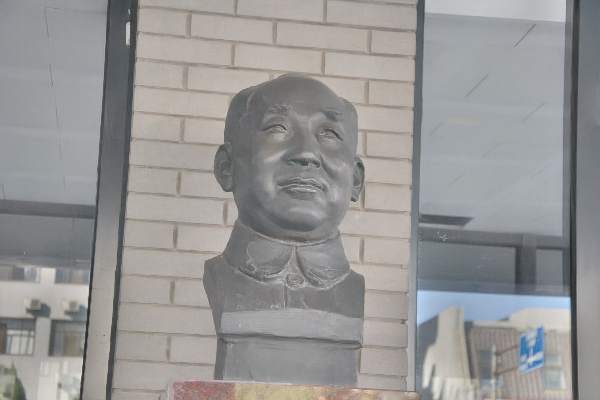}}
\quad
\subfloat[\centering w/o IEL $14.55/0.710$]{\includegraphics[width=0.145\textwidth]{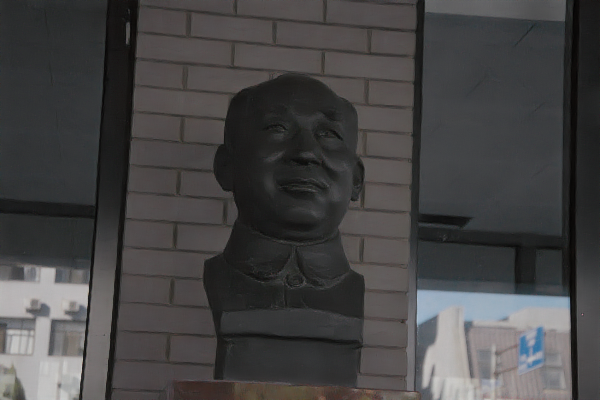}}
\quad
\subfloat[\centering w/o CDL $13.71/0.657$]{\includegraphics[width=0.145\textwidth]{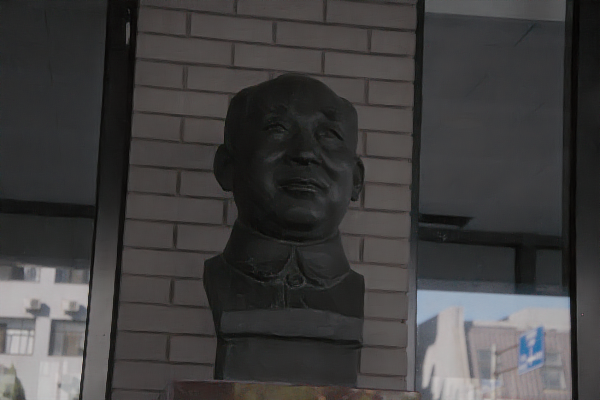}}
\quad
\subfloat[\centering Full LCA $20.80/0.848$]{\includegraphics[width=0.145\textwidth]{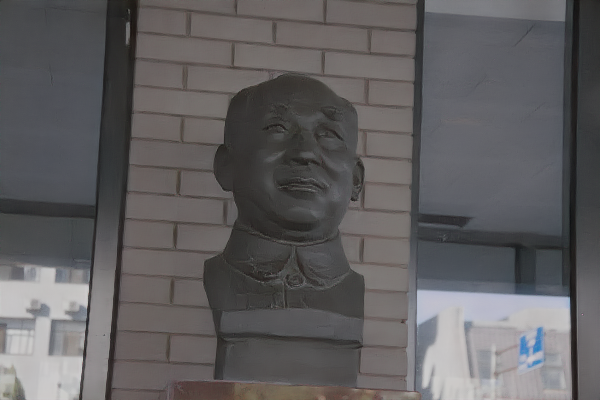}}
\quad
\subfloat[\centering GroundTruth $\infty/1.0$]{\includegraphics[width=0.15\textwidth]{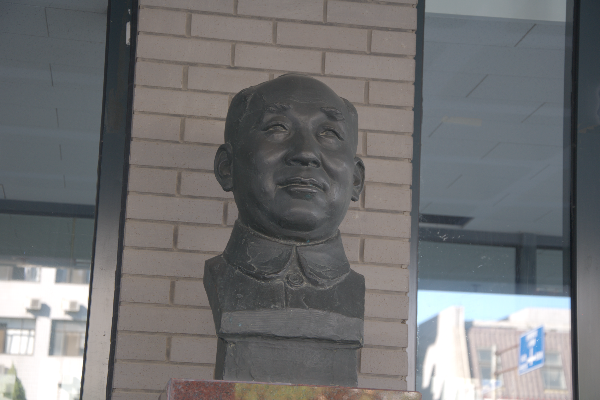}}
\caption{The visual quality comparison results on LOLv2-Real dataset with various LCA blocks (by removing submodules in the LCA). (e) Full LCA denotes the original design of the LCA block.} 
\label{fi3}
\end{figure*}

The Intensity Enhance Layer (\textcolor[RGB]{112,48,160}{IEL}) has exactly the same structure as \textcolor[RGB]{46,162,152}{CDL}, \textbf{but follows a different theory as Retinex }\cite{land1977retinex}, which can be decomposed to illumination $\mathbf{L}$ and reflectance $\mathbf{R}$ as
\begin{equation}
\begin{split}
\mathbf{I} = \mathbf{L} \odot \mathbf{R}
\end{split}
\end{equation}
where $\odot$ denotes the element-wise multiplication and $\mathbf{I}$ represent Intensity features. 
We follows RetinexFormer \cite{RetinexFormer} to model the corruptions of light Intensity as
\begin{equation}
\begin{split}
\hat{\mathbf{I}} &= \mathbf{I} + \Delta \mathbf{I}\\
&= (\mathbf{L}+ \Delta \mathbf{L}) \odot (\mathbf{R}+ \Delta \mathbf{R})
\end{split}
\label{eq:e1}
\end{equation}
where $ \Delta\mathbf{I}$ and $\Delta\mathbf{R}$ denote the perturbations. 
The generation formula for light-up Intensity-feature $\hat{\mathbf{I}}$ is similar to that for $\hat{\mathbf{P}}$ as Eq. \ref{eq:2}. 
Thus the network structure of the \textcolor[RGB]{112,48,160}{IEL} can reuse the structure of the \textcolor[RGB]{46,162,152}{CDL}, but with statistically utterly different laws.

\begin{table}[t]
    \centering
    \renewcommand{\arraystretch}{1.}
    \caption{Ablation of Cross Attention Block (CAB), Intensity Enhancement Layer (IEL) and Color Denoise Layer (CDL) in Lighten Cross-Attention (LCA).}
    \resizebox{\linewidth}{!}{
    \begin{tabular}{ccccccc}
    \Xhline{1.5pt}
         \cellcolor{gray!10}&
         \cellcolor{gray!10}CAB&	
         \cellcolor{gray!10}IEL&	
         \cellcolor{gray!10}CDL&	
         \cellcolor{gray!10}PSNR$\uparrow$&	
         \cellcolor{gray!10}SSIM$\uparrow$&	
         \cellcolor{gray!10}LPIPS$\downarrow$
\\
\Xhline{1.5pt}
         (1)&&$\surd$  & $\surd$ &19.938& 	0.835& 	0.138 
\\
         (2)&$\surd$&&$\surd$    &22.647& 	0.855& 	0.126 
\\
         (3)&$\surd$&$\surd$  &  &22.324& 	0.847& 	0.136 
\\
         (4)&$\surd$&$\surd$&$\surd$ &24.111& 	0.871& 0.108
\\
\Xhline{1.5pt}
    \end{tabular}
    }
    \label{tab:ab-LCA}
\end{table}

\subsection{Ablation study on LCA sub-modules}
The experiment is performed on LOLv2-Real dataset \cite{LOLv2} for fast convergence and stable performance. 
As Tab. \ref{tab:ab-LCA}, removing the CAB, CDL or IEL clearly shows a decrease effect of PSNR and SSIM, which demonstrates the effectiveness of sub-modules in LCA block. 
The visual quality comparisons are shown in Fig. \ref{fi3}. Specifically, removing CAB leads to unstable brightness enhancement, resulting in local overexposure and artifacts. On the other hand, removing IEL or CDL results in excessively dark brightness, thereby affecting the details.

\section{Additional Experiments and Details}

\subsection{Datasets Summarization}
All the datasets used in the paper are summarized in Tab. \ref{tab:dataset}.
We evaluated the model's ability to enhance low-light image using the LOL dataset and five additional unpaired datasets. 
To assess its generalization on large-scale datasets, we conducted experiments on the SICE and SID datasets. 
Furthermore, to verify the effectiveness of the HVI-CIDNet method in other tasks, we utilized the LOL-Blur dataset to test its performance on low-light enhancement combined with deblurring and the SIDD dataset to evaluate its image denoising capability in this supplementary.

\subsection{Implementation Details}
\textbf{Padding.} 
Since CIDNet contains three times the downsampling process, both the length and width of the processed image must be divisible by eight.
Therefore, for input images that are not integrable, we pad the input images to be a multiplier of $8\times8$ using reflect padding on both sides. 
After that, we crop the padded image back to its original size. 

\textbf{Clip Operation.} 
Since there may exist outliers in the output tensor of CIDNet, we apply a simple clip operation in the PHVIT module, \ie, $\mathbb{D}=\{p=(h,v,i)|~h^2 + v^2 \le \sin^\frac{2}{k}(\frac{\pi i}{2}),~0 \le i \le 1\}$, where $p=(h,v,i)$ is the three-dimensional coordinates in the HVI color space, and $k$ denotes the density-$k$ in Eq. \ref{eq:2}.

\begin{table}[t]

\centering

\renewcommand{\arraystretch}{1.2}
\label{tab:datasets}
\resizebox{\linewidth}{!}{
\begin{tabular}{l|l|rr|c}

\cellcolor{gray!10}\textbf{Dataset} & \cellcolor{gray!10}\textbf{Subsets} & \cellcolor{gray!10}\textbf{\#Train} & \cellcolor{gray!10}\textbf{\#Test} & \cellcolor{gray!10}\textbf{Resolutions ($H\times W$)} \\
\midrule
\midrule

\multirow{3}{*}{\textbf{LOL}} & v1 \cite{RetinexNet} & 485 & 15 & $400\times600$ \\
& v2-Real \cite{LOLv2} & 689 & 100 & $400\times600$ \\
& v2-Synthetic \cite{LOLv2}& 900 & 100 & $384\times384$ \\
\midrule

\multirow{5}{*}{\makecell{\textbf{Unpaired} \\\textbf{Datasets}}} & DICM \cite{DICM} & 0 & 69 & Various \\
& LIME \cite{LIME} & 0 & 10 & Various \\
& MEF \cite{MEF} & 0 & 17 & Various \\
& NPE \cite{NPE} & 0 & 8 & Various \\
& VV \cite{VV} & 0 & 24 & Various \\
\midrule

\multirow{3}{*}{\textbf{SICE}} & Original \cite{SICE} & 4800 & 0 & Various \\
& Mix \cite{SICE-Mix} & 0 & 589 & $600\times900$ \\
& Grad \cite{SICE-Mix} & 0 & 589 & $600\times900$ \\
\midrule

\textbf{SID }\cite{SID}& Sony-Total-Dark & 2099 & 598 & $1424\times2128$ \\
\midrule
\textbf{LOL-Blur} \cite{LEDNet}& \makecell{low-blur and\\high-sharp-scaled} & 10200 & 1800 & $640\times1120$ \\
\midrule
\textbf{SIDD} \cite{abdelhamed2018high} & (crop) & 30608 & 1280 & \makecell{$512\times512$ (train) \\ $256\times256$ (test)} \\
\bottomrule
\end{tabular}
}
\caption{Datasets summary on low-light image enhancement, joint low-light image deblurring and enhancement, and single image denoising task.}
\label{tab:dataset}
\end{table}

\begin{table*}[htp]
    \centering
    \renewcommand{\arraystretch}{1.}
    \caption{Quantitative results of PSNR/SSIM$\uparrow$ and LPIPS$\downarrow$ on the LOL (v1 and v2) datasets. Due to the limited number of test set in LOLv1, we use GT mean method during testing to minimize errors. Since the GLARE \cite{Han_ECCV24_GLARE} method did not provide pre-trained weights or training code for the LOLv2-Synthetic dataset, we were unable to conduct testing on this dataset. The FLOPs is tested on a single $256\times256$ image. The best result is in \textcolor{red}{red} color.}
\vspace{-1.8mm}
    \resizebox{\textwidth}{!}{
    \begin{tabular}{c|c|cc|ccc|ccc|ccc}
        \Xhline{1.5pt}
        \multirow{2}{*}{\textbf{Methods}}&\multirow{2}{*}{\textbf{Color Model}}&\multicolumn{2}{c|}{\textbf{Complexity}}&\multicolumn{3}{c|}{\textbf{LOLv1}} & \multicolumn{3}{c|}{\textbf{LOLv2-Real}} & \multicolumn{3}{c}{\textbf{LOLv2-Synthetic}}\\
        
        ~&~&Params/M&FLOPs/G&PSNR$\uparrow$&SSIM$\uparrow$&LPIPS$\downarrow$&PSNR$\uparrow$&SSIM$\uparrow$&LPIPS$\downarrow$&PSNR$\uparrow$&SSIM$\uparrow$&LPIPS$\downarrow$\\
        \hline

        CoLIE \cite{chobola2024fast} & HSV& 0.12 & 8.06 & 20.275 & 0.490 & 0.365&15.787& 0.497 & 0.340& 14.986 & 0.685 & 0.233
\\
        Zero-IG \cite{shi2024zero} & Retinex& 0.08 & 30.19 & 26.077 & 0.794 & 0.191&18.132& 0.746 & 0.248& 15.777 & 0.762 & 0.259
\\
        LightenDiff \cite{Jiang_2024_ECCV} & Retinex& 26.54 & 2257.42& 23.620 & 0.829 & 0.180&22.878& 0.855 & 0.166& 21.582 & 0.869 & 0.153
\\
        GLARE \cite{Han_ECCV24_GLARE}& RGB& 59.48&508.42 & 27.451& 0.883 & 0.081&22.511 & \color{red}{0.871} & \color{red}{0.105}& -&- &-
\\

        \textbf{CIDNet(Ours)}& HVI& 1.88 & 7.57 & 	\color{red}{28.201} &	\color{red}{0.889} &\color{red}{0.079}	&\color{red}{24.111}& 	\color{red}{0.871} &0.108 &	\textcolor{red}{25.705} &	\color{red}{0.942} & \color{red}{0.045}
\\
         \Xhline{1.5pt}
    \end{tabular}
    }
    \label{tab:table-LOL2}
\end{table*}

\textbf{GT Mean.} 
When the number of test images in a paired dataset is too small, resulting in significant metric fluctuations that hinder the evaluation of model performance, we often employ the GT Mean method as an auxiliary tool for testing various metrics. 
During evaluation, GT Mean aligns the overall brightness of the output image with the GroundTruth by adjusting the average luminance, allowing for a more precise numerical comparison of non-luminance-related attributes such as noise in details, color consistency, and structural stability. 
This evaluation approach has been widely adopted in recent state-of-the-art LLIE methods \cite{GSAD,Han_ECCV24_GLARE,LLFlow,pydiff}.
Specifically, GT Mean first transfer the output image and GroundTruth to gray image ($\mathbf{I}_{output}$ and $\mathbf{I}_{gt}$) \cite{gevers2012color}.
Then, we calculated the light adjustment value ($q$) as
\begin{equation}
    q = \frac{\frac{1}{MN} \sum_{i=1}^{M} \sum_{j=1}^{N} \mathbf{I}_{output}(i, j)}{\frac{1}{MN} \sum_{i=1}^{M} \sum_{j=1}^{N} \mathbf{I}_{gt}(i, j)},
\end{equation}
and the refined output image $\mathbf{I}_{q}$ can be formulated as $\mathbf{I}_{q}=q\mathbf{I}_{output}$ and clip to $[0,255]$ as an normal sRGB image.
We replaced the original method $\mathbf{I}_{output}$ with $\mathbf{I}_{q}$ to evaluate image metrics, aiming to minimize unnecessary luminance fluctuations and enable a more accurate comparison of different models' performance. 
This approach was applied to test all methods on the LOLv1 \cite{RetinexNet} test set (only 15 standard definition images).

\textbf{Loss Details.} To integrate the advantages of HVI  space and the sRGB space, the loss function consists both color spaces.
In HVI color space, we utilize L1 loss $L_1$, SSIM loss $L_d$ \cite{SSIM}, edge loss $L_e$ \cite{Edgeloss}, and perceptual loss $L_p$ \cite{johnson2016perceptual} for the low-light enhancement task. It can be expressed as 
\begin{equation}
\begin{split}
l(\hat{X}_{HVI},X_{HVI})&=\lambda_{1} \cdot L_1(\hat{X}_{HVI},X_{HVI}) \\
&+ \lambda_{d} \cdot L_d(\hat{X}_{HVI},X_{HVI})\\
&+ \lambda_{e} \cdot L_e(\hat{X}_{HVI},X_{HVI})\\
&+ \lambda_{p} \cdot L_p(\hat{X}_{HVI},X_{HVI}),
\end{split}
\label{eq:14}
\end{equation}
where $\lambda_{1},\lambda_{d},\lambda_{e},\lambda_{p}$ are all the weight to trade-off the loss function $l(\cdot)$.
In sRGB color space, we employ the same loss function as $l(\hat{I}, I)$.
Therefore, our overall loss function $L$ is represented by 
\begin{equation}
L= \lambda_{c}\cdot l(\mathbf{\hat{I}_{HVI}},\mathbf{I_{HVI}}) + l(\mathbf{\hat{I}},\mathbf{I}),
\end{equation}
where $\lambda_{c}$ is the weight to balance the loss in different color spaces.

\subsection{More Comparison and Discussion with Updated Methods on LOL Datasets}
\begin{figure}
\centering
\begin{minipage}[t]{0.237\linewidth}
    \centering
        \vspace{1pt}
        \centerline{\includegraphics[width=\textwidth]{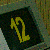}}
    \centerline{\small (a) CoLIE \cite{chobola2024fast}}
\end{minipage}
\hfill
\begin{minipage}[t]{0.237\linewidth}
    \centering
        \vspace{1pt}
        \centerline{\includegraphics[width=\textwidth]{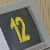}}
    \centerline{\small (b) GLARE \cite{Han_ECCV24_GLARE}}
\end{minipage}
\hfill
\begin{minipage}[t]{0.237\linewidth}
    \centering
        \vspace{1pt}
        \centerline{\includegraphics[width=\textwidth]{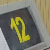}}
    \centerline{\small (c) CIDNet}
\end{minipage}
\hfill
\begin{minipage}[t]{0.237\linewidth}
    \centering
        \vspace{1pt}
        \centerline{\includegraphics[width=\textwidth]{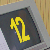}}
    \centerline{\small (d) GroundTruth}
\end{minipage} 
\vspace{-0.18cm}
\caption{Qualitative results of CoLIE \cite{chobola2024fast}, GLARE \cite{Han_ECCV24_GLARE} and our CIDNet on LOL dataset. Zoom in to see more details of the differences between GLARE and CIDNet.}
\label{fig:new}
\end{figure}

\textbf{Quantitative Result.} We selected four state-of-the-art low-light enhancement methods published in \textbf{ECCV 2024} \cite{leonardis2024computer} and \textbf{CVPR 2024}, namely CoLIE \cite{chobola2024fast}, Zero-IG \cite{shi2024zero}, LightenDiff \cite{Jiang_2024_ECCV}, and GLARE \cite{Han_ECCV24_GLARE}. 
The selection criteria required that each method provides open-source code and pre-trained weights. For methods without available pre-trained weights, we retrained them on the LOL datasets before testing. 
The quantitative results are presented in Tab. \ref{tab:table-LOL2}. Compared to these methods, our CIDNet achieves the most competitive metrics with the lowest FLOPs. 
However, on the LOLv2-real dataset, our LPIPS score is slightly worse than that of GLARE. This is because GLARE, being based on a Flow model, produces smoother outputs (see Figs. \ref{fig:new}(b) and (c)) that improve LPIPS scores but at the cost of losing finer details, making the results appear less realistic.
This is why the GLARE method has a high SSIM and a lower PSNR than ours.

\begin{figure}
    \centering
    \includegraphics[width=1\linewidth]{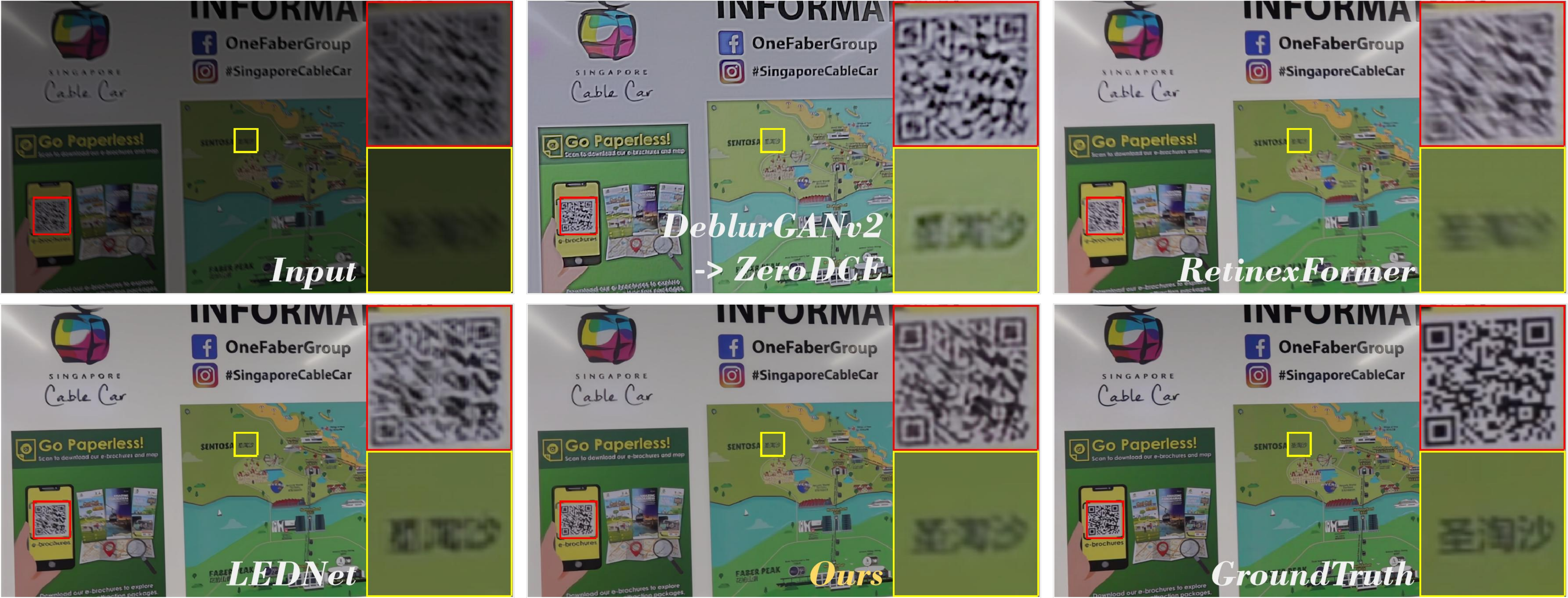}
    \caption{Visual comparison on LOL-Blur dataset. Compared to other methods, our CIDNet is closer to GroundTruth and more dominant in visual recognition.  (\textbf{Zoom in for best view.}) }
    \label{fig:blur}
\end{figure}

\textbf{Additional Discussion.} In Tab. \ref{tab:table-LOL2}, we observe that, apart from GLARE, the other three models utilize color models derived from physical optics theory. 
The CoLIE method enhances the Value axis in HSV color space but does not address noise reduction or color shift correction in the Hue and Saturation axes, resulting in a notably low SSIM score. 
As shown in Fig. \ref{fig:new}(a), CoLIE does not include a denoising process, so the output image can be seen as very strong noise, which seriously affects the visual quality.
Both Zero-IG and LightenDiff employ Retinex theory to decouple illumination and reflectance. By enhancing the illumination space and denoising the reflectance space, these methods achieve better metrics and visual results.
However, the metrics of CoLIE, Zero-IG, and LightenDiff do not surpass those of the RGB-based GLARE method. The primary reason is that these methods are Zero-shot approaches, which prioritize cross-dataset generalization at the cost of reduced performance on a single dataset \cite{wang2024zero}. 

Additionally, CoLIE and Zero-IG have significantly fewer parameters compared to GLARE, limiting their ability to fit more precise results.
We look forward to future methods that further explore the potential of the HVI color space, enabling models to achieve both superior visual quality and enhanced generalization capability.

\subsection{Joint Low-light Image Deblurring and Enhancement}
Long exposures in dimly lit environments can result in photos that are prone to blurring. To verify the robustness ability of our model, we conduct experiments on the low-light blur dataset LOL-Blur.

In the first set, we perform lighting-up with ZeroDCE and then deblurring with MIMO \cite{MIMO}. In the second set, we perform deblurring with DeblurGAN-v2 \cite{DeblurGANv2} and then enahancement with ZeroDCE. 
In the third group, we have retrained on LOL-Blur with four methods, RetinexFormer, MIMO, and LEDNet, and compared them with our CIDNet. 
The results (see Tab. \ref{tab:blur}) show that the quantitative comparison of CIDNet against the current stage SOTA method LEDNet by 5.15\%, 3.61\%, and 14.89\% in PSNR, SSIM, and LPIPS metrics respectively. Not only that, the FLOPs of our model are the lowest among these methods.

As shown in Fig. \ref{fig:blur}, we have taken a set of blurred images, recovered them using different methods, and compared them with GroundTruth. The experimental results reveal that the image reconstructions achieved by CIDNet exhibit a notable improvement in visual comfort and perceptual recognition, thereby enhancing the overall quality and interpretability of the generated images.

Fig. \ref{fig:LOLBlur} shows additional comparisons of our model with competing methods on LOL-Blur dataset \cite{LEDNet} for not only enhancing but deblurring. It is evident that our developed approach yields sharper edges, leading to a more visually appealing outcome.

\begin{table}[tp]
    \centering
    \renewcommand{\arraystretch}{1.2}
    \caption{Quantitative evaluation on LOL-Blur dataset. PSNR$\uparrow$ and SSIM$\uparrow$: the higher, the better; LPIPS$\downarrow$ and FLOPs$\downarrow$: the lower, the better. The symbol ‘$\dag$’ indicates that we use DeblurGAN-v2 trained on RealBlur \protect\cite{RealBlur} dataset. ‘$\ddag$’ indicates the network is retrained on the LOL-Blur dataset. The highest result is in\textcolor{red}{~red} color.}
    \resizebox{\linewidth}{!}{
\begin{tabular}{l|ccc|c}
\Xhline{1.5pt}
\rowcolor{gray!10} Methods& PSNR$\uparrow$&SSIM$\uparrow$&LPIPS$\downarrow$ &FLOPs$\downarrow$ \\
\Xhline{1.5pt}
ZeroDCE \cite{Zero-DCE} $\rightarrow$ MIMO \cite{MIMO} & 17.680 & 0.542 & 0.422 & 67.19 \\

DeblurGAN\textsuperscript{$\dag$} \cite{DeblurGANv2} $\rightarrow$ ZeroDCE \cite{Zero-DCE} & 18.330 & 0.589 & 0.384 & 34.19\\
RetinexFormer\textsuperscript{$\ddag$} \cite{RetinexFormer} & 22.904 & 0.824 & 0.236 & 15.85 \\
MIMO\textsuperscript{$\ddag$} \cite{MIMO} & 24.410 & 0.835 & 0.183 & 62.36 \\
LEDNet\textsuperscript{$\ddag$} \cite{LEDNet} & 25.271 & 0.859 & 0.141 & 35.93 \\
\Xhline{1.5pt}
\textbf{CIDNet}\textsuperscript{$\ddag$} &  \color{red}{26.572} & \color{red}{0.890} & \color{red}{0.120} & \color{red}{7.57} \\
\Xhline{1.5pt}
\end{tabular}
}
    
    \label{tab:blur}
\end{table}

\begin{figure}[t]
\centering
\begin{minipage}[t]{0.31\linewidth}
    \centering
        \vspace{1pt}
        \centerline{\includegraphics[width=\textwidth]{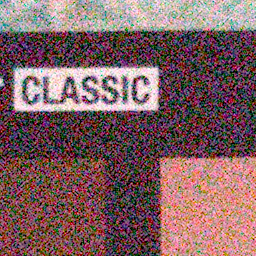}}
        \centerline{\includegraphics[width=\textwidth]{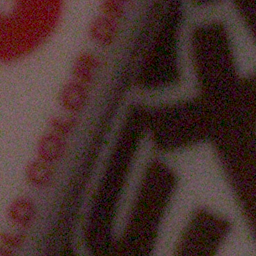}}
    \centerline{Input}
\end{minipage}
\hfill
\begin{minipage}[t]{0.31\linewidth}
    \centering
        \vspace{1pt}
        \centerline{\includegraphics[width=\textwidth]{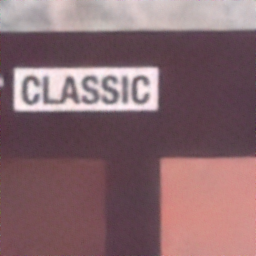}}
        \centerline{\includegraphics[width=\textwidth]{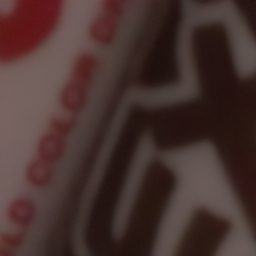}}
    \centerline{CIDNet}
\end{minipage}
\hfill
\begin{minipage}[t]{0.31\linewidth}
    \centering
        \vspace{1pt}
        \centerline{\includegraphics[width=\textwidth]{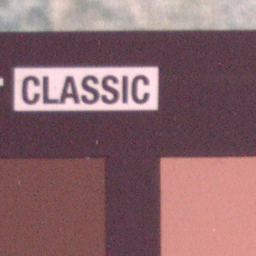}}
        \centerline{\includegraphics[width=\textwidth]{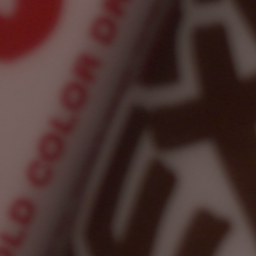}}
    \centerline{GroundTruth}
\end{minipage}
\caption{Visual comparison between inputs, outputs by our CIDNet, and GroundTruth. Obviously, our methods denoise a single image clearly with less color bias.}
\label{fig:SIDD}
\end{figure}

\subsection{Image Denoising}
Tab. \ref{tab:denoise} reports the results of real noise removal on SIDD dataset. We compare three state-of-the-art methods with our CIDNet. It is clear to see that our CIDNet achieves a PSNR of 39.88 dB, which is 0.12 dB above the MIRNet \cite{MIRNet}. The outcomes underscore the favorable performance of our method in the denoising task as Fig. \ref{fig:SIDD}, further elucidating that the notable results in the enhancement task are attributed to effectively addressing the severe noise introduced by low-light conditions.

\subsection{Variant HVIT Ablation Study and Discussion}
To investigate how the first two $3\times3$ convolution layers (in Fig. \ref{fig:CIDNet-pipeline}(b)) learned to generate I-feature and HV-feature, we further develop three different HVIT by changing the inputs. 
We hypothesize that the intensity features (hereinafter abbreviated as I-Feature) can be generated either from the HVI-Map (the concatenation of Intensity Map and HV Color Map) or the Intensity Map, while the HV-Feature can similarly be generated from either the HVI-Map or the HV-Map. To validate this, we conducted additional ablation experiments on three types of HVIT, with the results summarized in Tab. \ref{tab:ablation-c}.
It is clear that the Half-HVIT, which is our default version, achieves the best restoration results among the three HVIT models. 
The performance drop of the Separate-HVIT is more pronounced, which can be attributed to the lower information content in the HV-Feature compared to the HVI-Map, which lacks guidance from the intensity part. 
On the other hand, the performance decline of the Full-HVIT was due to the interference noise information in the HVI-Map for extracting I-Feature, leading to convolutional layers failing to accurately extract key features.

\begin{table}
    \centering
    \renewcommand{\arraystretch}{1.2}
    \caption{Quantitative evaluation on SIDD dataset for testing image denoising. The best result is in\textcolor{red}{~red} color.}
    \resizebox{\linewidth}{!}{
    \begin{tabular}{cccccc}
    \Xhline{1.5pt}
         \cellcolor{gray!10}Methods& 
         \cellcolor{gray!10}\makecell{BM3D \cite{4271520}}&	
         \cellcolor{gray!10}\makecell{DnCNN \cite{7839189}}&	 
         \cellcolor{gray!10}\makecell{RIDNet\cite{Anwar_2019_ICCV}}&	 
         \cellcolor{gray!10}\makecell{MIRNet\cite{MIRNet}}&	 
         \cellcolor{gray!10}\textbf{CIDNet}\\
    \Xhline{1.5pt}
         PSNR$\uparrow$&	25.65&  	23.66& 	38.71& 	39.72& \color{red}{39.88}
\\
         SSIM$\uparrow$&	0.685& 	 	0.583& 	0.951& 	\color{red}{0.959} & \color{red}{0.959}
\\
    \Xhline{1.5pt}
    \end{tabular}
    }
    
    \label{tab:denoise}
\end{table}

\begin{table}
    \centering
    \renewcommand{\arraystretch}{1.2}
    \caption{Ablation of three different types of inputs in Enhancement Network. Each distinct convolution layers will extract and generate corresponding intensity features (as I-Feature) and HV-Feature from different input maps (the type of input is indicated in the columns I-Feature and HV-Feature). The best result is in \textcolor{red}{red} color.}
    \resizebox{\linewidth}{!}{
    \begin{tabular}{l|cc|ccc}
    \Xhline{1.5pt}
         \cellcolor{gray!10}Types&\cellcolor{gray!10}I-Feature&\cellcolor{gray!10}HV-Feature&\cellcolor{gray!10}PSNR$\uparrow$&	\cellcolor{gray!10}SSIM$\uparrow$&	\cellcolor{gray!10}LPIPS$\downarrow$
\\
\Xhline{1.5pt}
         (1) Half-HVIT& intensity & HVI-Map&\textcolor{red}{24.111}& 	\textcolor{red}{0.868}& 	\textcolor{red}{0.108} 
\\
         (2) Separate-HVIT& intensity & HV-Map&23.734& 	0.857& 	0.141
\\
         (3) Full-HVIT& HVI-Map & HVI-Map&23.814& 	0.859& 	0.127 
\\

\Xhline{1.5pt}
    \end{tabular}
    }
    \label{tab:ablation-c}
\end{table}


\subsection{HVI Color General Ablation}
The HVI color space proposed in this study is derived from an optimized adaptation of the HSV color space, aiming to address the inherent limitations in handling chromatic artifacts, particularly black and red noise, at the color space level. By integrating the polarization and $\mathbf{C}_k$ function, the coupling between chromatic detail and low-light noise is substantially decoupled. This advancement enables the HVI-based CIDNet to achieve remarkable performance improvements in low-light image enhancement tasks. Furthermore, to rigorously validate the efficacy of the proposed HVI framework, a comprehensive experimental protocol was designed and conducted as outlined below.

\begin{figure}
\centering
\begin{minipage}[t]{0.19\linewidth}
    \centering
        \vspace{1pt}
        \centerline{\includegraphics[width=\textwidth]{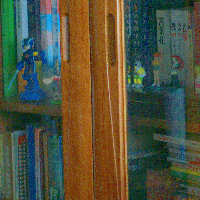}}
    \centerline{\small (a) Corrected}
\end{minipage}
\hfill
\begin{minipage}[t]{0.19\linewidth}
    \centering
        \vspace{1pt}
        \centerline{\includegraphics[width=\textwidth]{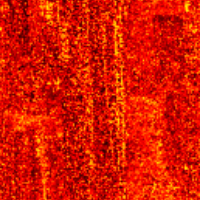}}
    \centerline{\small (b) $\mathbf{E}_{HS}$}
\end{minipage}
\hfill
\begin{minipage}[t]{0.19\linewidth}
    \centering
        \vspace{1pt}
        \centerline{\includegraphics[width=\textwidth]{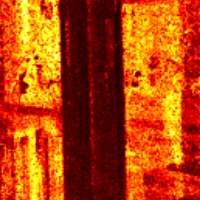}}
    \centerline{\small (c) $\mathbf{E}_{w/P}$}
\end{minipage}
\hfill
\begin{minipage}[t]{0.19\linewidth}
    \centering
        \vspace{1pt}
        \centerline{\includegraphics[width=\textwidth]{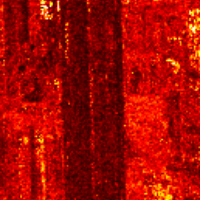}}
    \centerline{\small (d) $\mathbf{E}_{w/\mathbf{C}_k}$}
\end{minipage}
\hfill
\begin{minipage}[t]{0.19\linewidth}
    \centering
        \vspace{1pt}
        \centerline{\includegraphics[width=\textwidth]{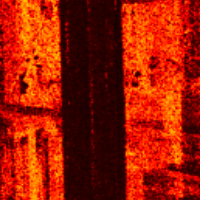}}
    \centerline{\small (e) $\mathbf{E}_{HV}$}
\end{minipage}
\caption{(a) Corrected Image: replacing the low-light Value Map ($\mathbf{V}$) with the GT’s, yet significant noise remains.
(b)-(e): Error Map between (a) and GT of different spaces. $w/P$ and $w/\mathbf{C}_k$ represents HVI with only Polarization and only $\mathbf{C}_k$ function, respectively.}
\label{fig:rebuttal1}
\end{figure}

We compare the error map of different color spaces to show the superiority of our methods.
In HSV, noise and chromatic information are often indistinguishable under low-light conditions, leading to low-quality images.
As Fig. \ref{fig:rebuttal1} (b), $\mathbf{E}_{HS}$ shows significant noise
and causes artifacts in the enhanced image (Fig. 1 of the manuscript).
The proposed polarization of HVI can decouple noise from red color details, 
which helps precisely recover the details and red region colors,
as shown in the bookcase in Fig. \ref{fig:rebuttal1} (c).
Furthermore, the proposed $\mathbf{C}_k$ function can cluster similar black tones, thus directly enhancing the SNR (see Fig. \ref{fig:rebuttal1} (d)).
Thanks to the polarization and $\mathbf{C}_k$ function, $\mathbf{E}_{HV}$ (Fig. \ref{fig:rebuttal1} (e)) has the lowest errors, demonstrating the effectiveness of the proposed method, achieving the highest PSNR of 27.115 dB (Tab. \ref{tab:rebuttal2}) on the LOL datasets.

\begin{table}
    \centering
    \resizebox{\linewidth}{!}{
    \begin{tabular}{ccccc}
    \toprule
        \cellcolor{gray!10}Color Space &\cellcolor{gray!10} HSV &\cellcolor{gray!10} w/ Polarization & \cellcolor{gray!10}w/ $\mathbf{C}_k$ &\cellcolor{gray!10} HVI \\
        Avg. PSNR$\uparrow$ & 14.346 & 20.632 & 25.046 & 27.115 \\
    \bottomrule
    \end{tabular}
    }
    \caption{The average PSNR$\uparrow$ of corrected image in different color spaces, across LOLv1 and v2 datasets.}
    \label{tab:rebuttal2}
\end{table}

\begin{table}
    \centering
    \resizebox{\linewidth}{!}{
    \begin{tabular}{ccccccc}
    \toprule
        \cellcolor{gray!10} Datasets &\cellcolor{gray!10} DICM \cite{DICM} &\cellcolor{gray!10} LIME \cite{LIME} & \cellcolor{gray!10} MEF \cite{MEF} &\cellcolor{gray!10} NPE \cite{NPE} & \cellcolor{gray!10} VV \cite{VV} & \cellcolor{gray!10} Avg. \\
        NIQE$\downarrow$ & 3.357 & 3.032 & 3.114 & 3.326 & 2.487 & 3.134 \\
    \bottomrule
    \end{tabular}
    }
    \caption{The NIQE$\downarrow$ metric of five unpaired datasets enhanced by our method within random gamma curve technique to improve generalization ability.}
    \label{tab:general}
\end{table}

\subsection{Random Gamma Curve Experiment}
We observed that employing the random gamma curve technique ($\mathbf{I_{Input}}=\mathbf{I_{LQ}}^\gamma$, where $\gamma$ is the random number between 0.6 and 1.2) for data pre-processing enhances the generalization capability of the model. 
To improve performance on the five unpaired datasets, we created LOLv2+, combining paired images from LOLv2's real and synthetic subsets, and applied the random gamma curve technique for pre-processing (Tab. \ref{tab:general}). 
This reduced the NIQE metric by 0.389 (Tab. \ref{tab:SID}), demonstrating enhanced generalization on unseen data. 
As other methods in Tab. \ref{tab:SID} did not use this technique, we avoid direct comparisons and instead present our optimal results to highlight the approach's effectiveness in the main text.

\subsection{More Visual Comparisons}
\begin{figure*}
\centering
 \begin{minipage}{0.245\linewidth}
 \centering
        \vspace{3pt}
 	\centerline{\includegraphics[width=\textwidth]{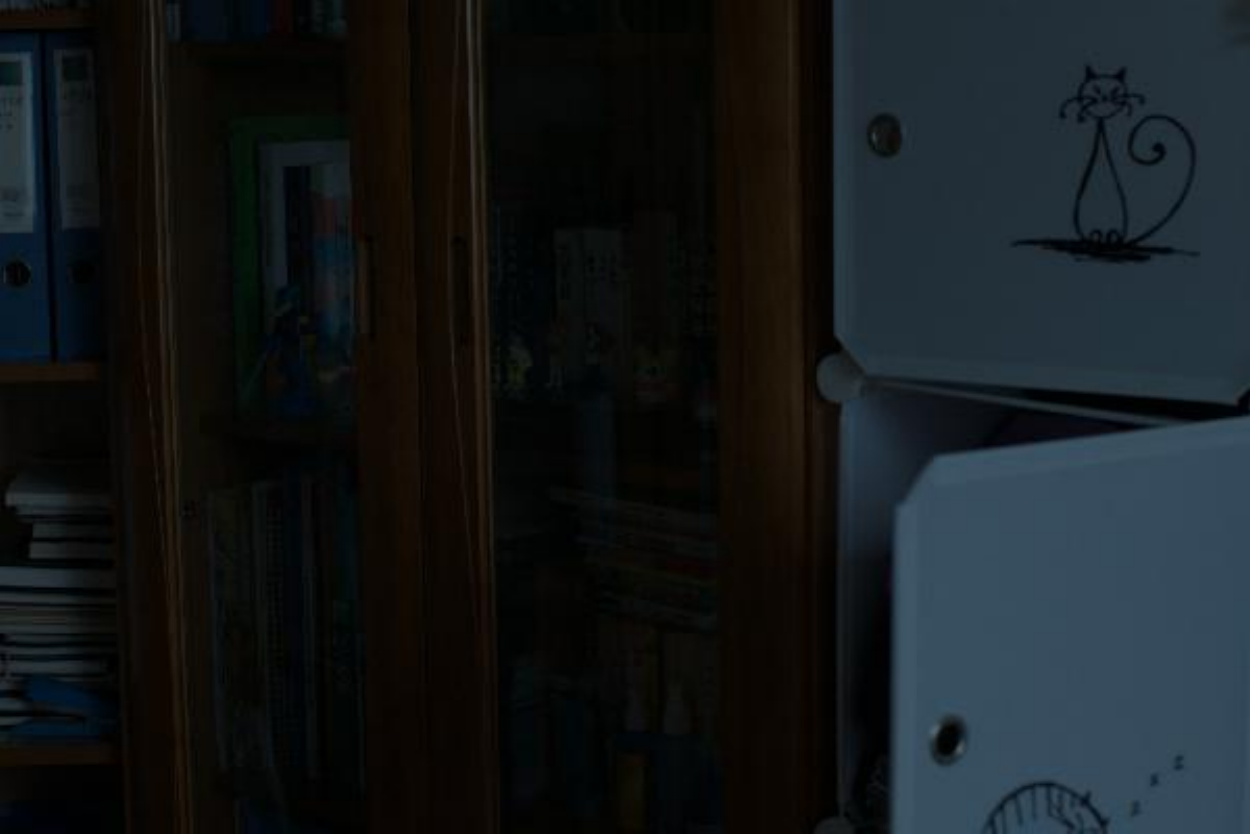}}
 	\centerline{Input}
 	\centerline{\includegraphics[width=\textwidth]{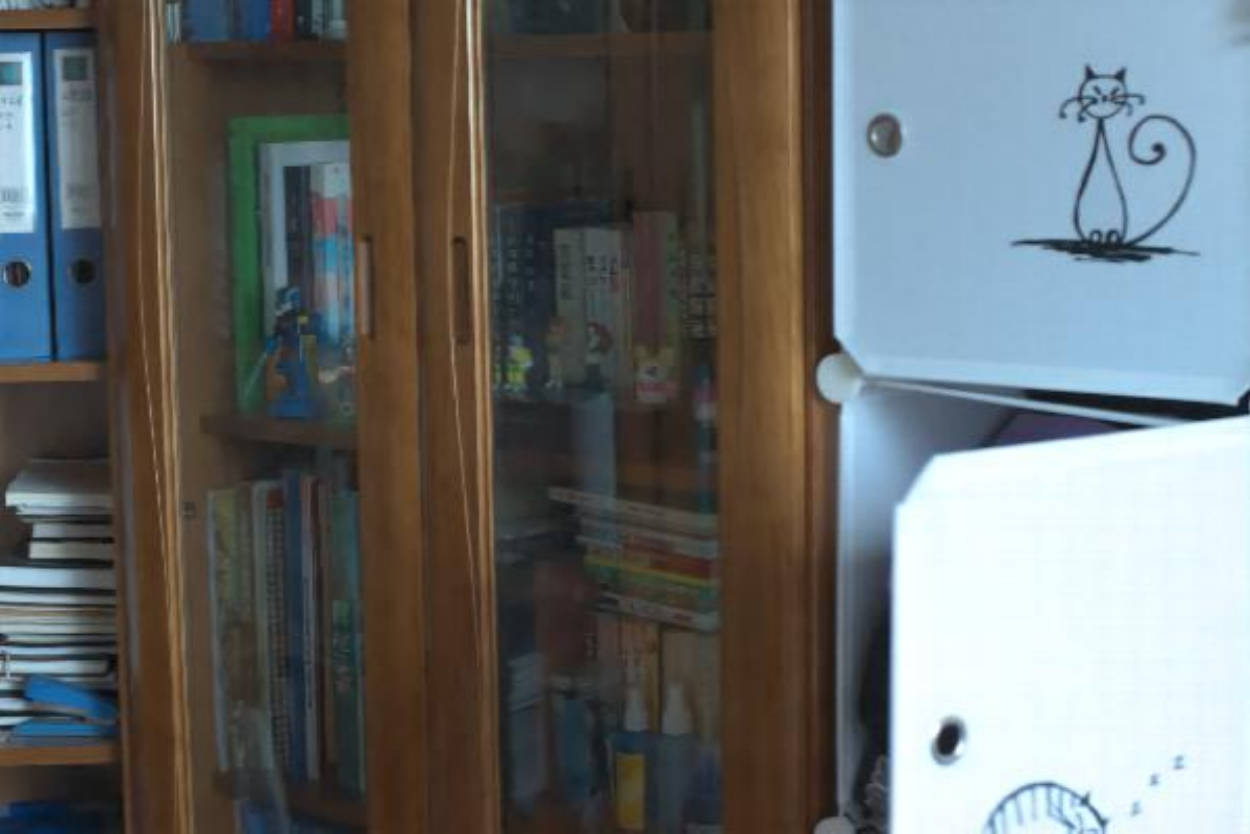}}
 	\centerline{SNR-Aware}
        \vspace{3pt}
 	\centerline{\includegraphics[width=\textwidth]{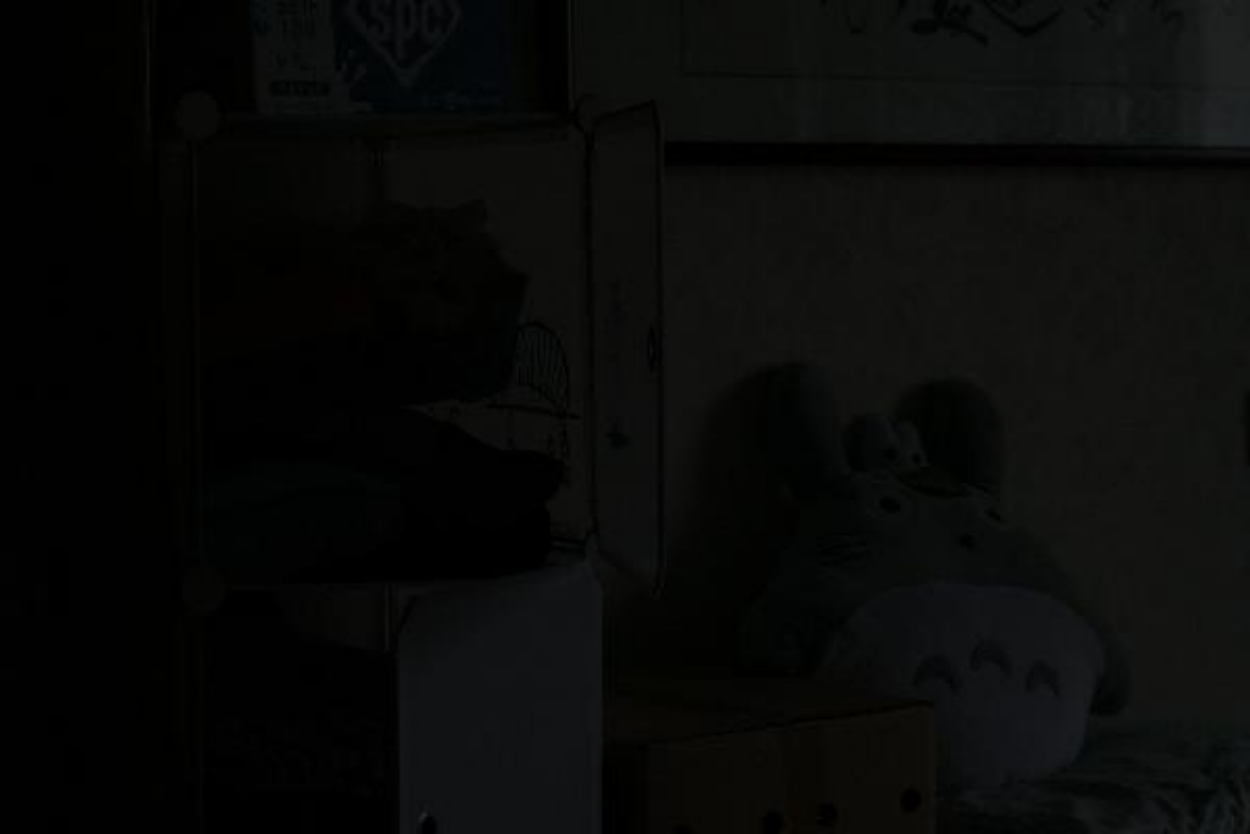}}
 	\centerline{Input}
 	\centerline{\includegraphics[width=\textwidth]{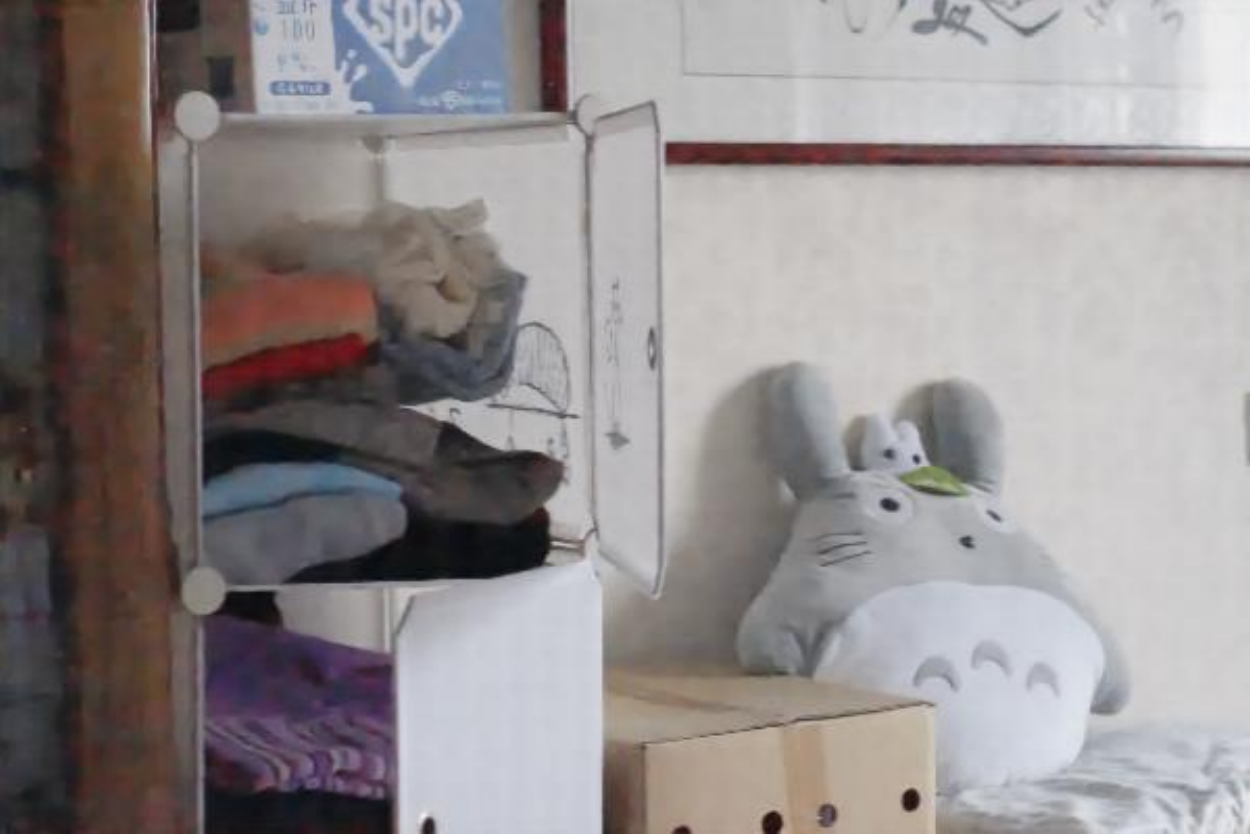}}
 	\centerline{SNR-Aware}
        \vspace{3pt}
 	\centerline{\includegraphics[width=\textwidth]{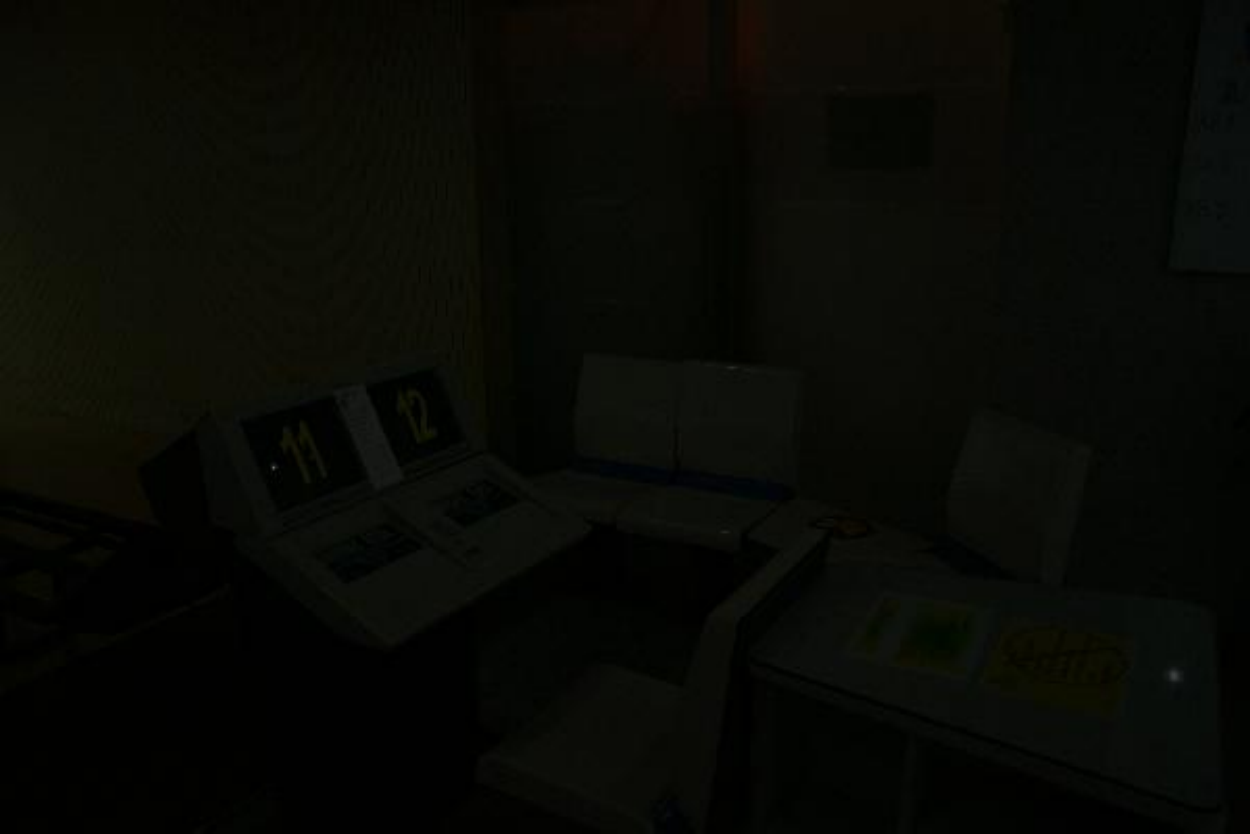}}
 	\centerline{Input}
 	\centerline{\includegraphics[width=\textwidth]{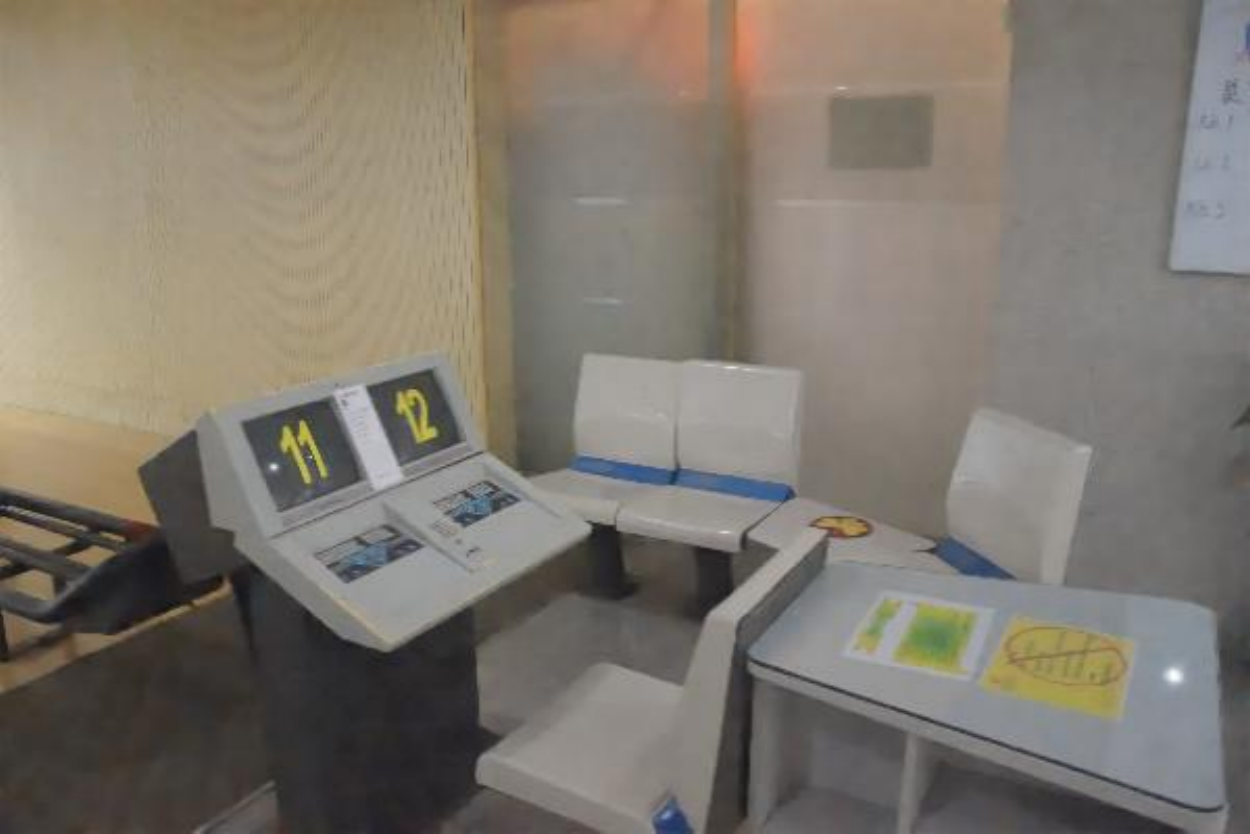}}
 	\centerline{SNR-Aware}
        \vspace{3pt}
\end{minipage}
\begin{minipage}{0.245\linewidth}
        \vspace{3pt}
 	\centerline{\includegraphics[width=\textwidth]{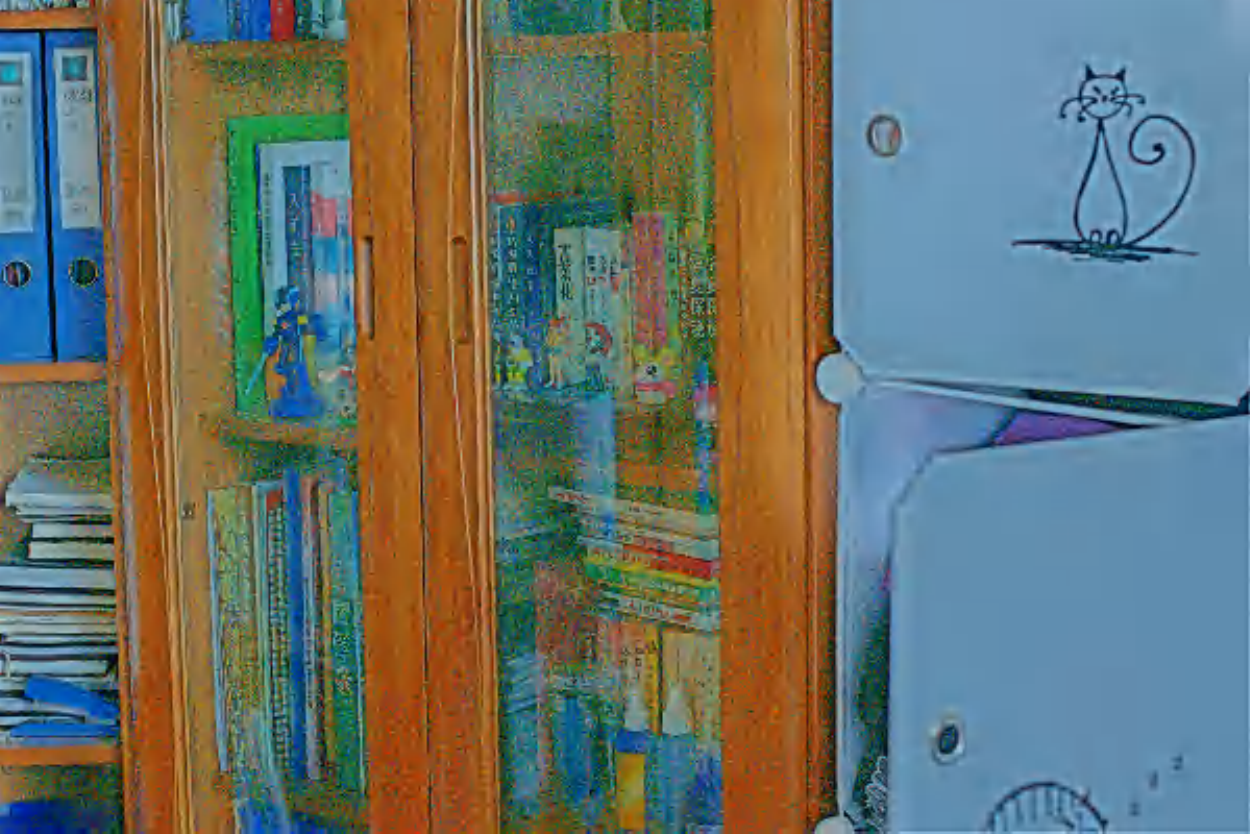}}
 	\centerline{RetinexNet}
        \vspace{2pt}
 	\centerline{\includegraphics[width=\textwidth]{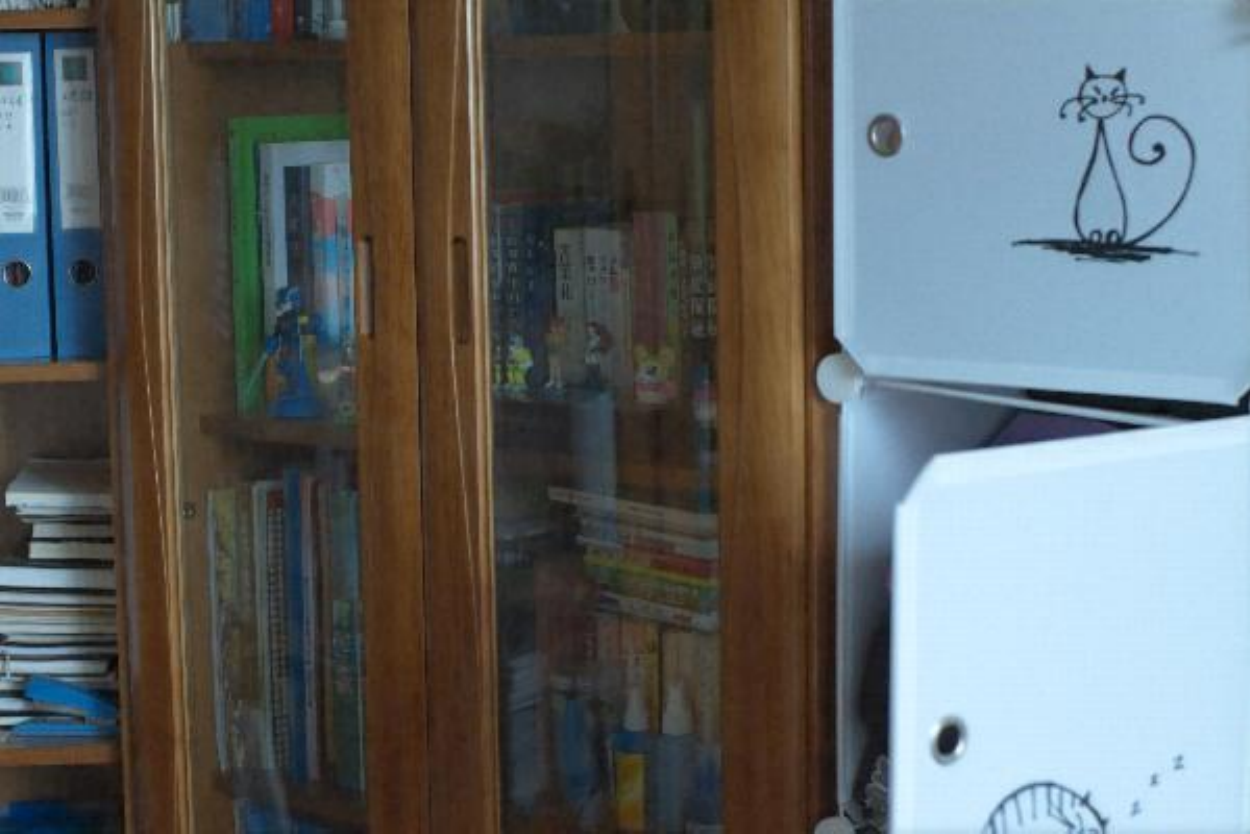}}
 	\centerline{RetinexFormer}
        \vspace{3pt}
 	\centerline{\includegraphics[width=\textwidth]{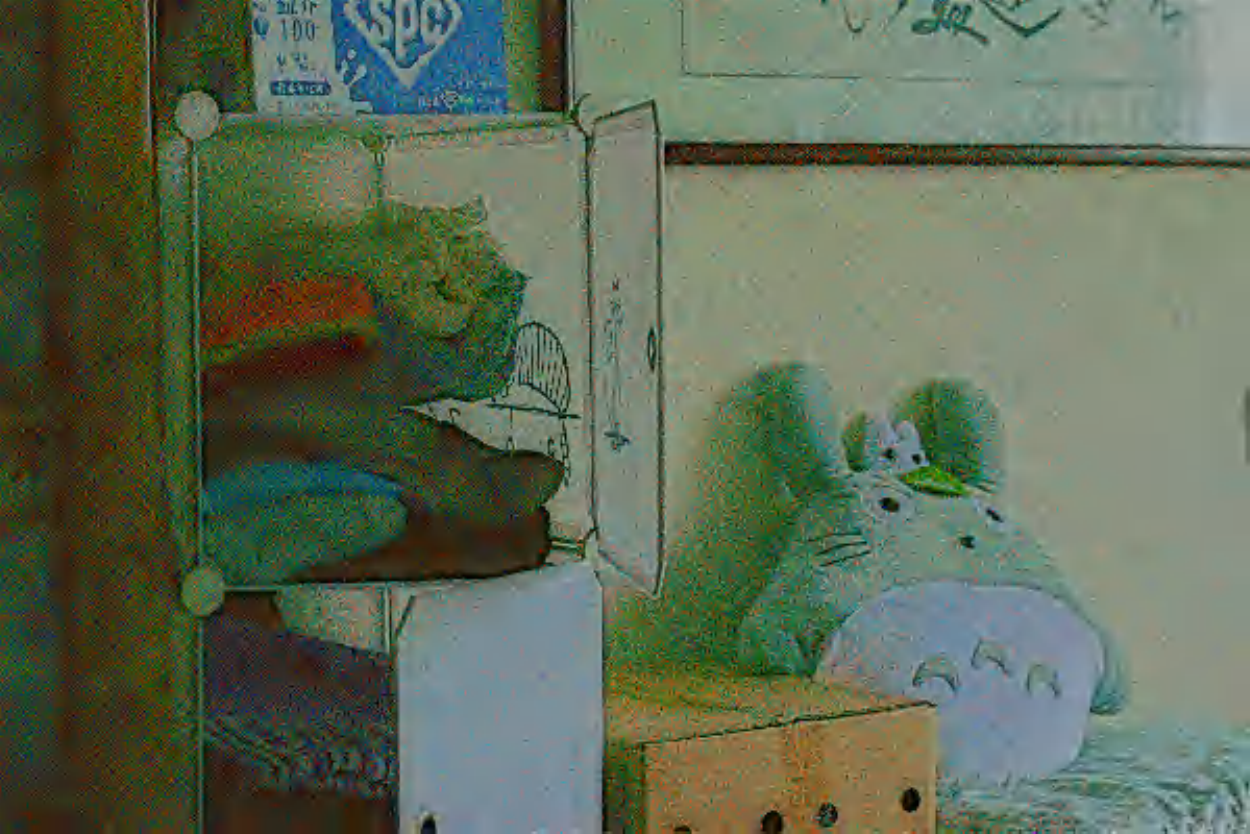}}
 	\centerline{RetinexNet}
        \vspace{2pt}
 	\centerline{\includegraphics[width=\textwidth]{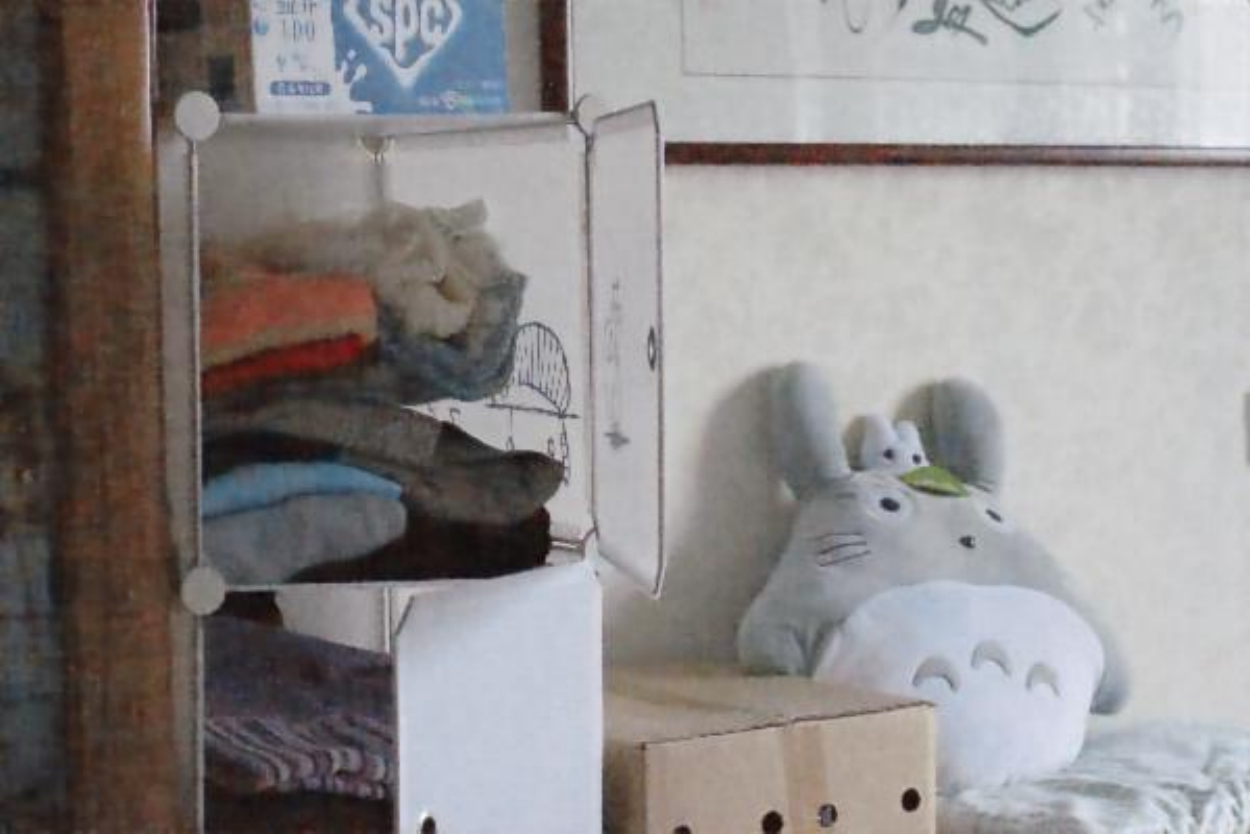}}
 	\centerline{RetinexFormer}
        \vspace{3pt}
 	\centerline{\includegraphics[width=\textwidth]{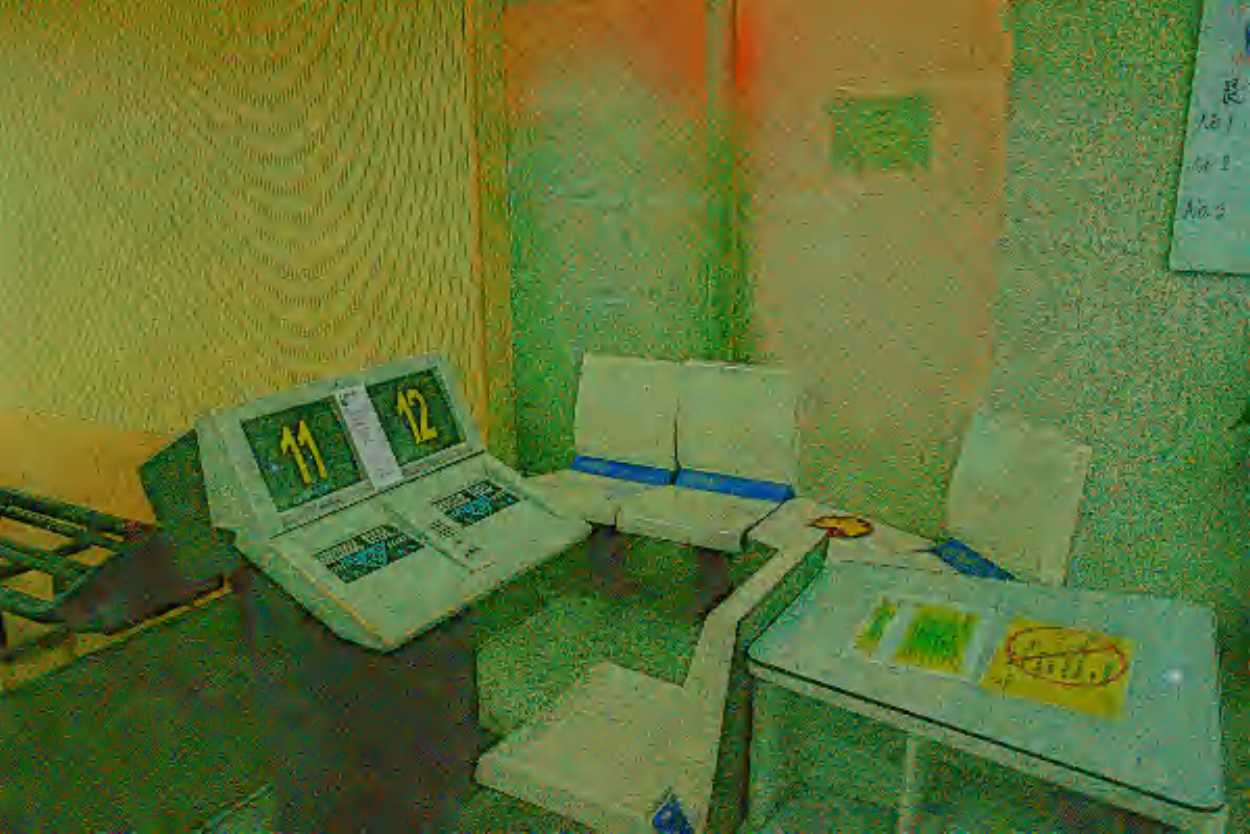}}
 	\centerline{RetinexNet}
        \vspace{2pt}
 	\centerline{\includegraphics[width=\textwidth]{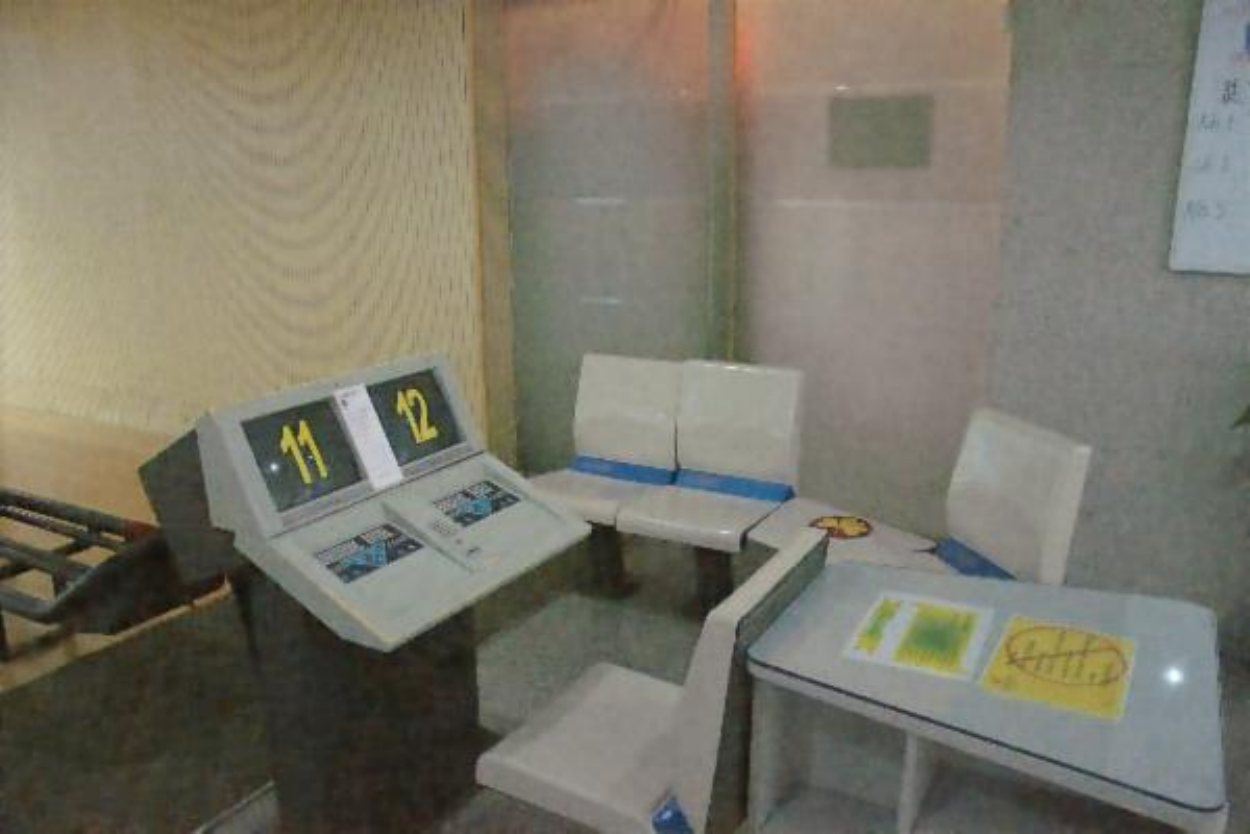}}
 	\centerline{RetinexFormer}
        \vspace{3pt}
 \end{minipage}
\begin{minipage}{0.245\linewidth}
 	\vspace{3pt}
 	\centerline{\includegraphics[width=\textwidth]{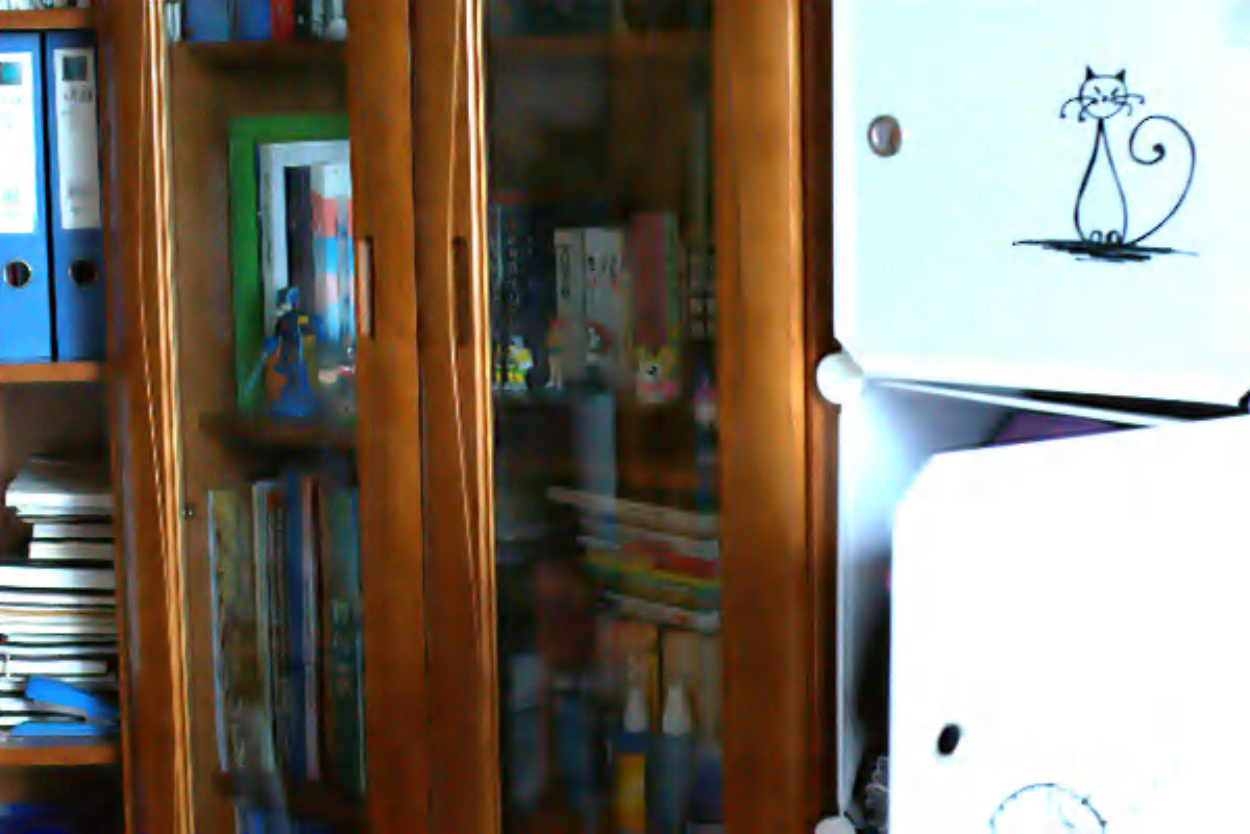}}
 	\centerline{RUAS}
 	\vspace{2pt}
 	\centerline{\includegraphics[width=\textwidth]{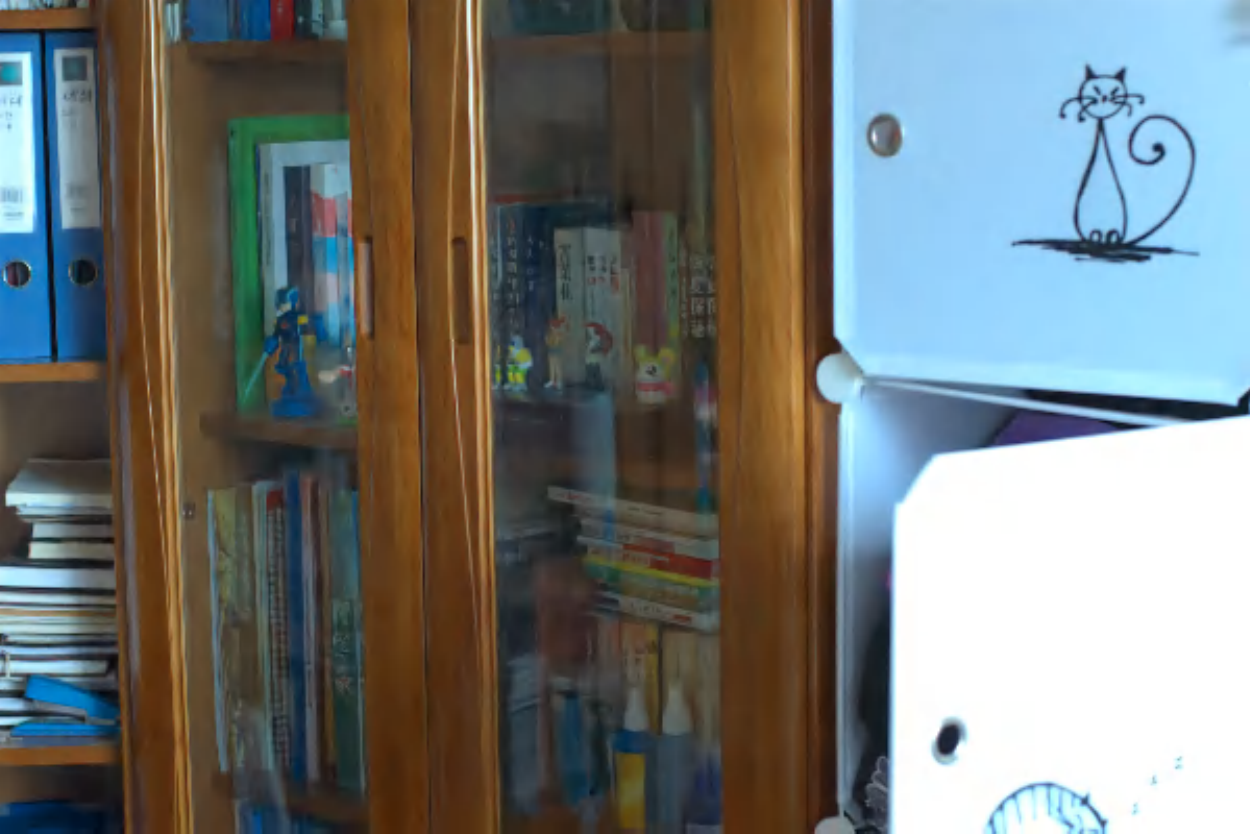}}
 	\centerline{CIDNet}
 	\vspace{3pt}
 	\centerline{\includegraphics[width=\textwidth]{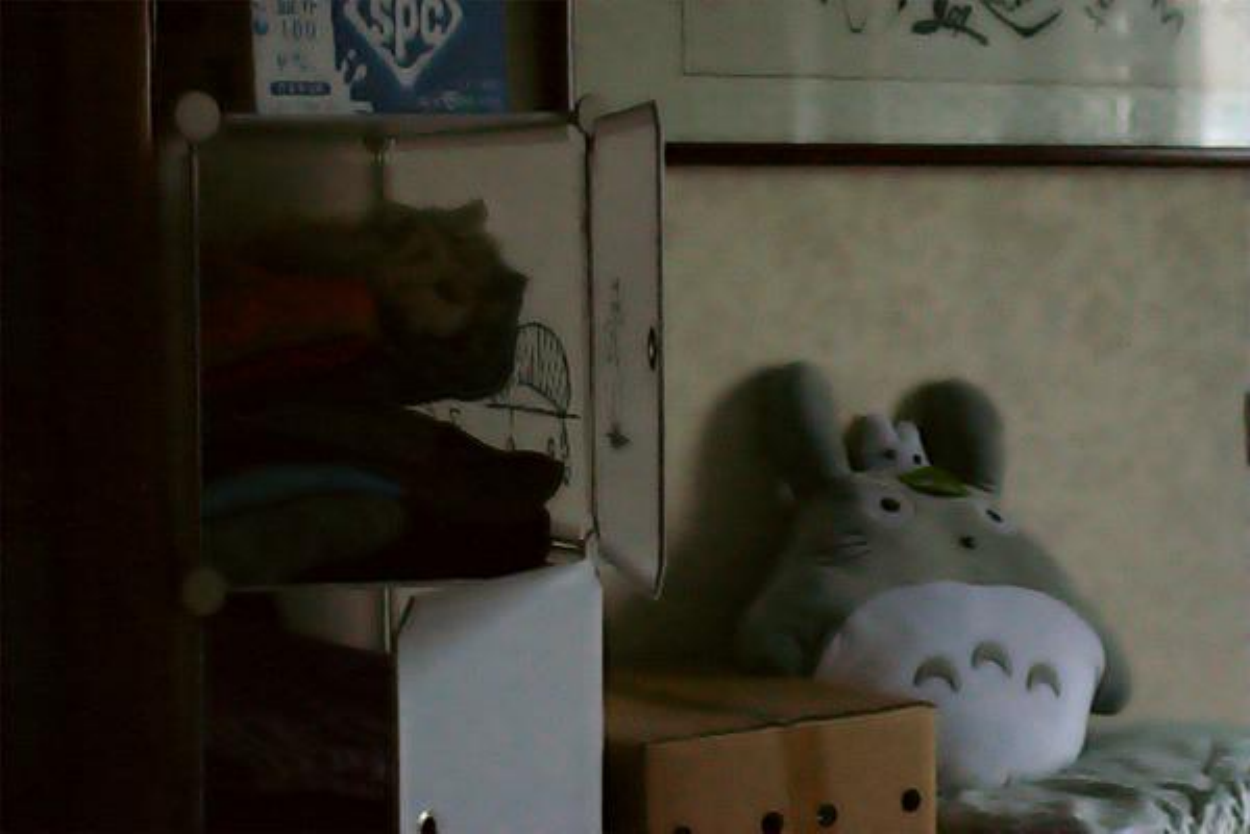}}
 	\centerline{RUAS}
 	\vspace{2pt}
 	\centerline{\includegraphics[width=\textwidth]{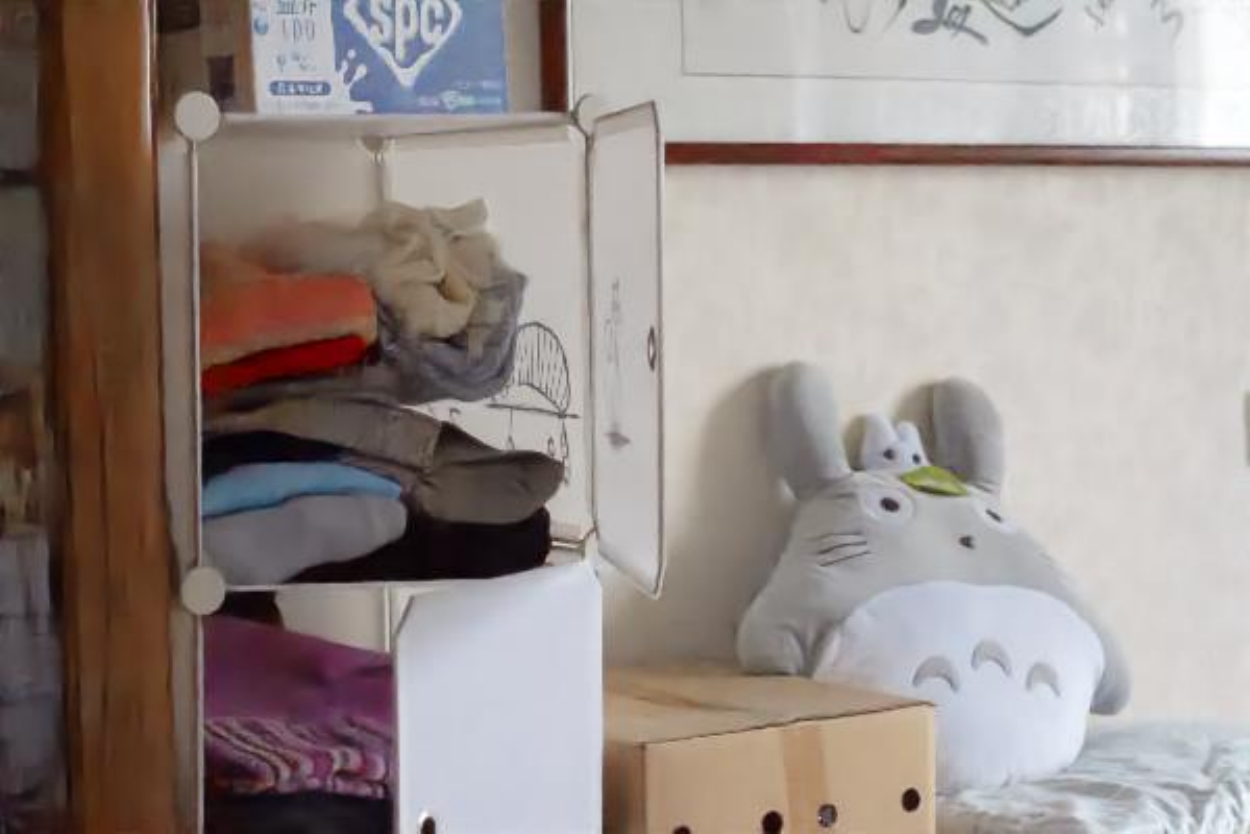}}
 	\centerline{CIDNet}
 	\vspace{3pt}
 	\centerline{\includegraphics[width=\textwidth]{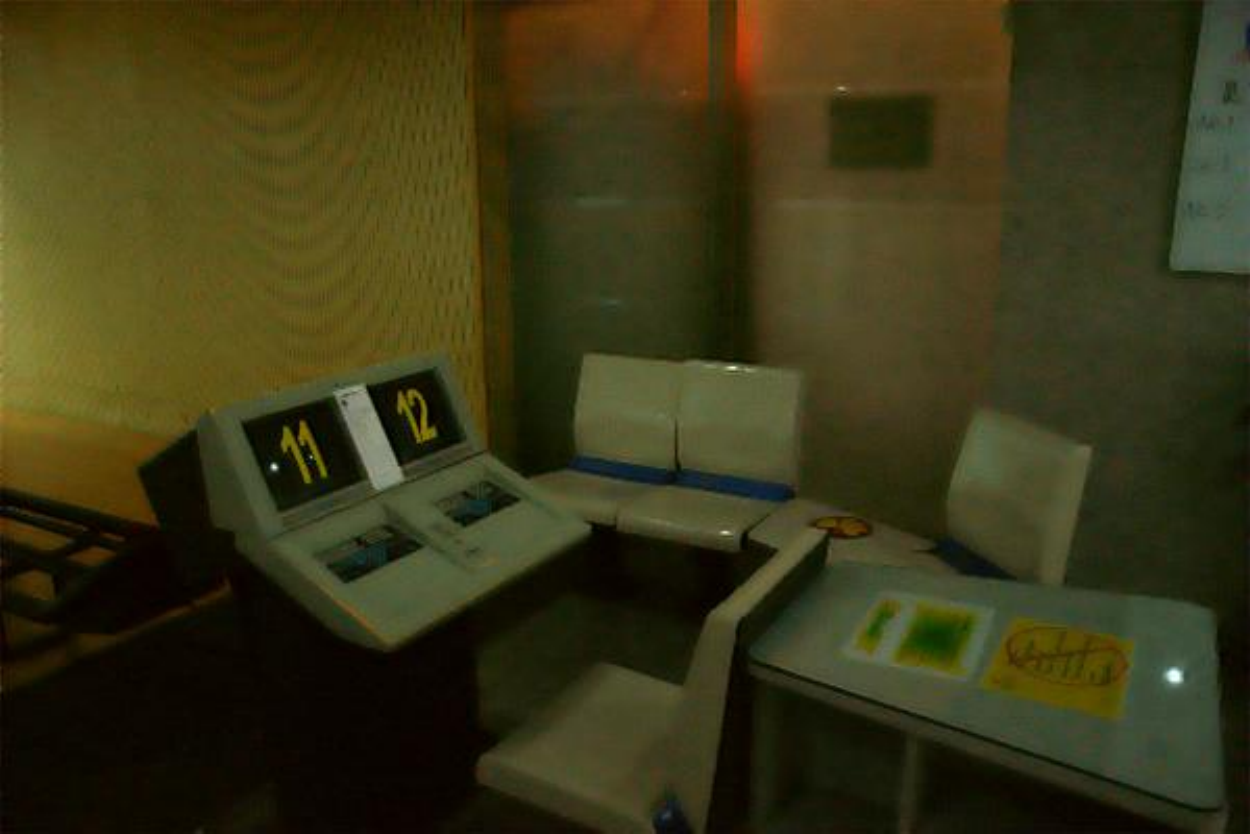}}
 	\centerline{RUAS}
 	\vspace{2pt}
 	\centerline{\includegraphics[width=\textwidth]{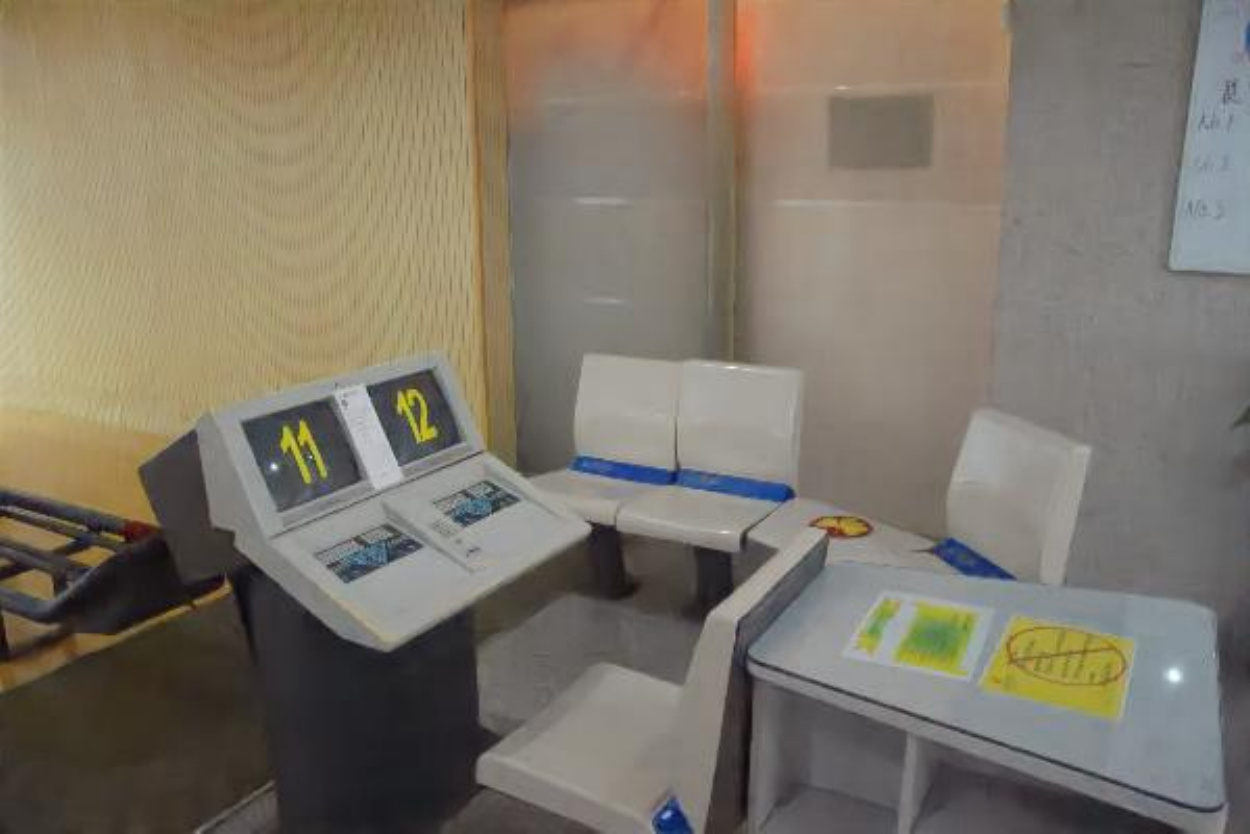}}
 	\centerline{CIDNet}
        \vspace{3pt}
 \end{minipage}
\begin{minipage}{0.245\linewidth}
 	\vspace{3pt}
 	\centerline{\includegraphics[width=\textwidth]{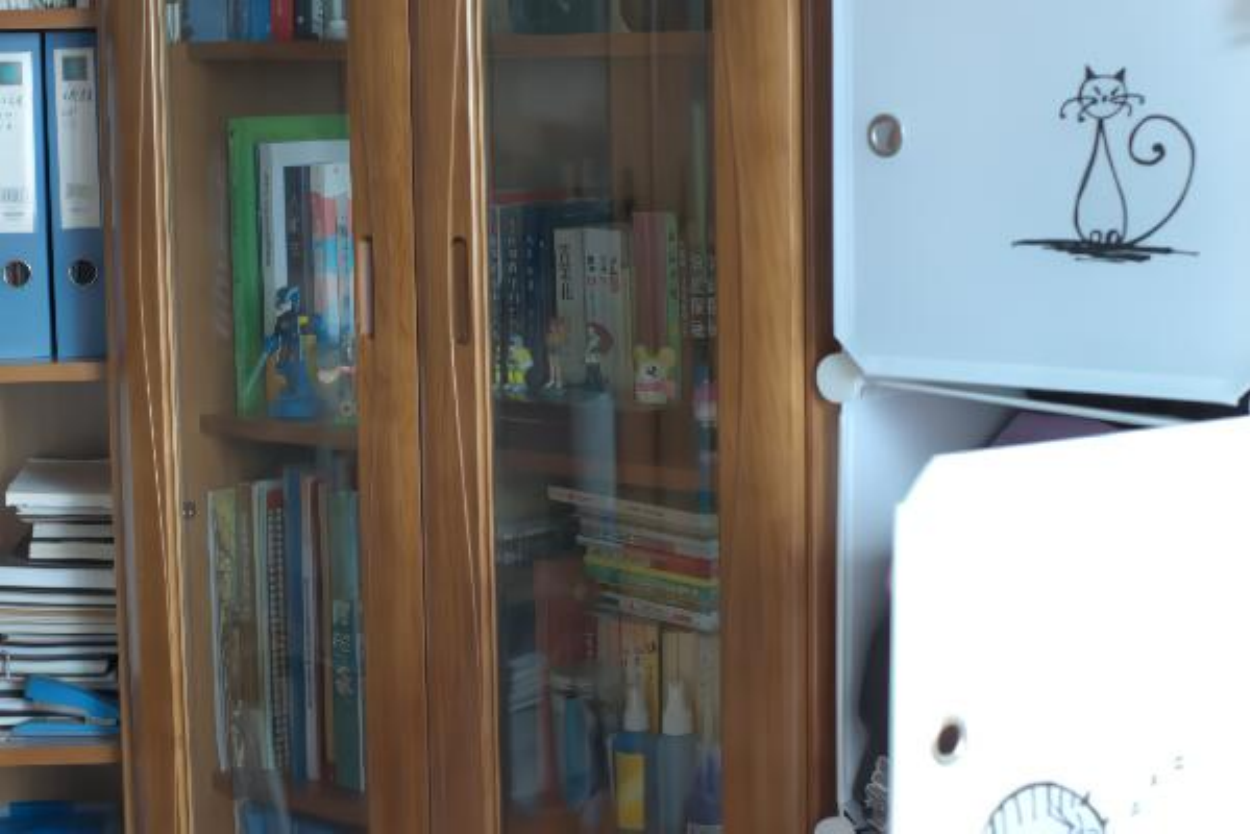}}
 	\centerline{LLFlow}
 	\vspace{2pt}
 	\centerline{\includegraphics[width=\textwidth]{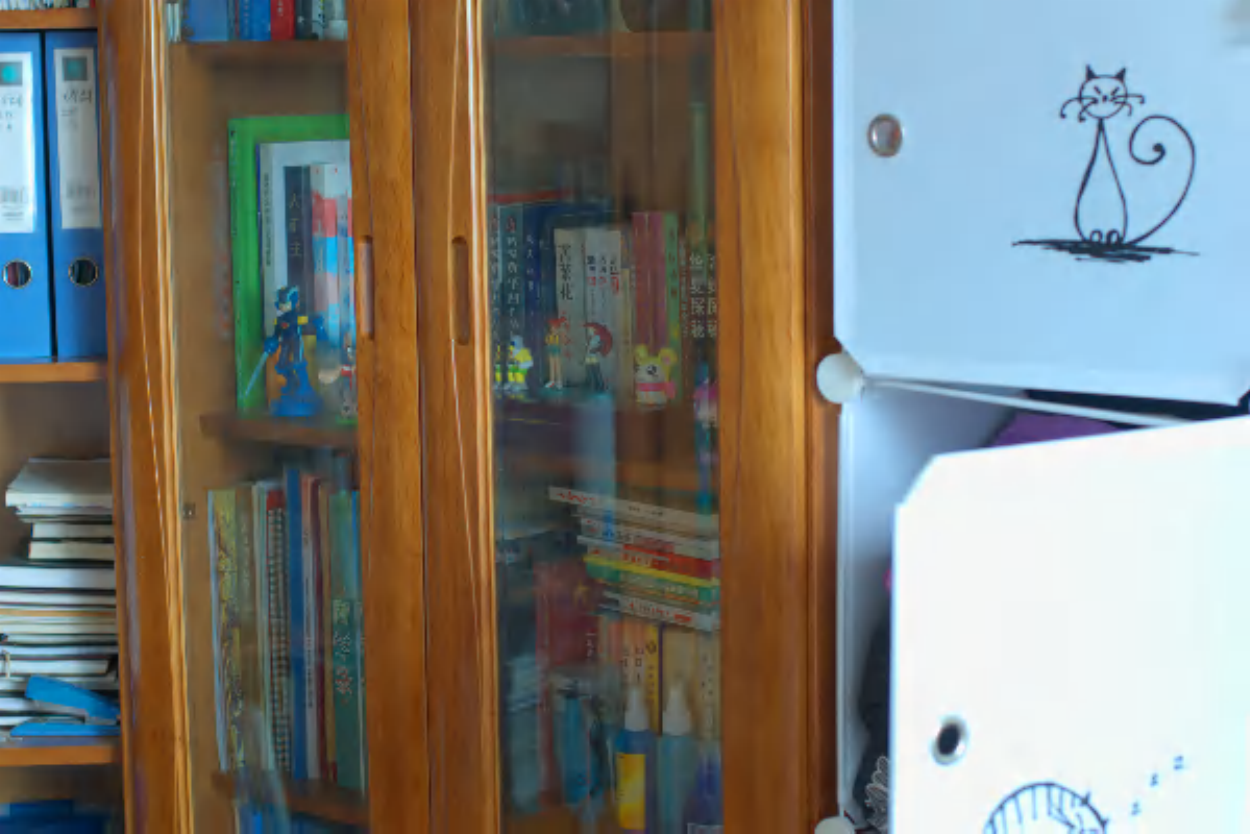}}
 	\centerline{GroundTruth}
 	\vspace{3pt}
 	\centerline{\includegraphics[width=\textwidth]{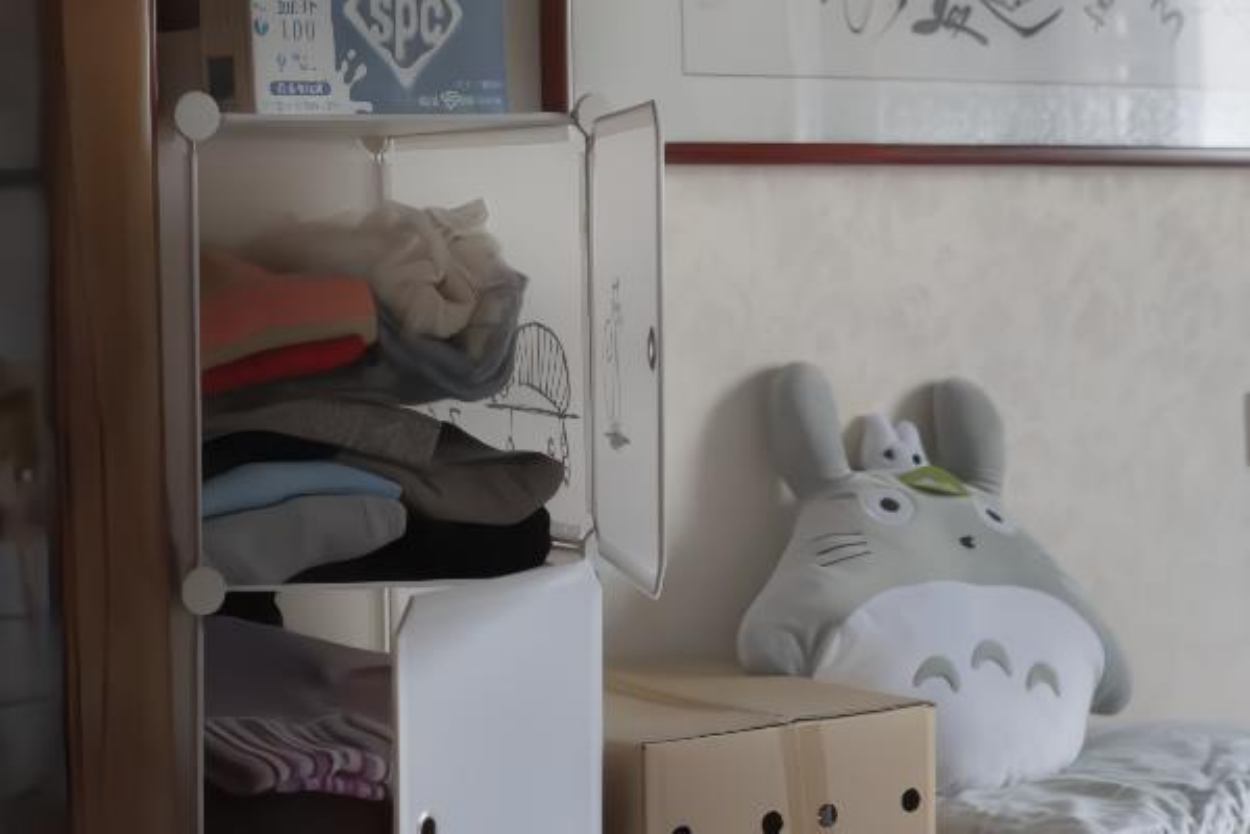}}
 	\centerline{LLFlow}
 	\vspace{2pt}
 	\centerline{\includegraphics[width=\textwidth]{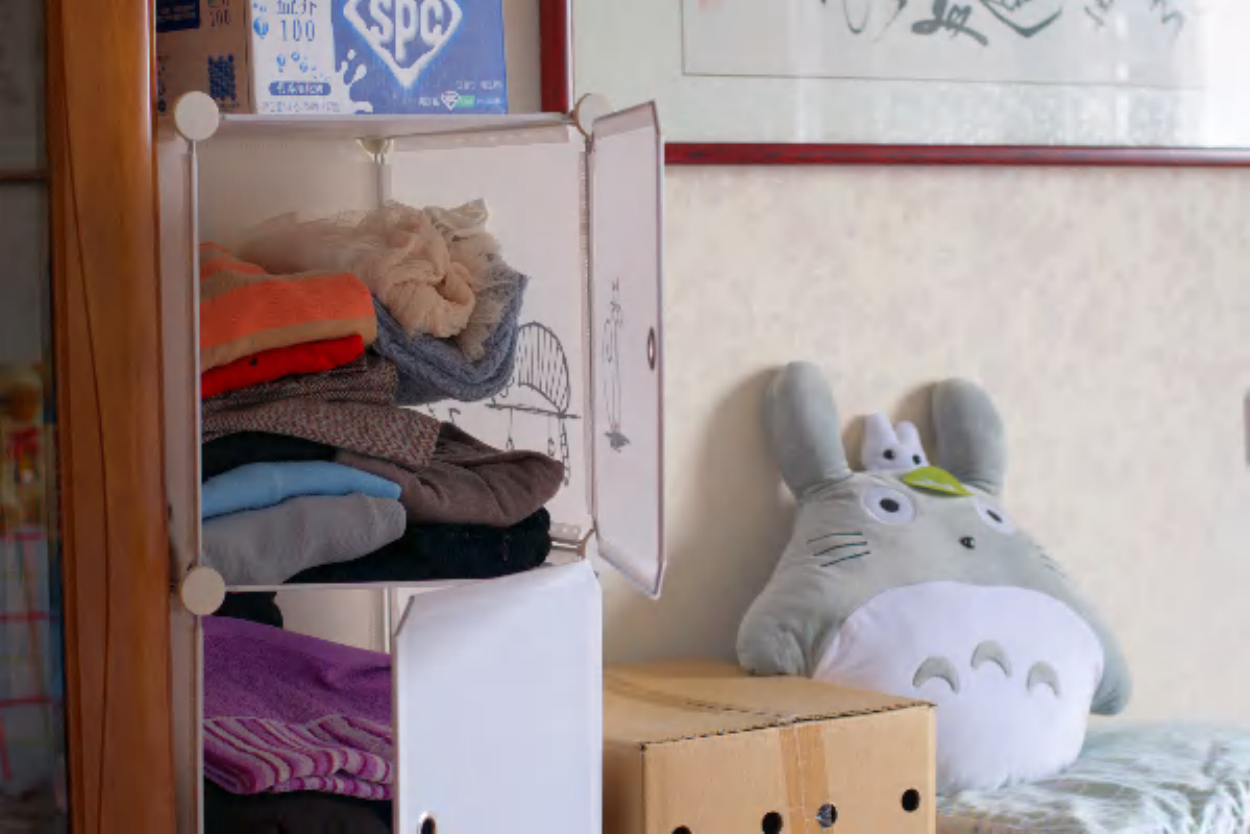}}
 	\centerline{GroundTruth}
 	\vspace{3pt}
 	\centerline{\includegraphics[width=\textwidth]{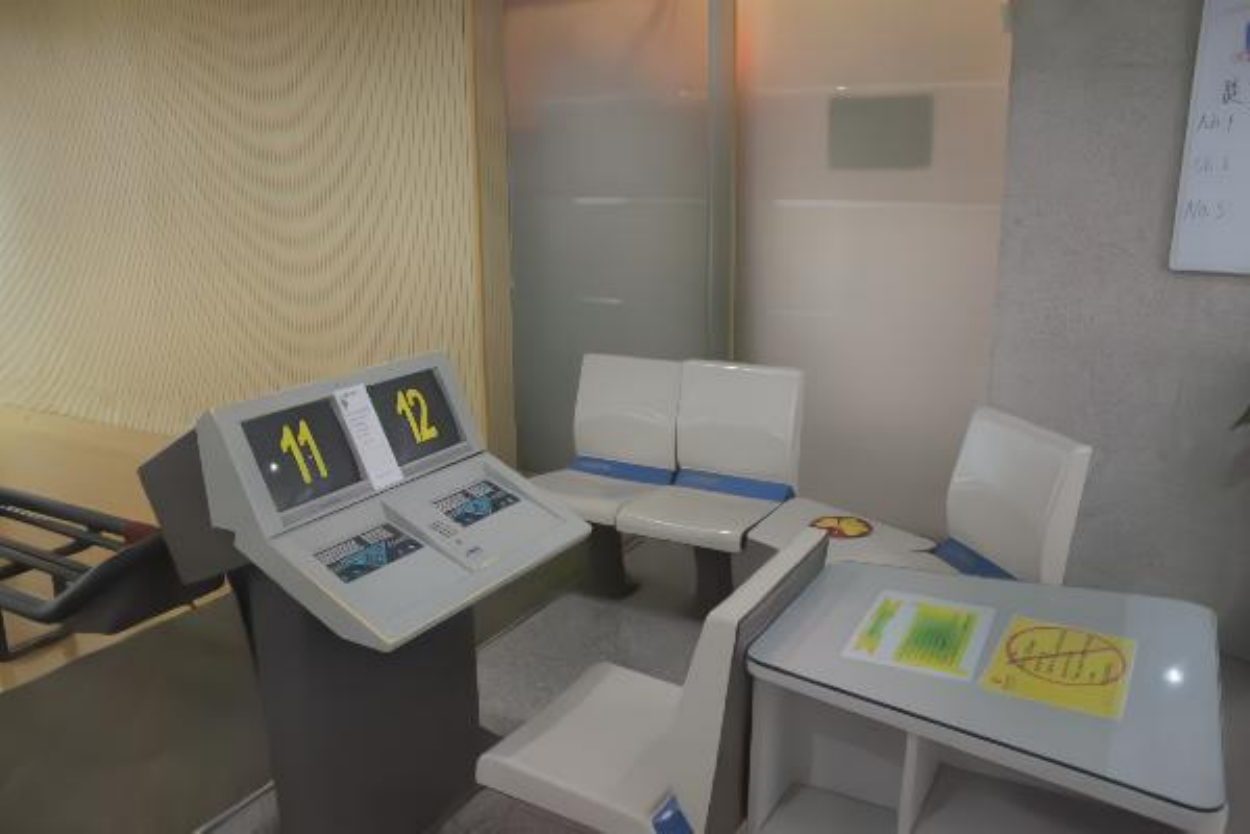}}
 	\centerline{LLFlow}
 	\vspace{2pt}
 	\centerline{\includegraphics[width=\textwidth]{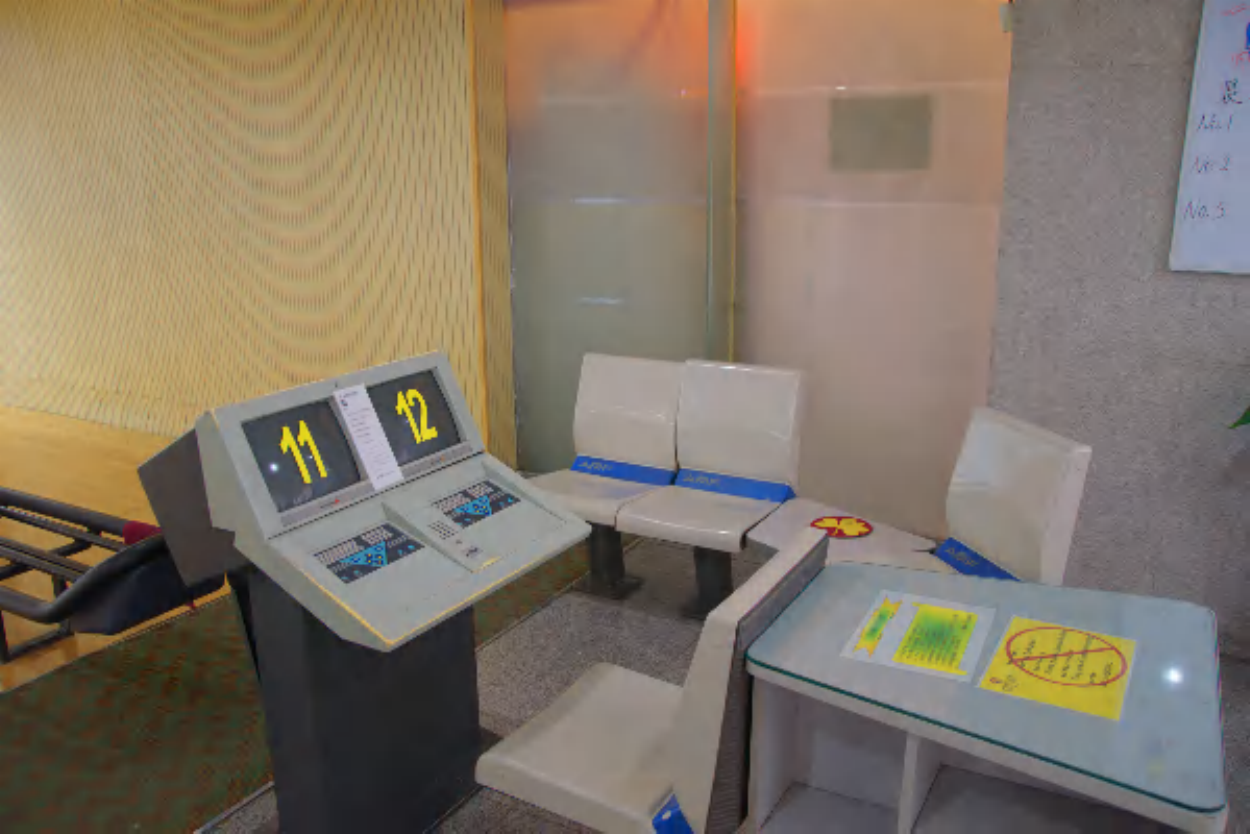}}
 	\centerline{GroundTruth}
        \vspace{3pt}
 \end{minipage}
 \caption{Visual examples for low-light enhancement on LOLv1 dataset \cite{RetinexNet} among RetinexNet \cite{RetinexNet}, RUAS \cite{RUAS}, LLFlow \cite{LLFlow}, SNR-Aware \cite{SNR-Aware}, RetinexFormer \cite{RetinexFormer}, and our CIDNet. Our model clearly removed real low-light noise while enhanced well-light and low color bias.}
 \label{fig:v1}
\end{figure*}

\begin{figure*}
\centering
 \begin{minipage}{0.245\linewidth}
 \centering
        \vspace{3pt}
 	\centerline{\includegraphics[width=\textwidth]{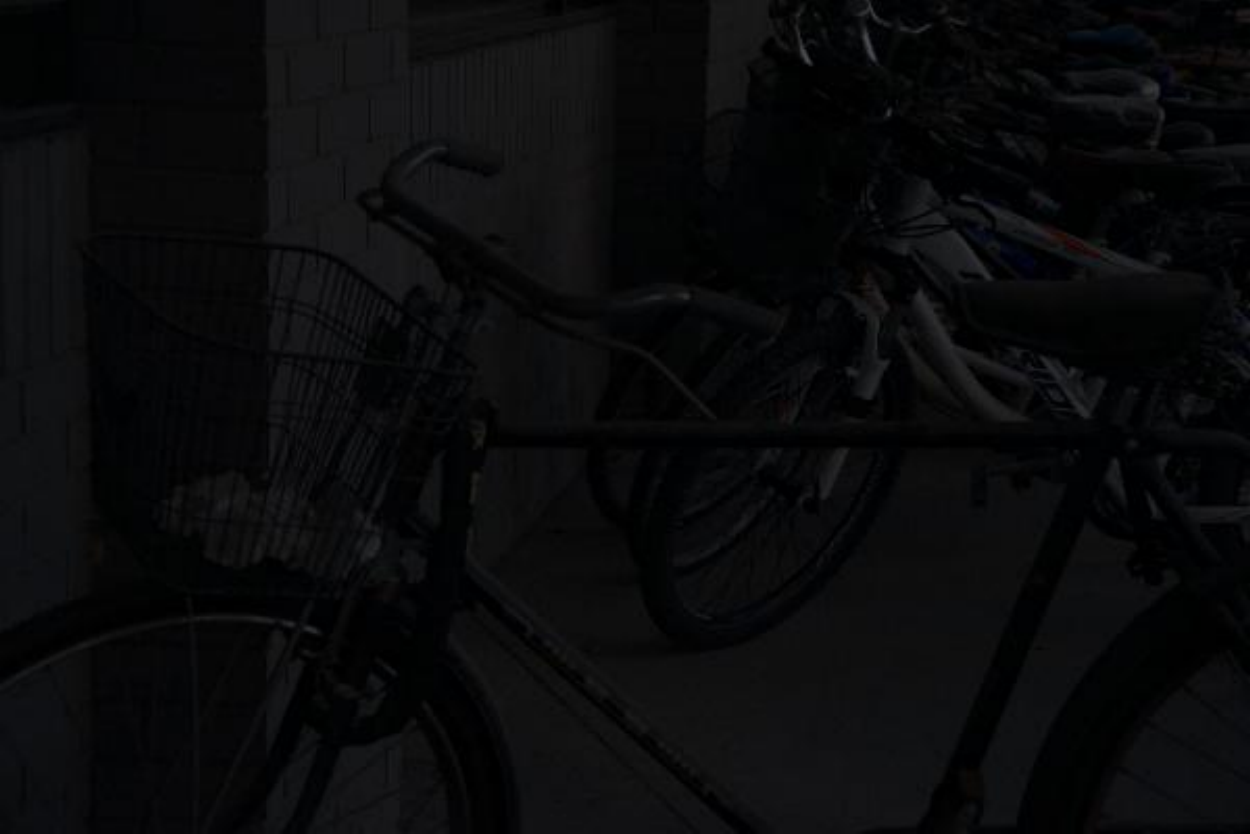}}
 	\centerline{Input}
 	\centerline{\includegraphics[width=\textwidth]{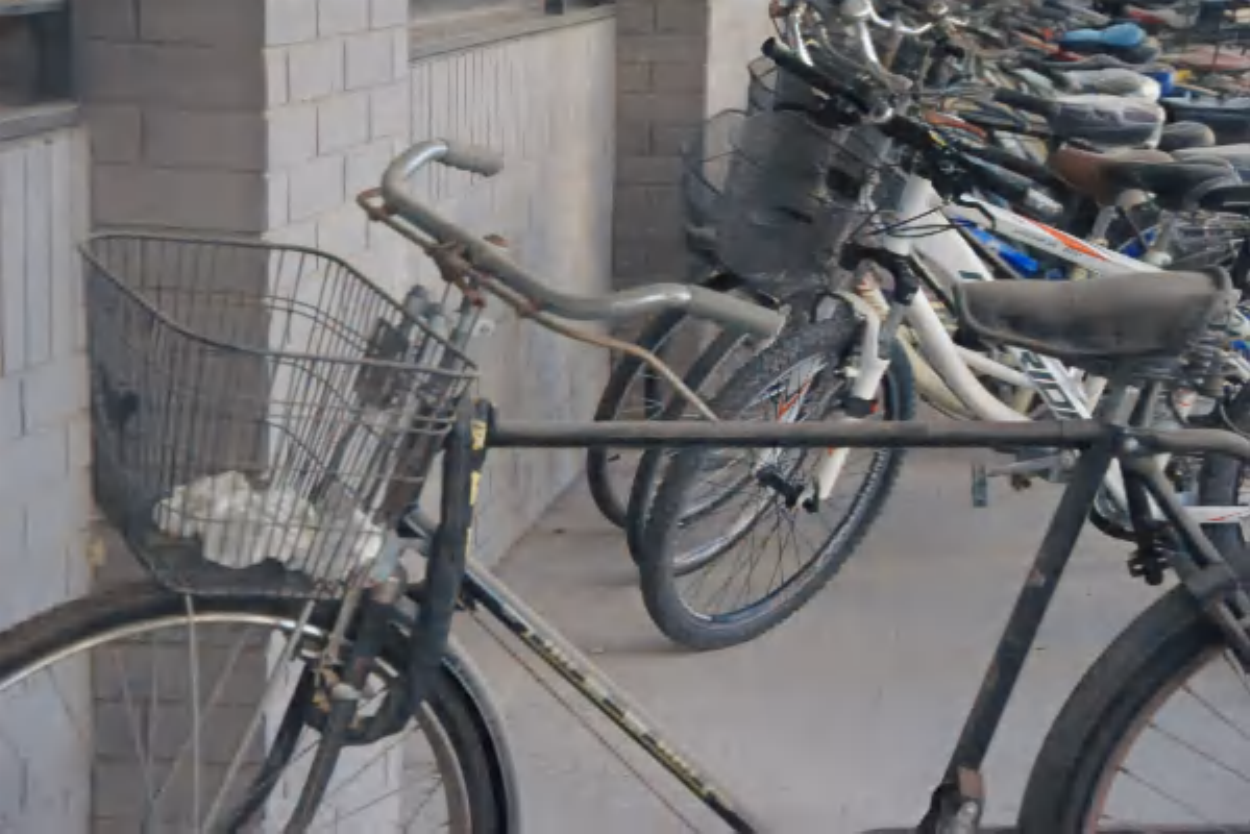}}
 	\centerline{SNR-Aware}
        \vspace{3pt}
 	\centerline{\includegraphics[width=\textwidth]{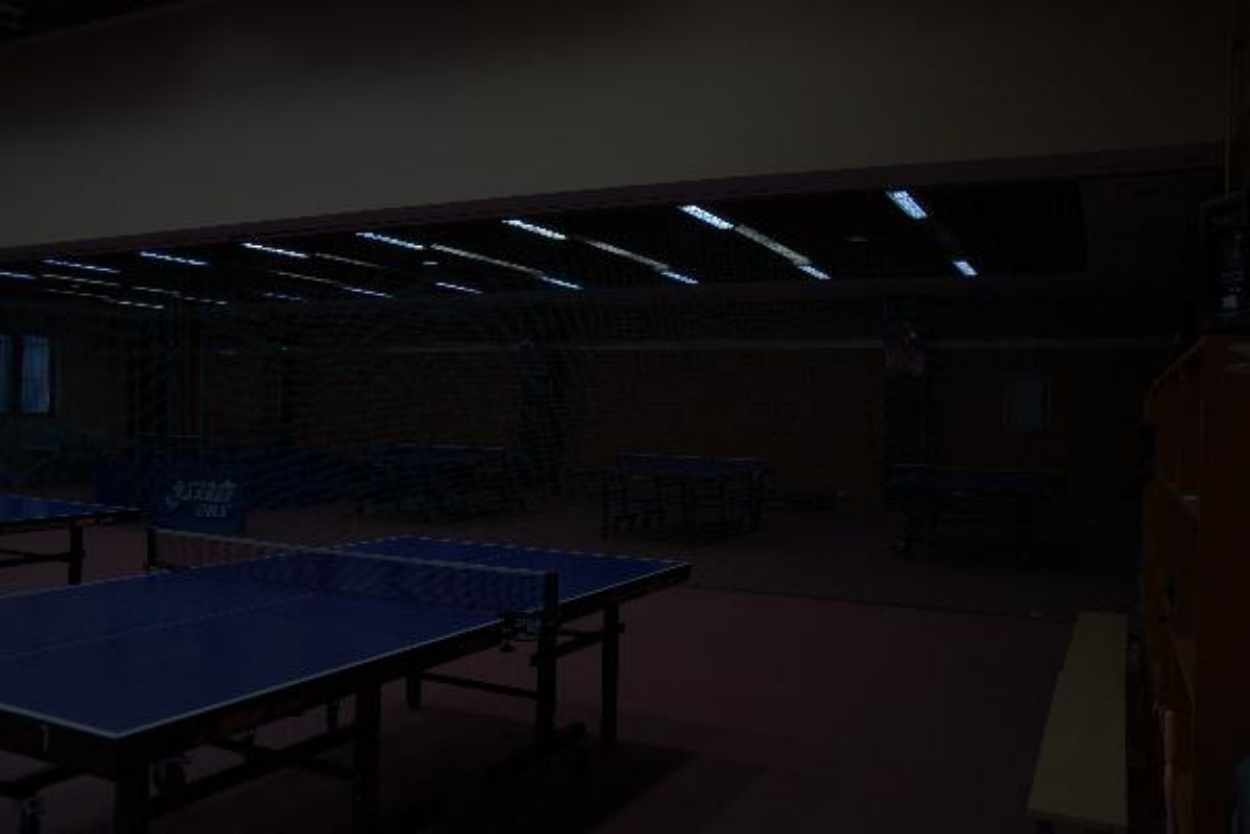}}
 	\centerline{Input}
 	\centerline{\includegraphics[width=\textwidth]{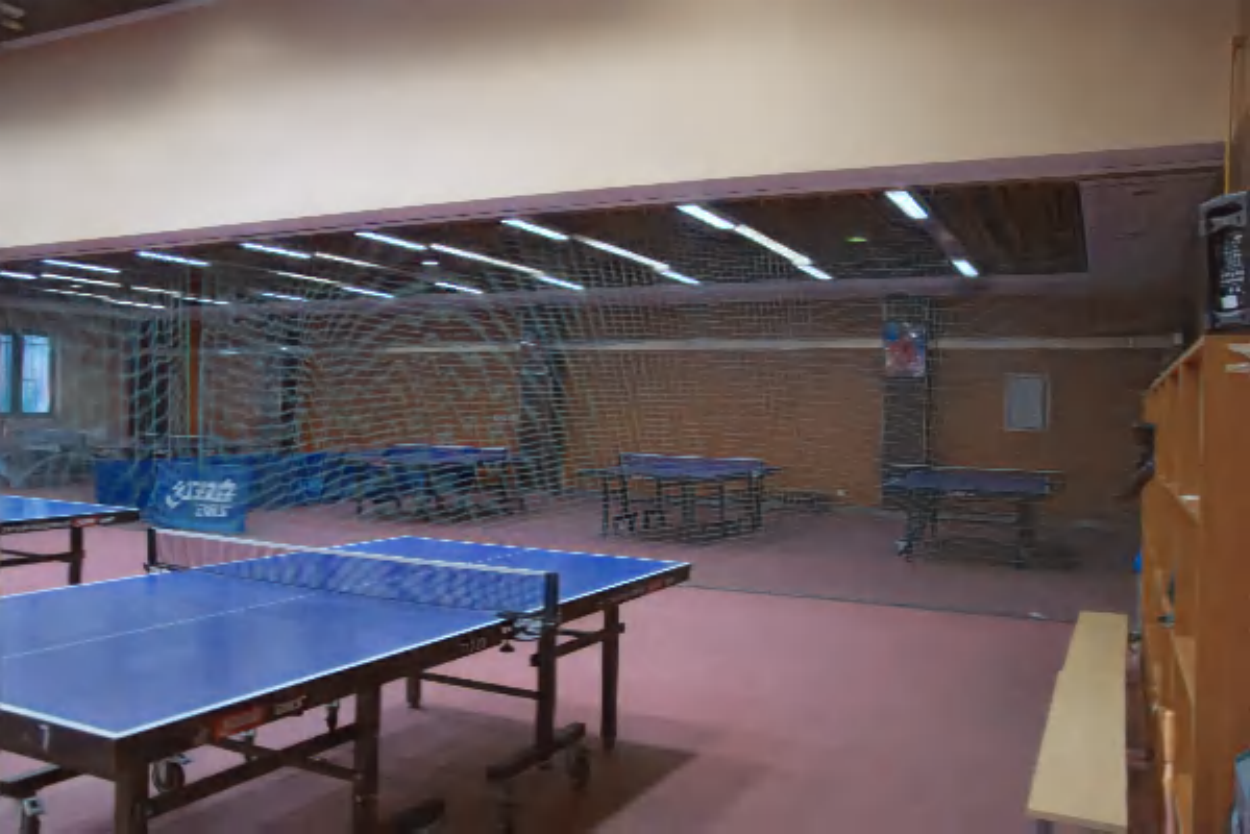}}
 	\centerline{SNR-Aware}
        \vspace{3pt}
 	\centerline{\includegraphics[width=\textwidth]{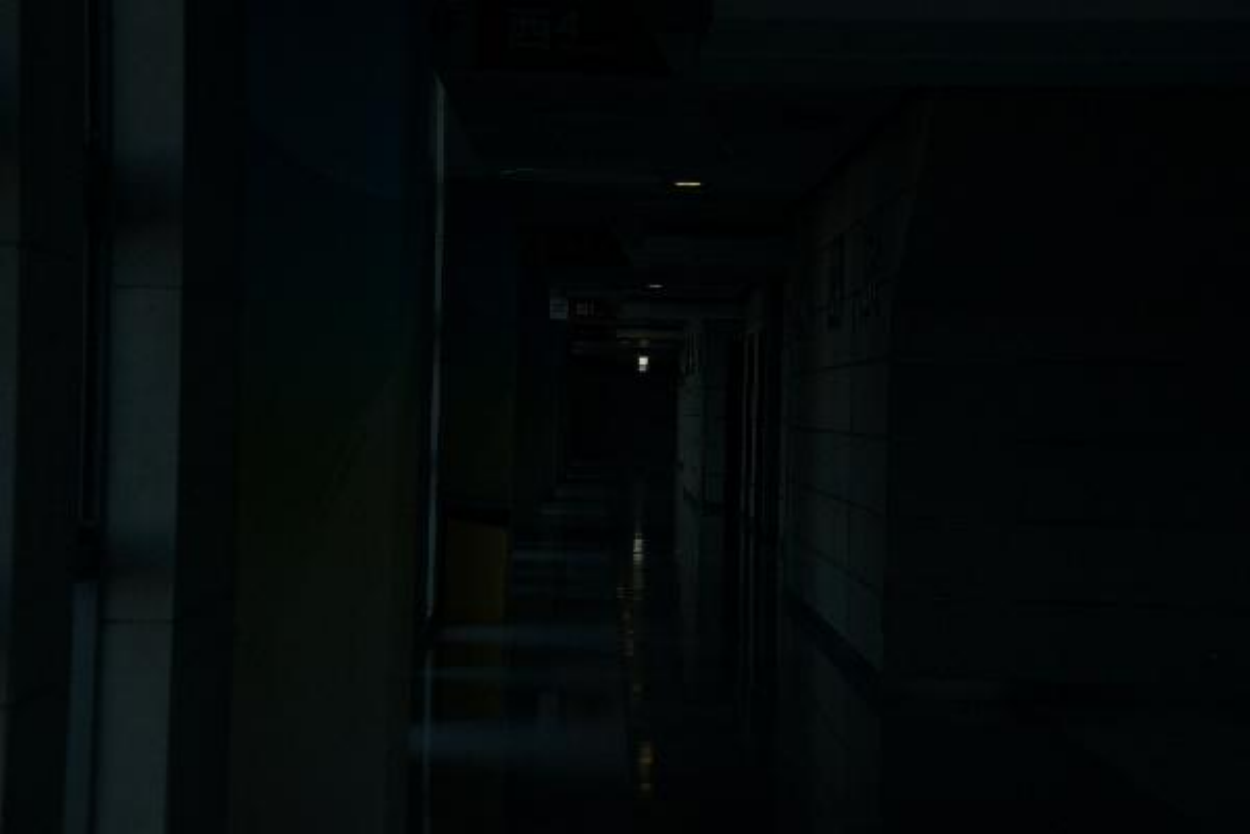}}
 	\centerline{Input}
 	\centerline{\includegraphics[width=\textwidth]{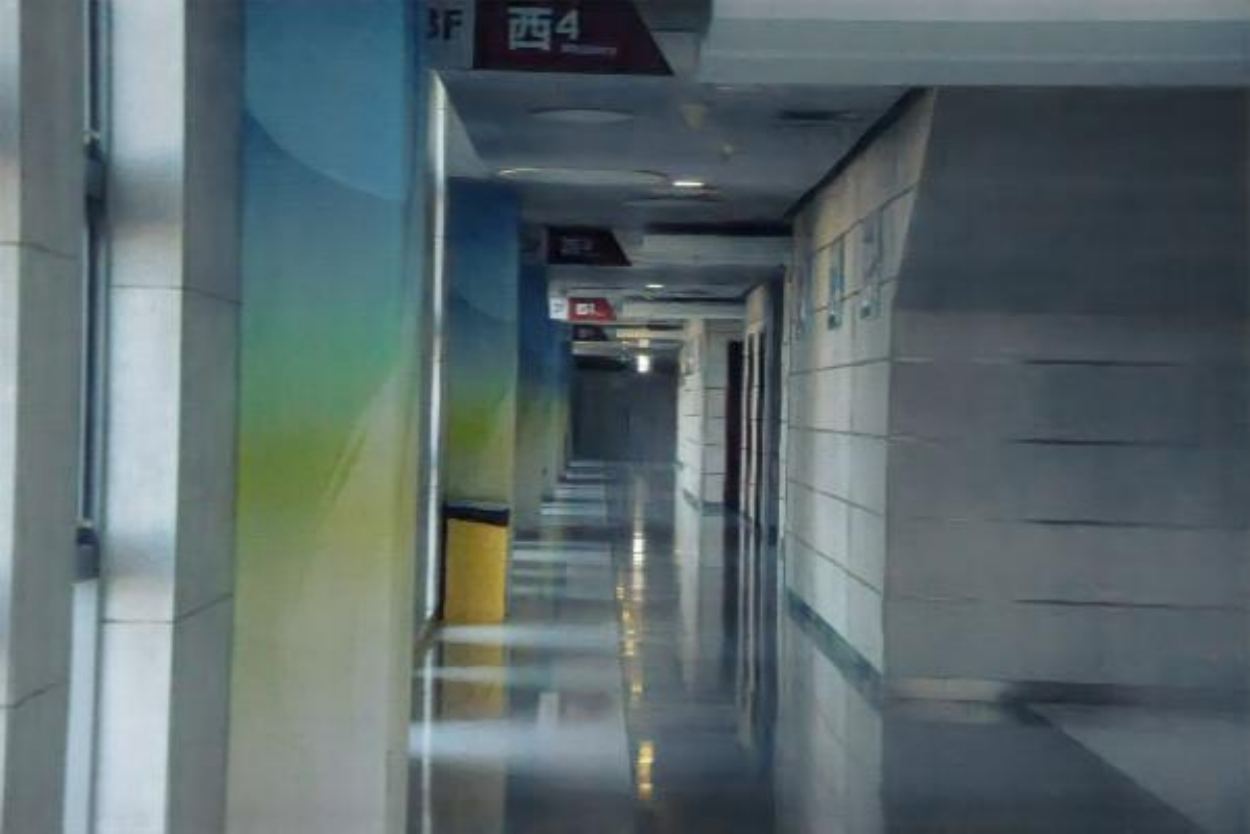}}
 	\centerline{SNR-Aware}
        \vspace{3pt}
\end{minipage}
\begin{minipage}{0.245\linewidth}
        \vspace{3pt}
 	\centerline{\includegraphics[width=\textwidth]{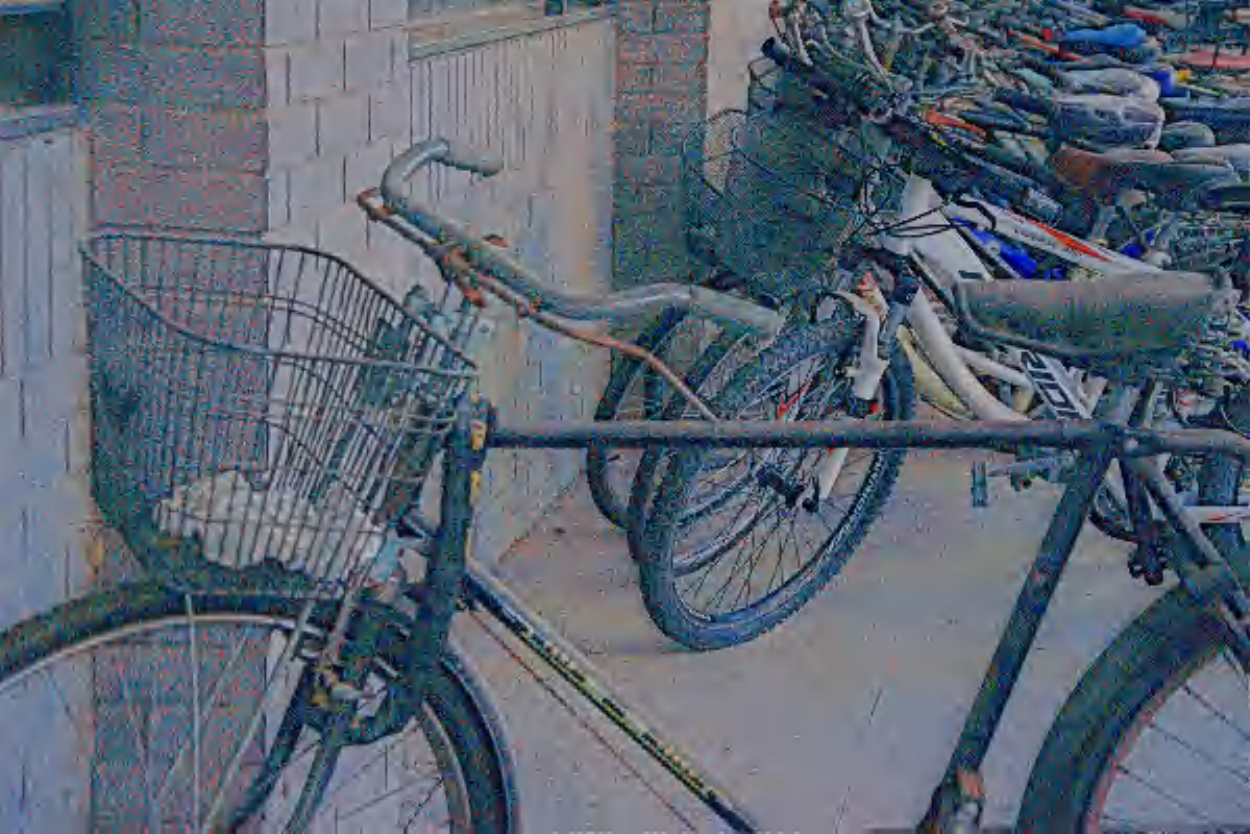}}
 	\centerline{RetinexNet}
        \vspace{2pt}
 	\centerline{\includegraphics[width=\textwidth]{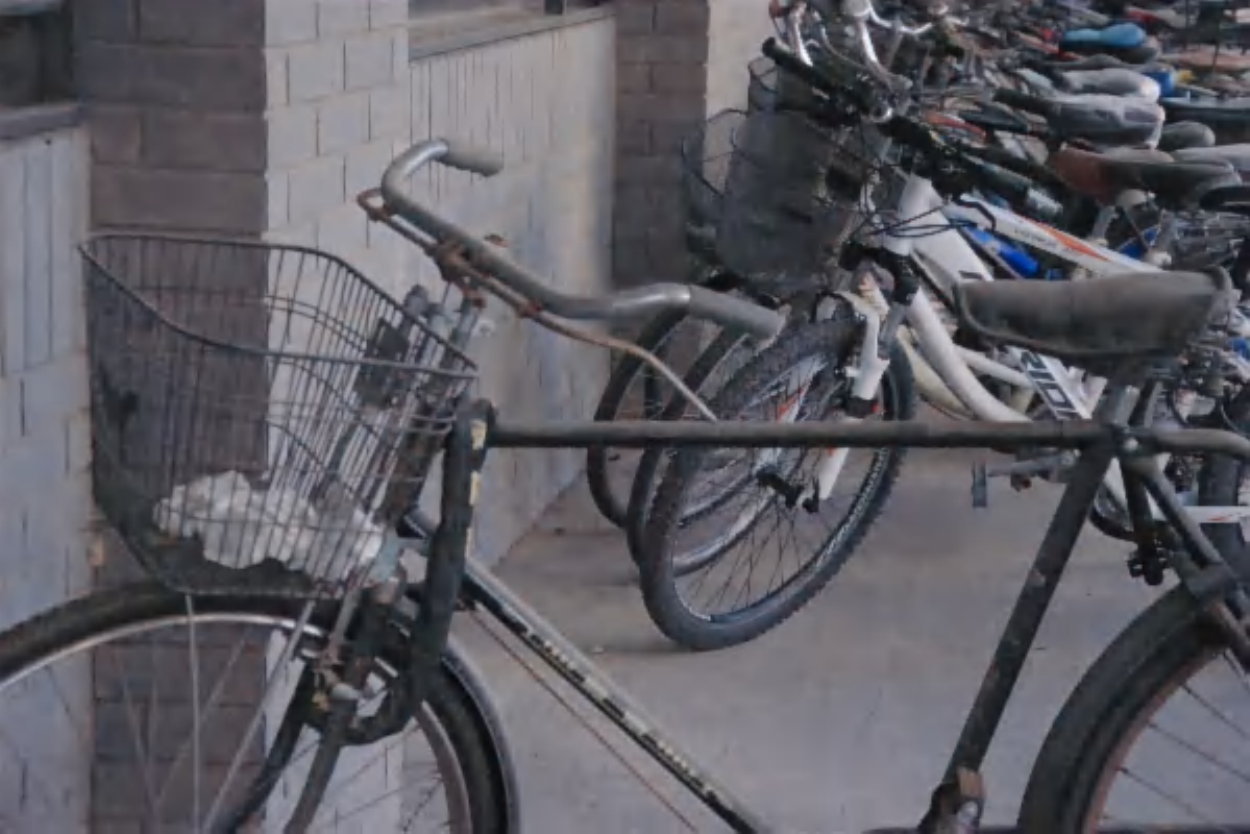}}
 	\centerline{RetinexFormer}
        \vspace{3pt}
 	\centerline{\includegraphics[width=\textwidth]{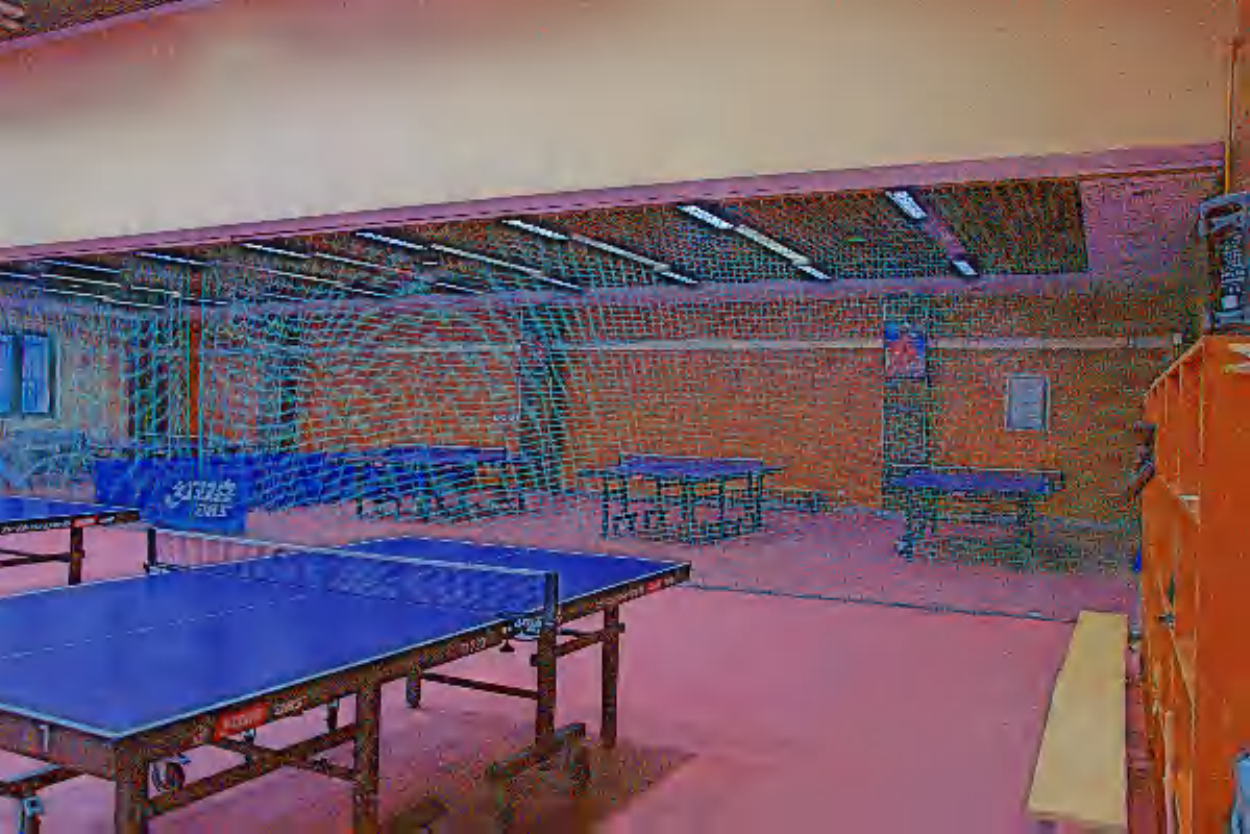}}
 	\centerline{RetinexNet}
        \vspace{2pt}
 	\centerline{\includegraphics[width=\textwidth]{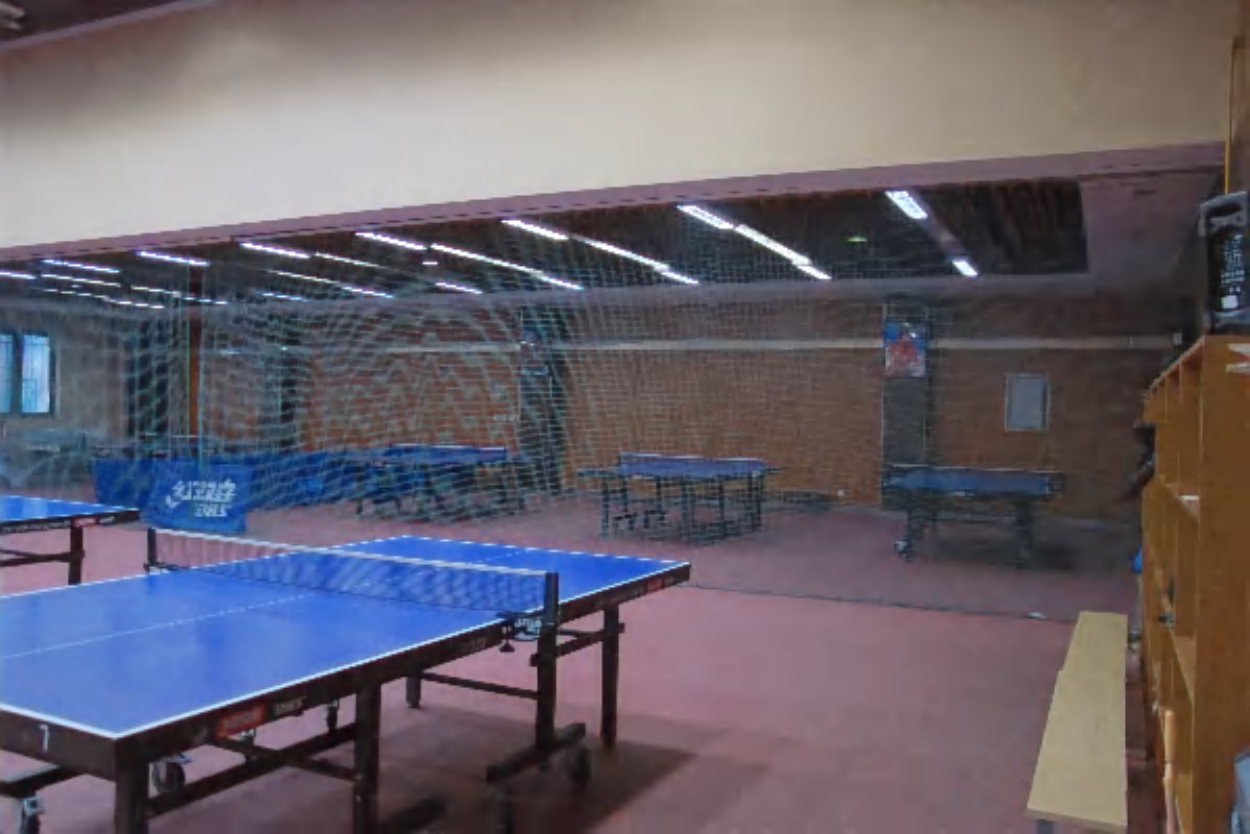}}
 	\centerline{RetinexFormer}
        \vspace{3pt}
 	\centerline{\includegraphics[width=\textwidth]{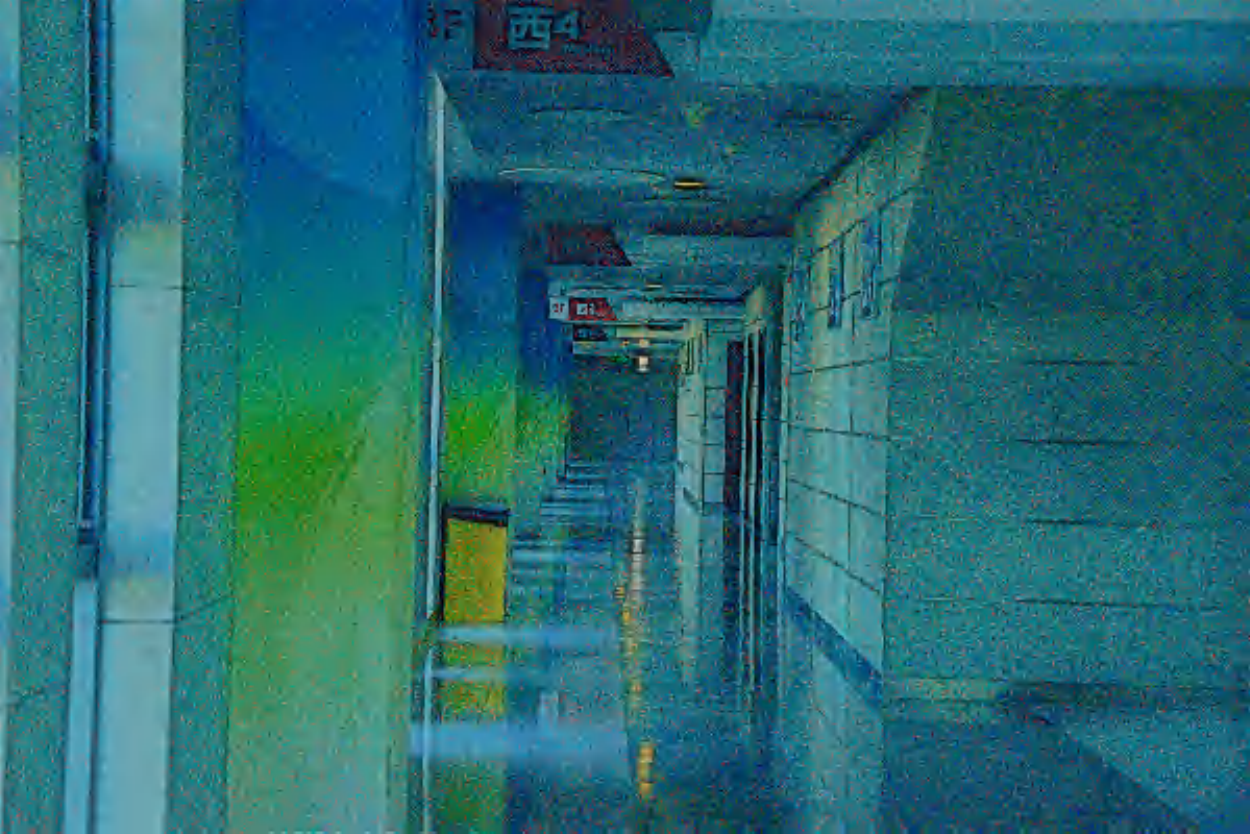}}
 	\centerline{RetinexNet}
        \vspace{2pt}
 	\centerline{\includegraphics[width=\textwidth]{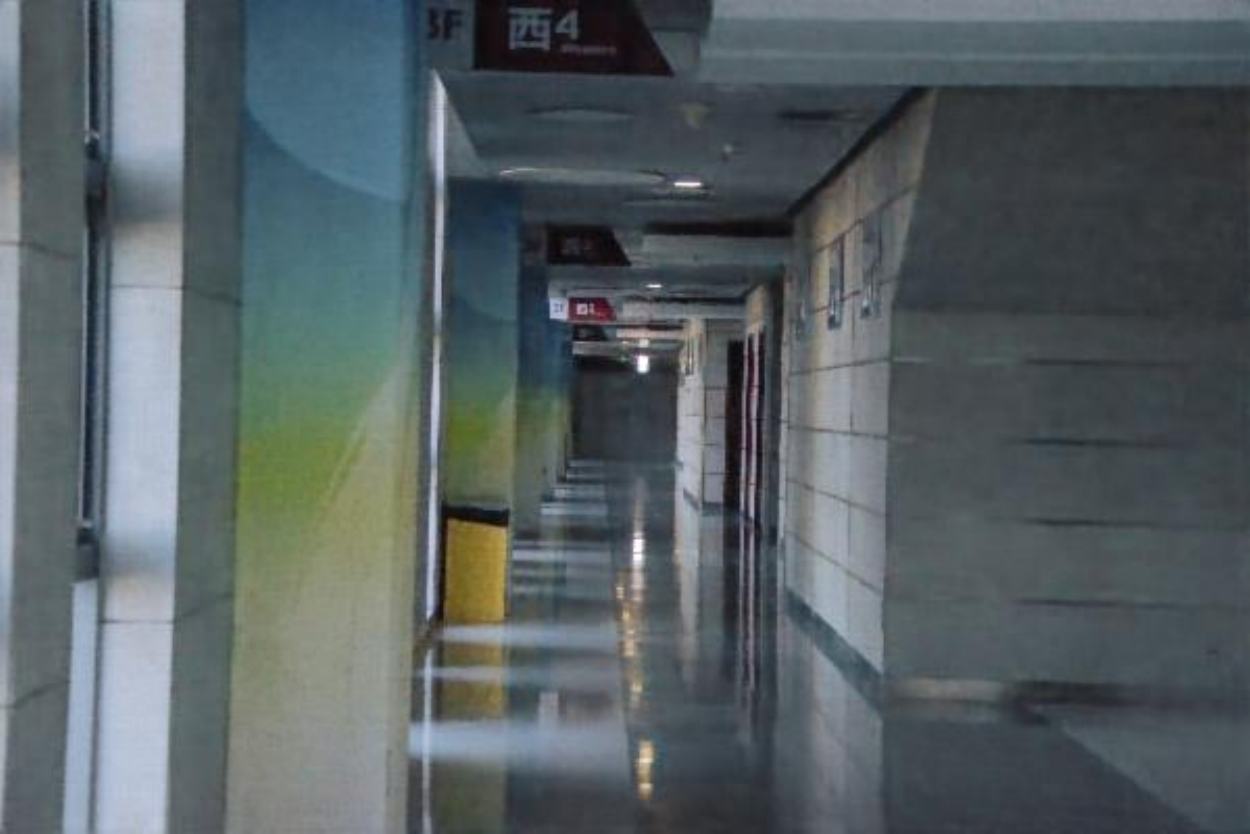}}
 	\centerline{RetinexFormer}
        \vspace{3pt}
 \end{minipage}
\begin{minipage}{0.245\linewidth}
 	\vspace{3pt}
 	\centerline{\includegraphics[width=\textwidth]{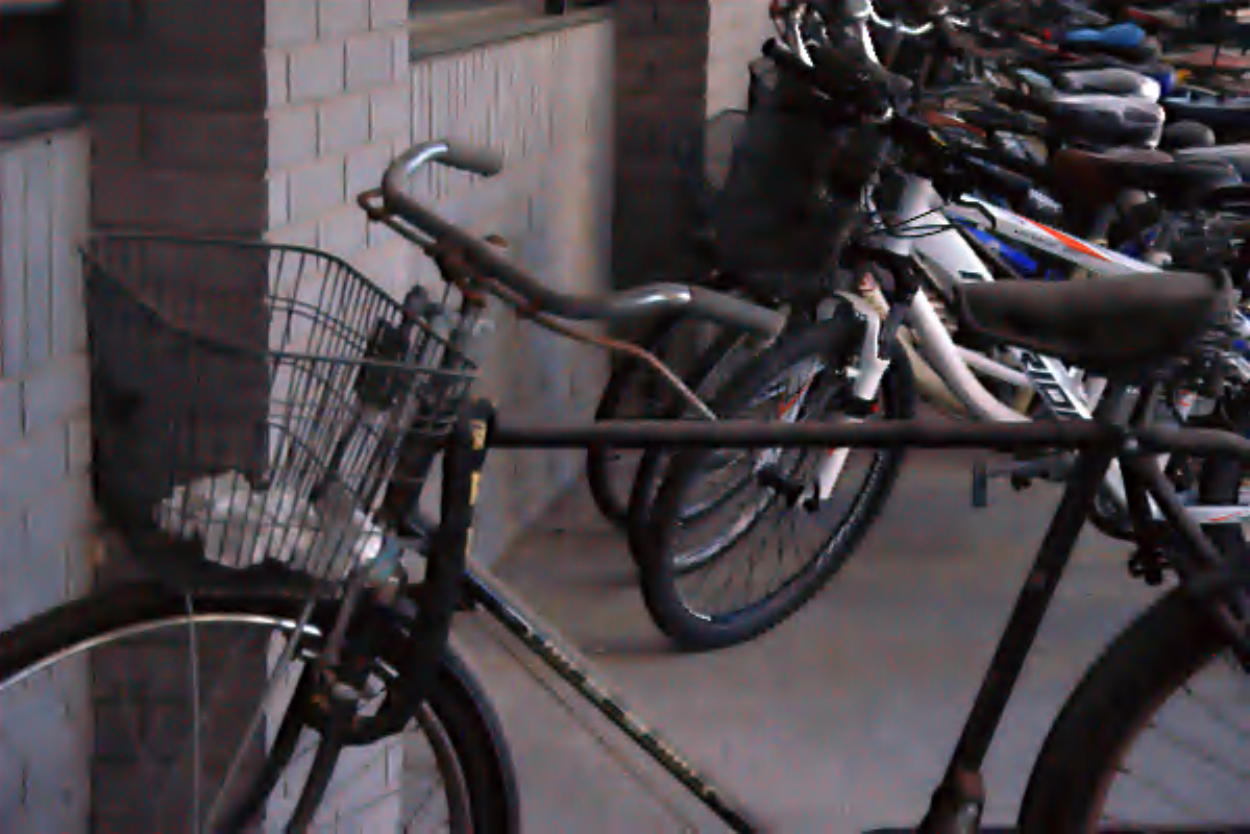}}
 	\centerline{RUAS}
 	\vspace{2pt}
 	\centerline{\includegraphics[width=\textwidth]{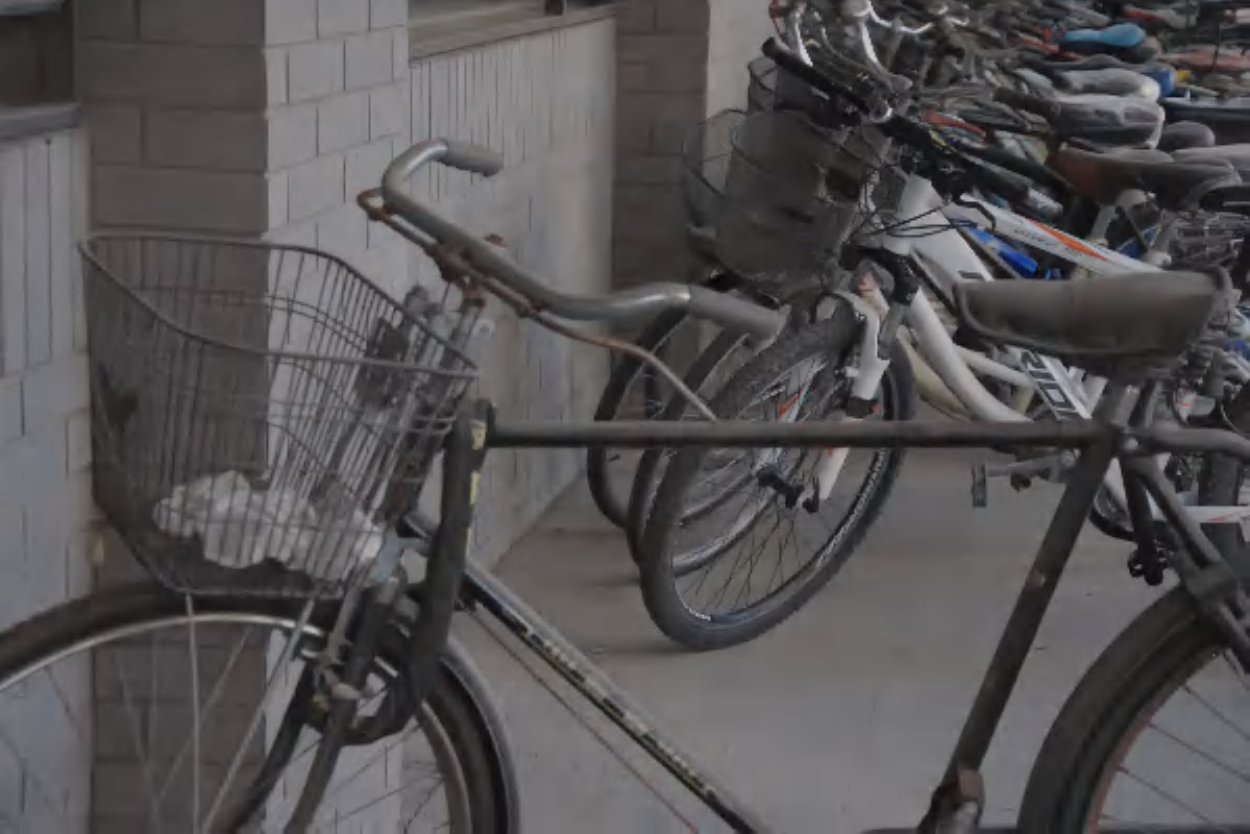}}
 	\centerline{CIDNet}
 	\vspace{3pt}
 	\centerline{\includegraphics[width=\textwidth]{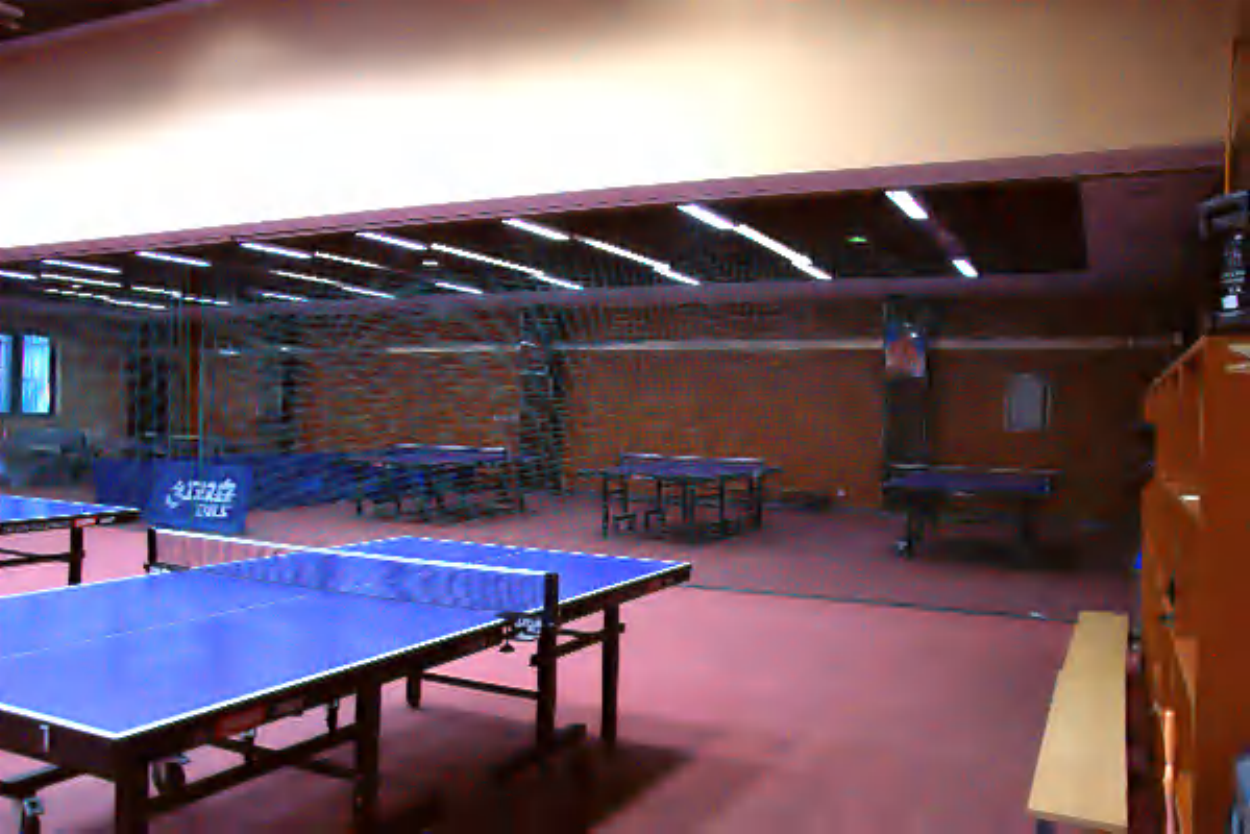}}
 	\centerline{RUAS}
 	\vspace{2pt}
 	\centerline{\includegraphics[width=\textwidth]{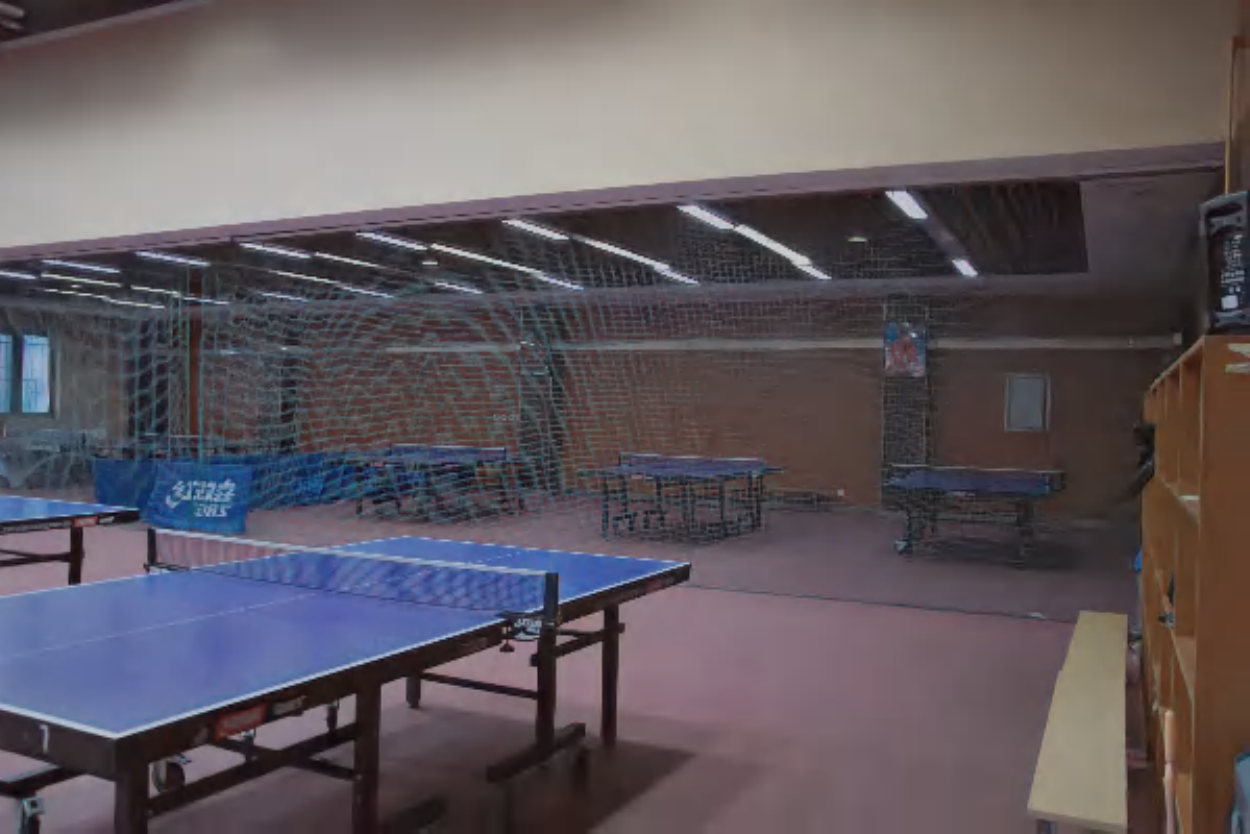}}
 	\centerline{CIDNet}
 	\vspace{3pt}
 	\centerline{\includegraphics[width=\textwidth]{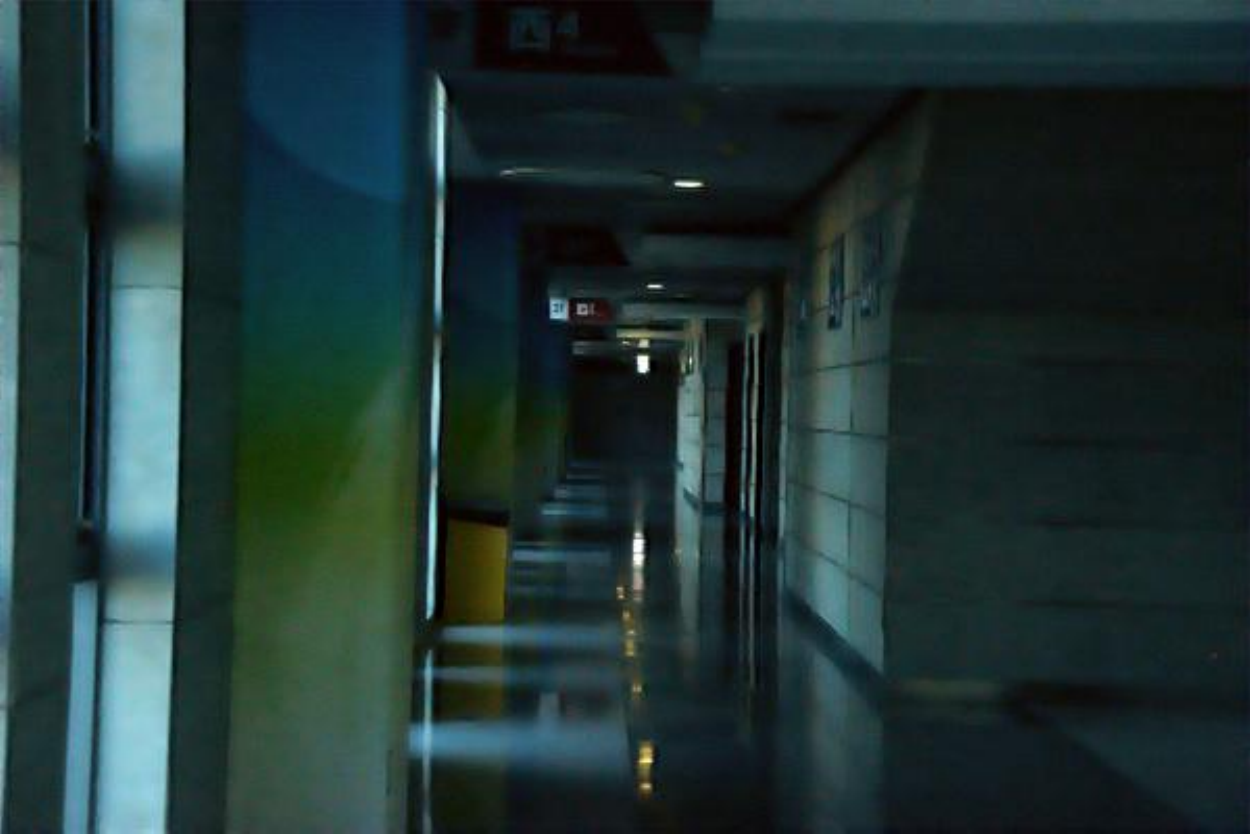}}
 	\centerline{RUAS}
 	\vspace{2pt}
 	\centerline{\includegraphics[width=\textwidth]{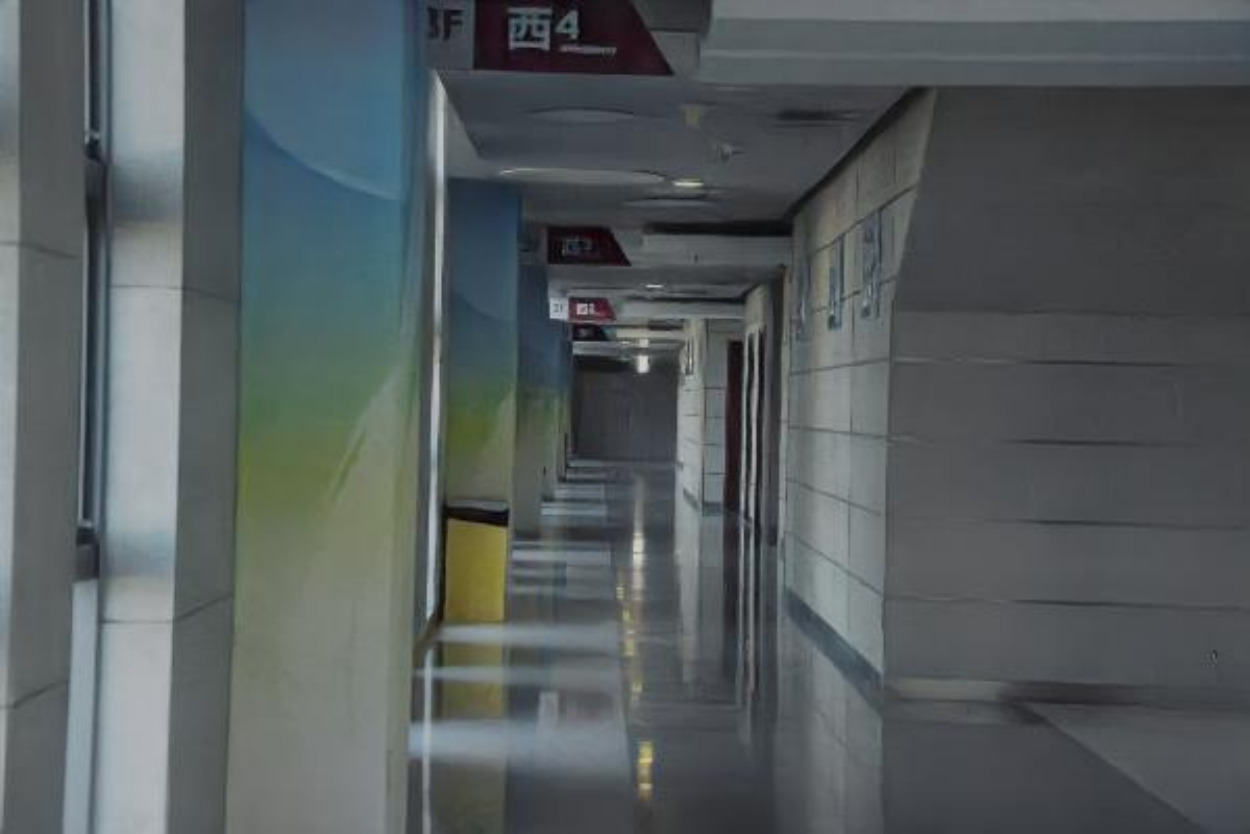}}
 	\centerline{CIDNet}
        \vspace{3pt}
 \end{minipage}
\begin{minipage}{0.245\linewidth}
 	\vspace{3pt}
 	\centerline{\includegraphics[width=\textwidth]{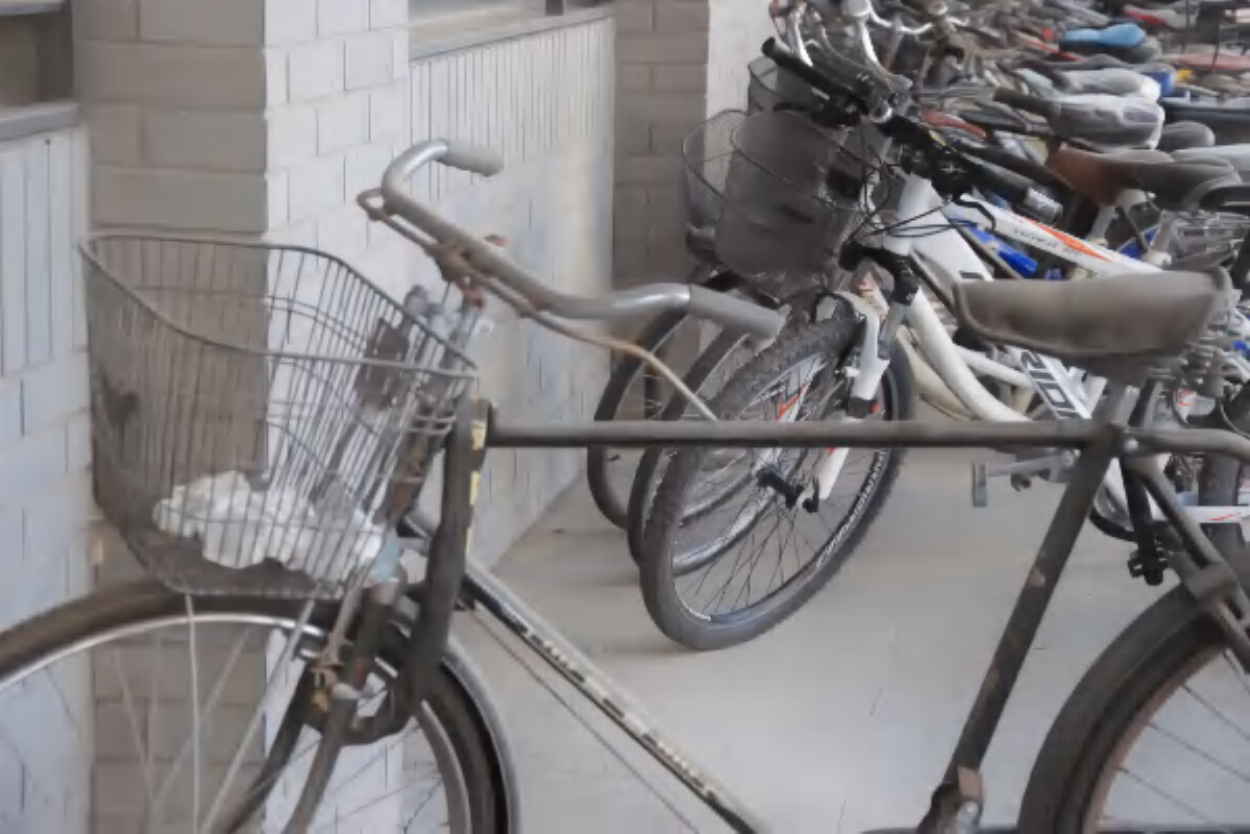}}
 	\centerline{LLFlow}
 	\vspace{2pt}
 	\centerline{\includegraphics[width=\textwidth]{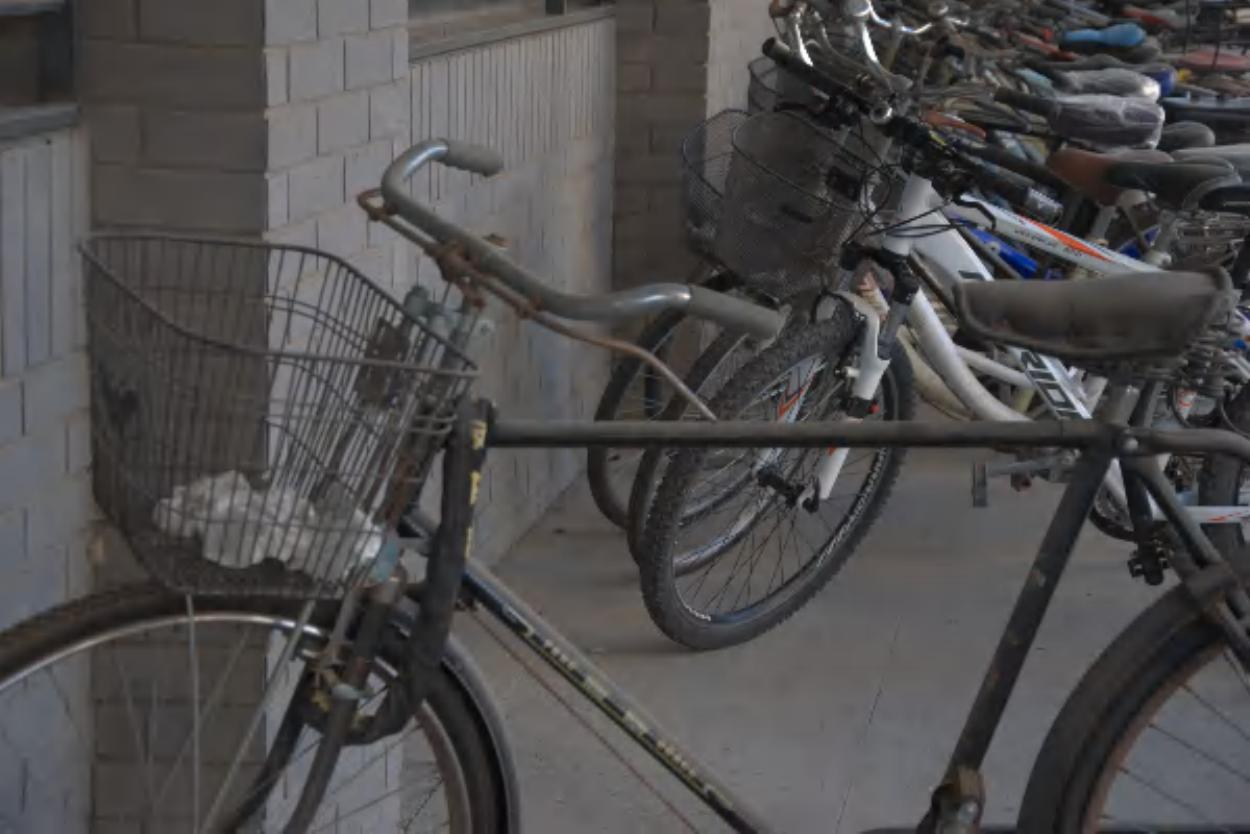}}
 	\centerline{GroundTruth}
 	\vspace{3pt}
 	\centerline{\includegraphics[width=\textwidth]{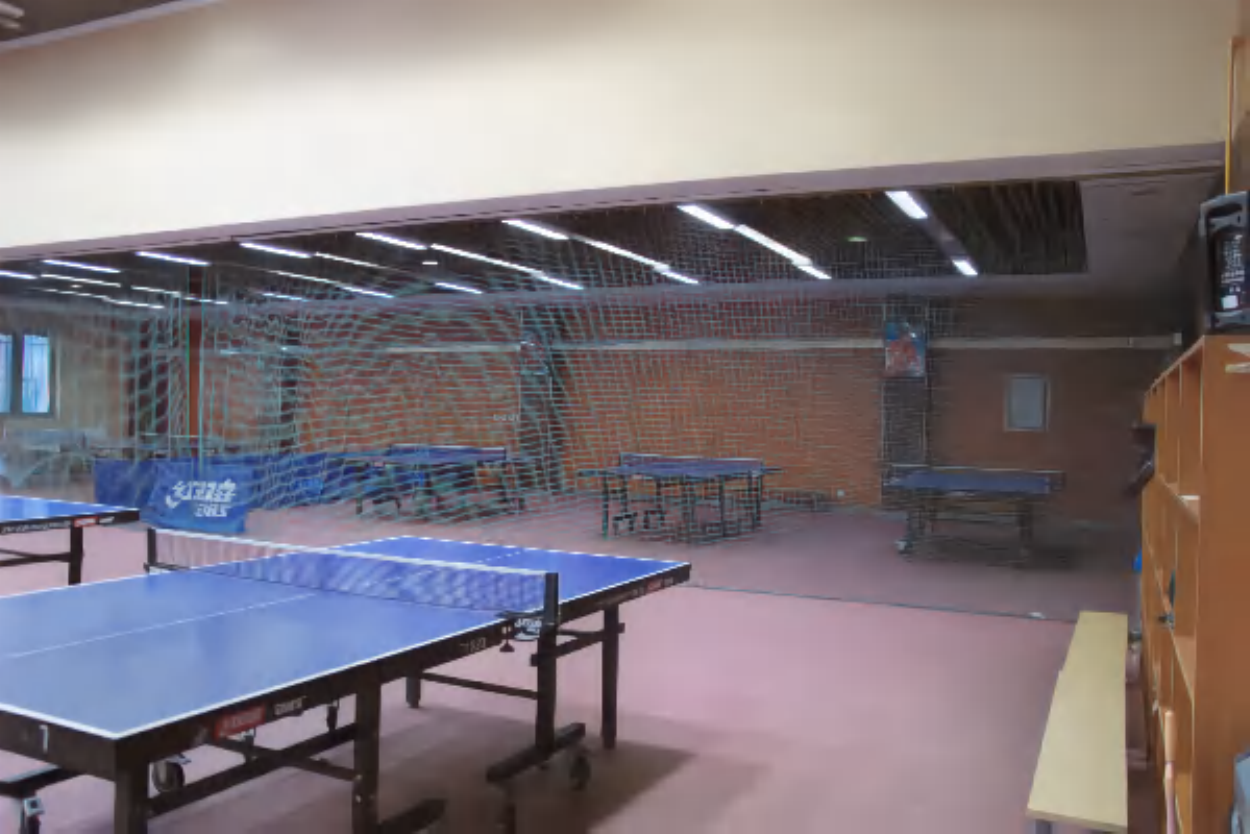}}
 	\centerline{LLFlow}
 	\vspace{2pt}
 	\centerline{\includegraphics[width=\textwidth]{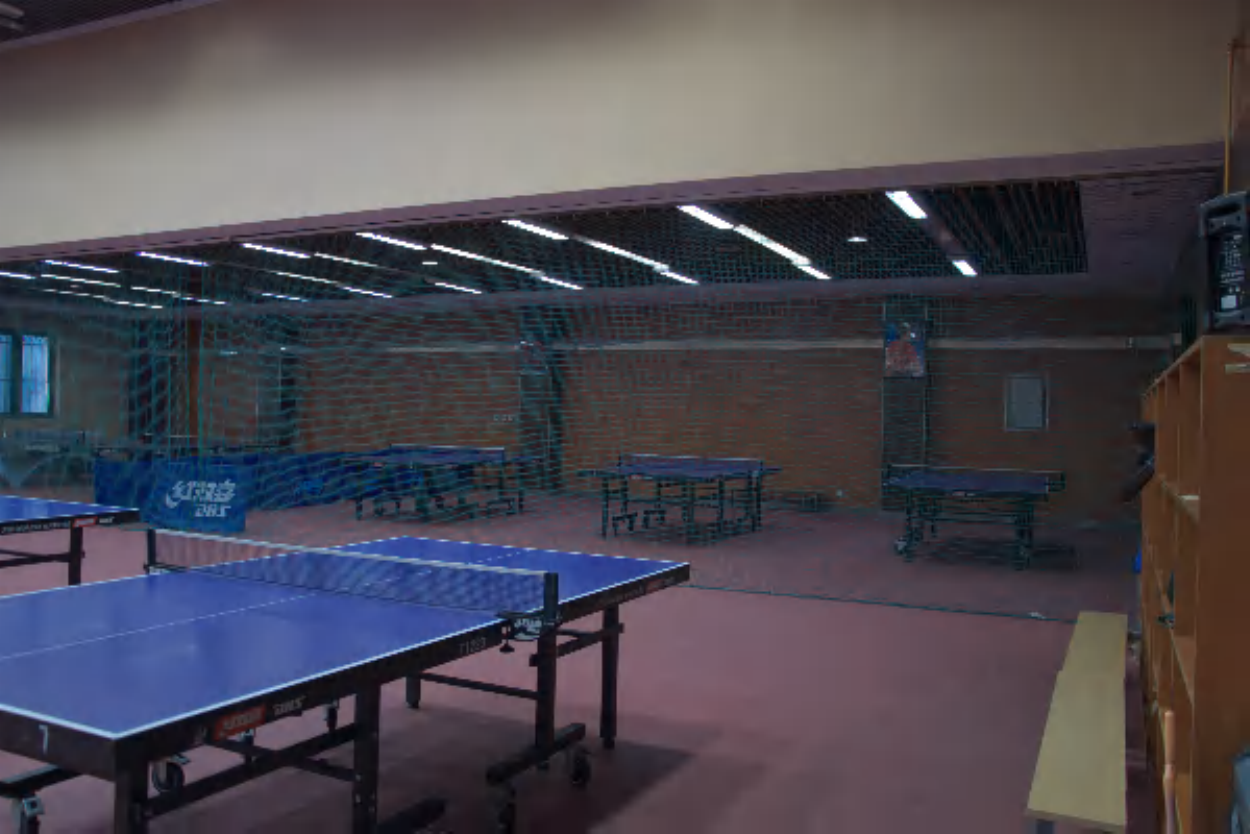}}
 	\centerline{GroundTruth}
 	\vspace{3pt}
 	\centerline{\includegraphics[width=\textwidth]{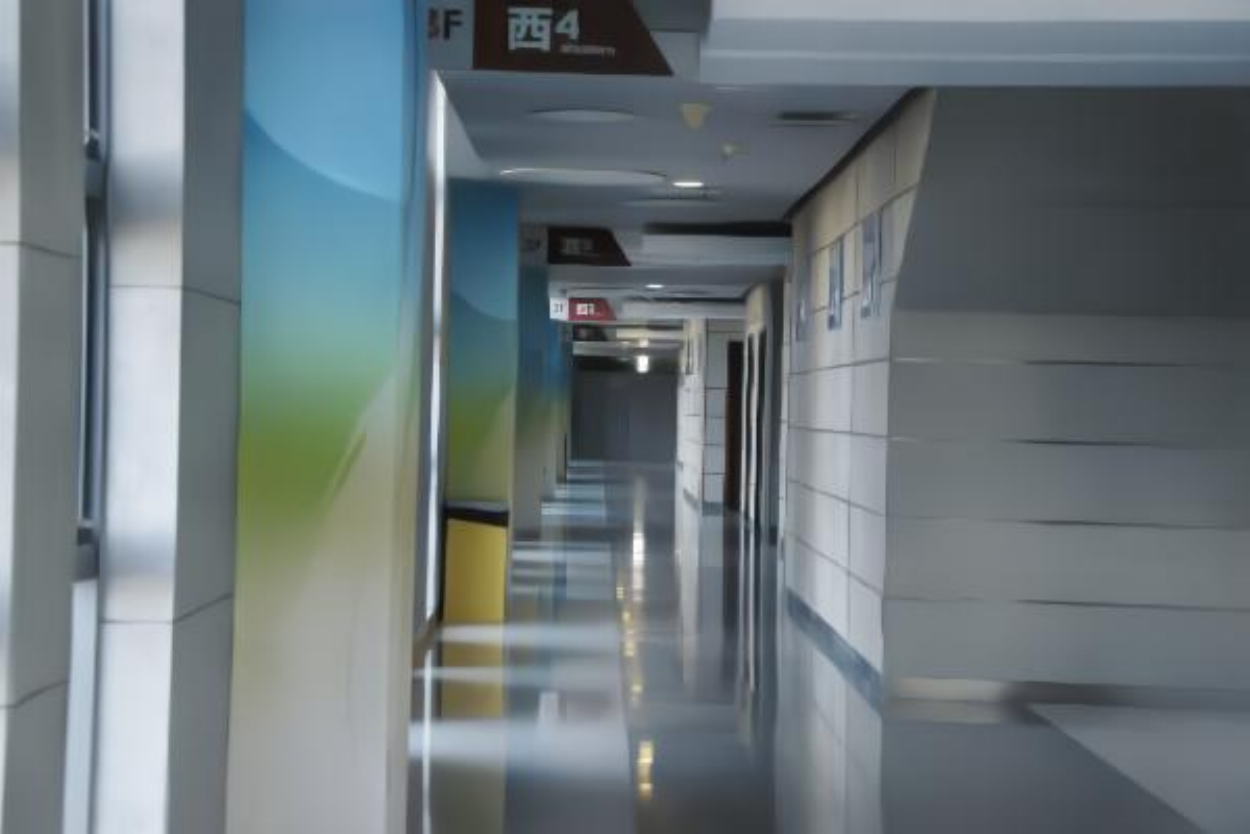}}
 	\centerline{LLFlow}
 	\vspace{2pt}
 	\centerline{\includegraphics[width=\textwidth]{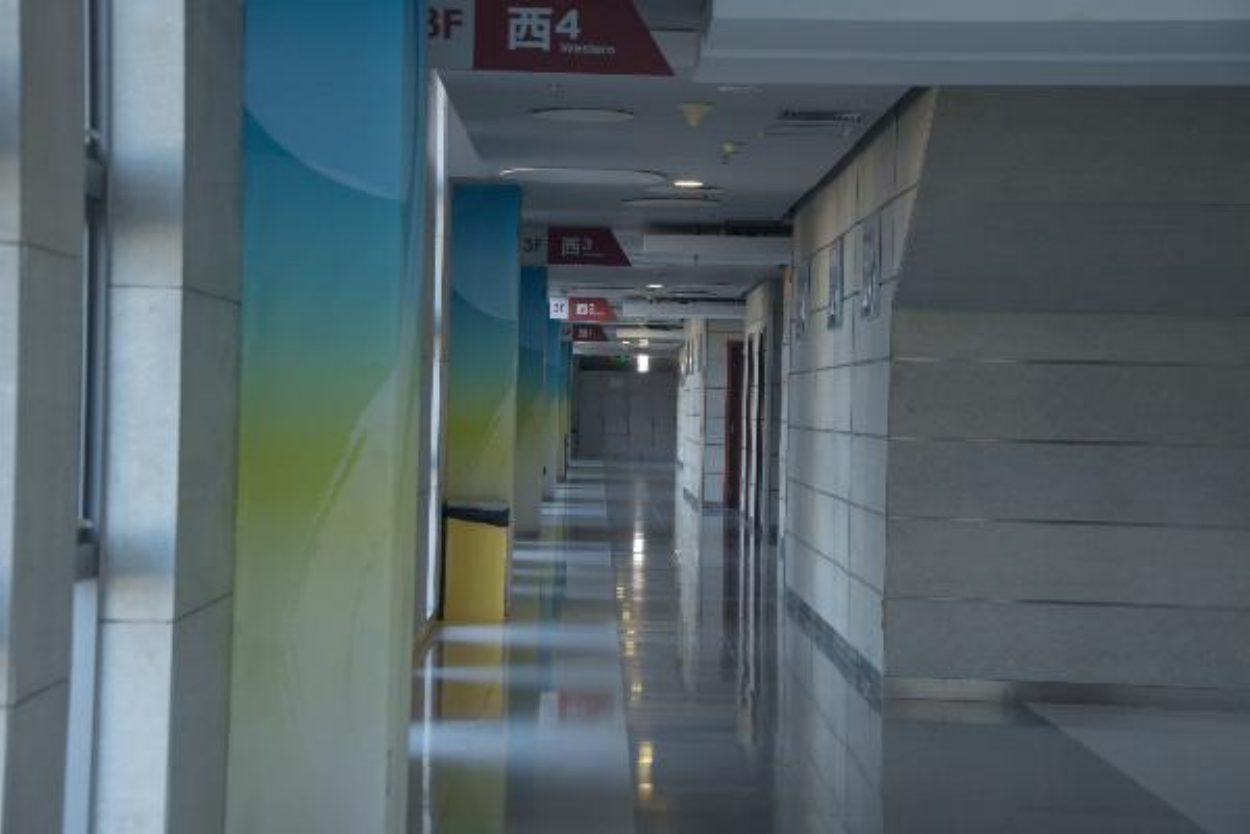}}
 	\centerline{GroundTruth}
        \vspace{3pt}
 \end{minipage}
 \caption{Visual examples for low-light enhancement on LOLv2-Real dataset \cite{LOLv2} among RetinexNet \cite{RetinexNet}, RUAS \cite{RUAS}, LLFlow \cite{LLFlow}, SNR-Aware \cite{SNR-Aware}, RetinexFormer \cite{RetinexFormer}, and our CIDNet. Our model demonstrates superior capabilities in effectively correcting brightness and color differences compared to existing methodologies.}
 \label{fig:v2real}
\end{figure*}

\begin{figure*}
\centering
 \begin{minipage}{0.18\linewidth}
 \centering
 	\centerline{\includegraphics[width=0.85\textwidth]{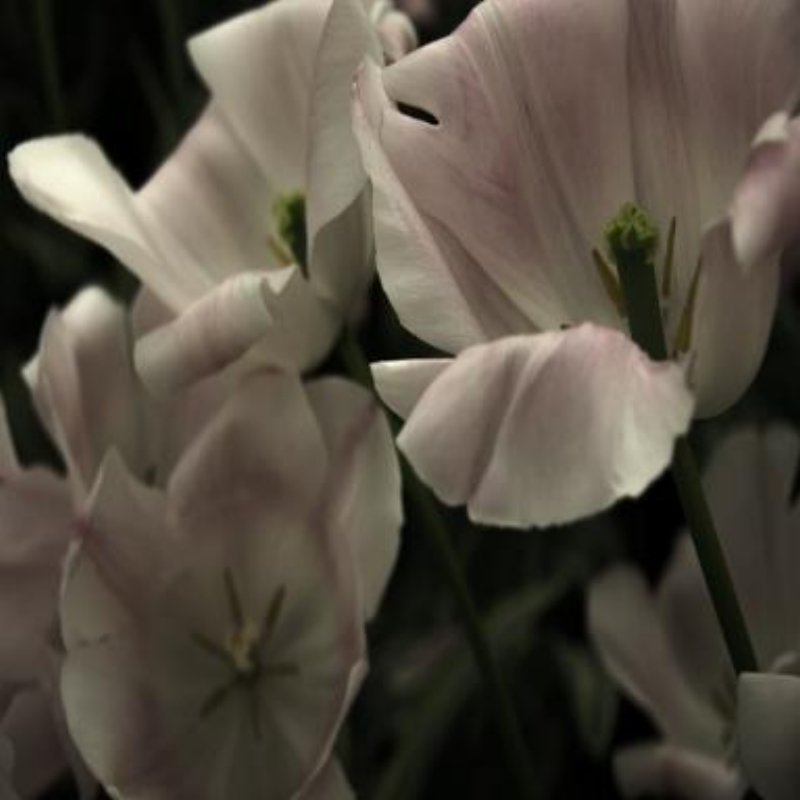}}
 	\centerline{Input}
 	\centerline{\includegraphics[width=0.85\textwidth]{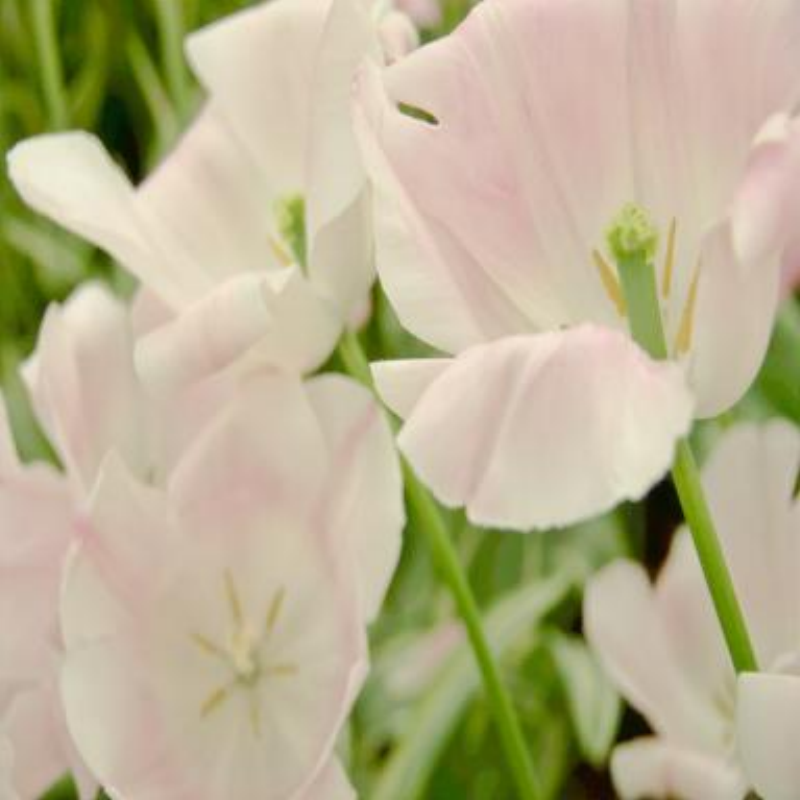}}
 	\centerline{SNR-Aware}
 	\centerline{\includegraphics[width=0.85\textwidth]{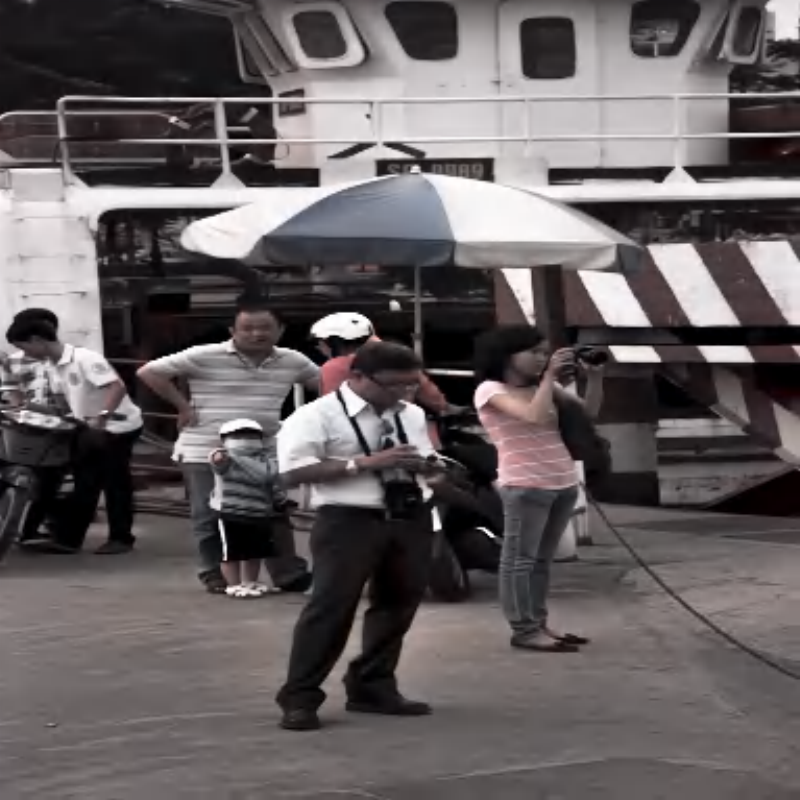}}
 	\centerline{Input}
 	\centerline{\includegraphics[width=0.85\textwidth]{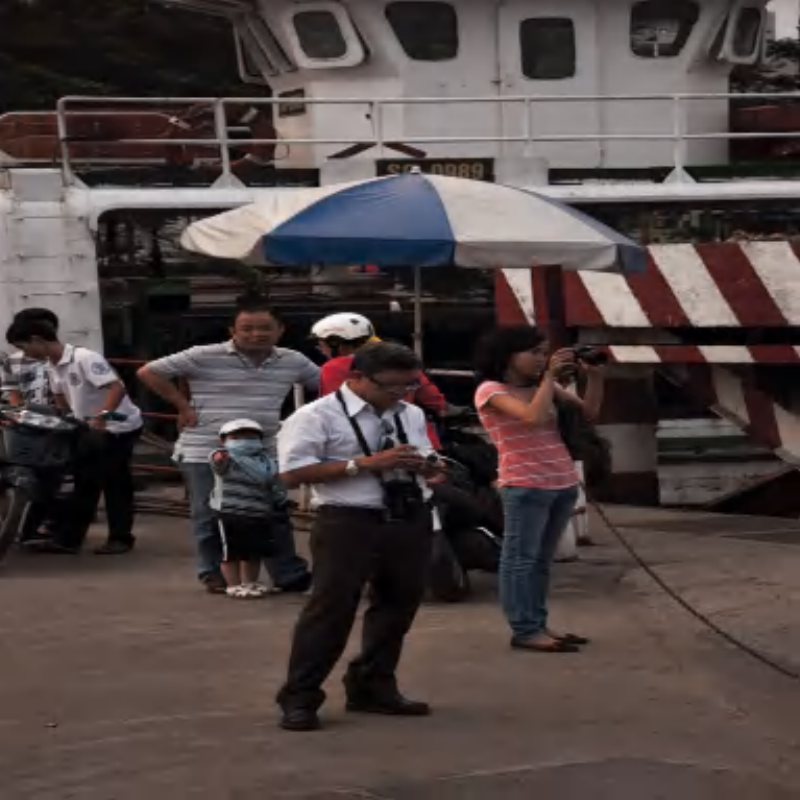}}
 	\centerline{SNR-Aware}
 	\centerline{\includegraphics[width=0.85\textwidth]{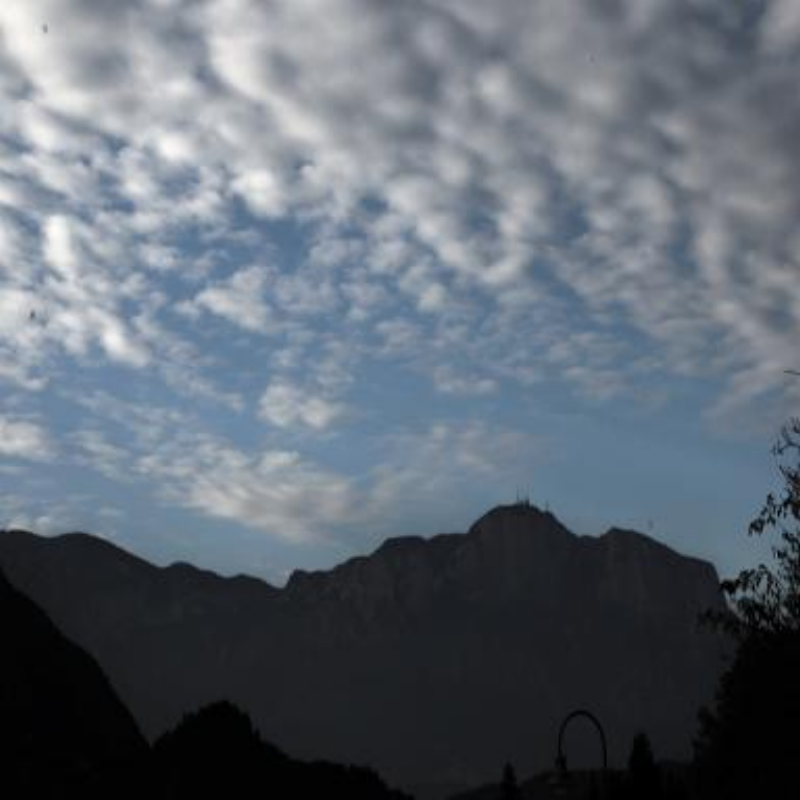}}
 	\centerline{Input}
 	\centerline{\includegraphics[width=0.85\textwidth]{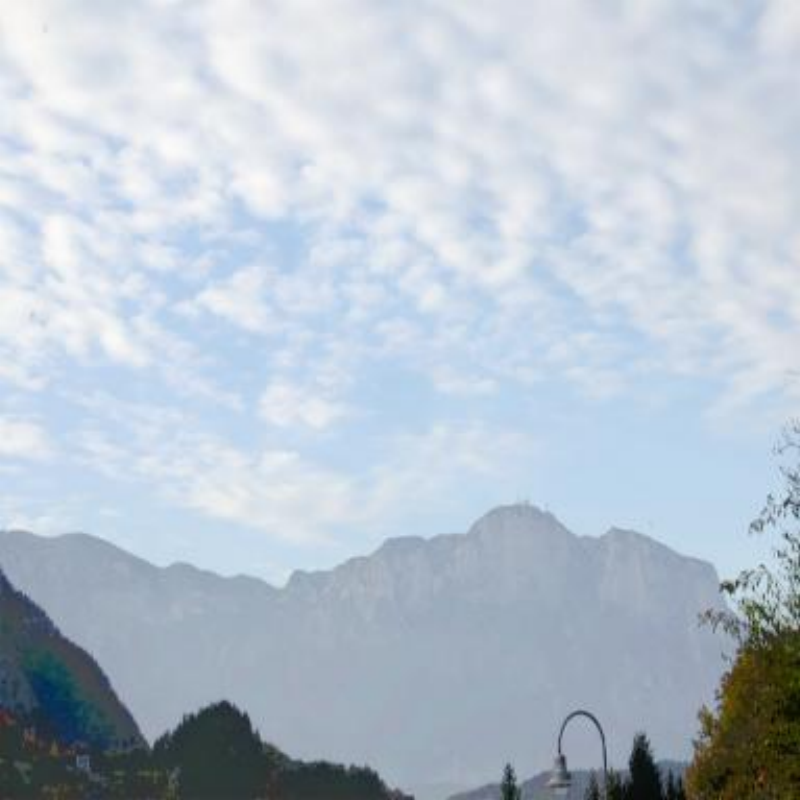}}
 	\centerline{SNR-Aware}
\end{minipage}
\begin{minipage}{0.18\linewidth}
 	\centerline{\includegraphics[width=0.85\textwidth]{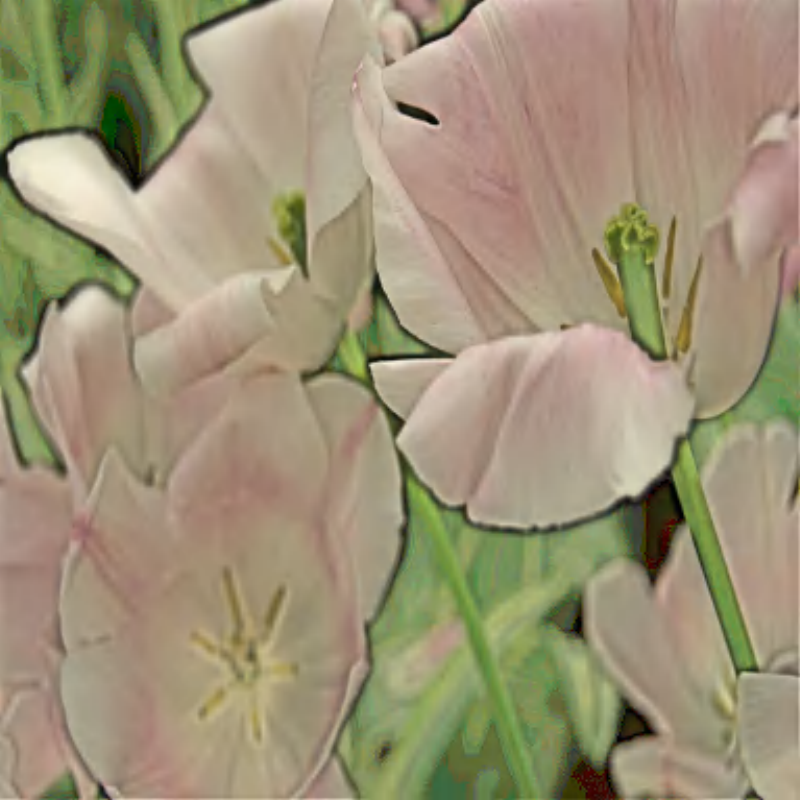}}
 	\centerline{RetinexNet}
 	\centerline{\includegraphics[width=0.85\textwidth]{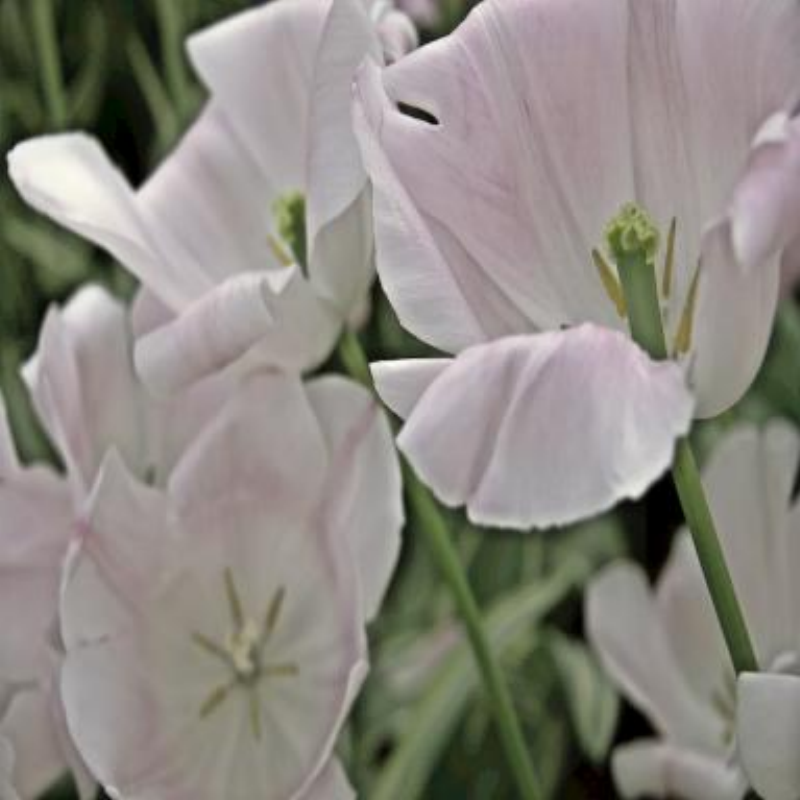}}
 	\centerline{ZeroDCE}
 	\centerline{\includegraphics[width=0.85\textwidth]{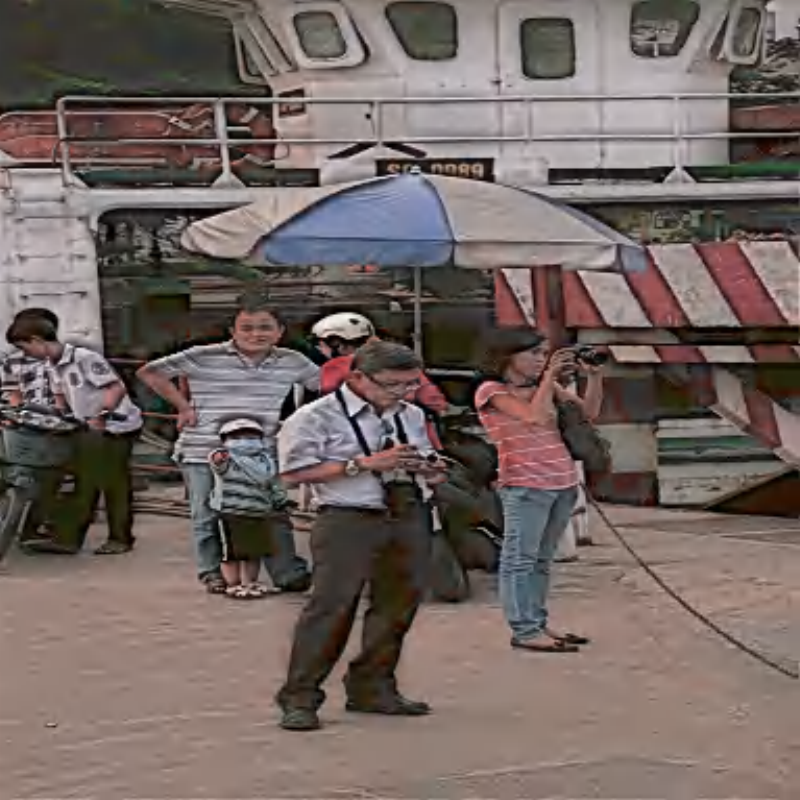}}
 	\centerline{RetinexNet}
 	\centerline{\includegraphics[width=0.85\textwidth]{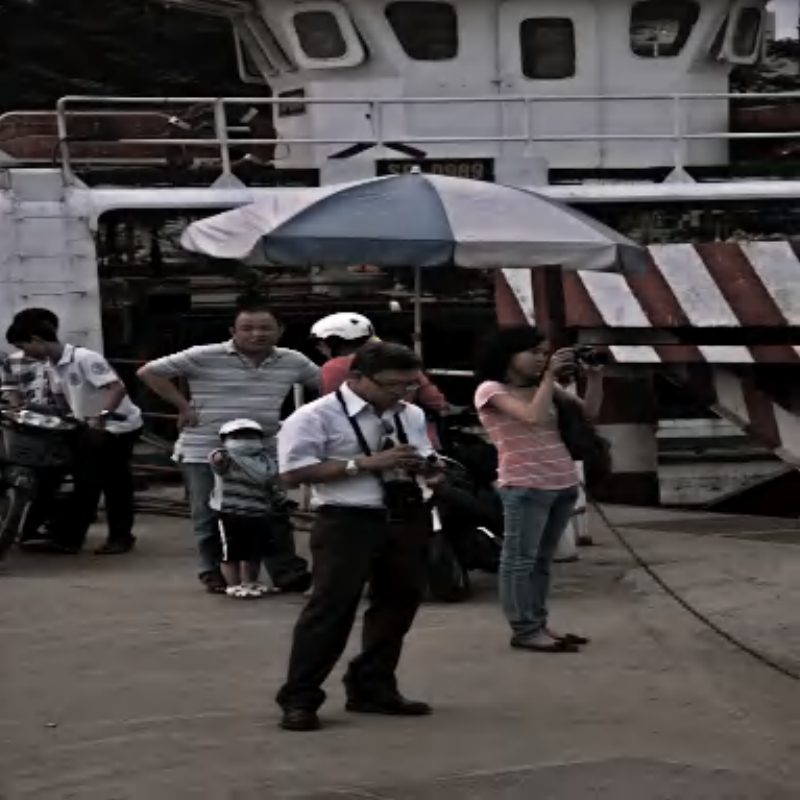}}
 	\centerline{ZeroDCE}
 	\centerline{\includegraphics[width=0.85\textwidth]{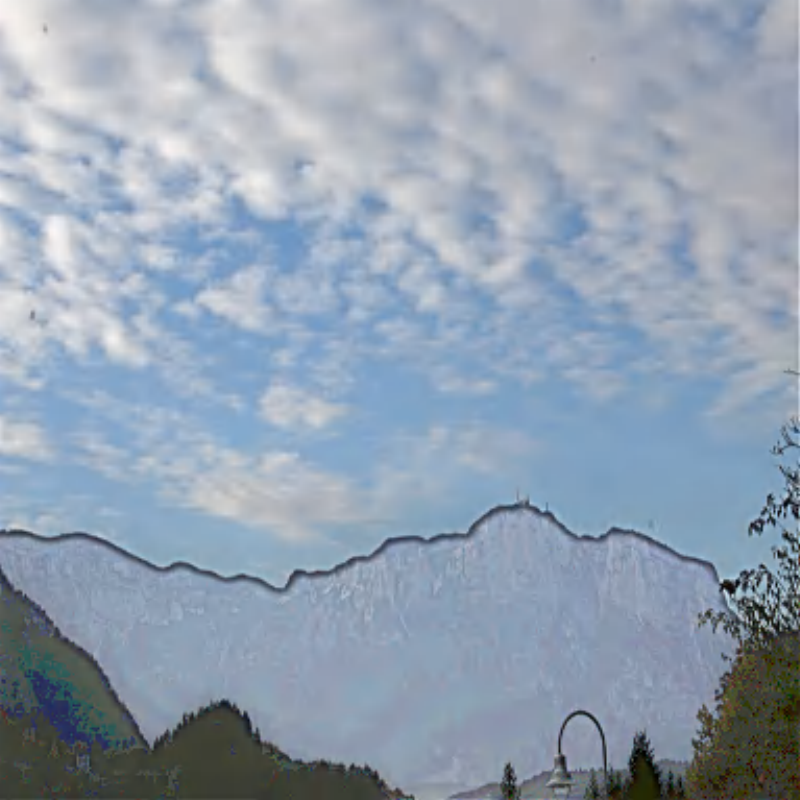}}
 	\centerline{RetinexNet}
 	\centerline{\includegraphics[width=0.85\textwidth]{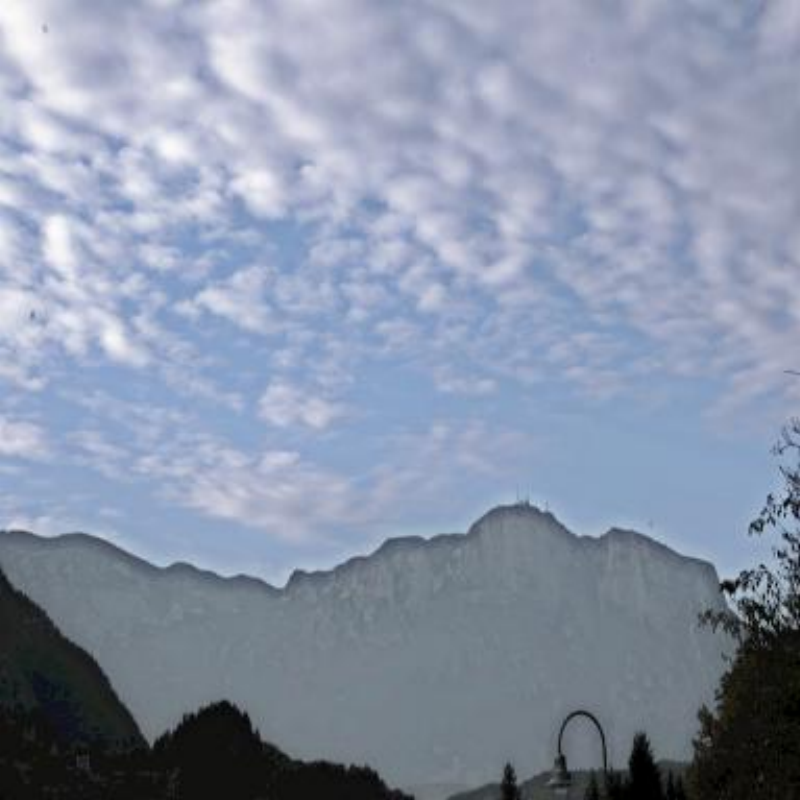}}
 	\centerline{ZeroDCE}
 \end{minipage}
\begin{minipage}{0.18\linewidth}
 	\centerline{\includegraphics[width=0.85\textwidth]{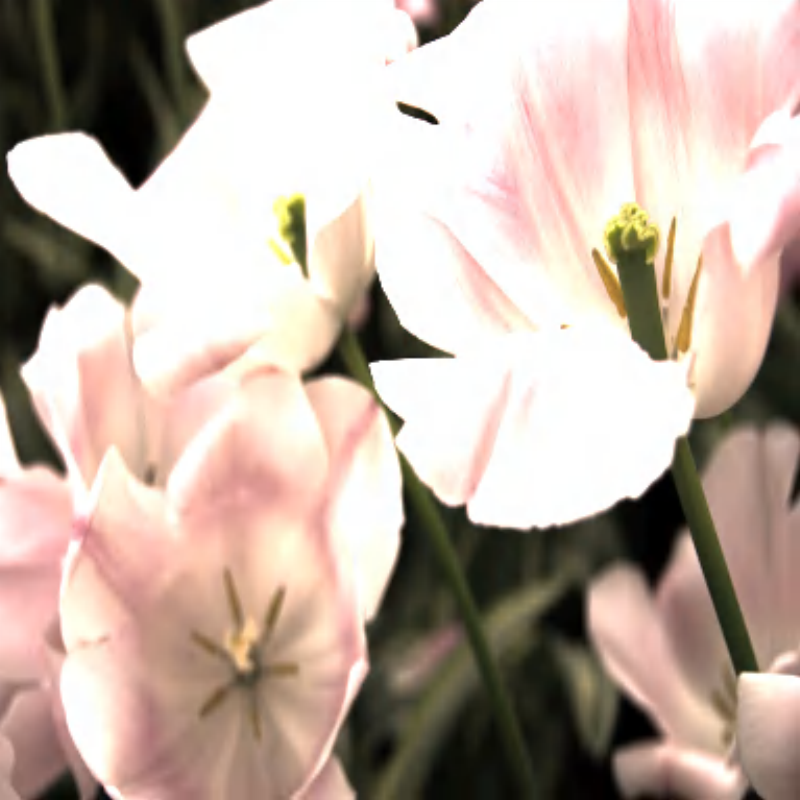}}
 	\centerline{RUAS}
 	\centerline{\includegraphics[width=0.85\textwidth]{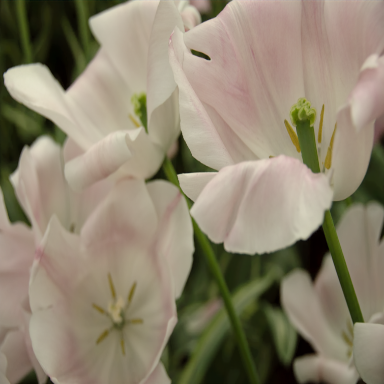}}
 	\centerline{GSAD}
 	\centerline{\includegraphics[width=0.85\textwidth]{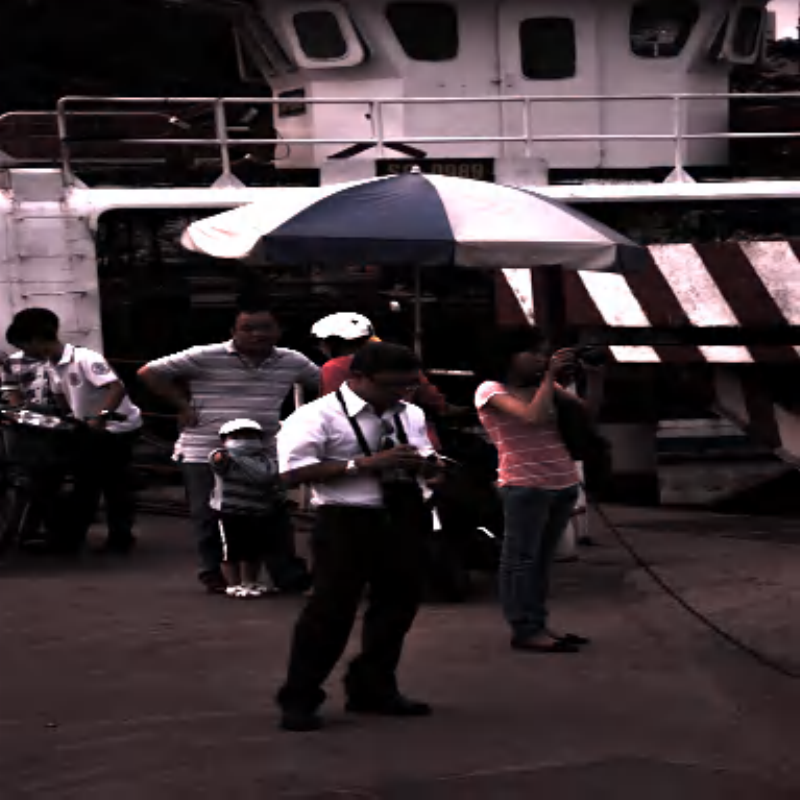}}
 	\centerline{RUAS}
 	\centerline{\includegraphics[width=0.85\textwidth]{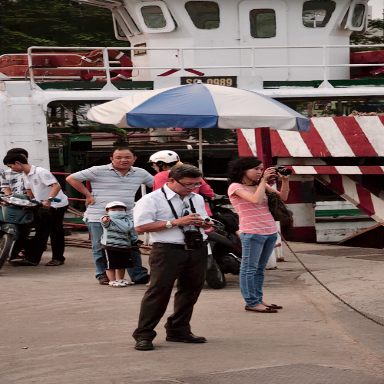}}
 	\centerline{GSAD}
 	\centerline{\includegraphics[width=0.85\textwidth]{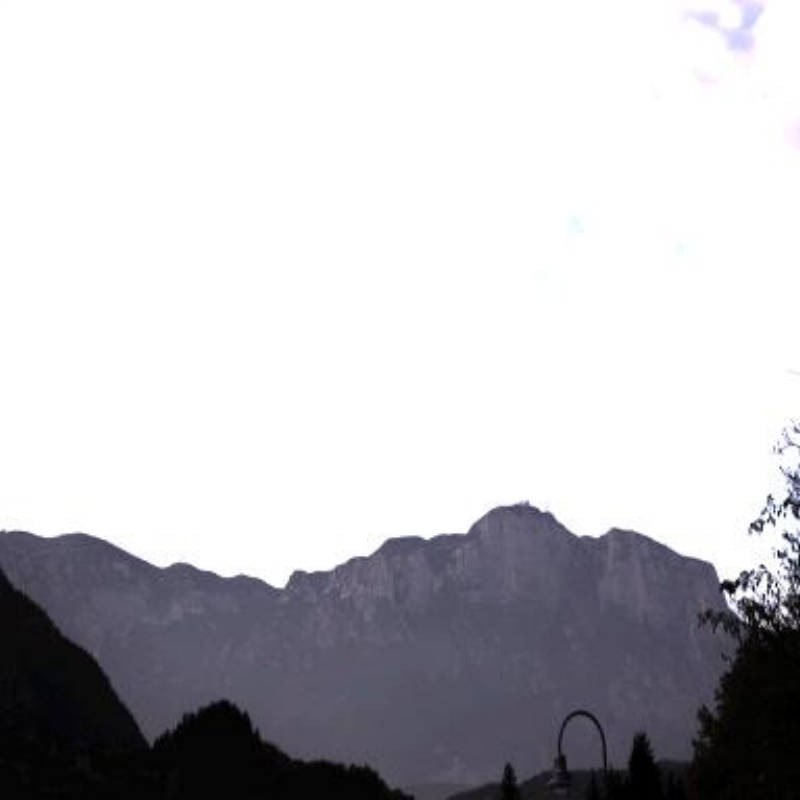}}
 	\centerline{RUAS}
 	\centerline{\includegraphics[width=0.85\textwidth]{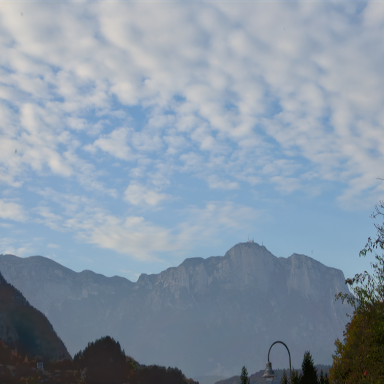}}
 	\centerline{GSAD}
 \end{minipage}
\begin{minipage}{0.18\linewidth}
 	\centerline{\includegraphics[width=0.85\textwidth]{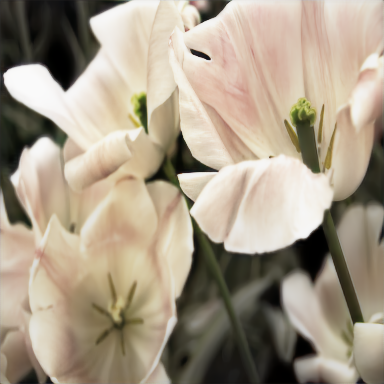}}
 	\centerline{KinD}
 	\centerline{\includegraphics[width=0.85\textwidth]{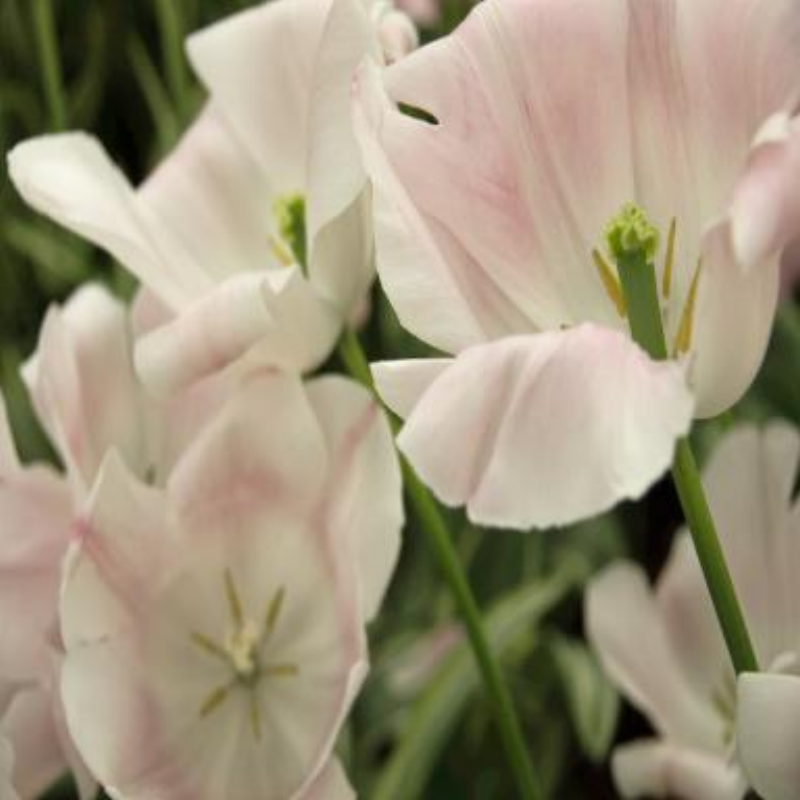}}
 	\centerline{CIDNet}
 	\centerline{\includegraphics[width=0.85\textwidth]{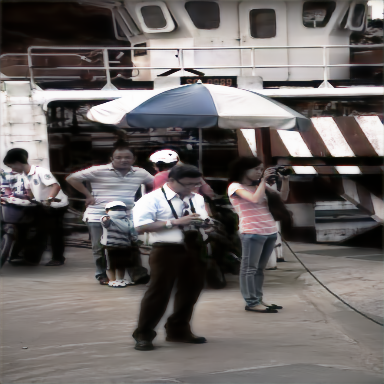}}
 	\centerline{KinD}
 	\centerline{\includegraphics[width=0.85\textwidth]{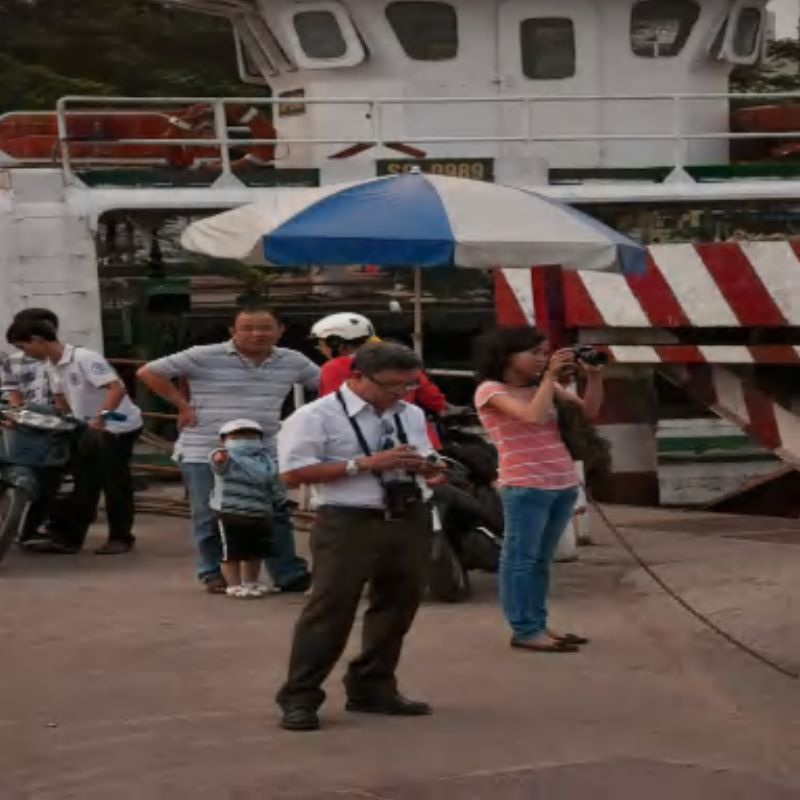}}
 	\centerline{CIDNet}
 	\centerline{\includegraphics[width=0.85\textwidth]{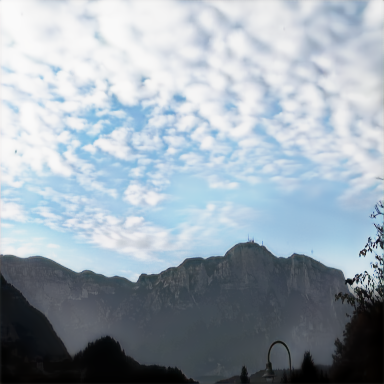}}
 	\centerline{KinD}
 	\centerline{\includegraphics[width=0.85\textwidth]{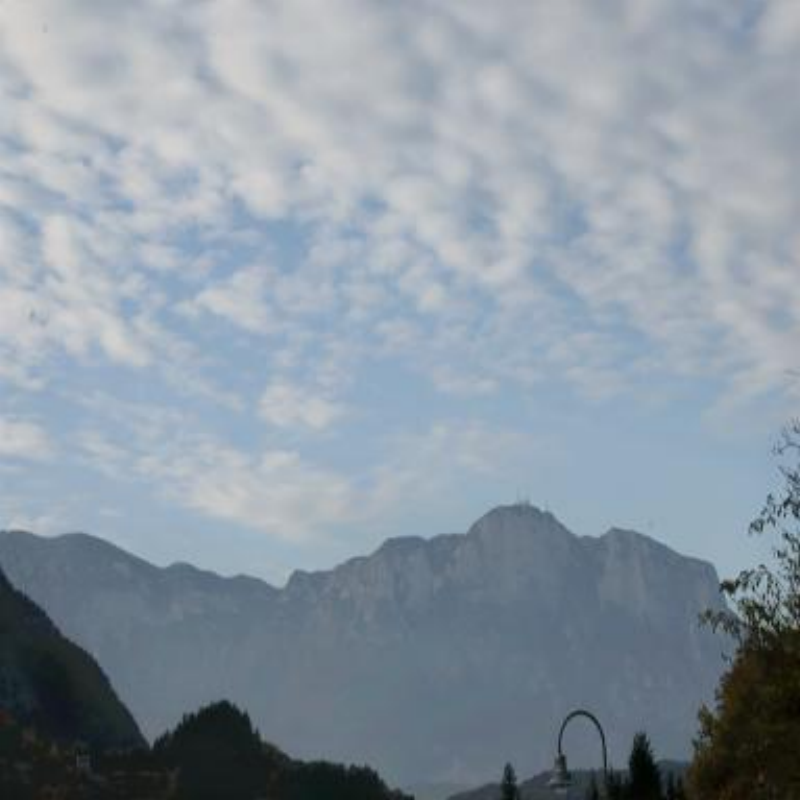}}
 	\centerline{CIDNet}
 \end{minipage}
 \begin{minipage}{0.18\linewidth}
 	\centerline{\includegraphics[width=0.85\textwidth]{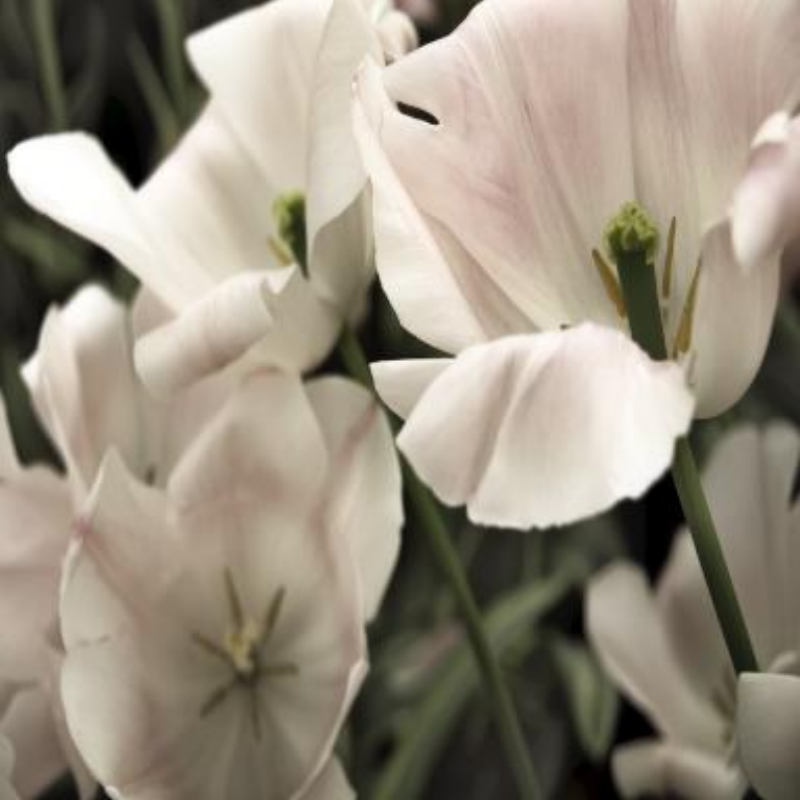}}
 	\centerline{LLFlow}
 	\centerline{\includegraphics[width=0.85\textwidth]{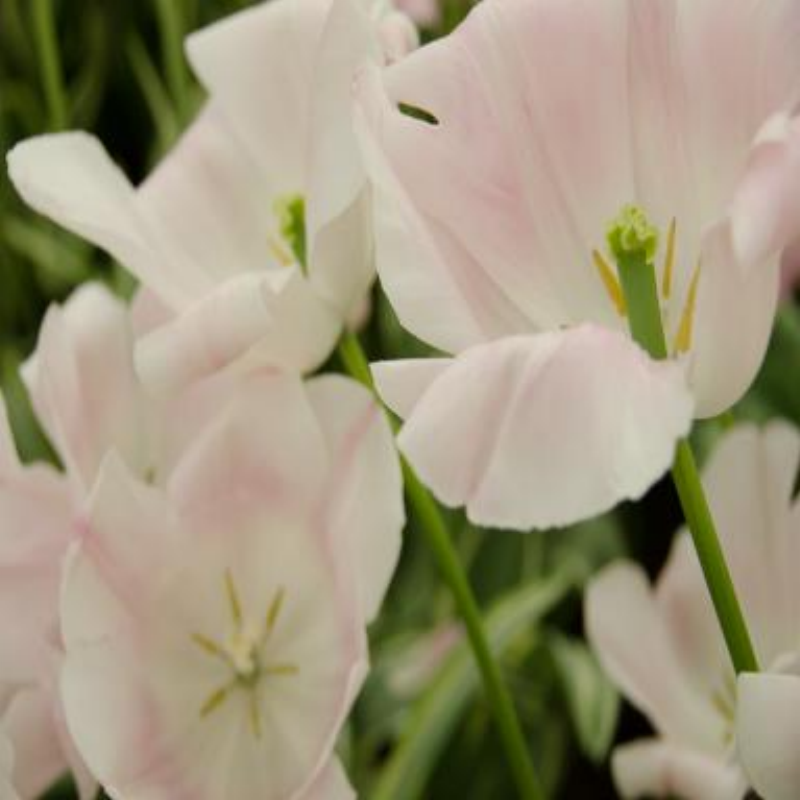}}
 	\centerline{GroundTruth}
 	\centerline{\includegraphics[width=0.85\textwidth]{pic/v2syn/2/c-LLFlow.pdf}}
 	\centerline{LLFlow}
 	\centerline{\includegraphics[width=0.85\textwidth]{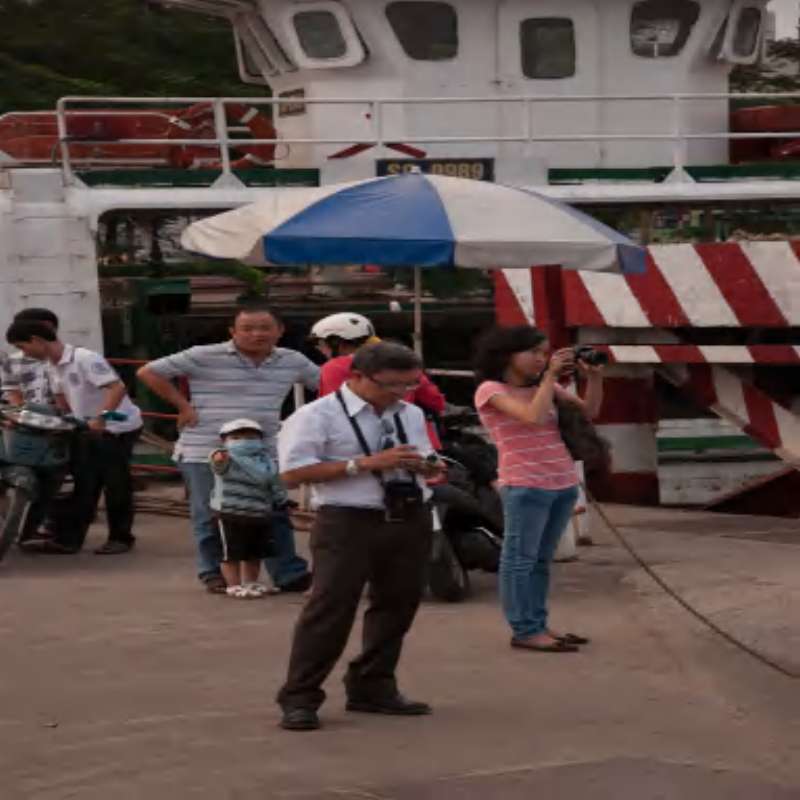}}
 	\centerline{GroundTruth}
 	\centerline{\includegraphics[width=0.85\textwidth]{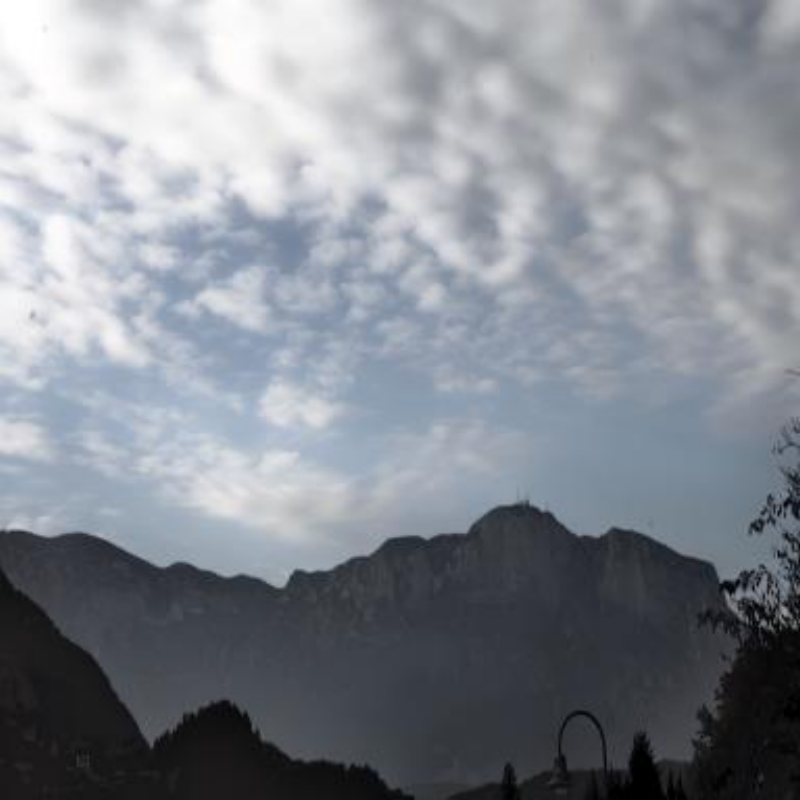}}
 	\centerline{LLFlow}
 	\centerline{\includegraphics[width=0.85\textwidth]{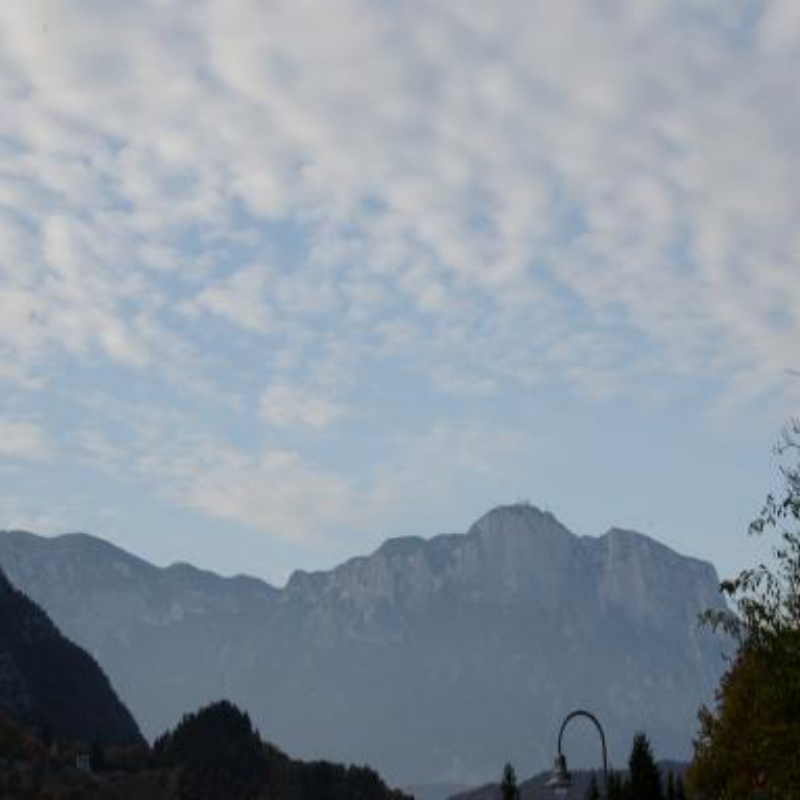}}
 	\centerline{GroundTruth}
 \end{minipage}
 \caption{Visual examples for low-light enhancement on LOLv2-Synthetic dataset \cite{LOLv2} among RetinexNet \cite{RetinexNet}, RUAS \cite{RUAS}, KinD \cite{KinD}, LLFlow \cite{LLFlow}, SNR-Aware \cite{SNR-Aware}, ZeroDCE \cite{Zero-DCE}, GSAD \cite{GSAD}, and our CIDNet. Although the output images of the GSAD method and CIDNet are relatively similar, it is evident that CIDNet produces more visually pleasing results.}
 \label{fig:v2syn}
\end{figure*}

\textbf{LOL.} Fig. \ref{fig:v1}, Fig. \ref{fig:v2real} and Fig. \ref{fig:v2syn} shows the low-light single image enhancement results of three subsets of LOL dataset, respectively.

\begin{figure*}
\centering
 \begin{minipage}{0.24\linewidth}
 \centering
        \vspace{3pt}
 	\centerline{\includegraphics[width=1\textwidth]{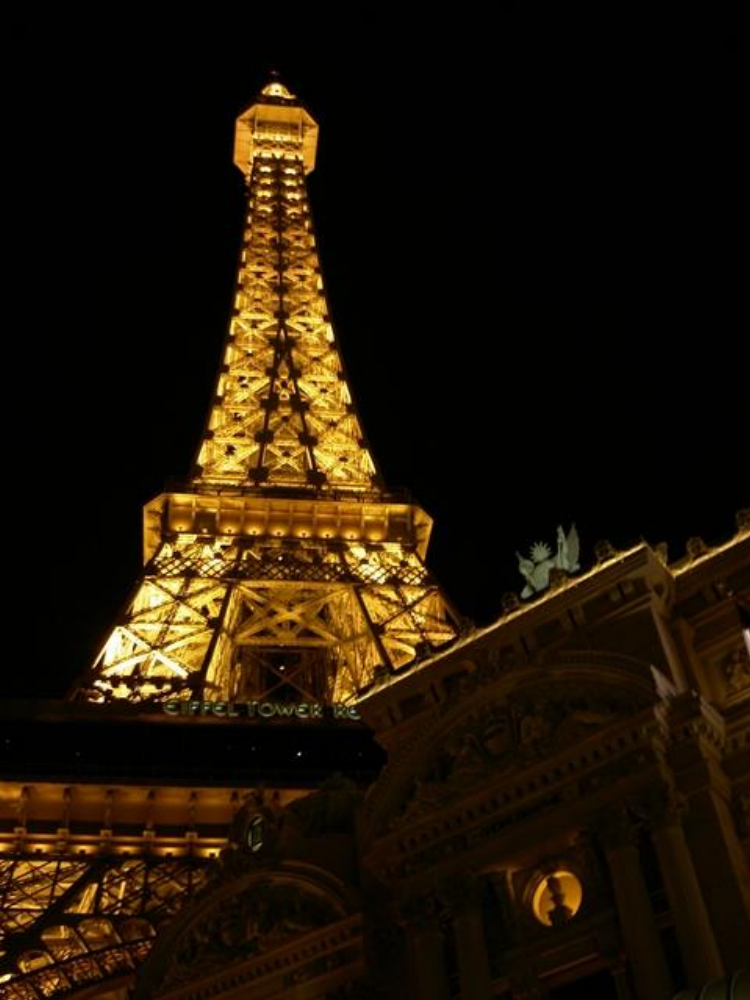}}
 	\centerline{Input}
 	\centerline{\includegraphics[width=1\textwidth]{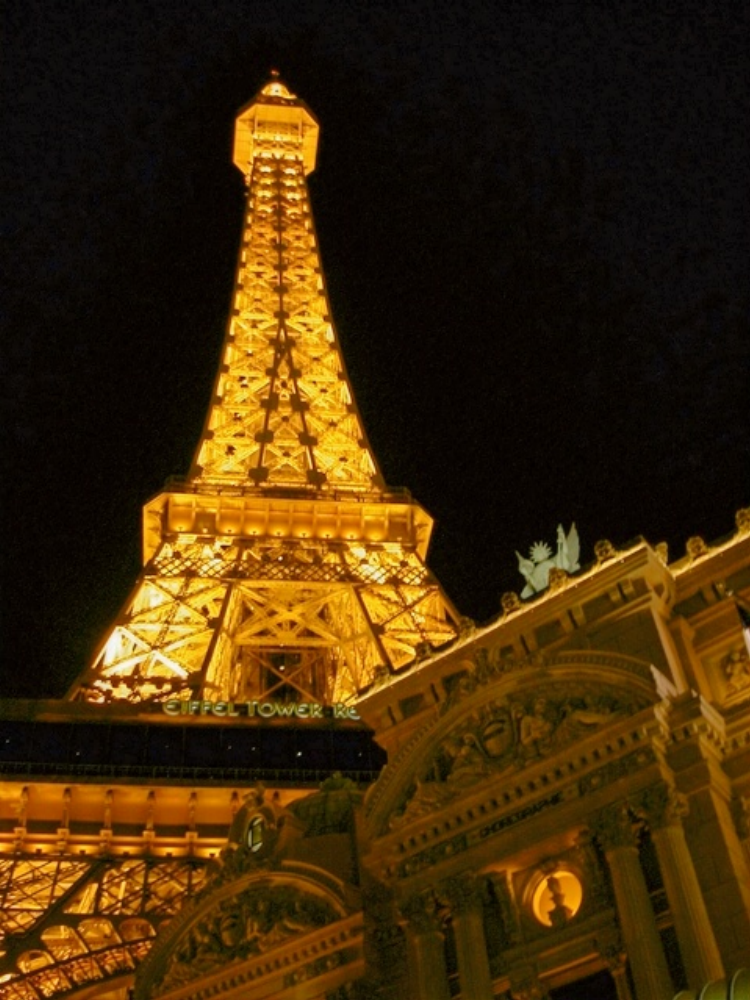}}
 	\centerline{RetinexFormer}
        \vspace{3pt}
 	\centerline{\includegraphics[width=1\textwidth]{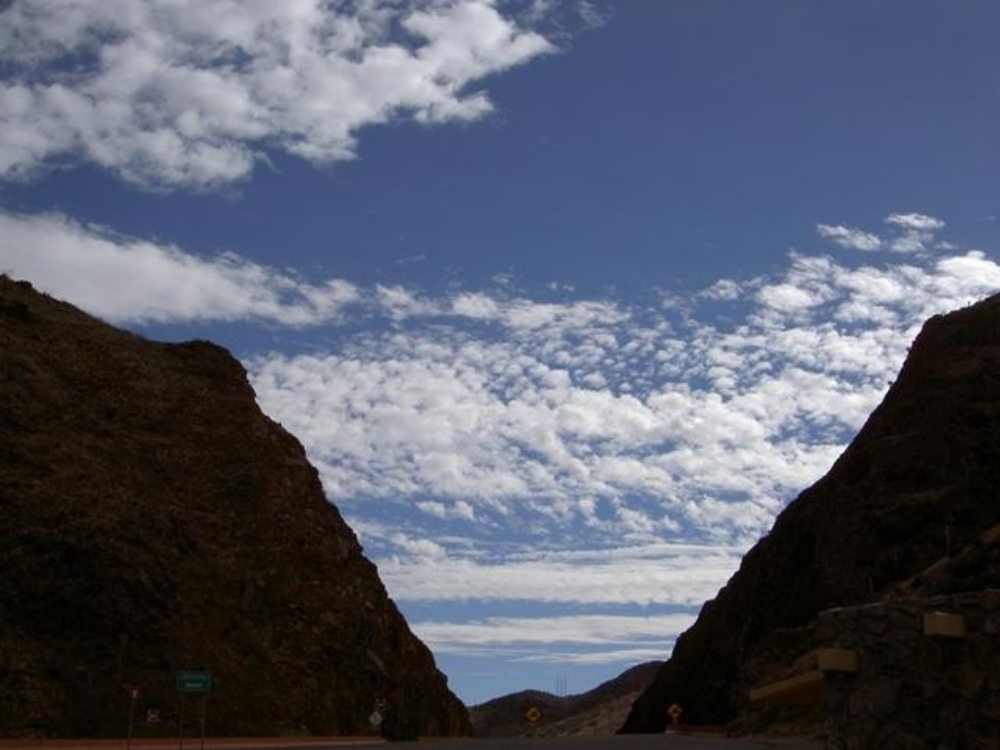}}
 	\centerline{Input}
 	\centerline{\includegraphics[width=1\textwidth]{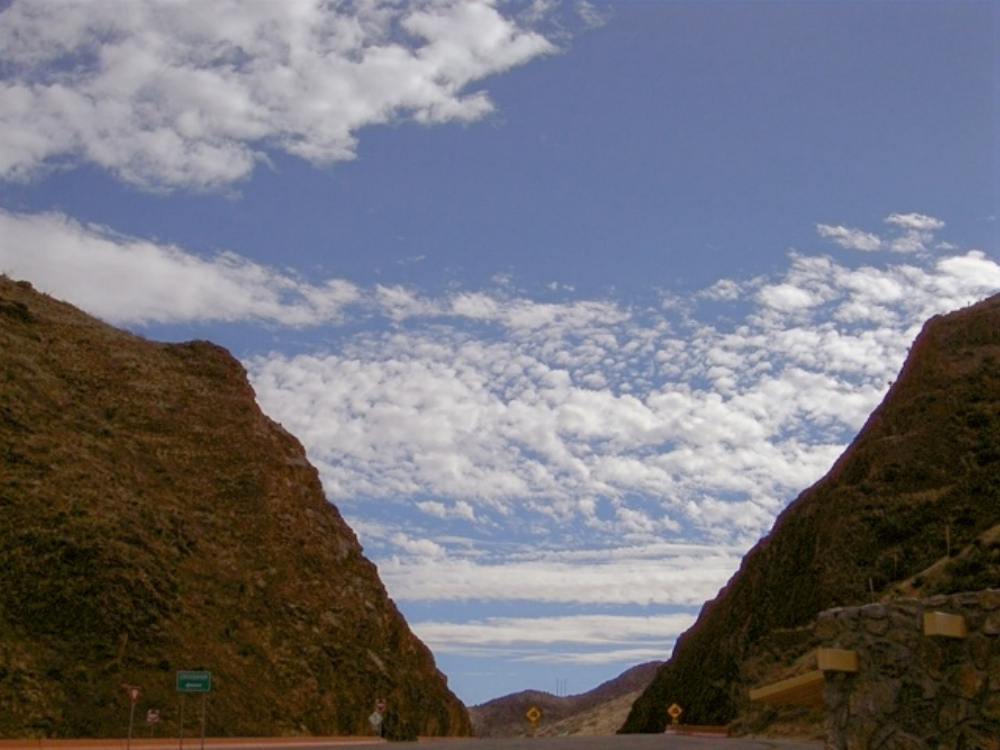}}
 	\centerline{RetinexFormer}
        \vspace{3pt}
\end{minipage}
\begin{minipage}{0.24\linewidth}
        \vspace{3pt}
 	\centerline{\includegraphics[width=1\textwidth]{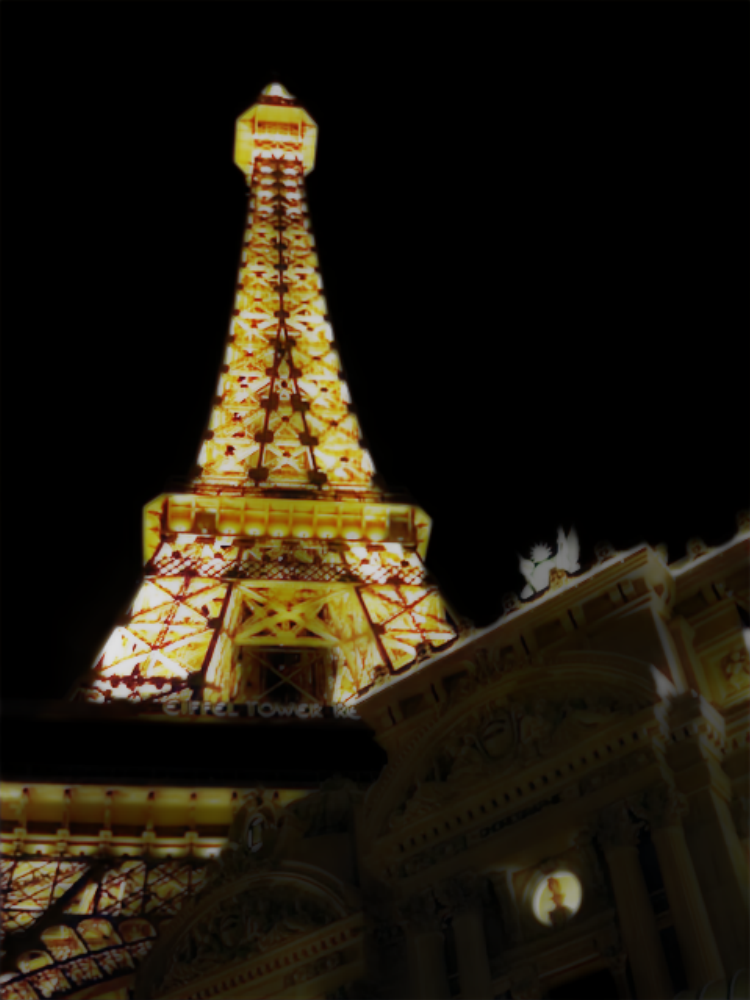}}
 	\centerline{KinD}
        \vspace{2pt}
 	\centerline{\includegraphics[width=1\textwidth]{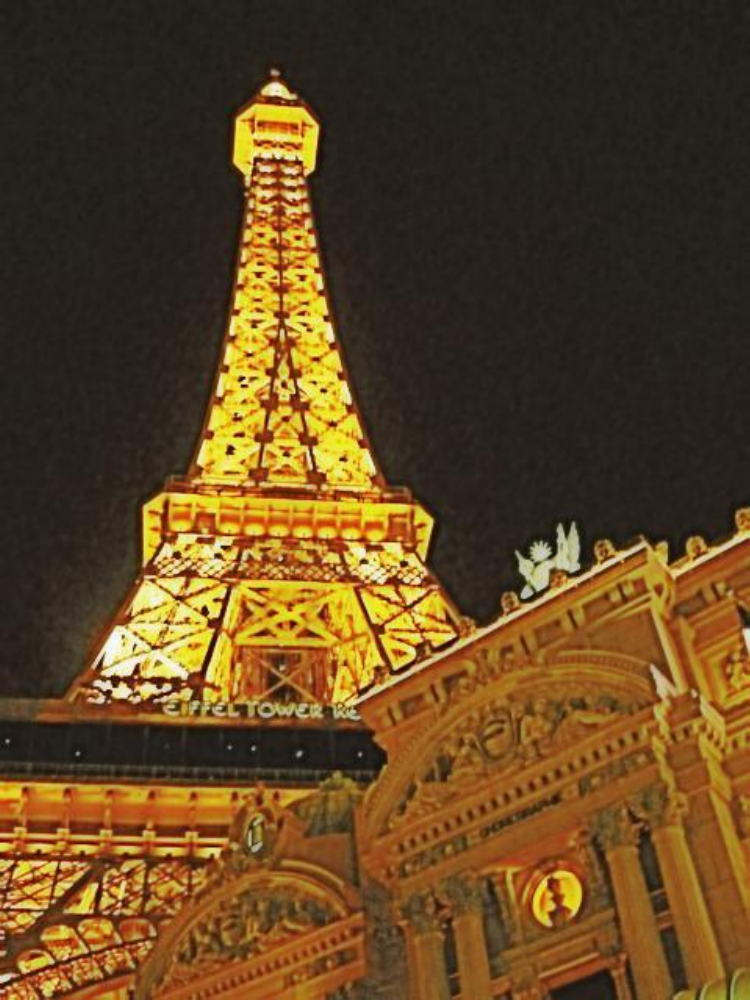}}
 	\centerline{PairLIE}
        \vspace{3pt}
 	\centerline{\includegraphics[width=1\textwidth]{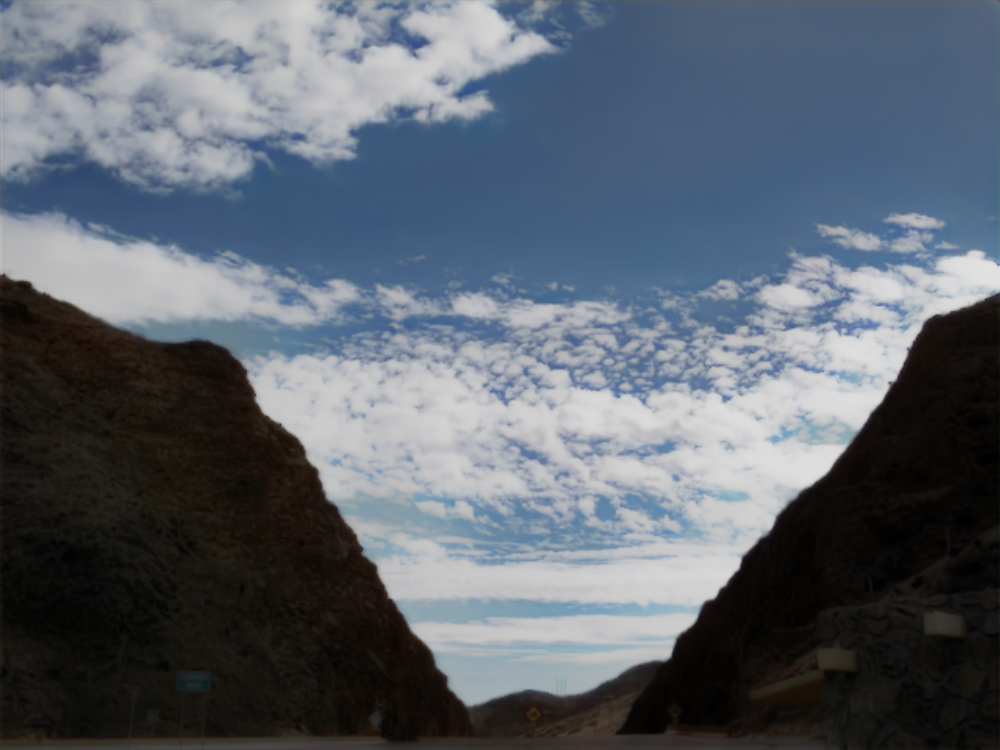}}
 	\centerline{KinD}
        \vspace{2pt}
 	\centerline{\includegraphics[width=1\textwidth]{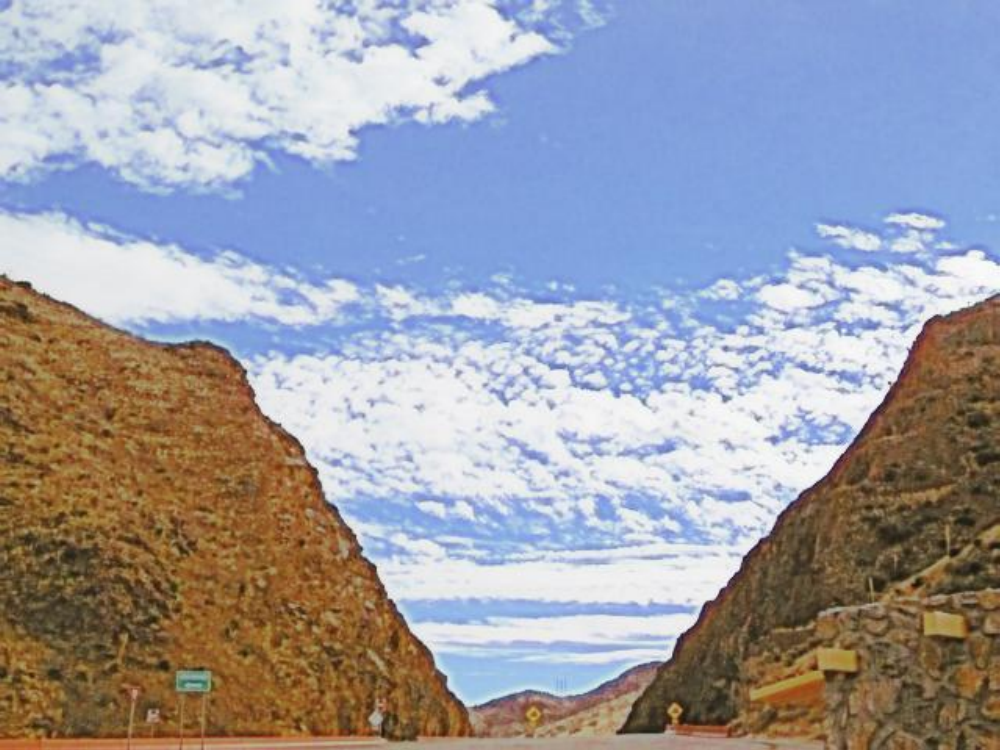}}
 	\centerline{PairLIE}
        \vspace{3pt}
 \end{minipage}
\begin{minipage}{0.24\linewidth}
 	\vspace{3pt}
 	\centerline{\includegraphics[width=1\textwidth]{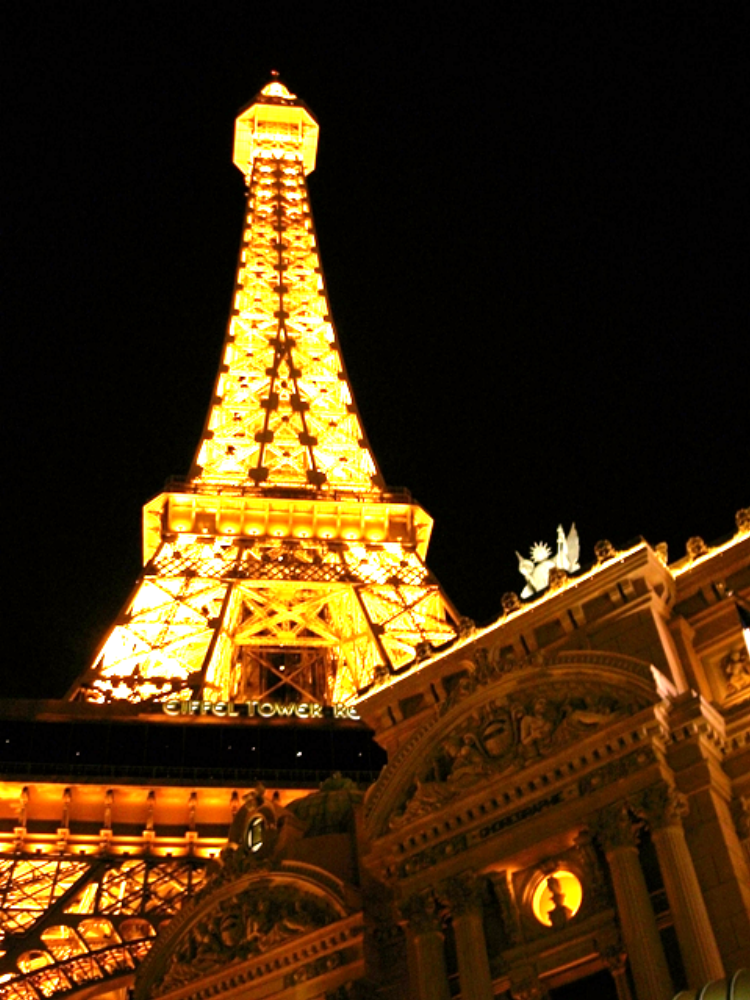}}
 	\centerline{RUAS}
 	\vspace{2pt}
 	\centerline{\includegraphics[width=1\textwidth]{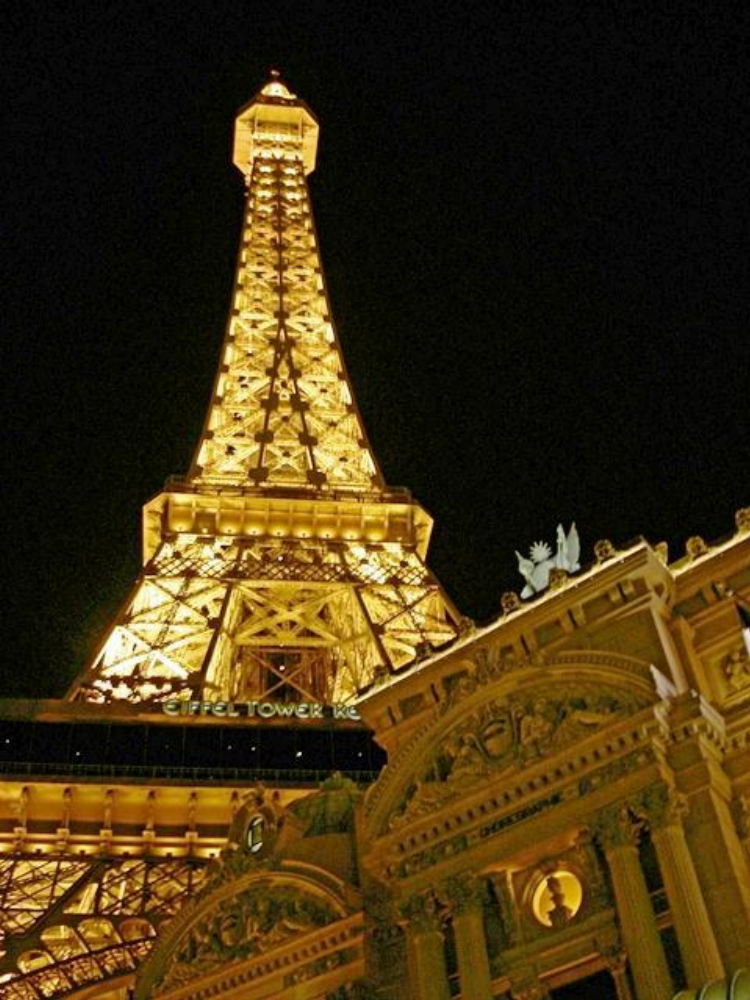}}
 	\centerline{ZeroDCE}
 	\vspace{3pt}
 	\centerline{\includegraphics[width=1\textwidth]{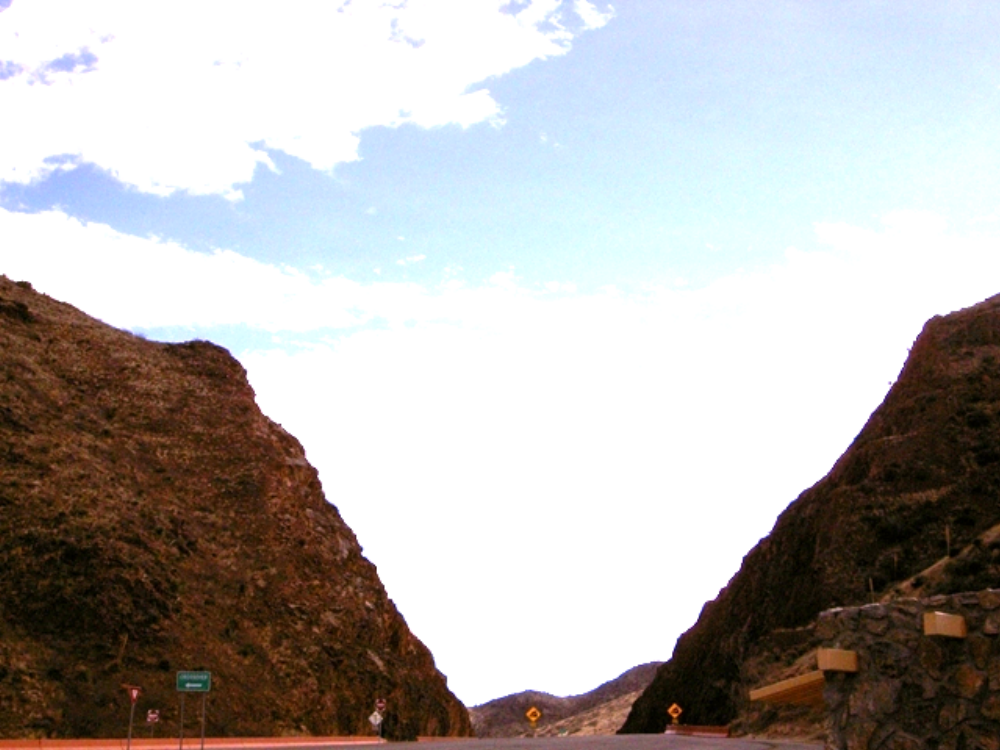}}
 	\centerline{RUAS}
 	\vspace{2pt}
 	\centerline{\includegraphics[width=1\textwidth]{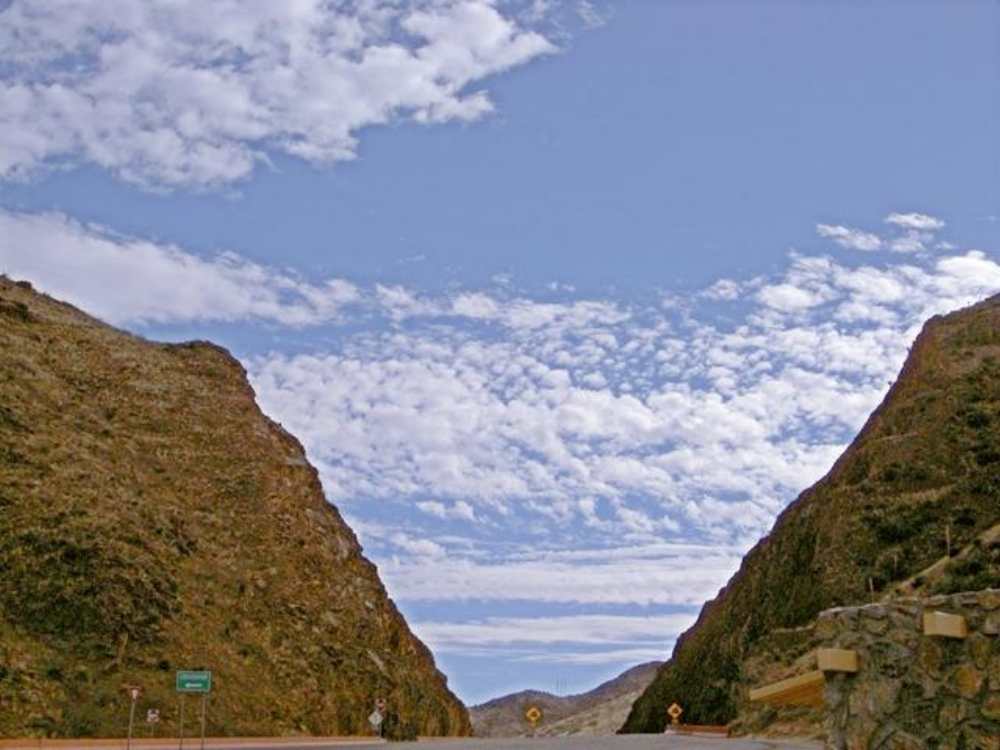}}
 	\centerline{ZeroDCE}
 	\vspace{3pt}
 \end{minipage}
\begin{minipage}{0.24\linewidth}
 	\vspace{3pt}
 	\centerline{\includegraphics[width=1\textwidth]{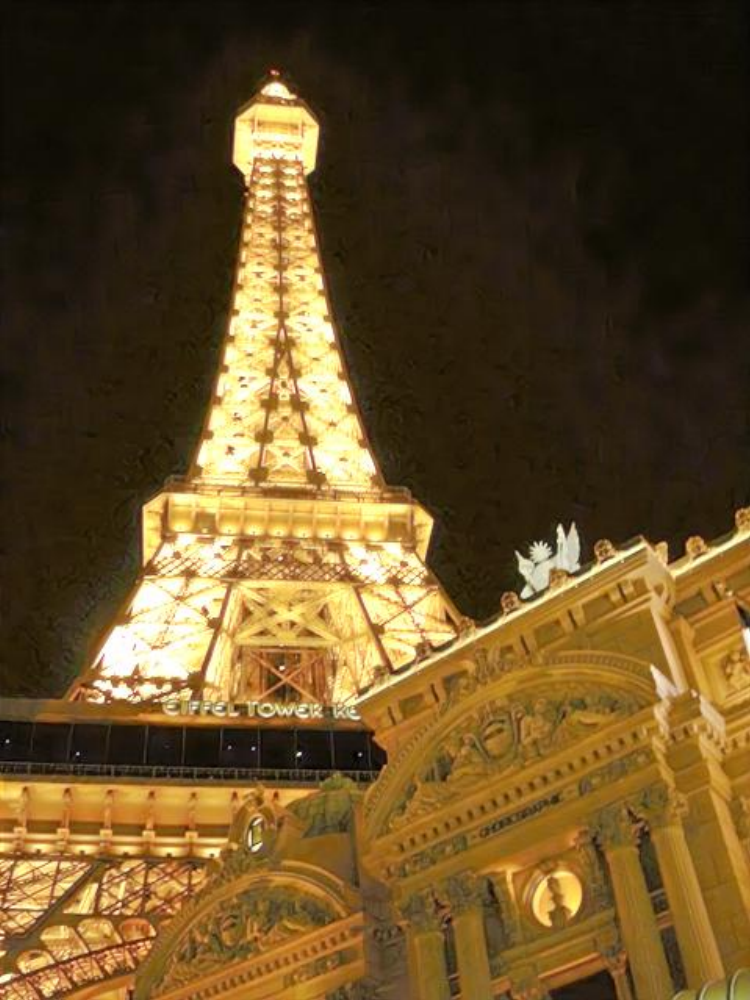}}
 	\centerline{URetinexNet}
 	\vspace{2pt}
 	\centerline{\includegraphics[width=1\textwidth]{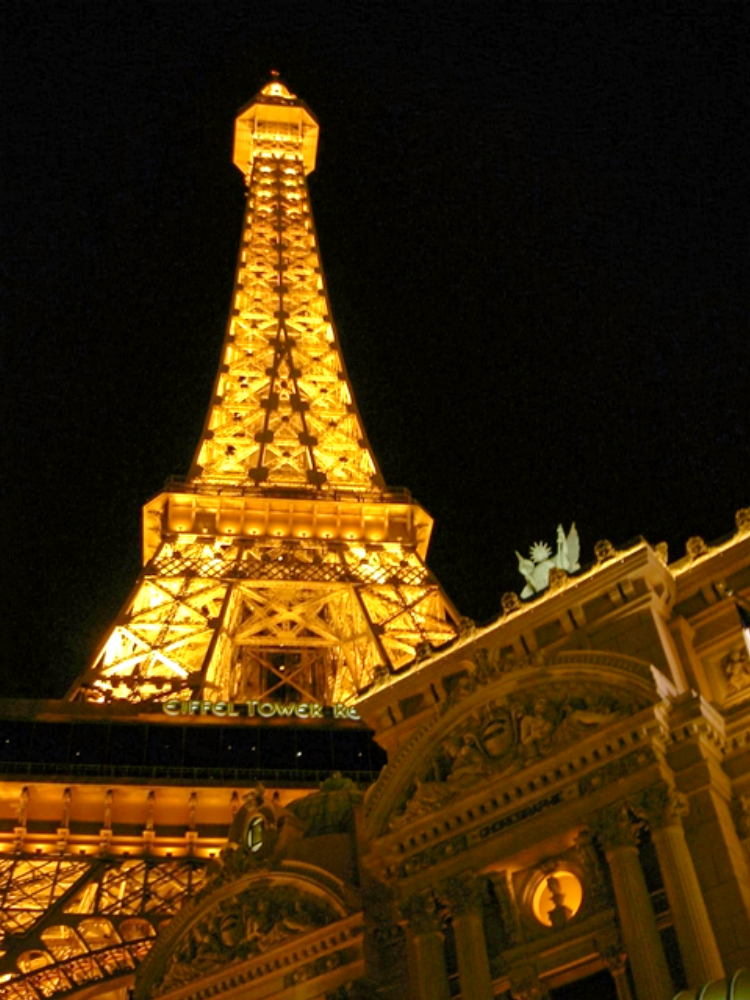}}
 	\centerline{CIDNet}
 	\vspace{3pt}
 	\centerline{\includegraphics[width=1\textwidth]{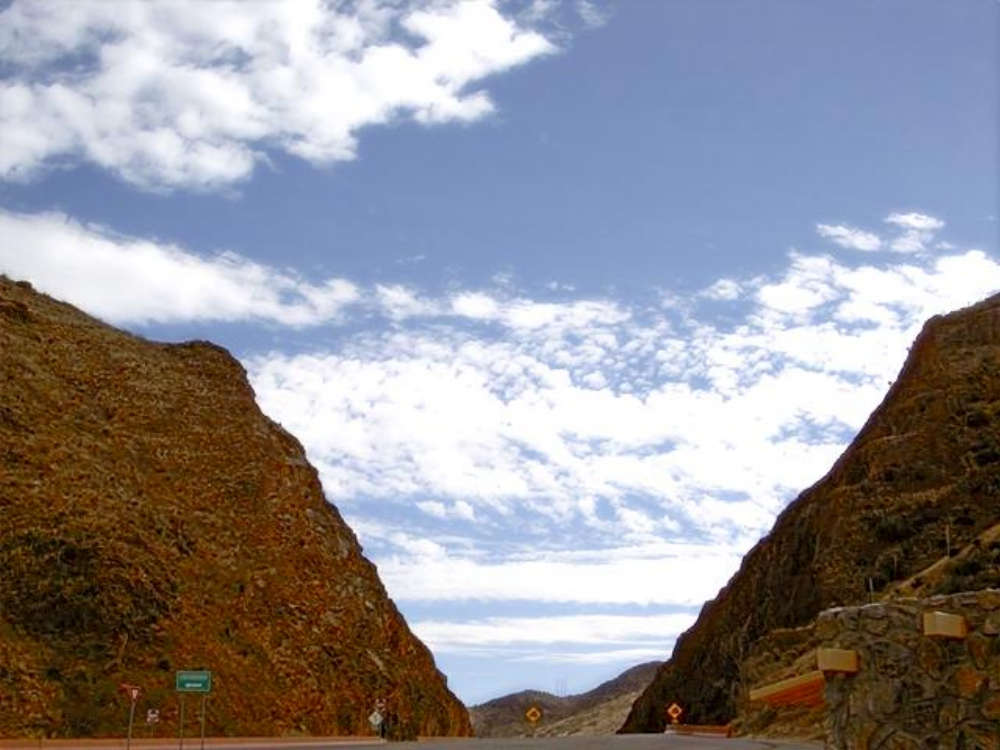}}
 	\centerline{URetinexNet}
 	\vspace{2pt}
 	\centerline{\includegraphics[width=1\textwidth]{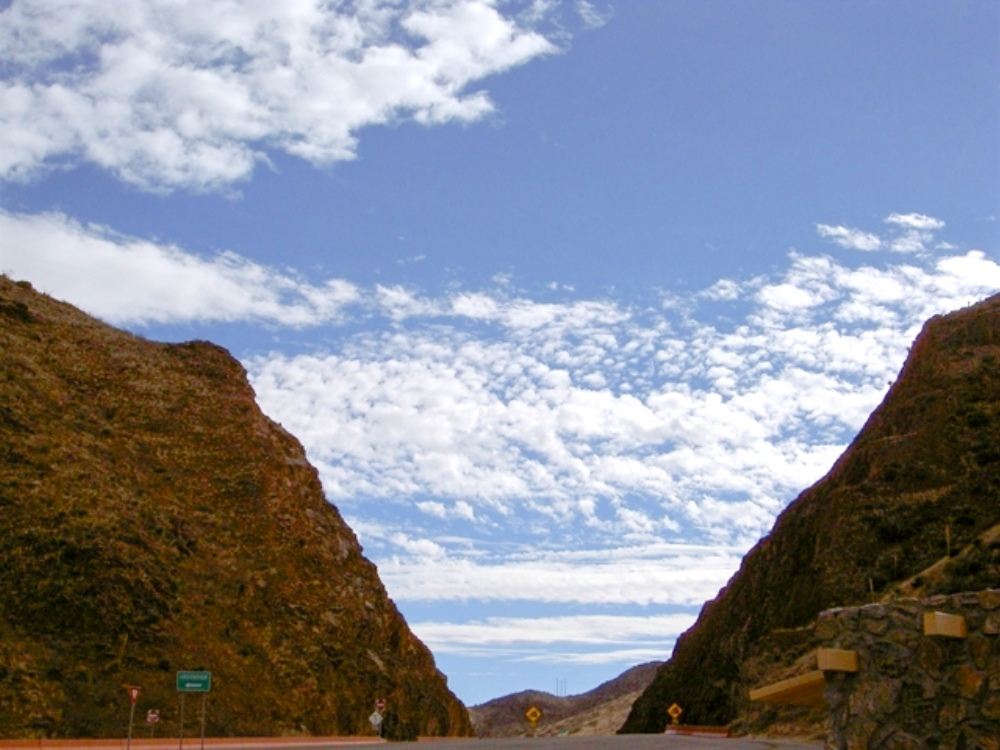}}
 	\centerline{CIDNet}
 	\vspace{3pt}
 \end{minipage}
 \caption{Visual examples for unpaired image enhancement on DICM dataset \cite{DICM} among KinD \cite{KinD}, RUAS \cite{RUAS}, URetinexNet \cite{URetinexNet}, RetinexFormer \cite{RetinexFormer}, PairLIE \cite{PairLIE}, ZeroDCE \cite{Zero-DCE}, and our CIDNet. Our method exhibits enhanced generalization capabilities, consequently resulting in more aesthetically pleasing outcomes with a higher degree of reality.}
 \label{fig:DICM}
\end{figure*}

\begin{figure*}
\centering
 \begin{minipage}{0.24\linewidth}
 \centering
        \vspace{3pt}
 	\centerline{\includegraphics[width=1\textwidth]{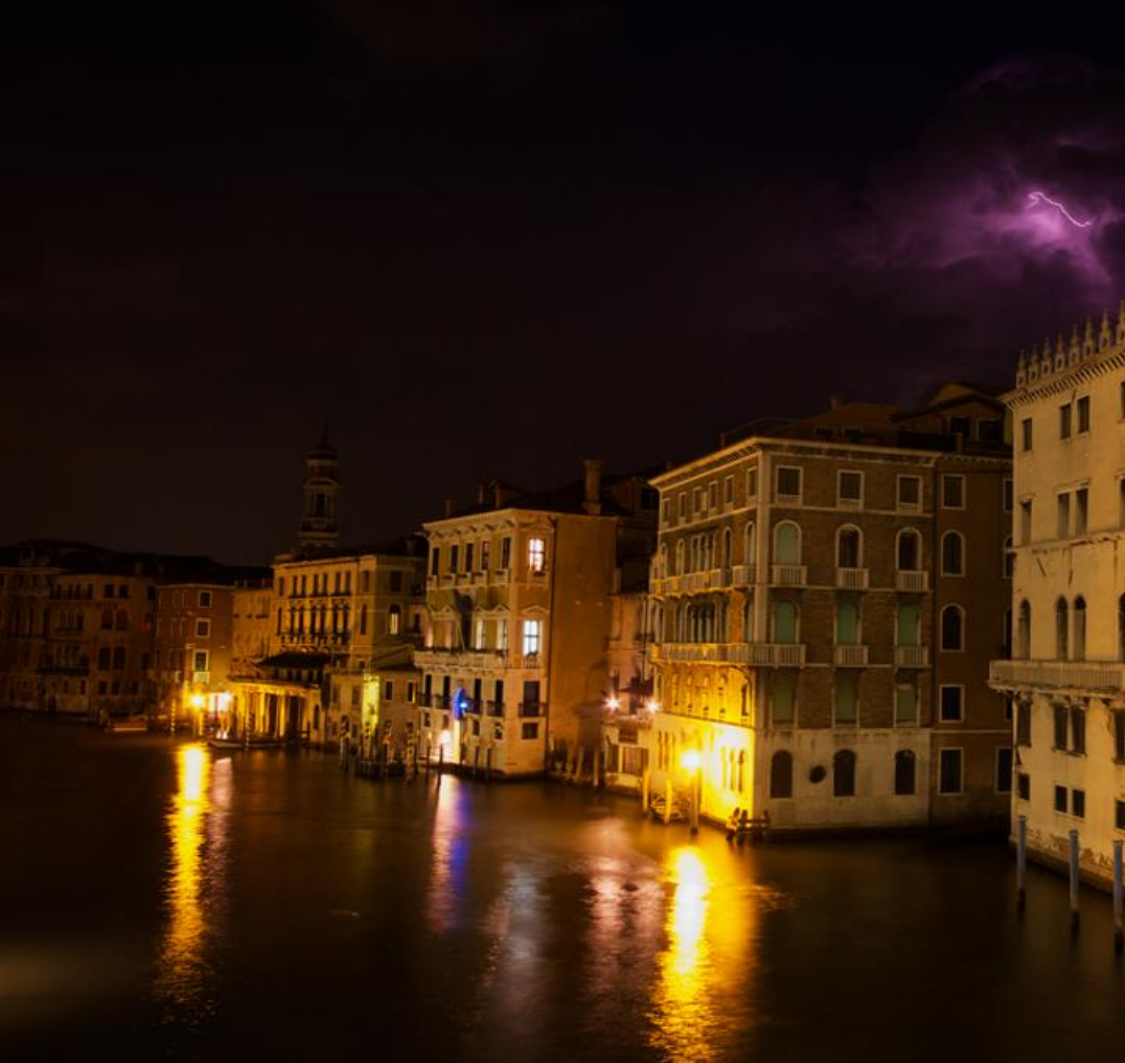}}
 	\centerline{Input}
 	\centerline{\includegraphics[width=1\textwidth]{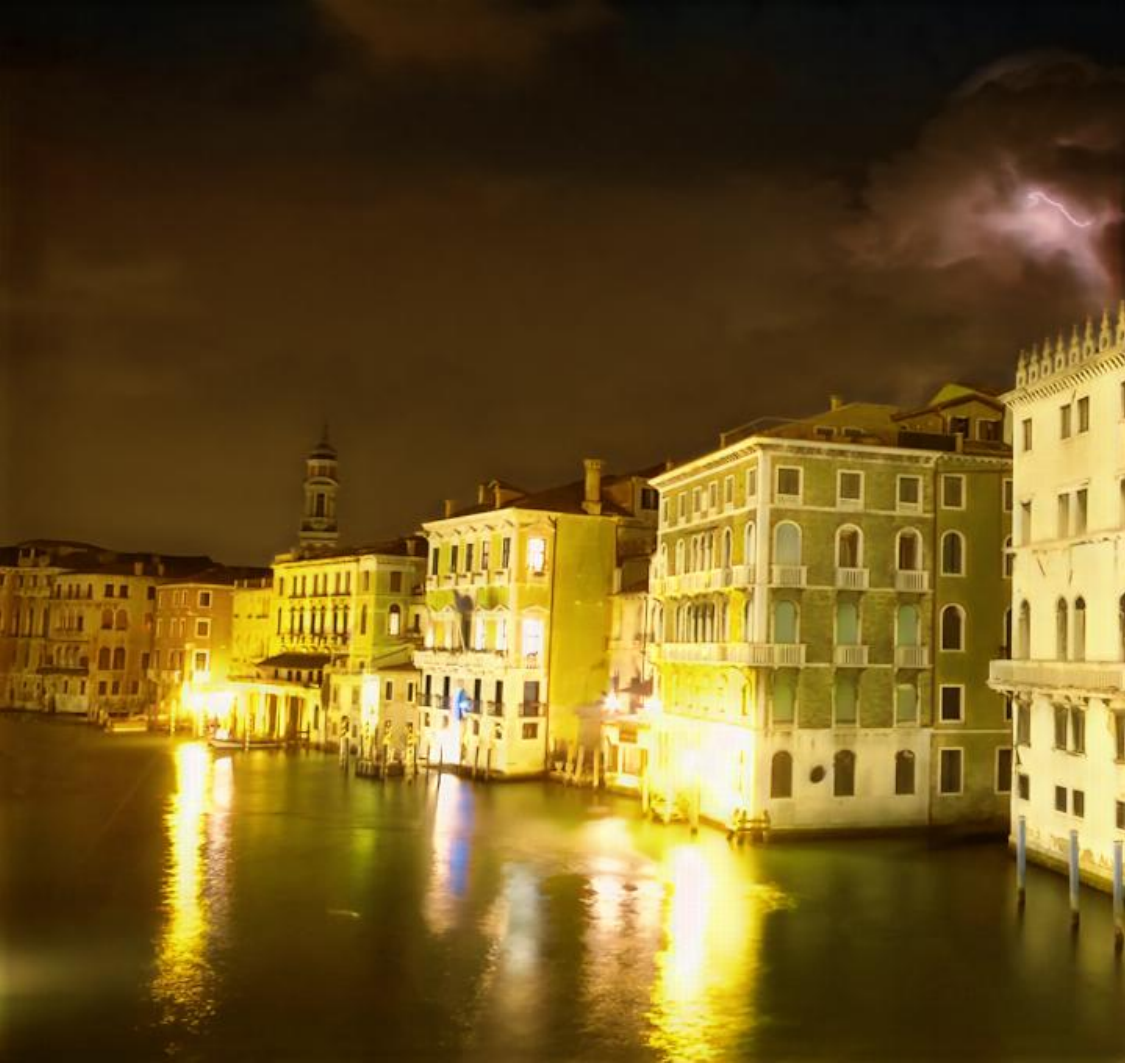}}
 	\centerline{RetinexFormer}
        \vspace{3pt}
 	\centerline{\includegraphics[width=1\textwidth]{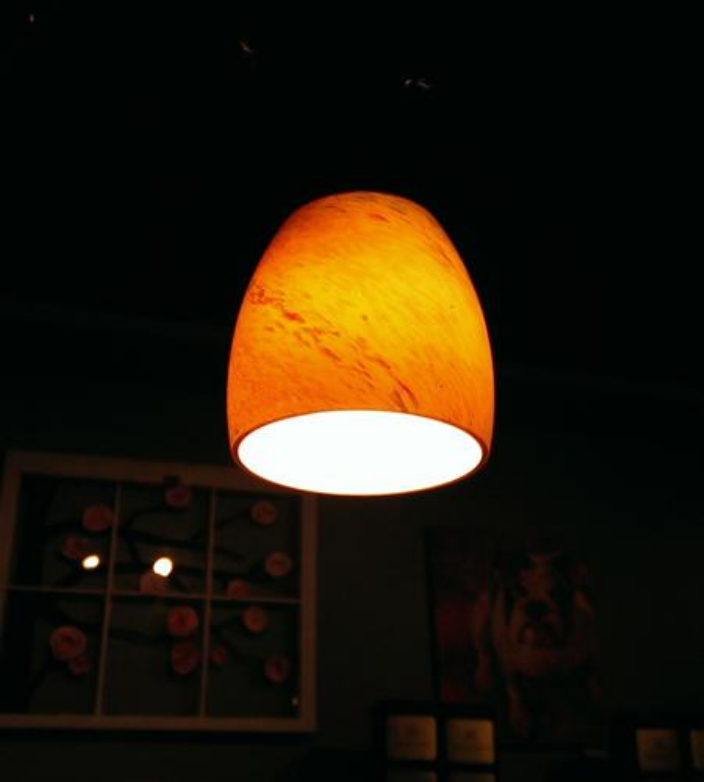}}
 	\centerline{Input}
 	\centerline{\includegraphics[width=1\textwidth]{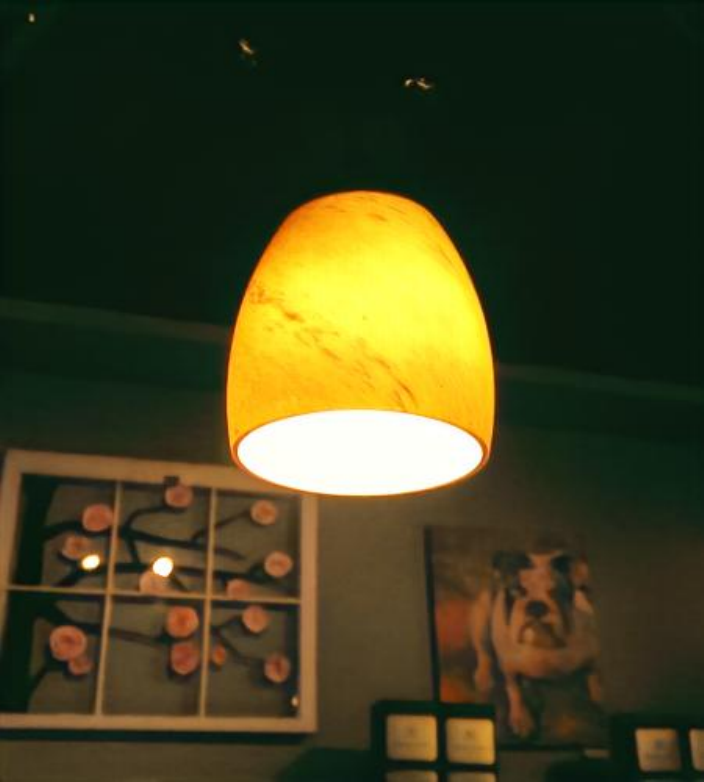}}
 	\centerline{RetinexFormer}
        \vspace{3pt}
\end{minipage}
\begin{minipage}{0.24\linewidth}
        \vspace{3pt}
 	\centerline{\includegraphics[width=1\textwidth]{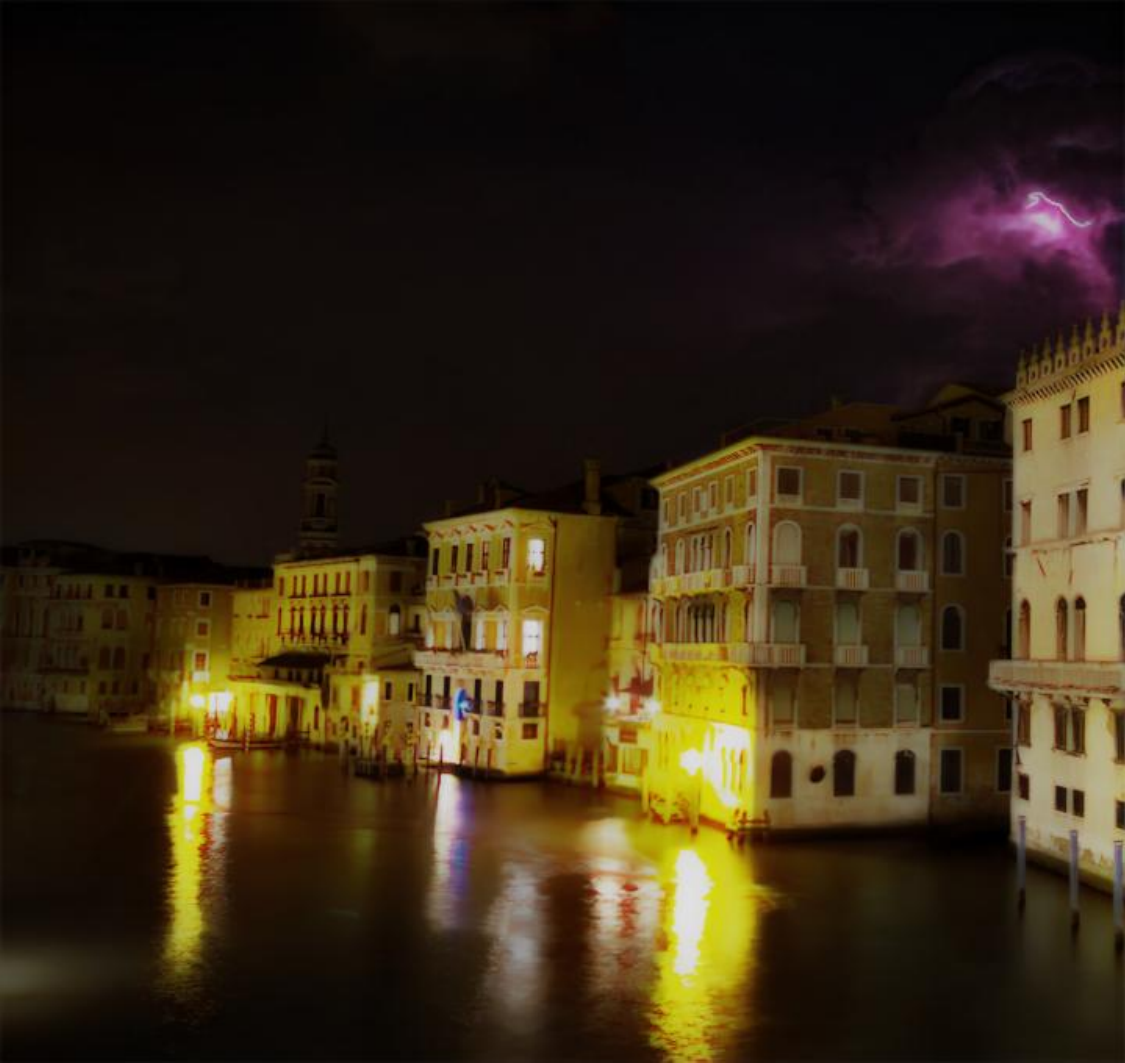}}
 	\centerline{KinD}
        \vspace{2pt}
 	\centerline{\includegraphics[width=1\textwidth]{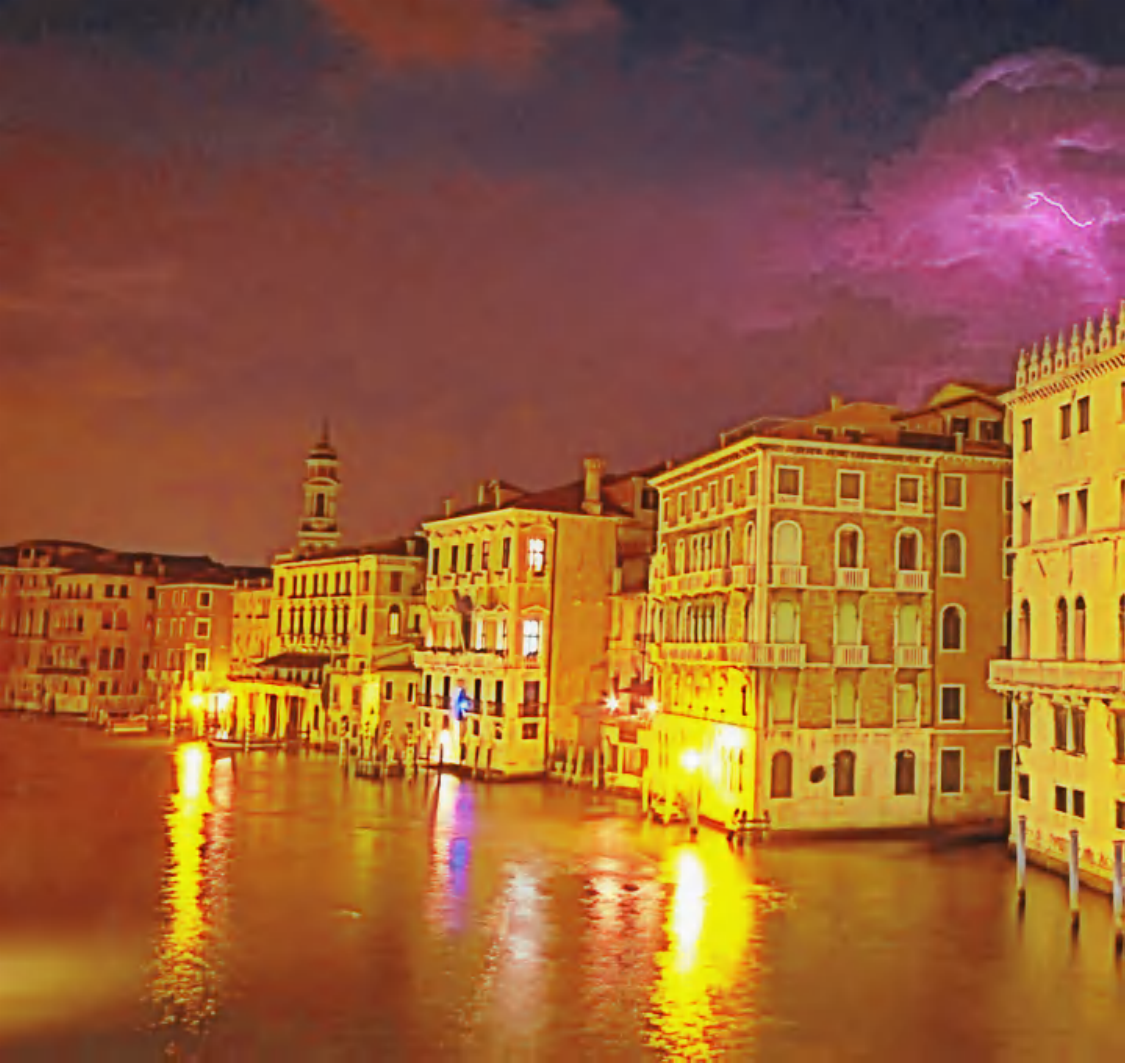}}
 	\centerline{PairLIE}
        \vspace{3pt}
 	\centerline{\includegraphics[width=1\textwidth]{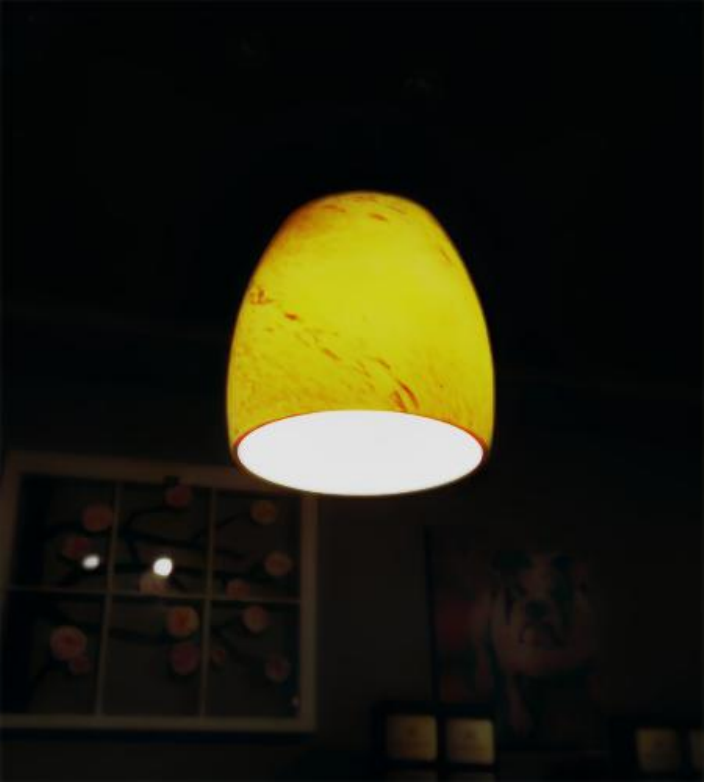}}
 	\centerline{KinD}
        \vspace{2pt}
 	\centerline{\includegraphics[width=1\textwidth]{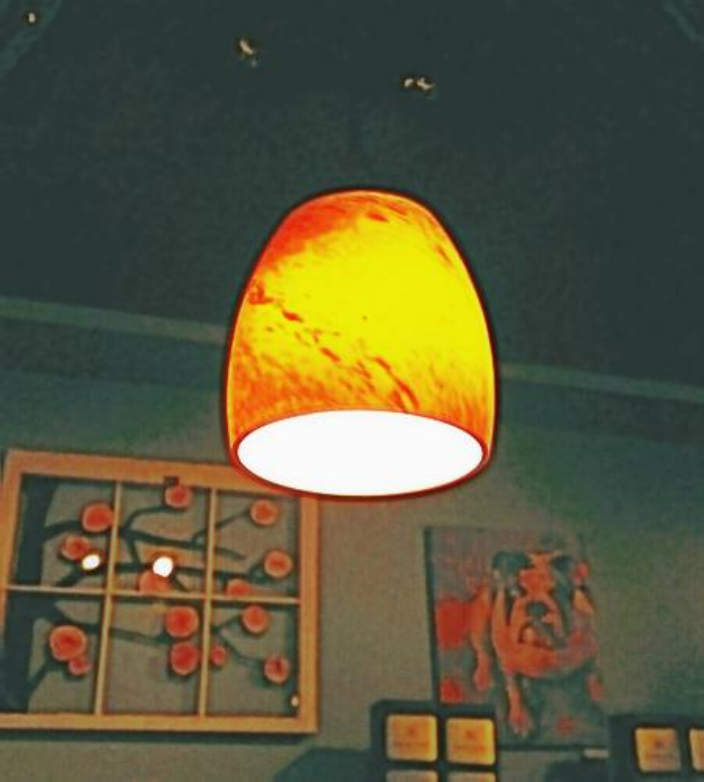}}
 	\centerline{PairLIE}
        \vspace{3pt}
 \end{minipage}
\begin{minipage}{0.24\linewidth}
 	\vspace{3pt}
 	\centerline{\includegraphics[width=1\textwidth]{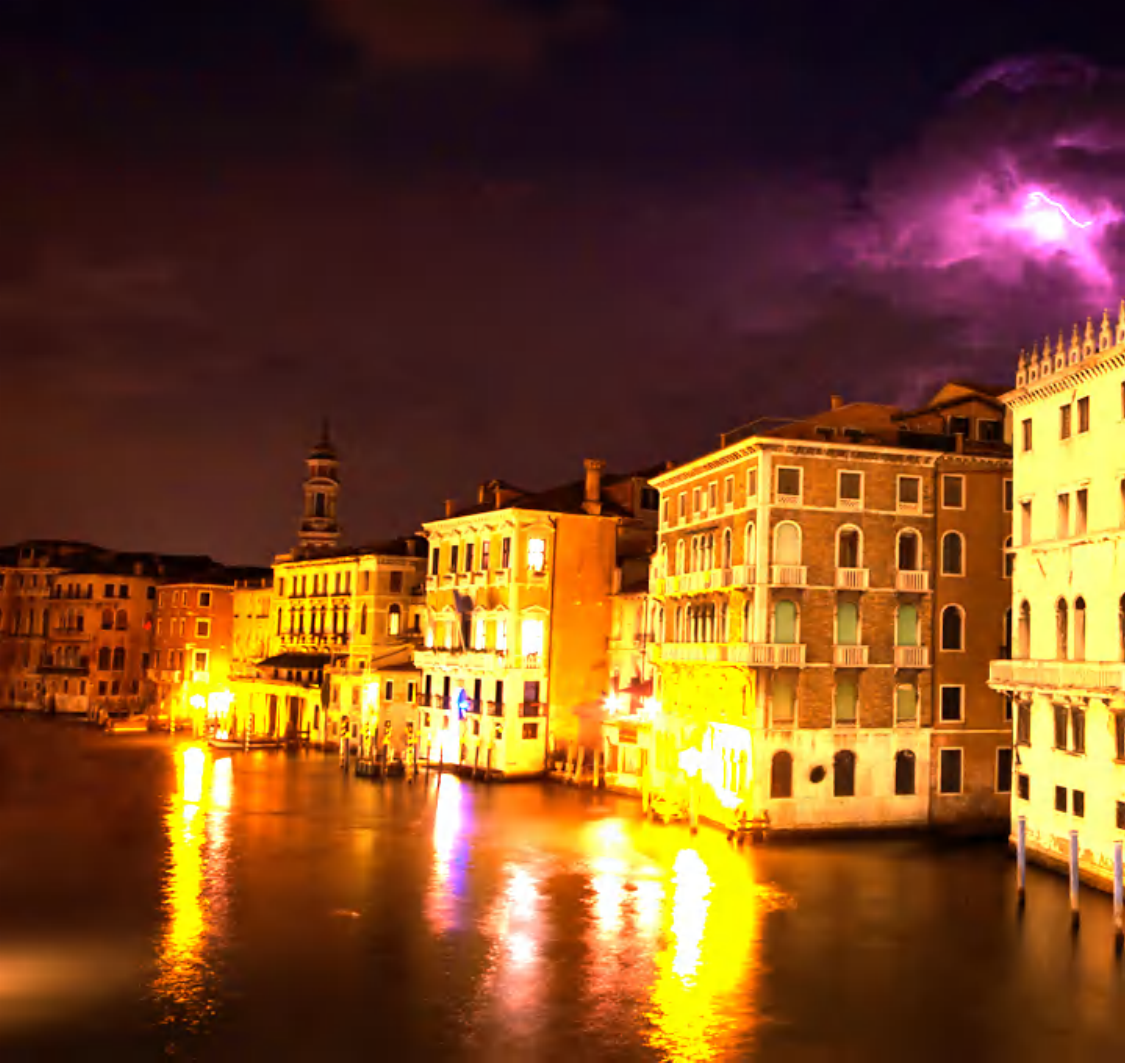}}
 	\centerline{RUAS}
 	\vspace{2pt}
 	\centerline{\includegraphics[width=1\textwidth]{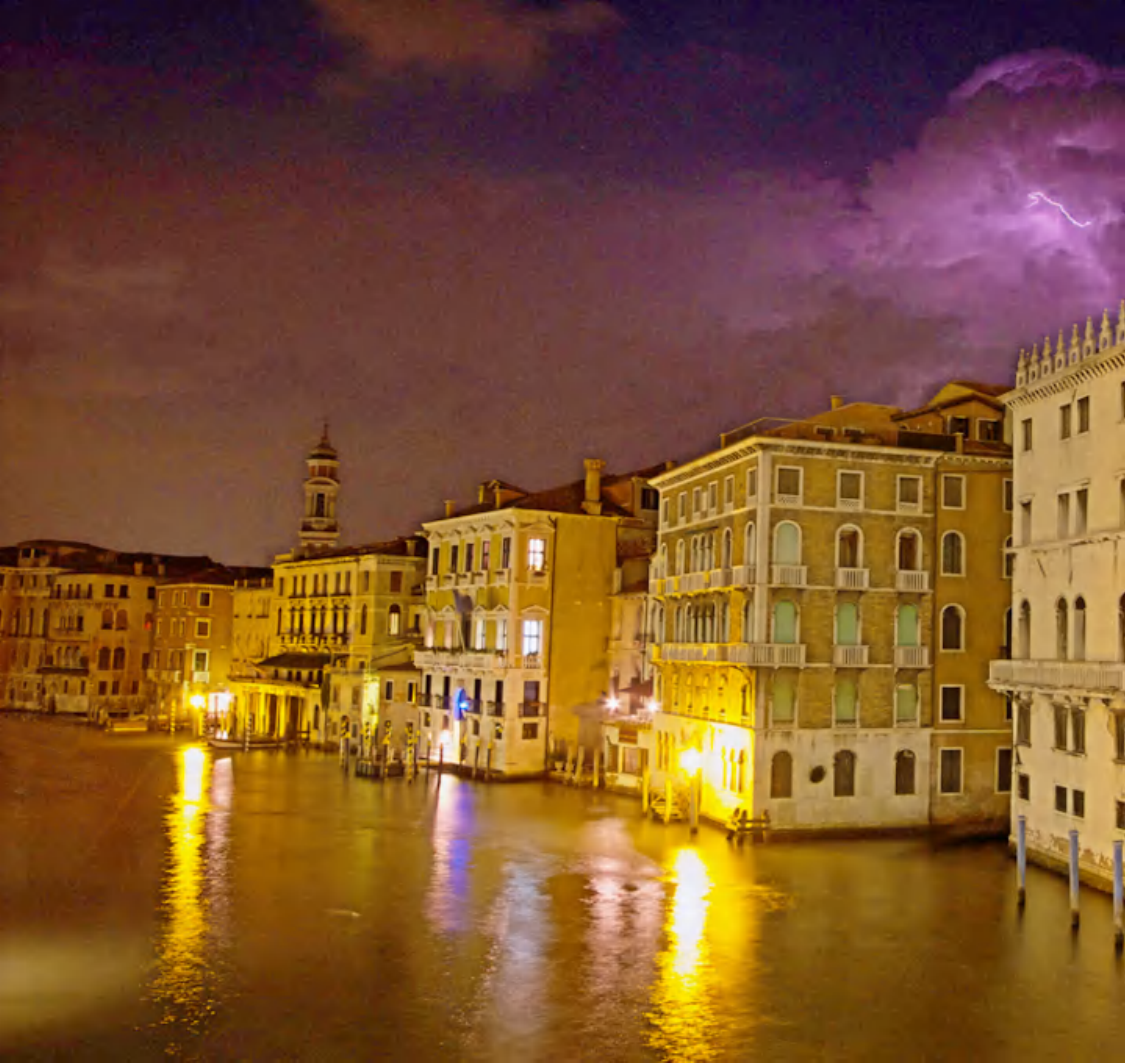}}
 	\centerline{ZeroDCE}
 	\vspace{3pt}
 	\centerline{\includegraphics[width=1\textwidth]{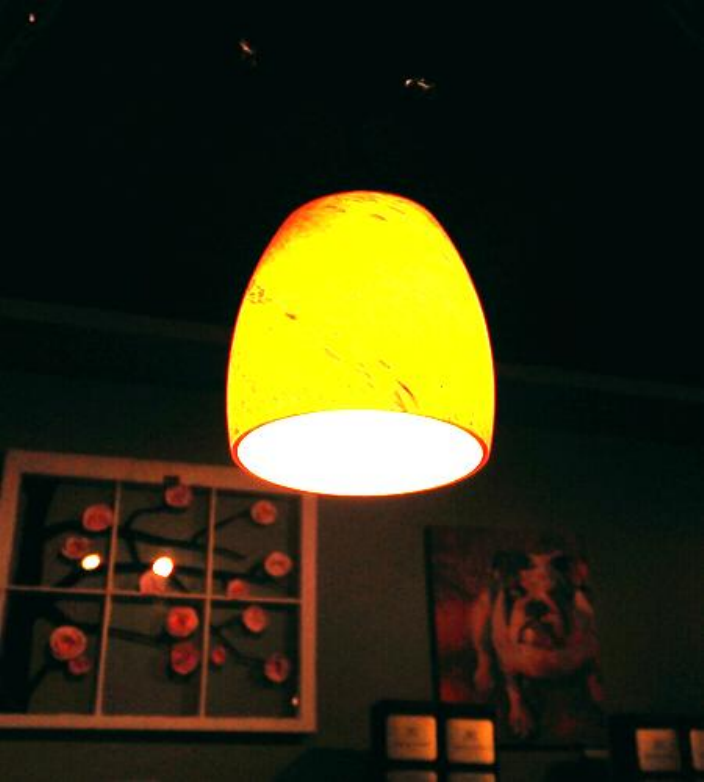}}
 	\centerline{RUAS}
 	\vspace{2pt}
 	\centerline{\includegraphics[width=1\textwidth]{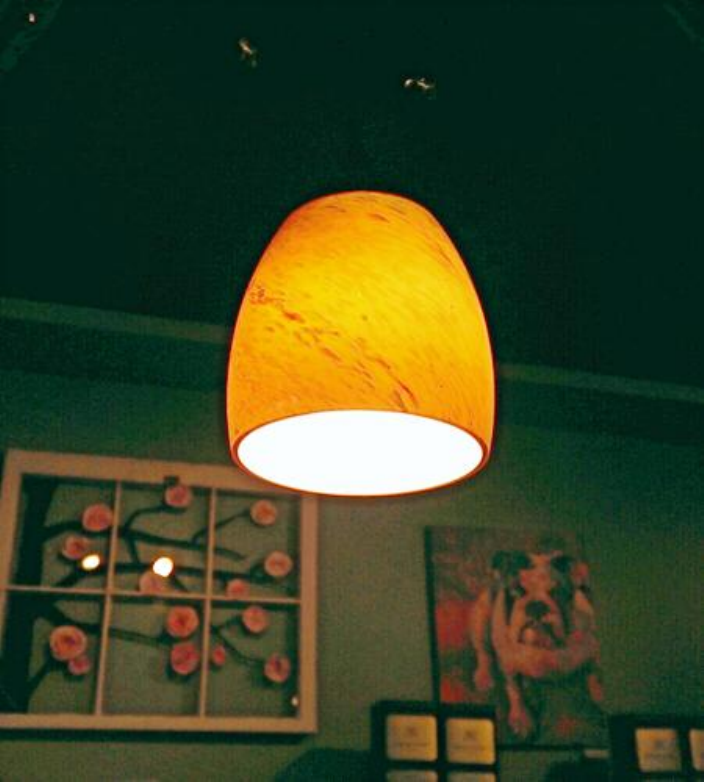}}
 	\centerline{ZeroDCE}
 	\vspace{3pt}
 \end{minipage}
\begin{minipage}{0.24\linewidth}
 	\vspace{3pt}
 	\centerline{\includegraphics[width=1\textwidth]{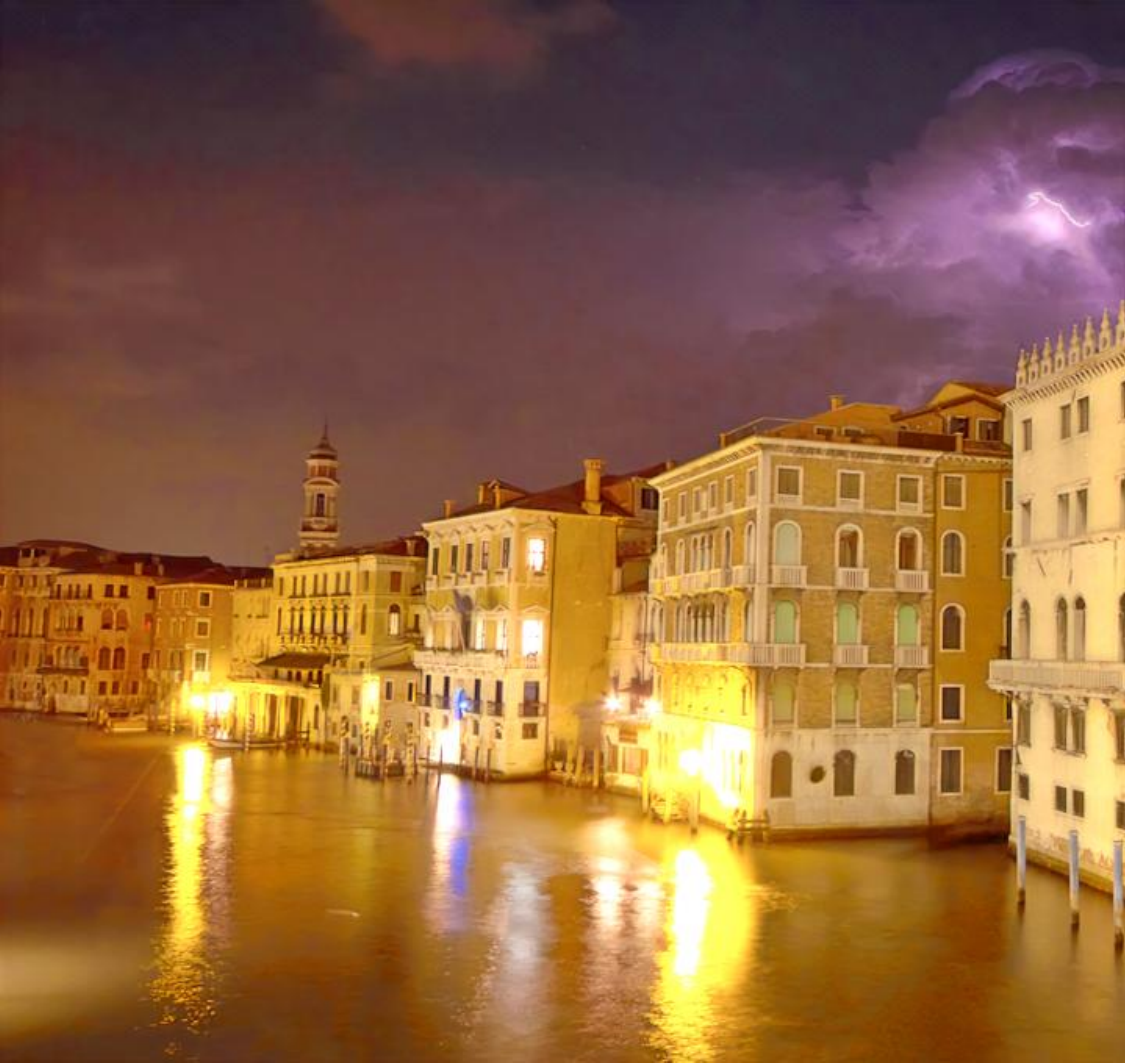}}
 	\centerline{URetinexNet}
 	\vspace{2pt}
 	\centerline{\includegraphics[width=1\textwidth]{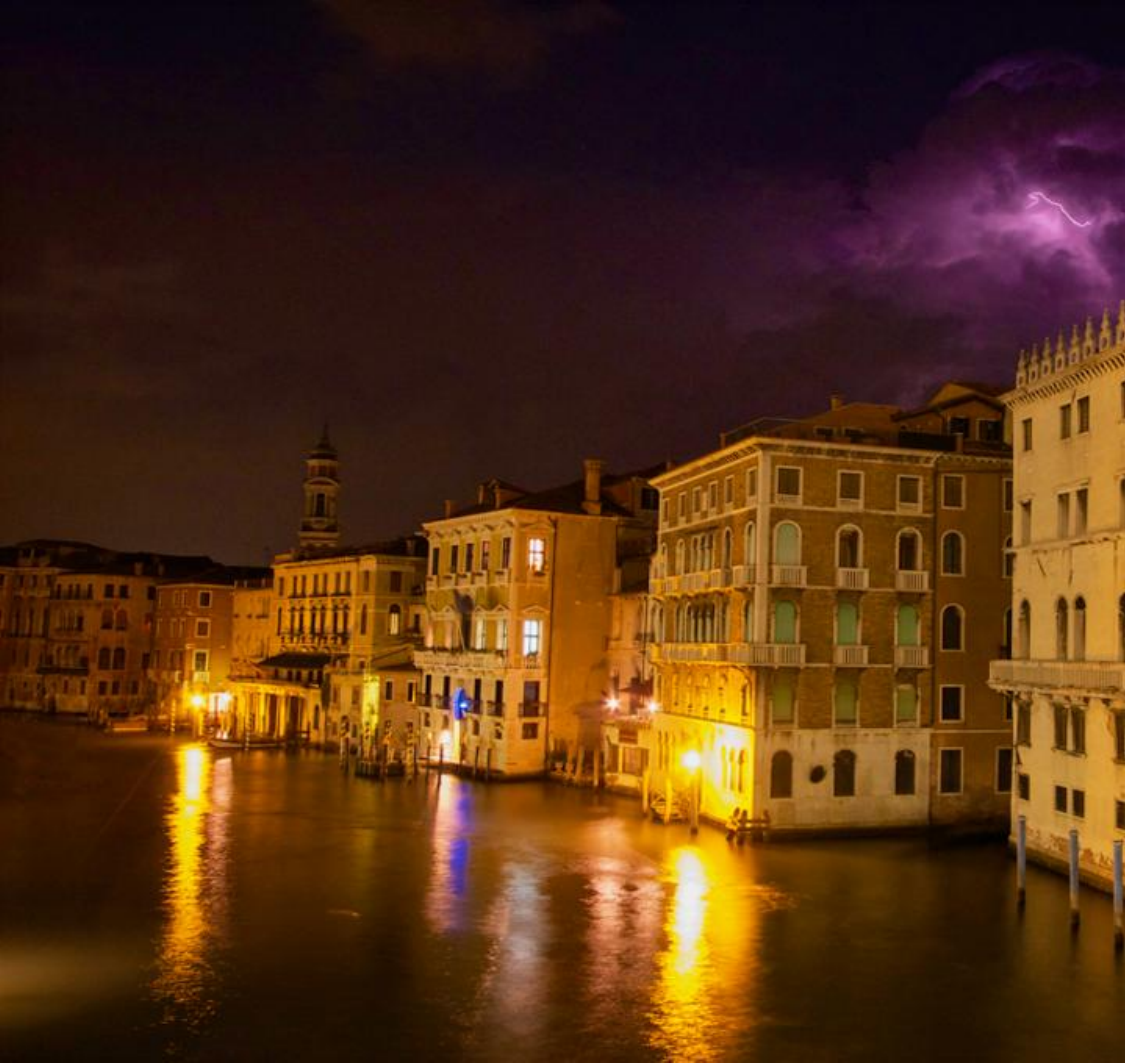}}
 	\centerline{CIDNet}
 	\vspace{3pt}
 	\centerline{\includegraphics[width=1\textwidth]{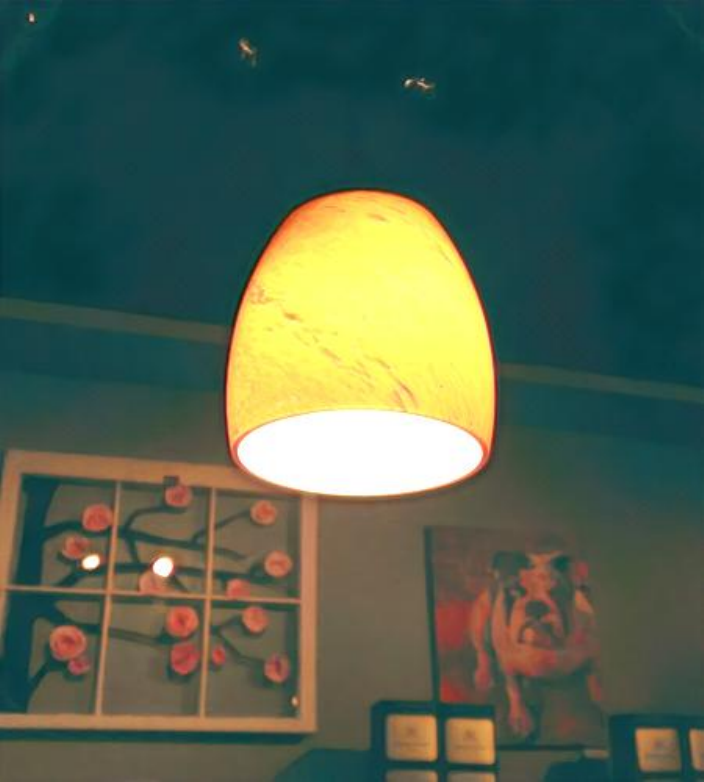}}
 	\centerline{URetinexNet}
 	\vspace{2pt}
 	\centerline{\includegraphics[width=1\textwidth]{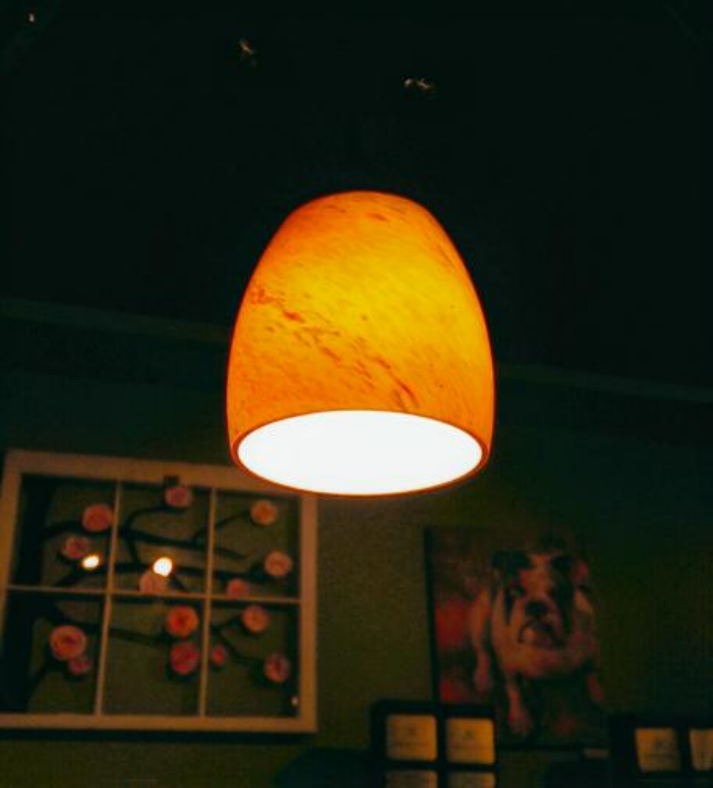}}
 	\centerline{CIDNet}
 	\vspace{3pt}
 \end{minipage}
 \caption{Visual examples for unpaired image enhancement on LIME dataset \cite{LIME} among KinD \cite{KinD}, RUAS \cite{RUAS}, URetinexNet \cite{URetinexNet}, RetinexFormer \cite{RetinexFormer}, PairLIE \cite{PairLIE}, ZeroDCE \cite{Zero-DCE}, and our CIDNet. Our method can suppress the color shift phenomenon while increasing brightness.}
 \label{fig:LIME}
\end{figure*}

\begin{figure*}
\centering
 \begin{minipage}{0.24\linewidth}
 \centering
        \vspace{3pt}
 	\centerline{\includegraphics[width=1\textwidth]{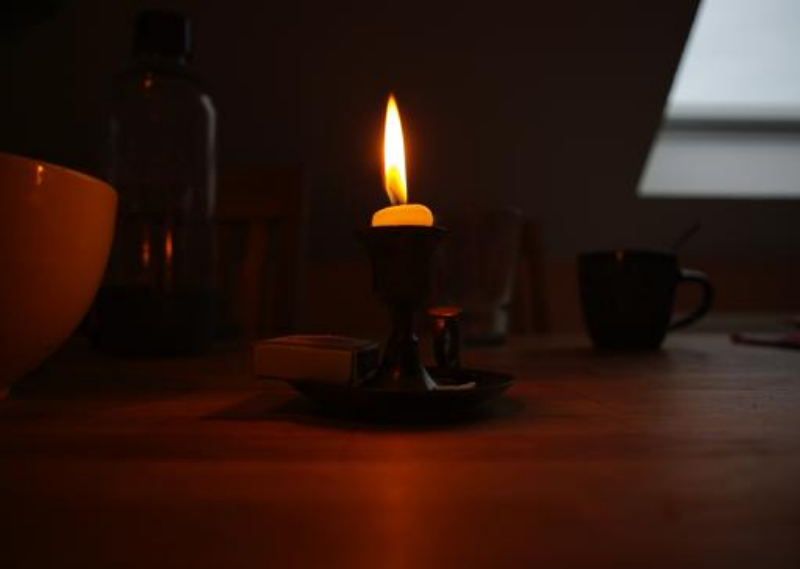}}
 	\centerline{Input}
 	\centerline{\includegraphics[width=1\textwidth]{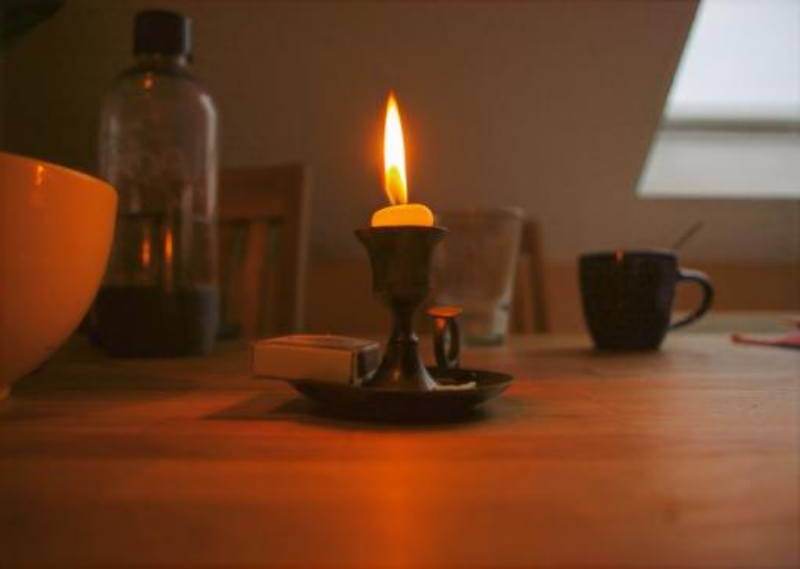}}
 	\centerline{RetinexFormer}
        \vspace{3pt}
 	\centerline{\includegraphics[width=1\textwidth]{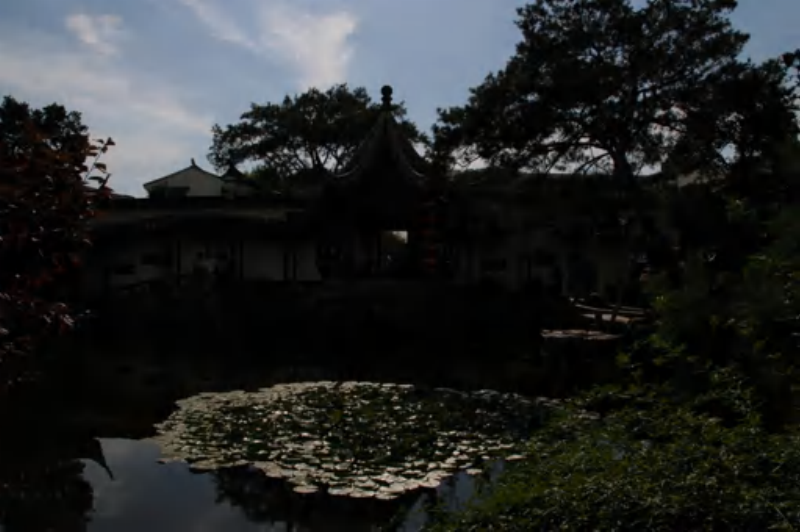}}
 	\centerline{Input}
 	\centerline{\includegraphics[width=1\textwidth]{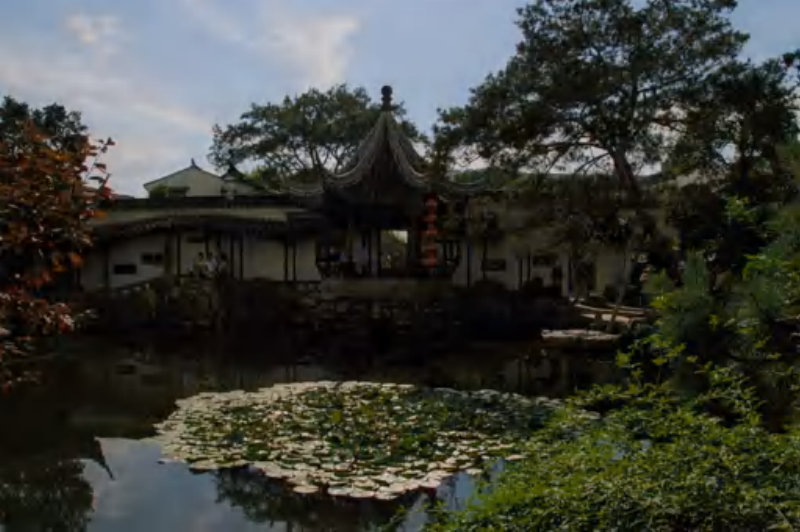}}
 	\centerline{RetinexFormer}
        \vspace{3pt}
 	\centerline{\includegraphics[width=1\textwidth]{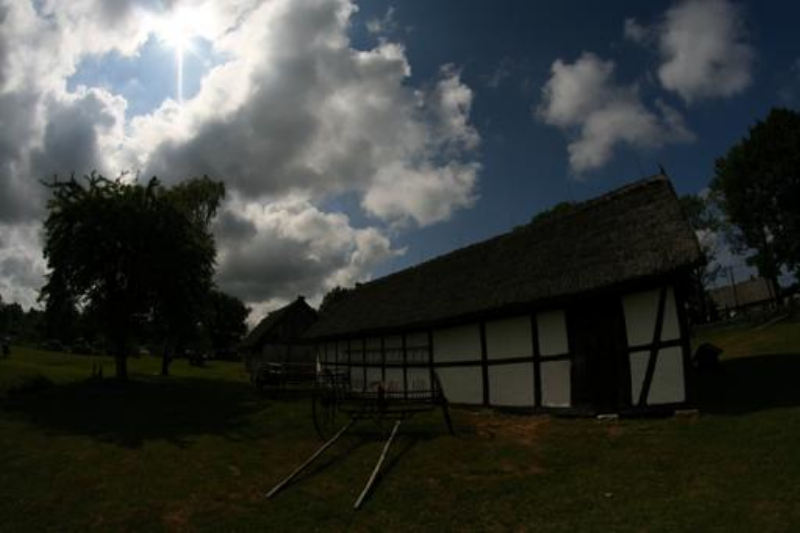}}
 	\centerline{Input}
 	\centerline{\includegraphics[width=1\textwidth]{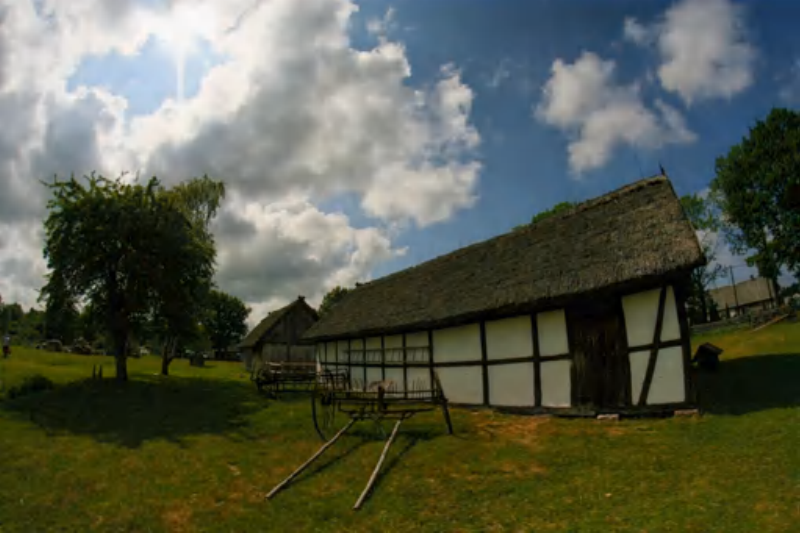}}
 	\centerline{RetinexFormer}
        \vspace{3pt}
\end{minipage}
\begin{minipage}{0.24\linewidth}
        \vspace{3pt}
 	\centerline{\includegraphics[width=1\textwidth]{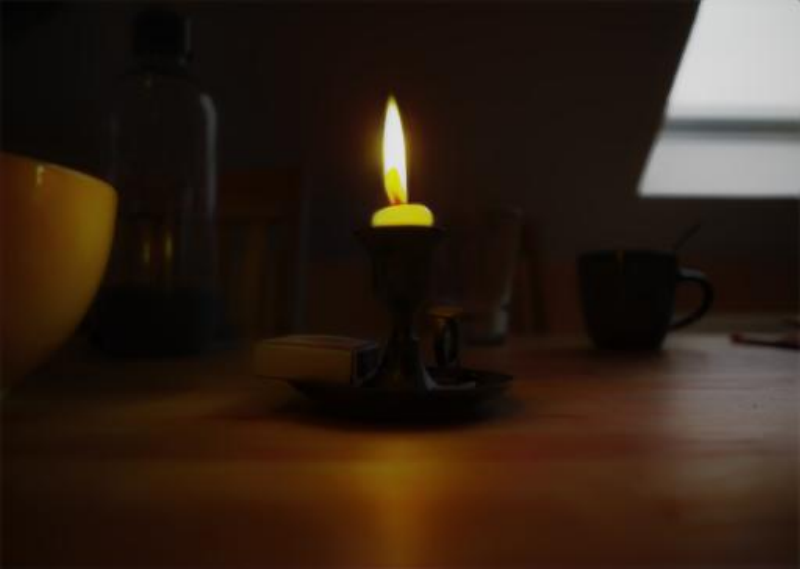}}
 	\centerline{KinD}
        \vspace{2pt}
 	\centerline{\includegraphics[width=1\textwidth]{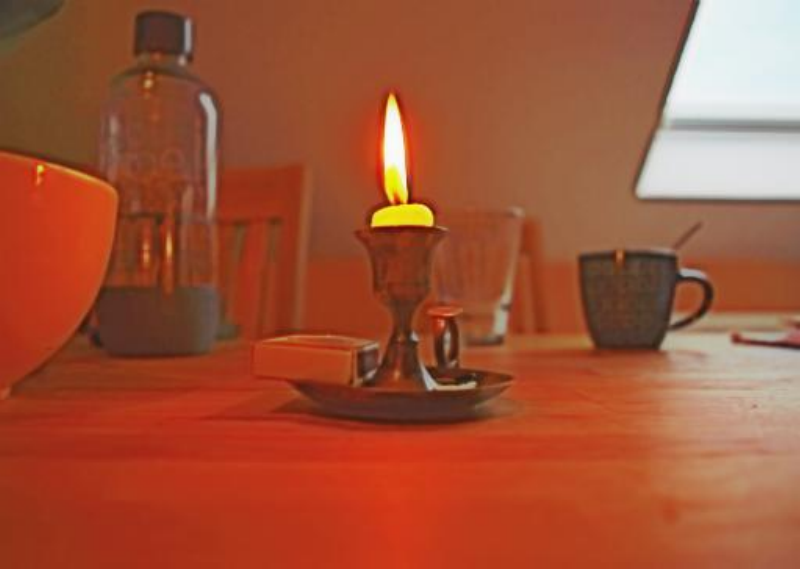}}
 	\centerline{PairLIE}
        \vspace{3pt}
 	\centerline{\includegraphics[width=1\textwidth]{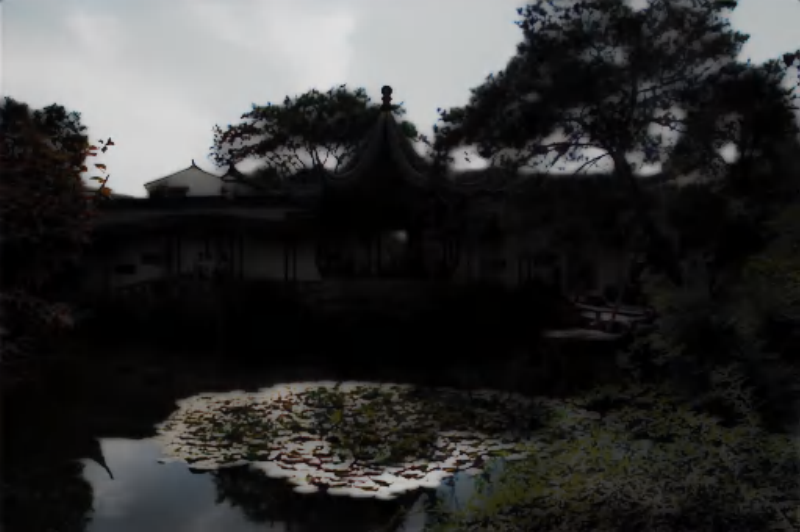}}
 	\centerline{KinD}
        \vspace{2pt}
 	\centerline{\includegraphics[width=1\textwidth]{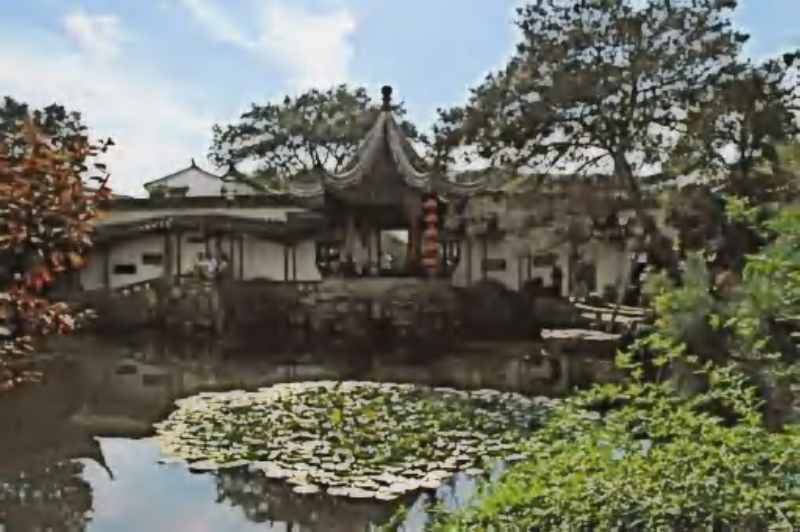}}
 	\centerline{PairLIE}
        \vspace{3pt}
 	\centerline{\includegraphics[width=1\textwidth]{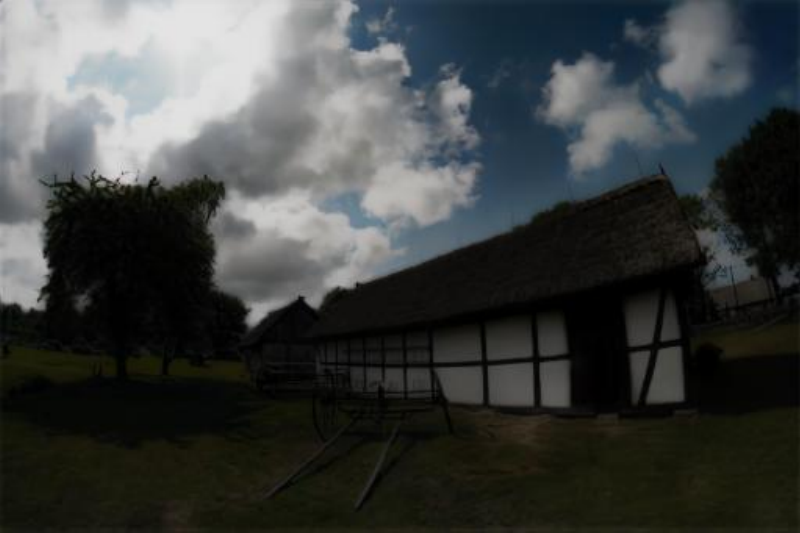}}
 	\centerline{KinD}
        \vspace{2pt}
 	\centerline{\includegraphics[width=1\textwidth]{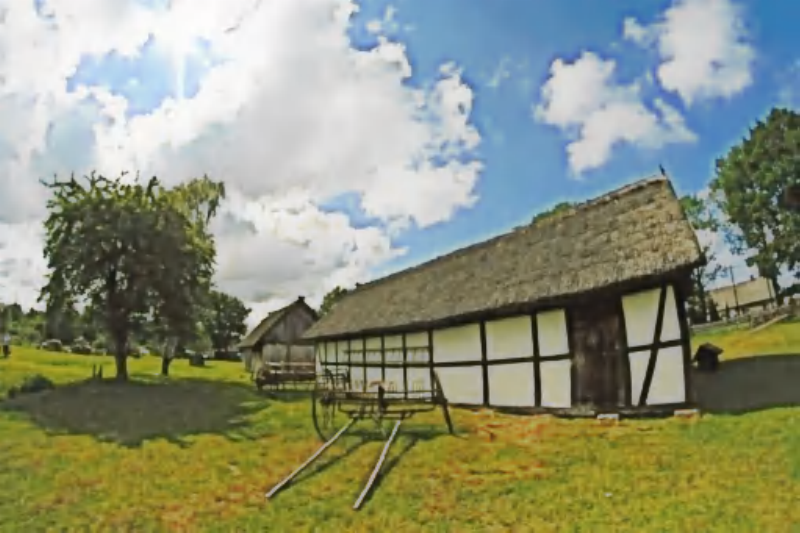}}
 	\centerline{PairLIE}
        \vspace{3pt}
 \end{minipage}
\begin{minipage}{0.24\linewidth}
 	\vspace{3pt}
 	\centerline{\includegraphics[width=1\textwidth]{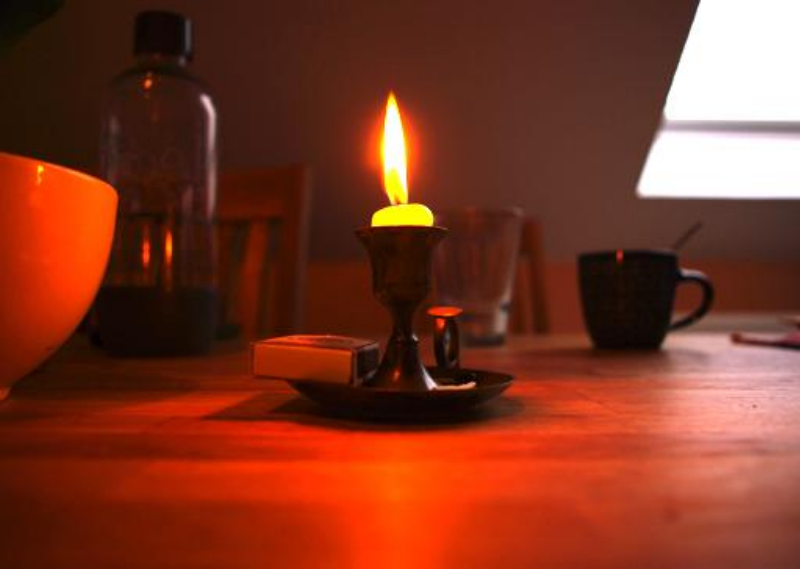}}
 	\centerline{RUAS}
 	\vspace{2pt}
 	\centerline{\includegraphics[width=1\textwidth]{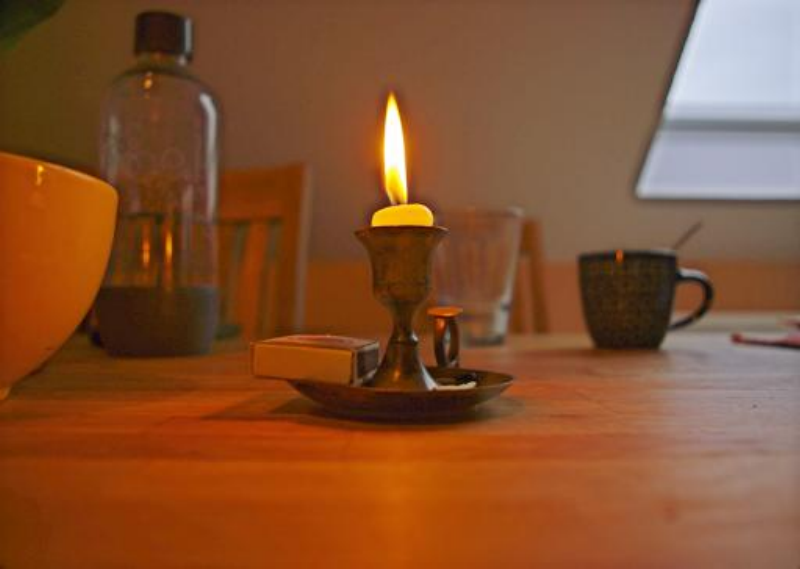}}
 	\centerline{ZeroDCE}
 	\vspace{3pt}
 	\centerline{\includegraphics[width=1\textwidth]{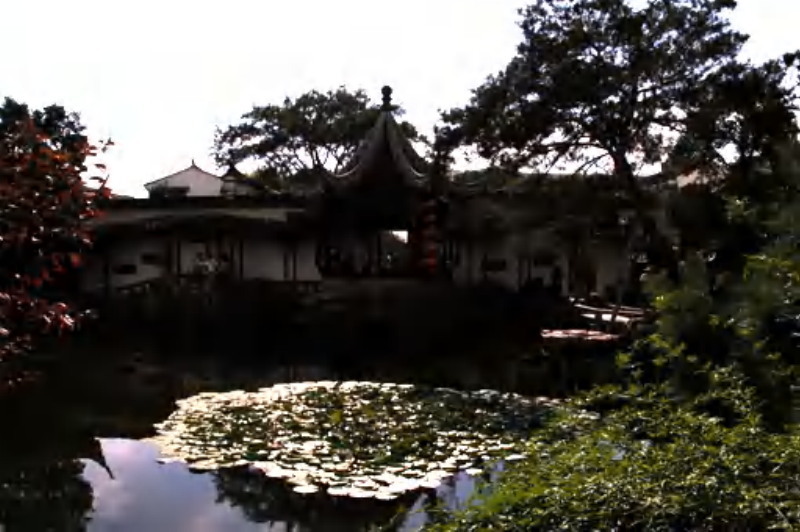}}
 	\centerline{RUAS}
 	\vspace{2pt}
 	\centerline{\includegraphics[width=1\textwidth]{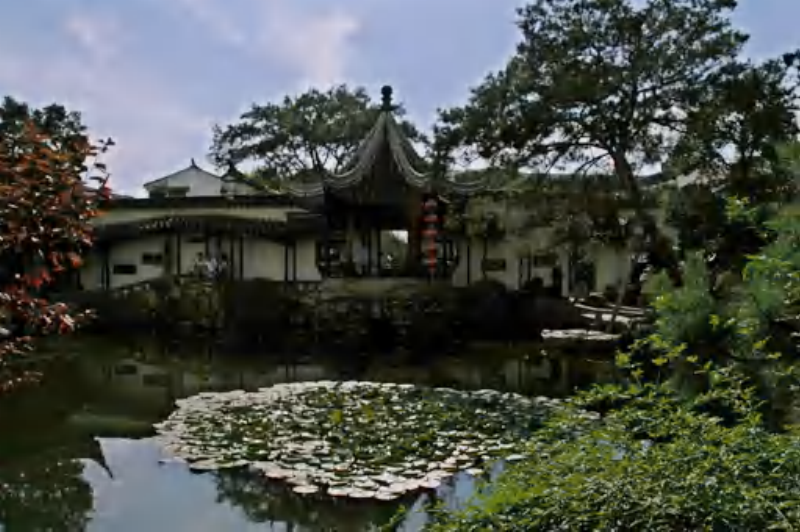}}
 	\centerline{ZeroDCE}
 	\vspace{3pt}
 	\centerline{\includegraphics[width=1\textwidth]{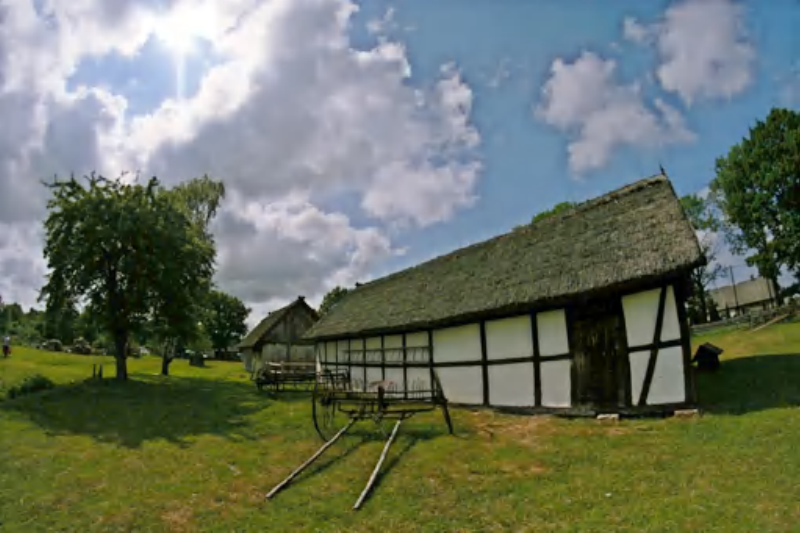}}
 	\centerline{RUAS}
 	\vspace{2pt}
 	\centerline{\includegraphics[width=1\textwidth]{pic/MEF/3/c-ZeroDCE.pdf}}
 	\centerline{ZeroDCE}
        \vspace{3pt}
 \end{minipage}
\begin{minipage}{0.24\linewidth}
 	\vspace{3pt}
 	\centerline{\includegraphics[width=1\textwidth]{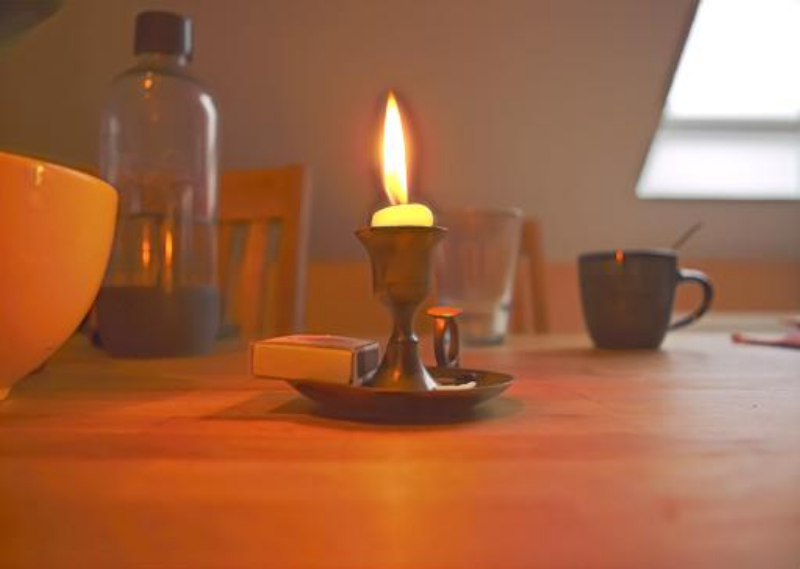}}
 	\centerline{URetinexNet}
 	\vspace{2pt}
 	\centerline{\includegraphics[width=1\textwidth]{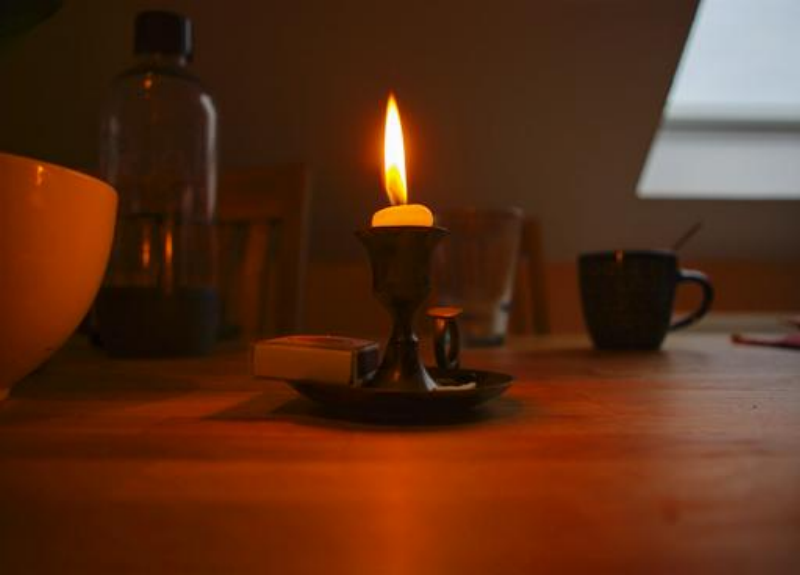}}
 	\centerline{CIDNet}
 	\vspace{3pt}
 	\centerline{\includegraphics[width=1\textwidth]{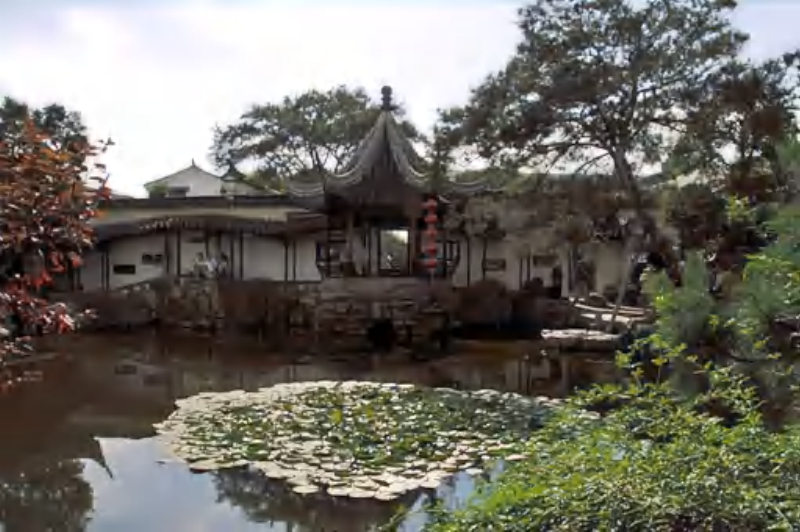}}
 	\centerline{URetinexNet}
 	\vspace{2pt}
 	\centerline{\includegraphics[width=1\textwidth]{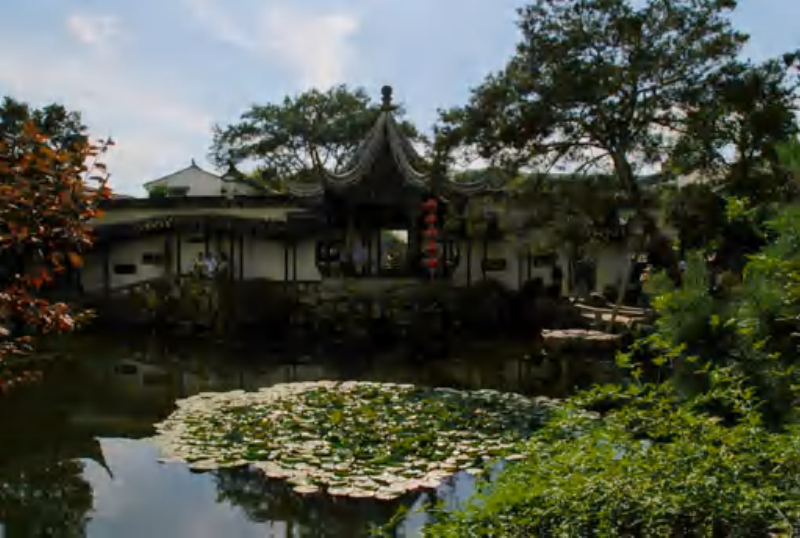}}
 	\centerline{CIDNet}
 	\vspace{3pt}
 	\centerline{\includegraphics[width=1\textwidth]{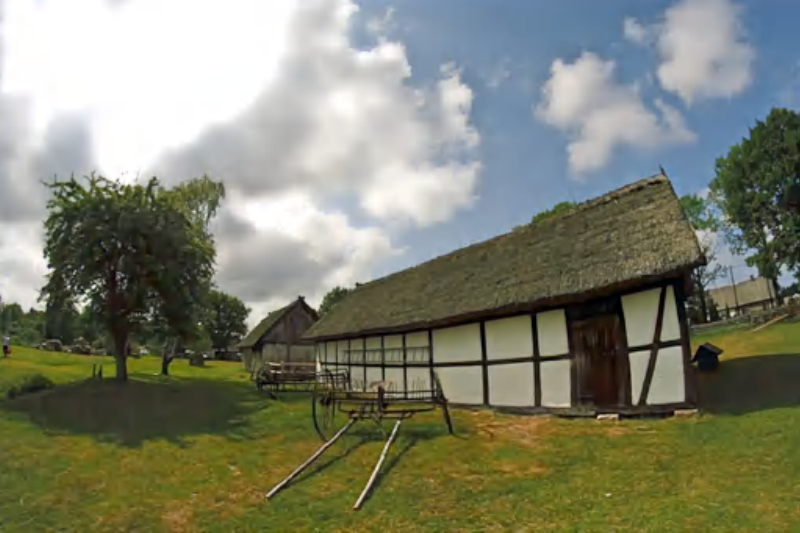}}
 	\centerline{URetinexNet}
 	\vspace{2pt}
 	\centerline{\includegraphics[width=1\textwidth]{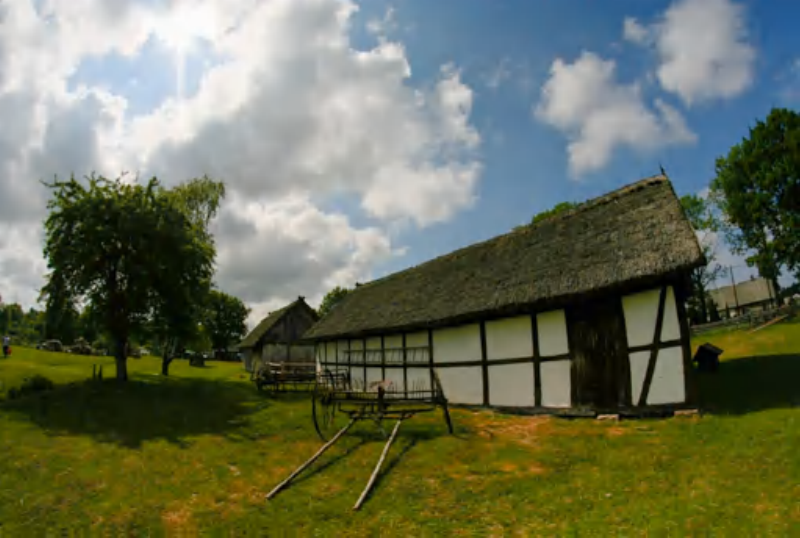}}
 	\centerline{CIDNet}
        \vspace{3pt}
 \end{minipage}
 \caption{Visual examples for unpaired image enhancement on MEF dataset \cite{MEF} among KinD \cite{KinD}, RUAS \cite{RUAS}, URetinexNet \cite{URetinexNet}, RetinexFormer \cite{RetinexFormer}, PairLIE \cite{PairLIE}, ZeroDCE \cite{Zero-DCE}, and our CIDNet. Our method maintains the reality colors while enhancing the brightness to a suitable threshold.}
 \label{fig:MEF}
\end{figure*}

\begin{figure*}
\centering
 \begin{minipage}{0.24\linewidth}
 \centering
        \vspace{3pt}
 	\centerline{\includegraphics[width=1\textwidth]{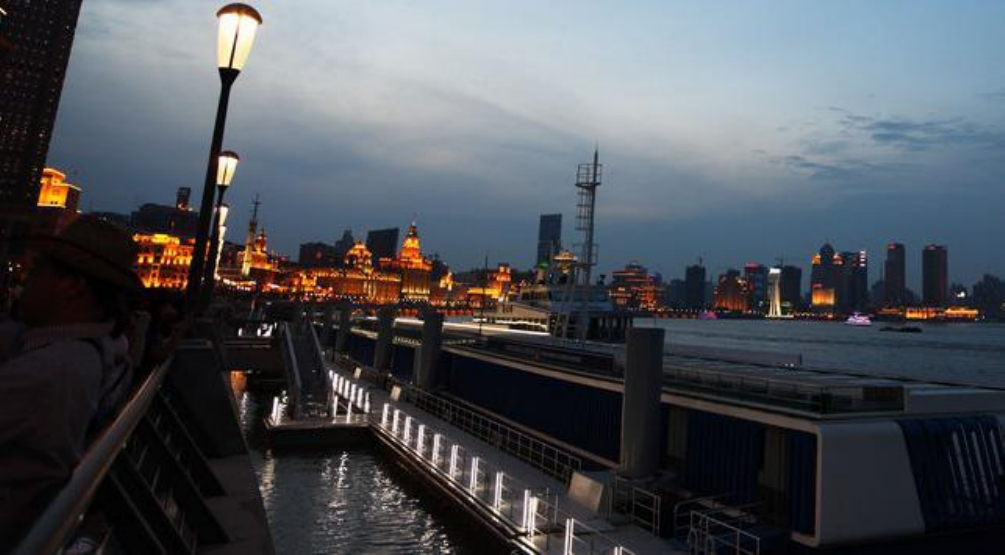}}
 	\centerline{Input}
 	\centerline{\includegraphics[width=1\textwidth]{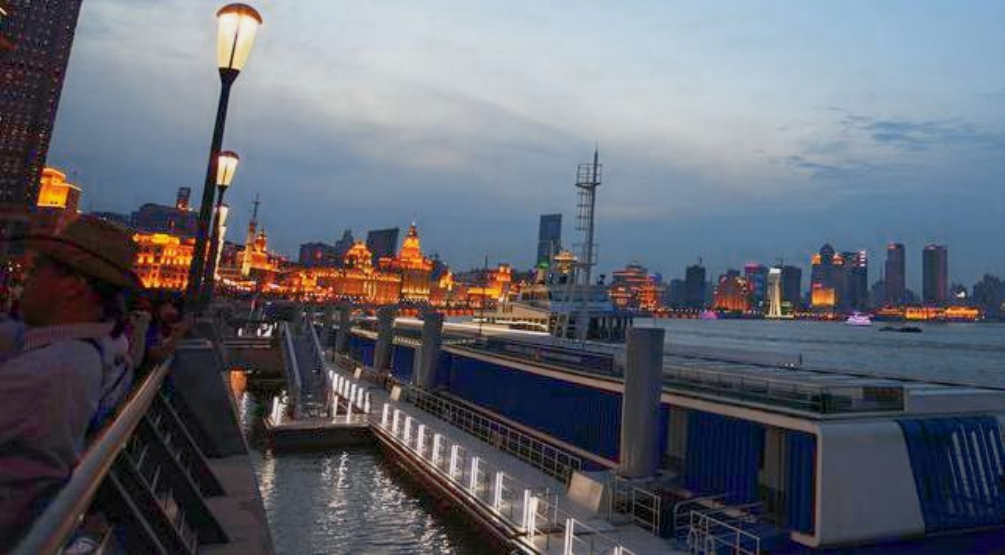}}
 	\centerline{RetinexFormer}
        \vspace{3pt}
 	\centerline{\includegraphics[width=1\textwidth]{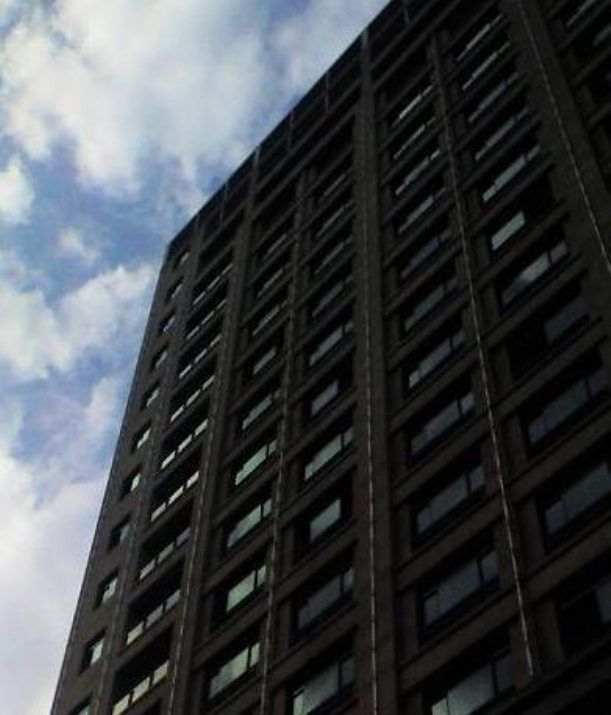}}
 	\centerline{Input}
 	\centerline{\includegraphics[width=1\textwidth]{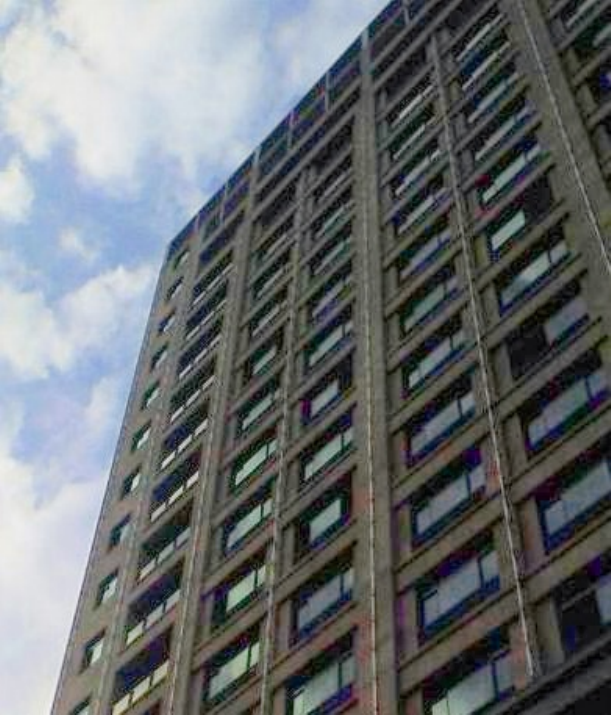}}
 	\centerline{RetinexFormer}
        \vspace{3pt}
\end{minipage}
\begin{minipage}{0.24\linewidth}
        \vspace{3pt}
 	\centerline{\includegraphics[width=1\textwidth]{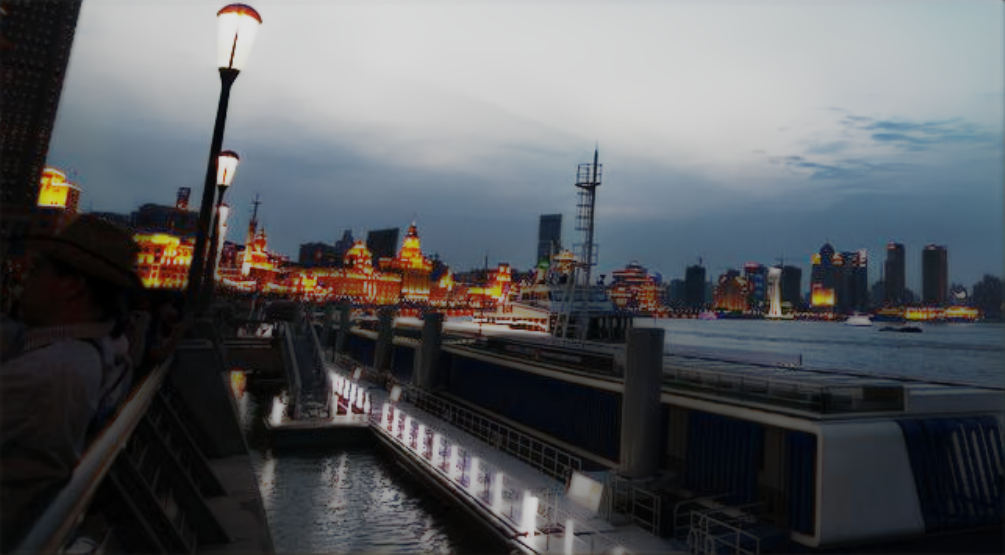}}
 	\centerline{KinD}
        \vspace{2pt}
 	\centerline{\includegraphics[width=1\textwidth]{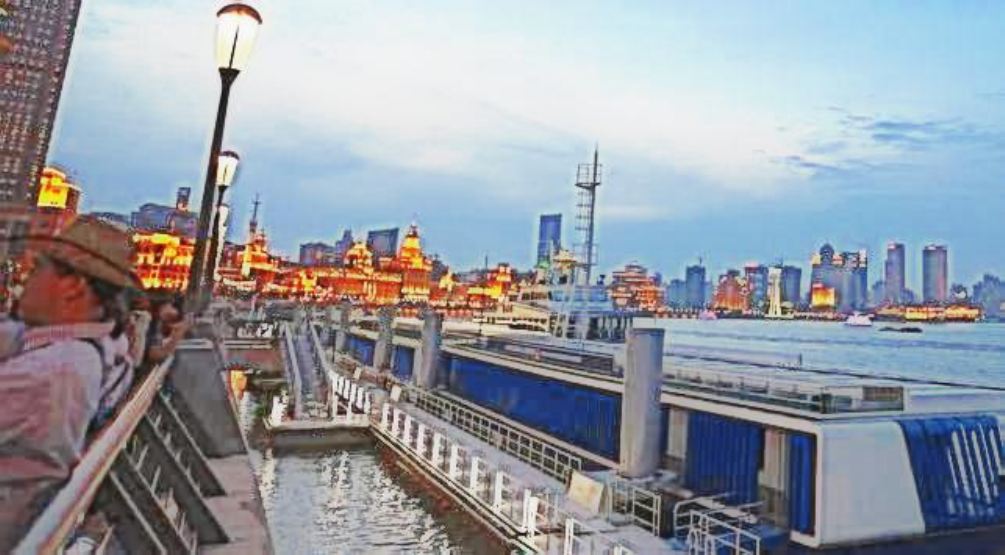}}
 	\centerline{PairLIE}
        \vspace{3pt}
 	\centerline{\includegraphics[width=1\textwidth]{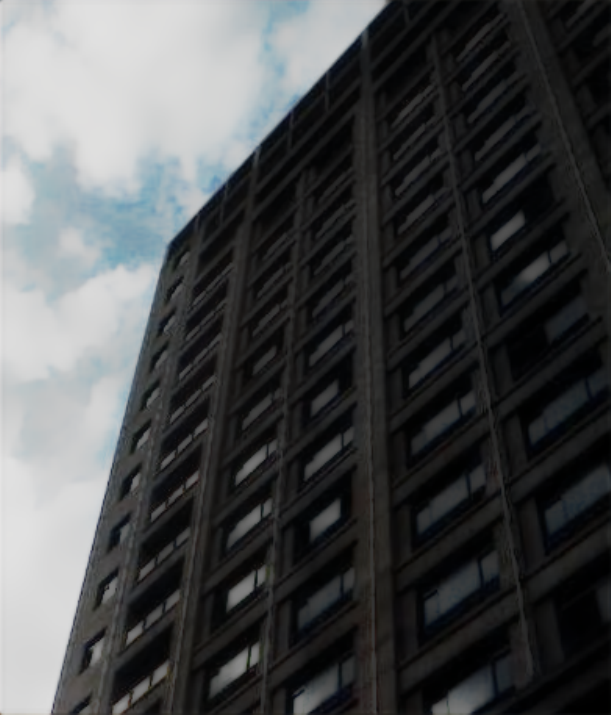}}
 	\centerline{KinD}
        \vspace{2pt}
 	\centerline{\includegraphics[width=1\textwidth]{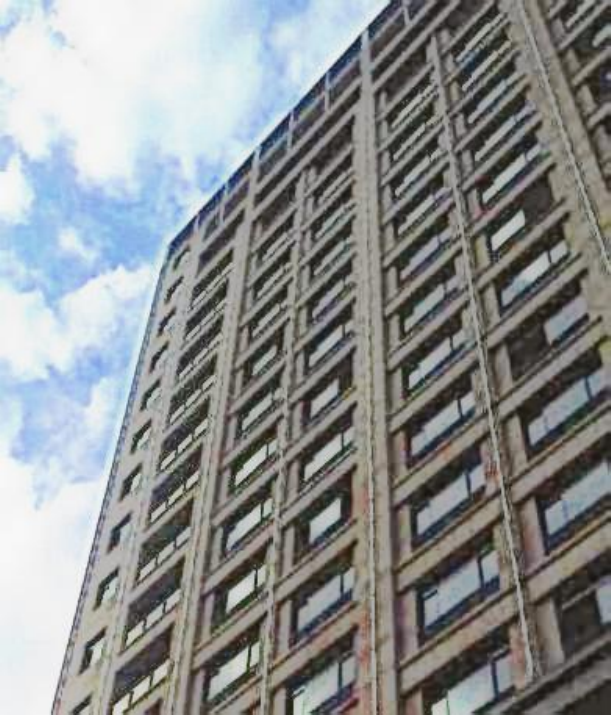}}
 	\centerline{PairLIE}
        \vspace{3pt}
 \end{minipage}
\begin{minipage}{0.24\linewidth}
 	\vspace{3pt}
 	\centerline{\includegraphics[width=1\textwidth]{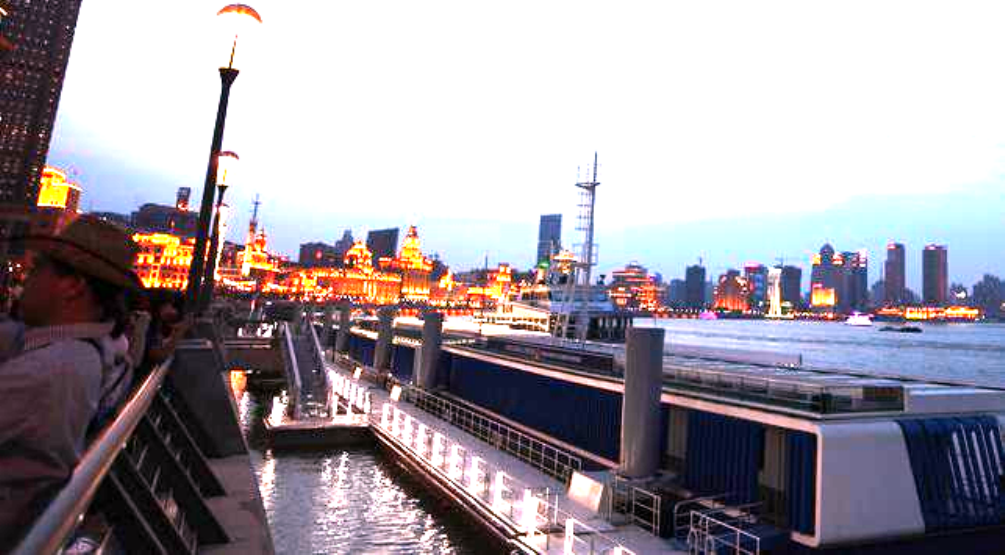}}
 	\centerline{RUAS}
 	\vspace{2pt}
 	\centerline{\includegraphics[width=1\textwidth]{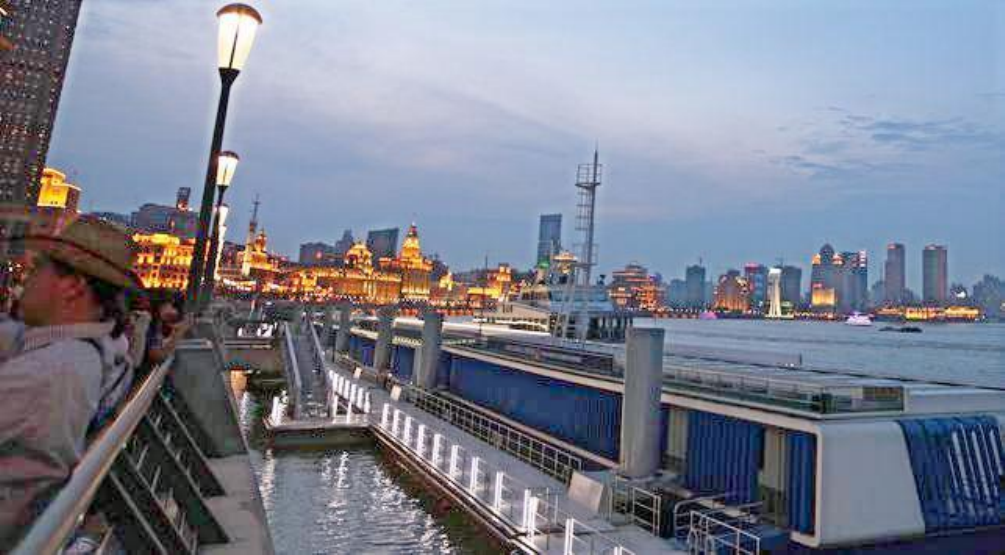}}
 	\centerline{ZeroDCE}
 	\vspace{3pt}
 	\centerline{\includegraphics[width=1\textwidth]{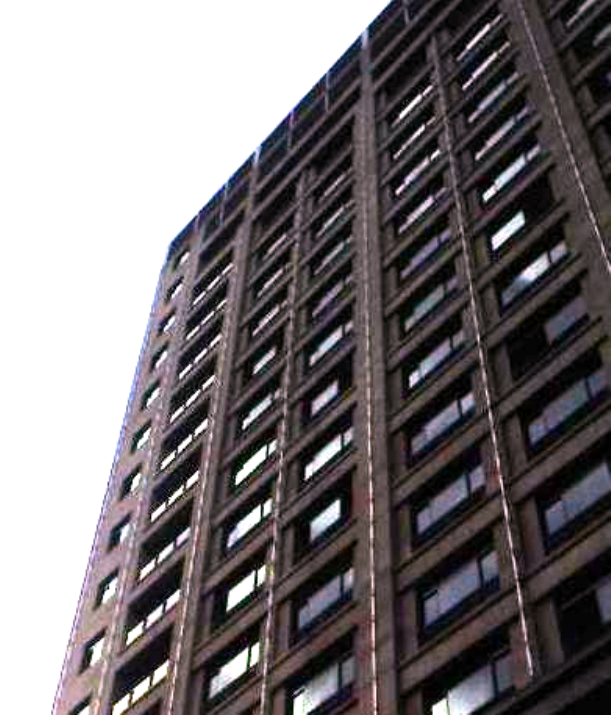}}
 	\centerline{RUAS}
 	\vspace{2pt}
 	\centerline{\includegraphics[width=1\textwidth]{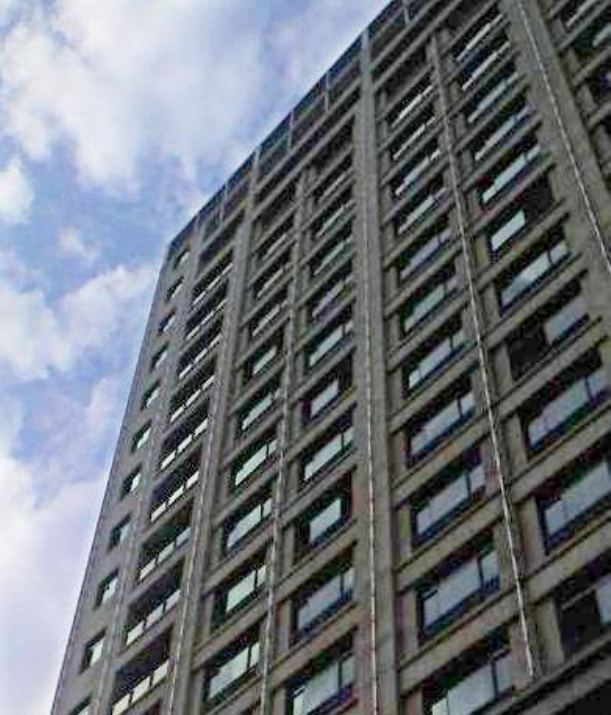}}
 	\centerline{ZeroDCE}
 	\vspace{3pt}
 \end{minipage}
\begin{minipage}{0.24\linewidth}
 	\vspace{3pt}
 	\centerline{\includegraphics[width=1\textwidth]{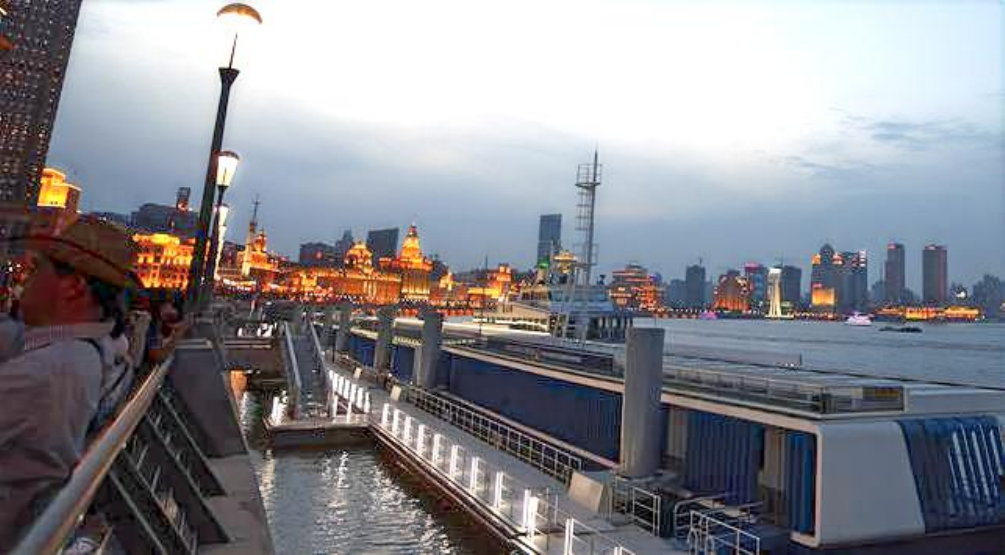}}
 	\centerline{URetinexNet}
 	\vspace{2pt}
 	\centerline{\includegraphics[width=1\textwidth]{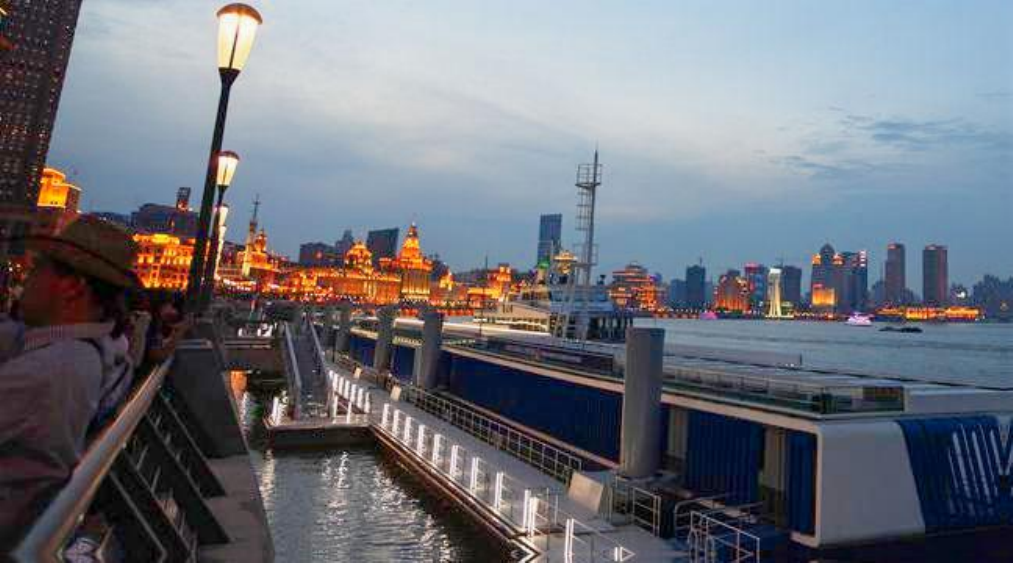}}
 	\centerline{CIDNet}
 	\vspace{3pt}
 	\centerline{\includegraphics[width=1\textwidth]{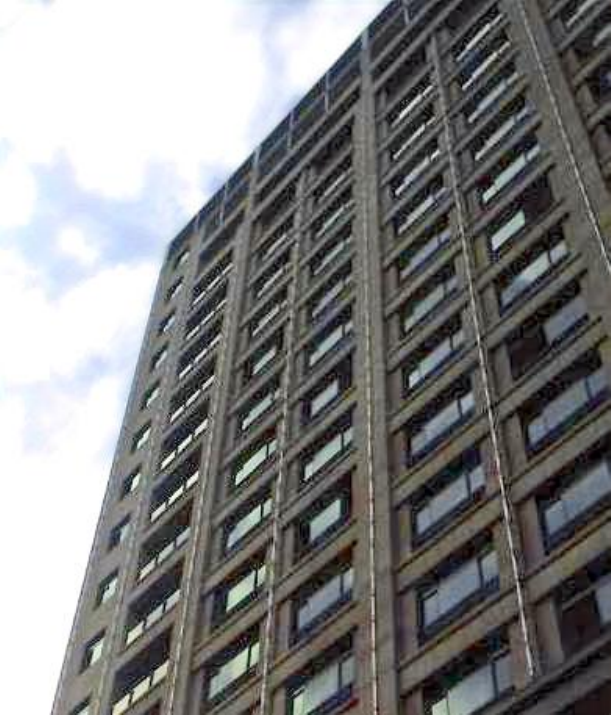}}
 	\centerline{URetinexNet}
 	\vspace{2pt}
 	\centerline{\includegraphics[width=1\textwidth]{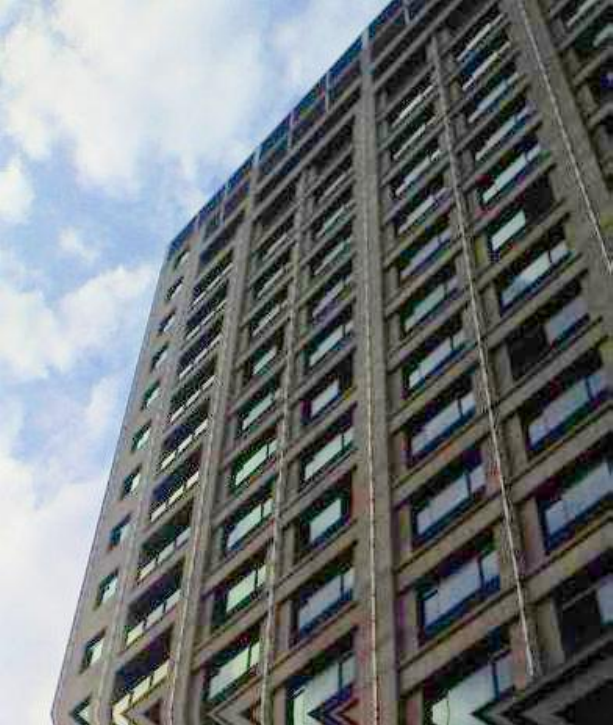}}
 	\centerline{CIDNet}
 	\vspace{3pt}
 \end{minipage}
 \caption{Visual examples for unpaired image enhancement on NPE dataset \cite{NPE} among KinD \cite{KinD}, RUAS \cite{RUAS}, URetinexNet \cite{URetinexNet}, RetinexFormer \cite{RetinexFormer}, PairLIE \cite{PairLIE}, ZeroDCE \cite{Zero-DCE}, and our CIDNet. Our method aligns more closely with the characteristics observed in real scenes with long exposures.}
 \label{fig:NPE}
\end{figure*}

\begin{figure*}
\centering
 \begin{minipage}{0.32\linewidth}
 \centering
        \vspace{3pt}
 	\centerline{\includegraphics[width=1\textwidth]{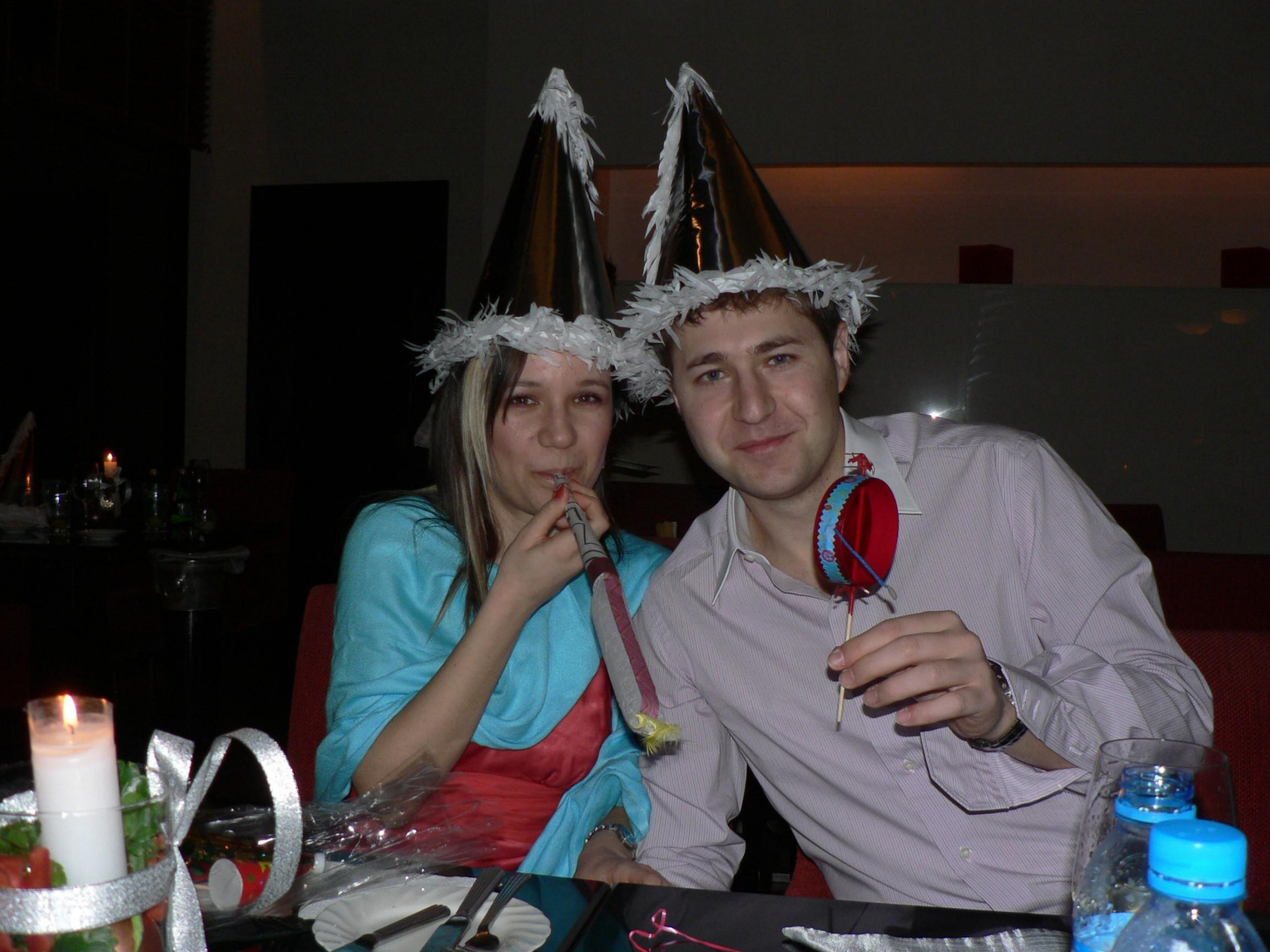}}
 	\centerline{Input}
 	\centerline{\includegraphics[width=1\textwidth]{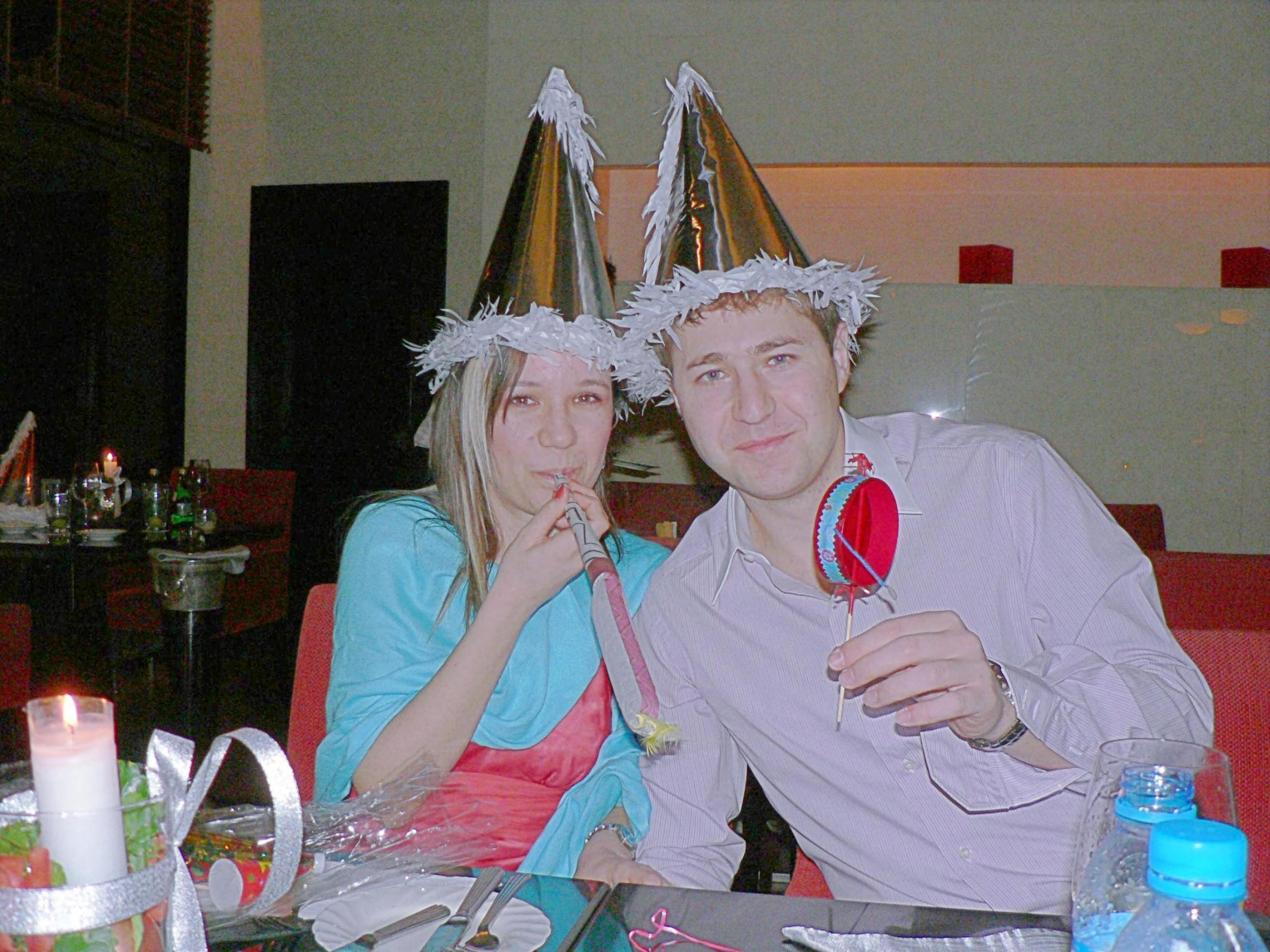}}
 	\centerline{ZeroDCE}
        \vspace{3pt}
 	\centerline{\includegraphics[width=1\textwidth]{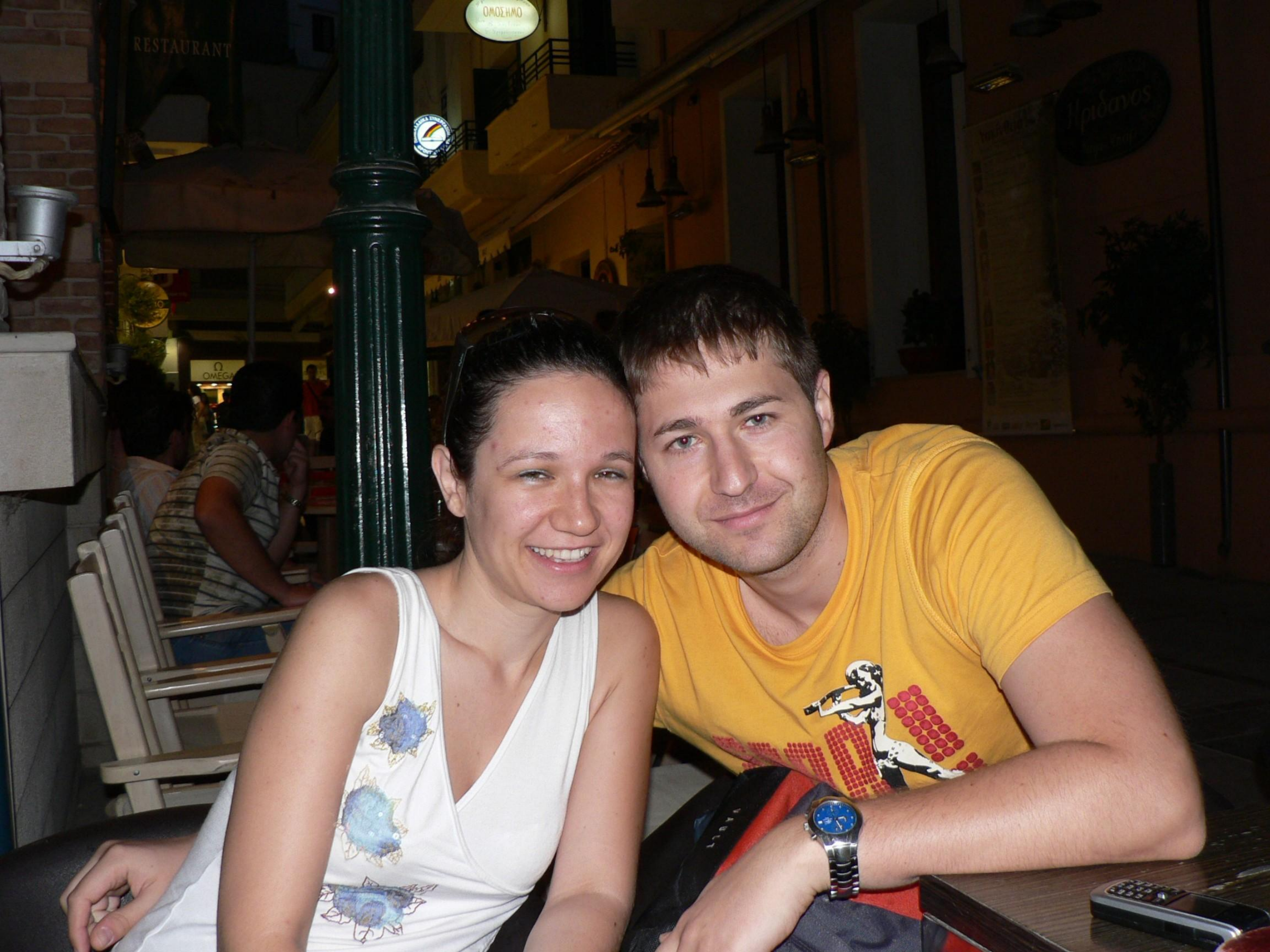}}
 	\centerline{Input}
 	\centerline{\includegraphics[width=1\textwidth]{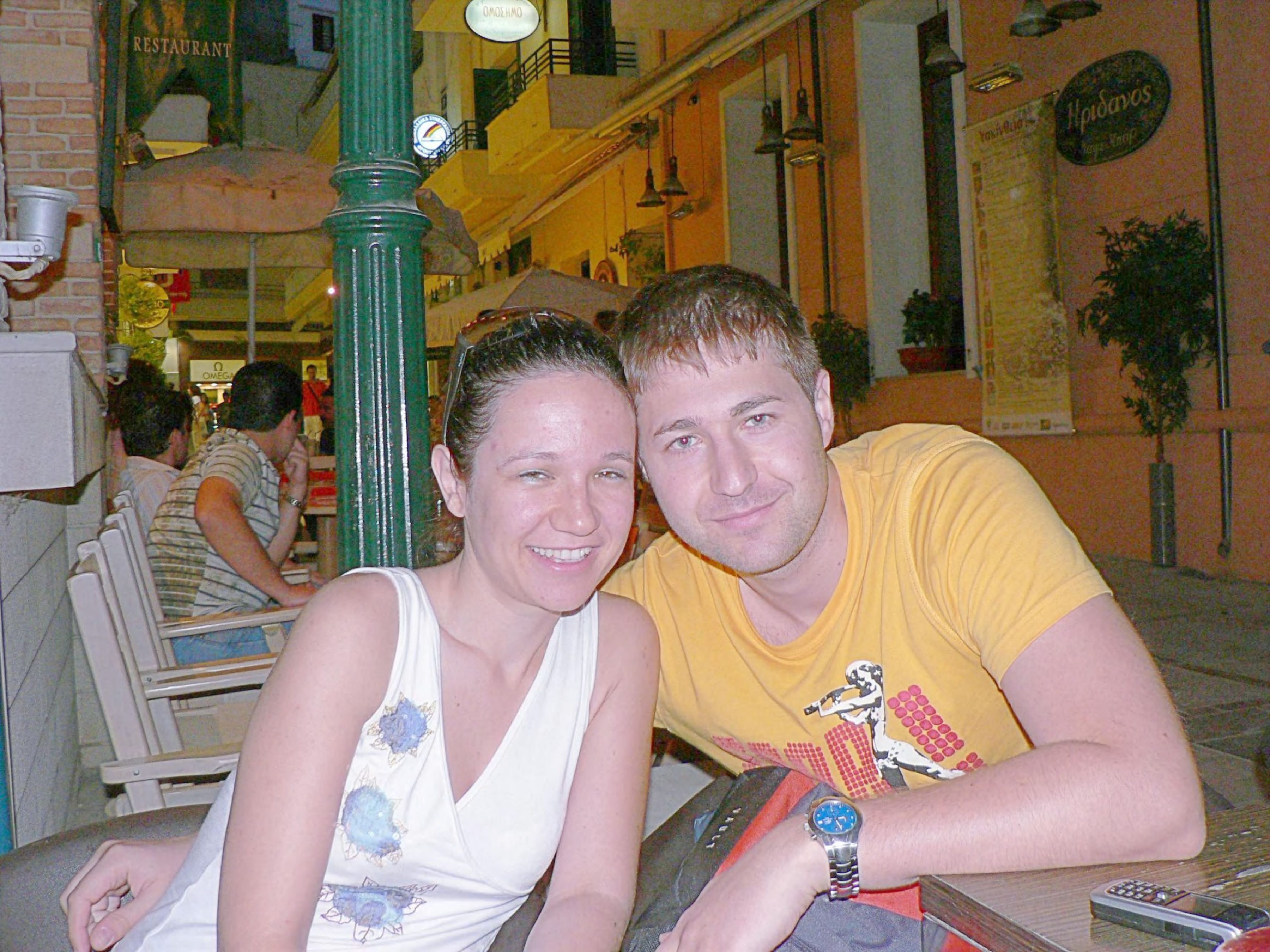}}
 	\centerline{ZeroDCE}
        \vspace{3pt}
\end{minipage}
\begin{minipage}{0.32\linewidth}
        \vspace{3pt}
 	\centerline{\includegraphics[width=1\textwidth]{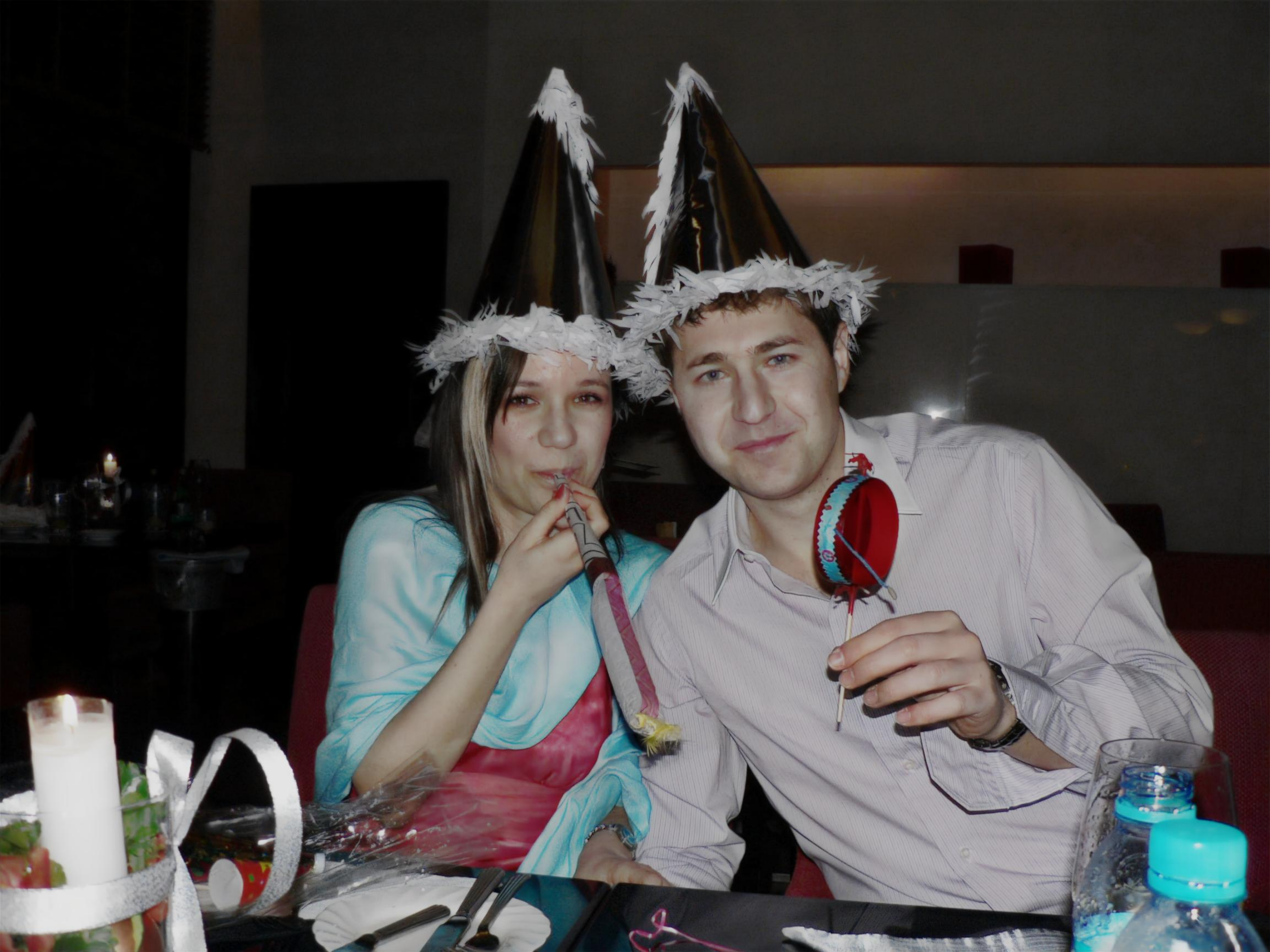}}
 	\centerline{KinD}
        \vspace{2pt}
 	\centerline{\includegraphics[width=1\textwidth]{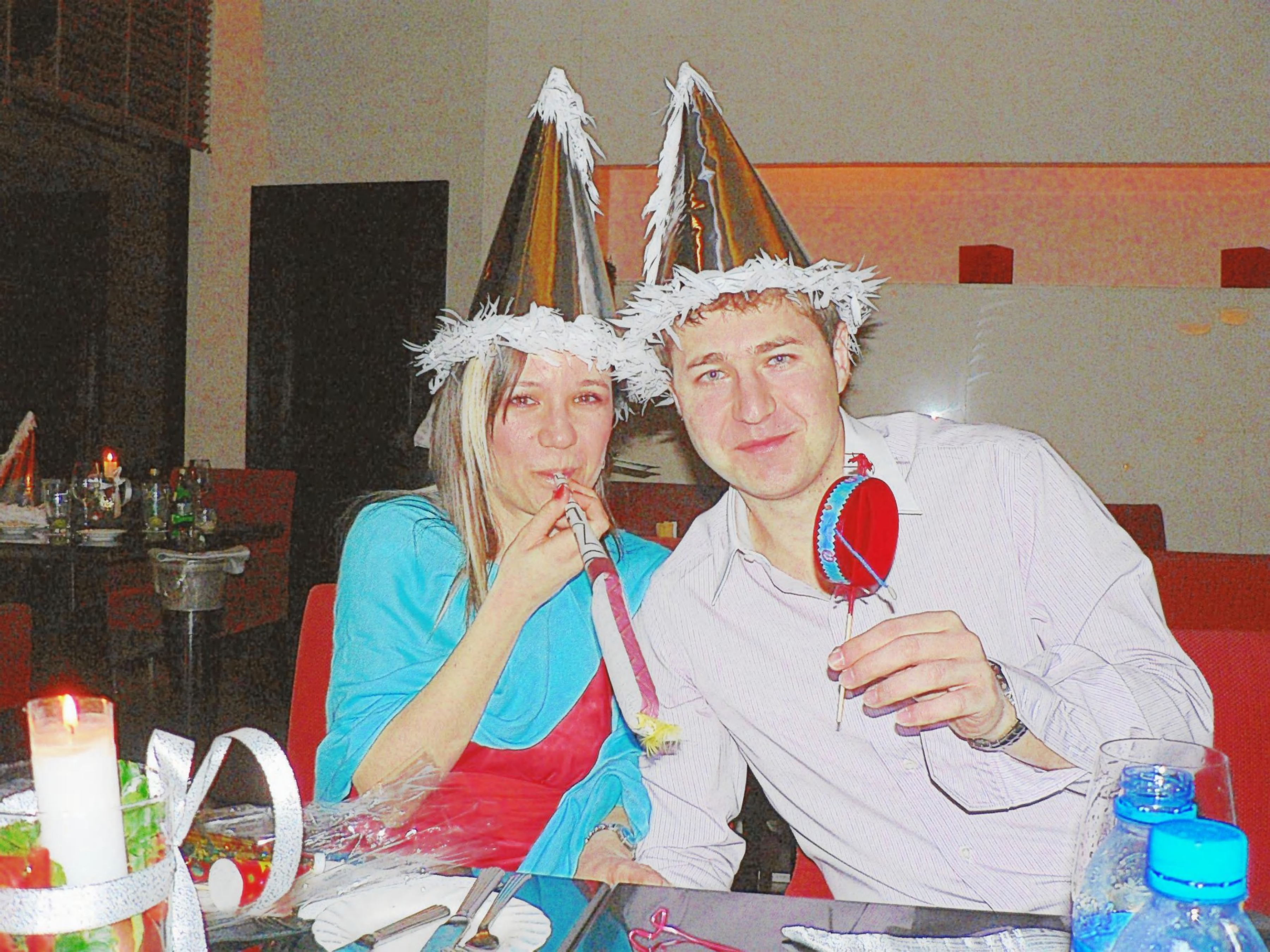}}
 	\centerline{PairLIE}
        \vspace{3pt}
 	\centerline{\includegraphics[width=1\textwidth]{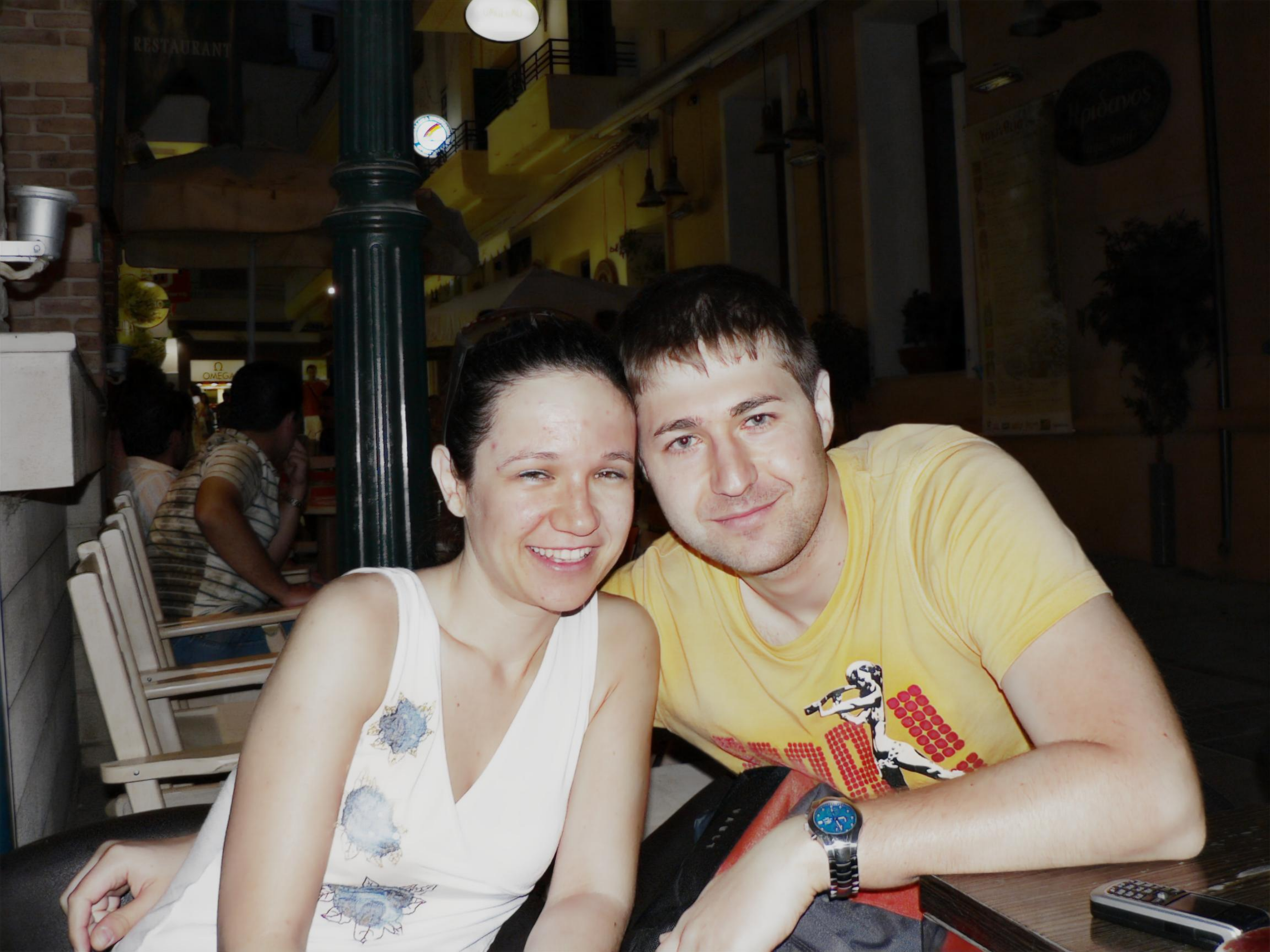}}
 	\centerline{KinD}
        \vspace{2pt}
 	\centerline{\includegraphics[width=1\textwidth]{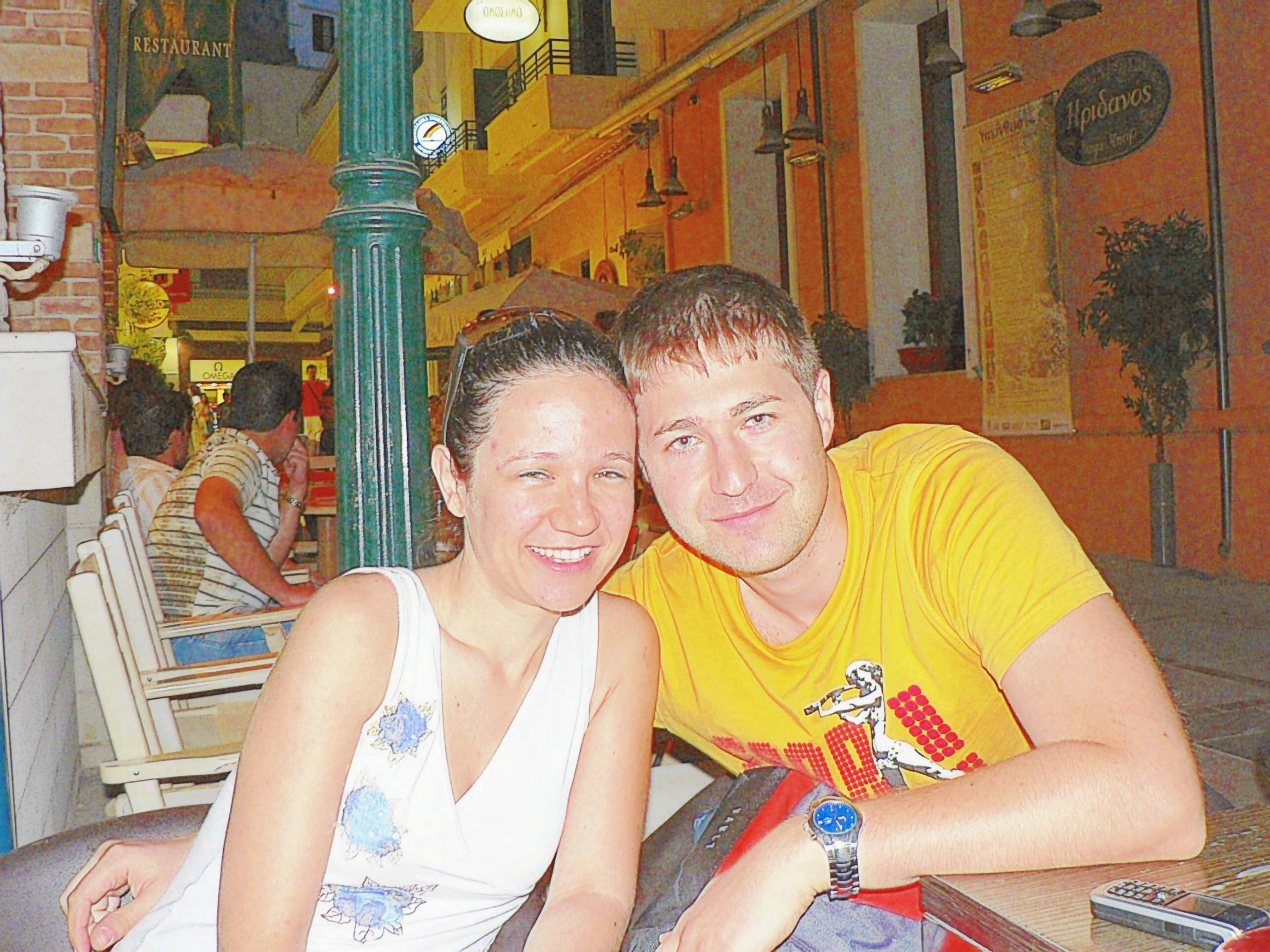}}
 	\centerline{PairLIE}
        \vspace{3pt}
 \end{minipage}
\begin{minipage}{0.32\linewidth}
 	\vspace{3pt}
 	\centerline{\includegraphics[width=1\textwidth]{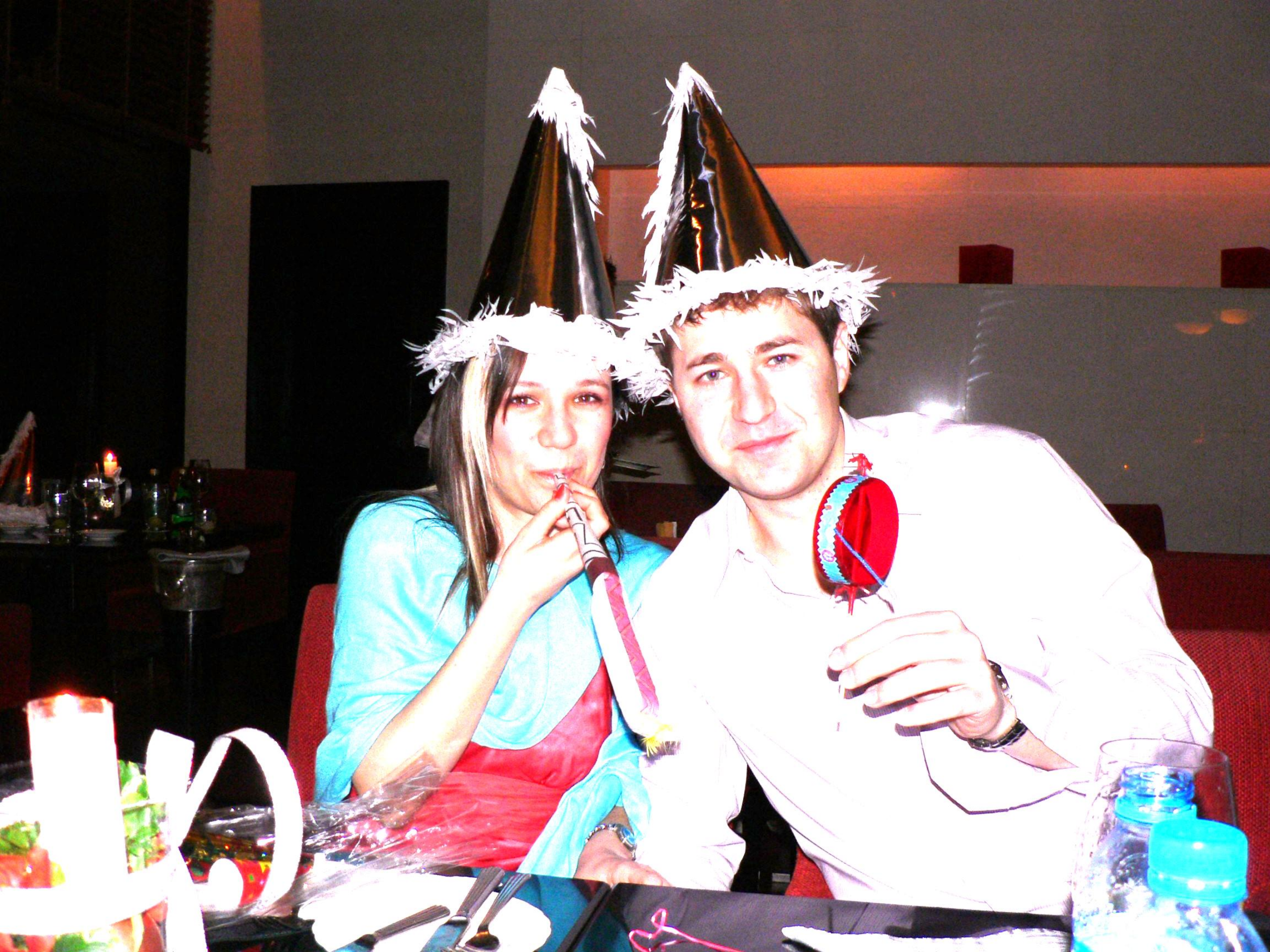}}
 	\centerline{RUAS}
 	\vspace{2pt}
 	\centerline{\includegraphics[width=1\textwidth]{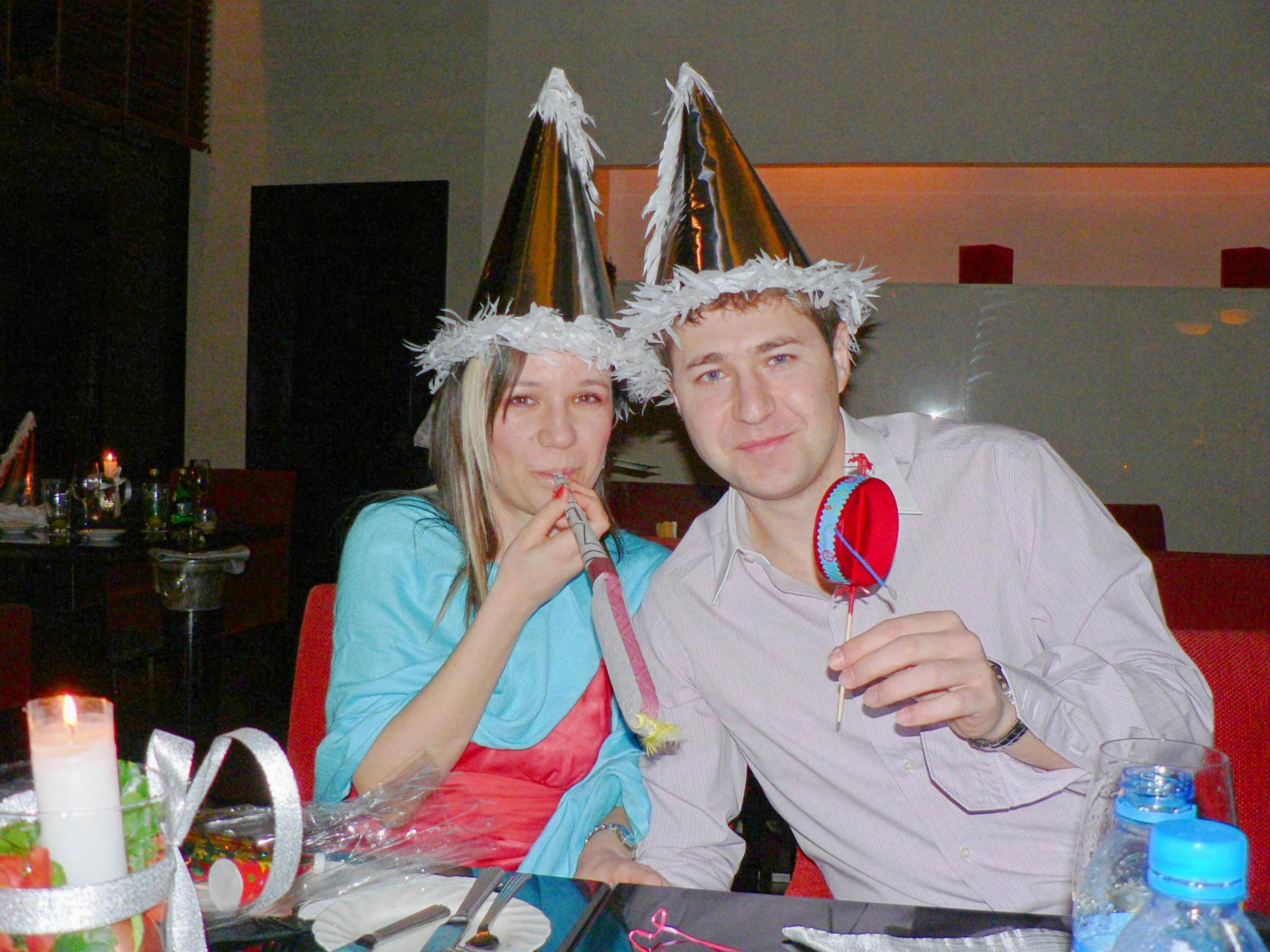}}
 	\centerline{CIDNet}
 	\vspace{3pt}
 	\centerline{\includegraphics[width=1\textwidth]{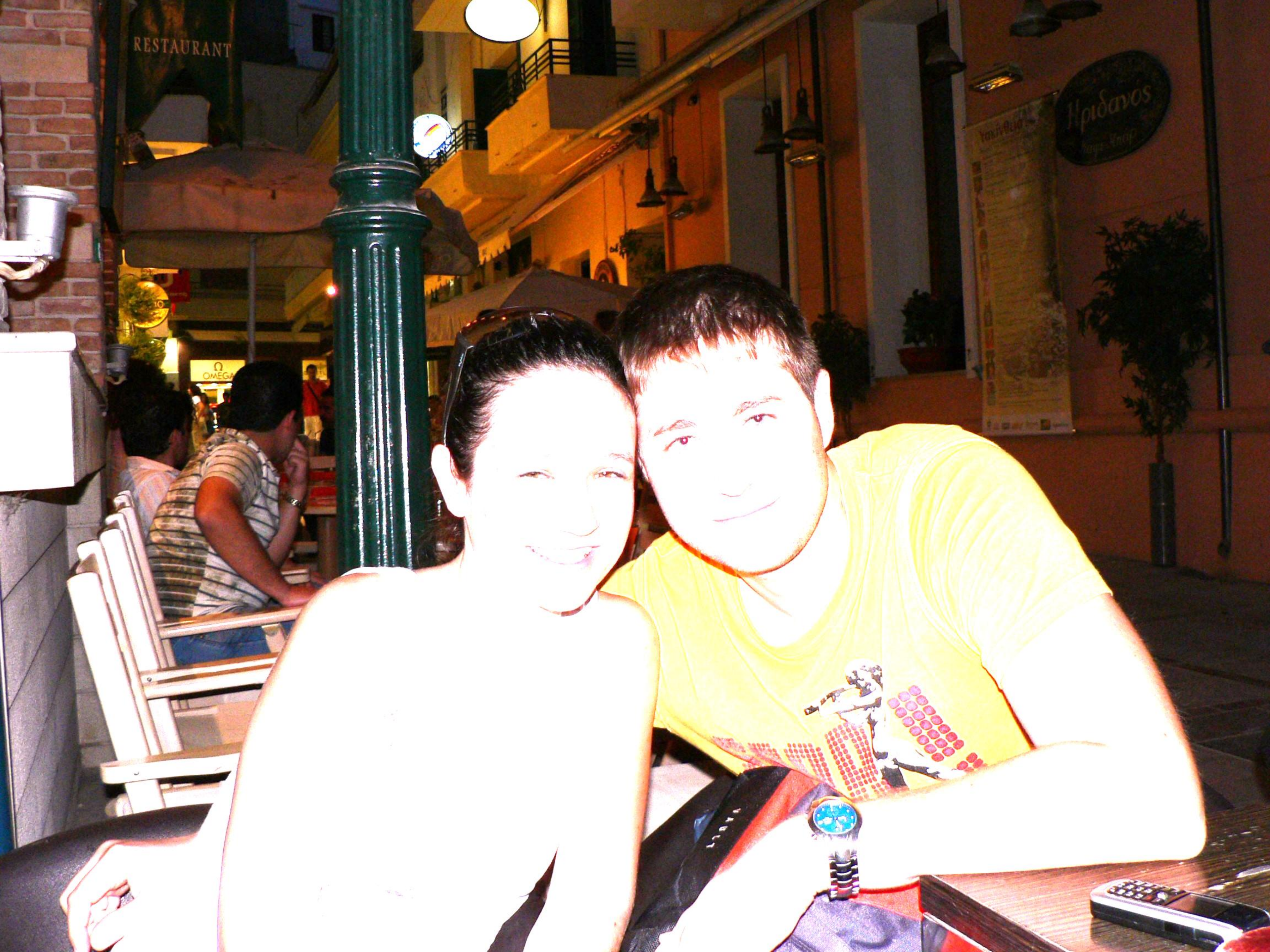}}
 	\centerline{RUAS}
 	\vspace{2pt}
 	\centerline{\includegraphics[width=1\textwidth]{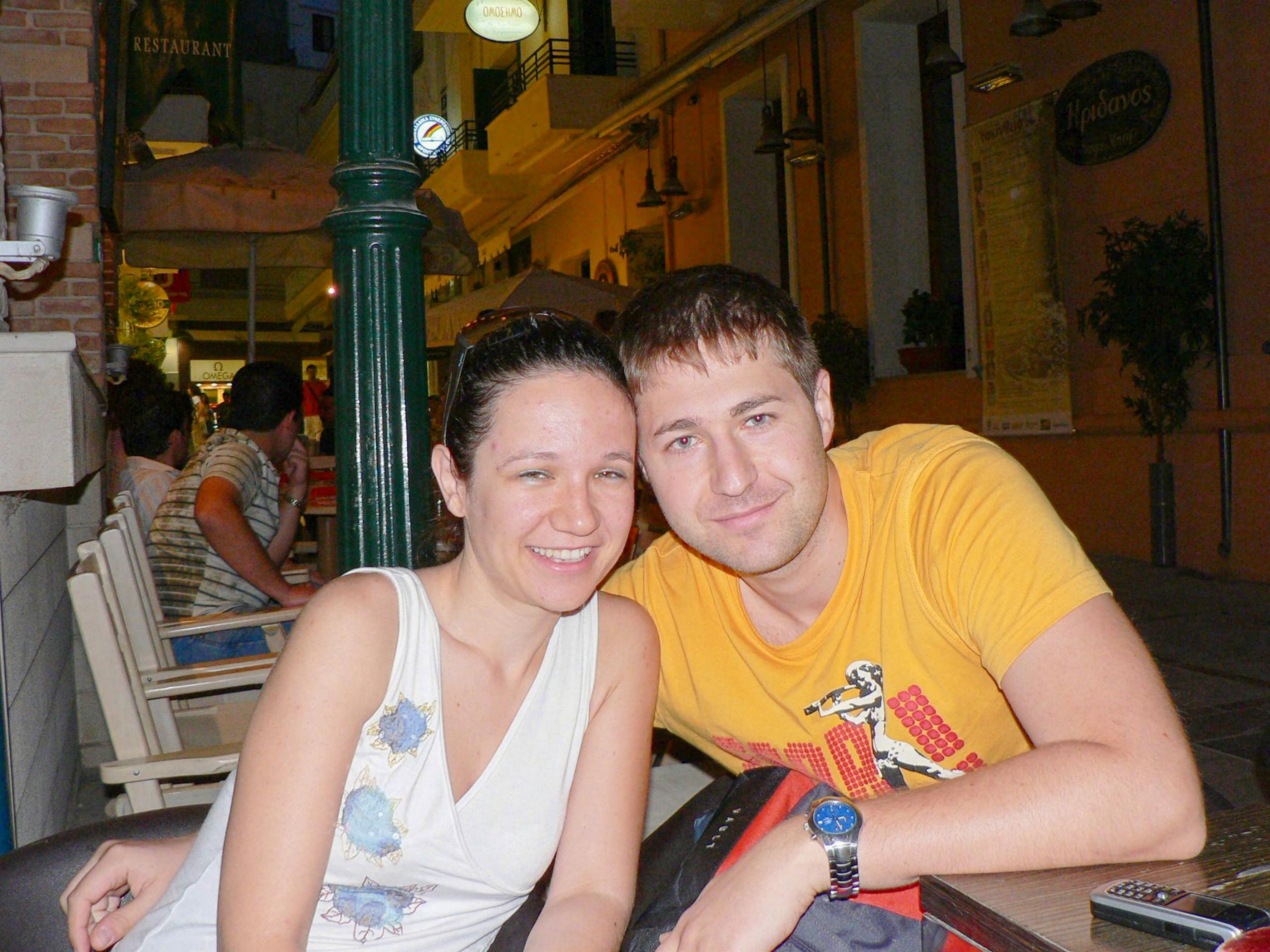}}
 	\centerline{CIDNet}
 	\vspace{3pt}
 \end{minipage}
 \caption{Visual examples for unpaired image enhancement on VV dataset \cite{VV} among KinD \cite{KinD}, RUAS \cite{RUAS}, URetinexNet \cite{URetinexNet}, RetinexFormer \cite{RetinexFormer}, PairLIE \cite{PairLIE}, ZeroDCE \cite{Zero-DCE}, and our CIDNet. Our method can suppress the color shift phenomenon while increasing brightness, and achieve more realistic result.}
 \label{fig:VV}
\end{figure*}

\textbf{Unpaired Datasets.} Qualitative visual comparisons of five unpaired datasets DICM \cite{DICM}, LIME \cite{LIME}, MEF \cite{MEF}, NPE \cite{NPE}, and VV \cite{VV} with our CIDNet against SOTA methods. They are shown in Fig. \ref{fig:DICM}, Fig. \ref{fig:LIME}, Fig. \ref{fig:MEF}, Fig. \ref{fig:NPE}, and Fig. \ref{fig:VV}.

\begin{figure*}
\centering
 \begin{minipage}{0.245\linewidth}
 \centering
        \vspace{3pt}
 	\centerline{\includegraphics[width=1\textwidth]{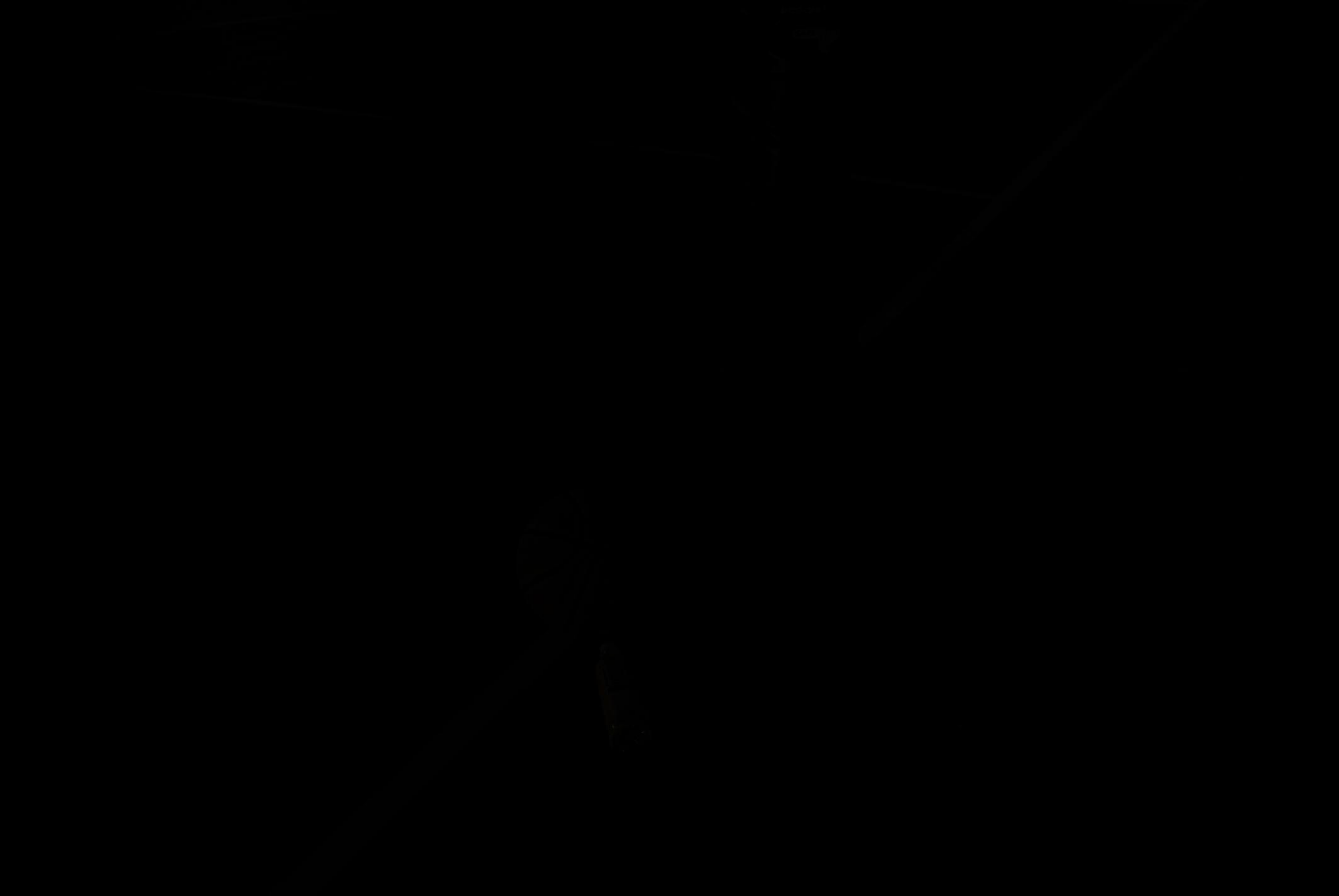}}
 	\centerline{Input}
 	\centerline{\includegraphics[width=1\textwidth]{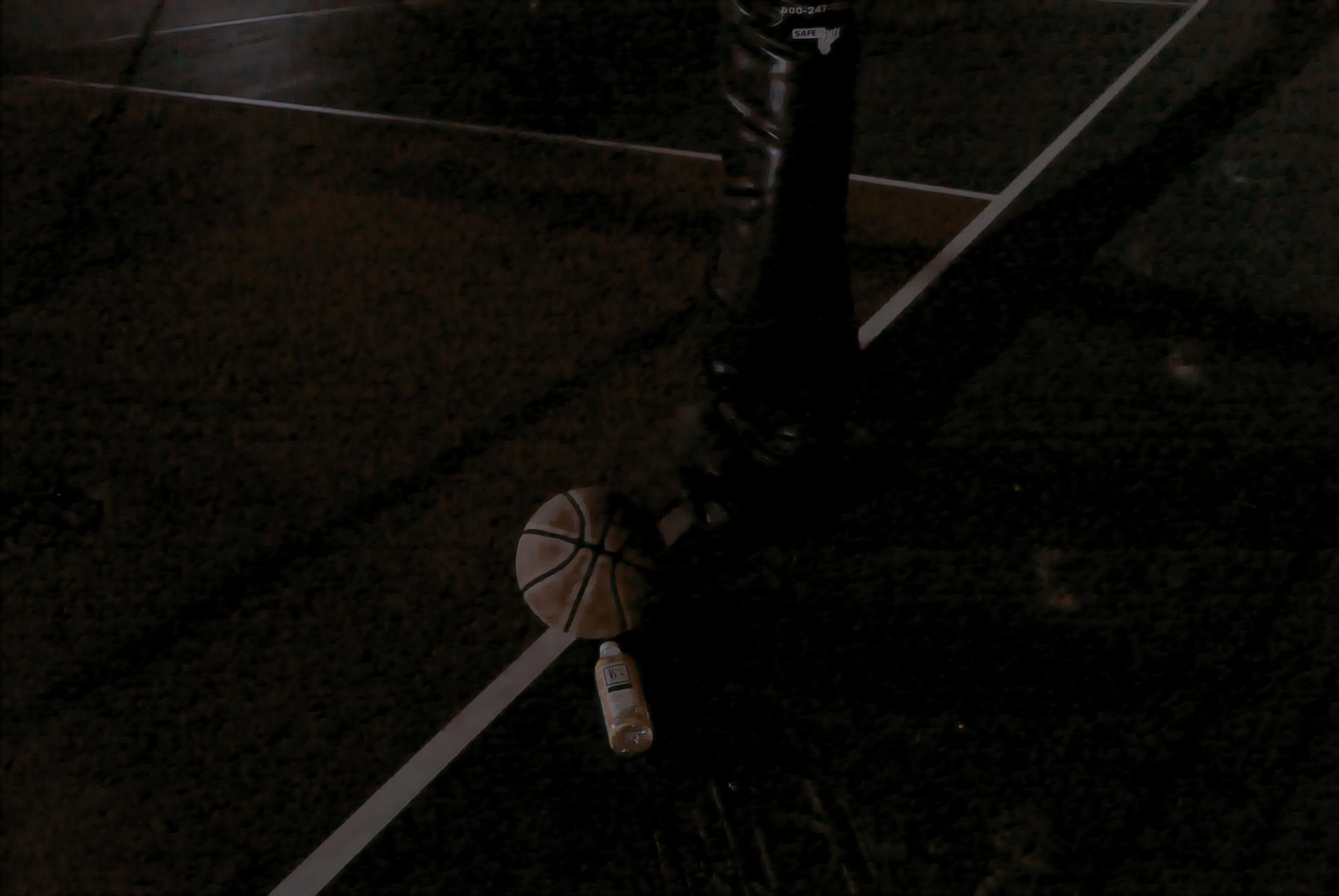}}
 	\centerline{URetinexNet}
        \vspace{3pt}
 	\centerline{\includegraphics[width=1\textwidth]{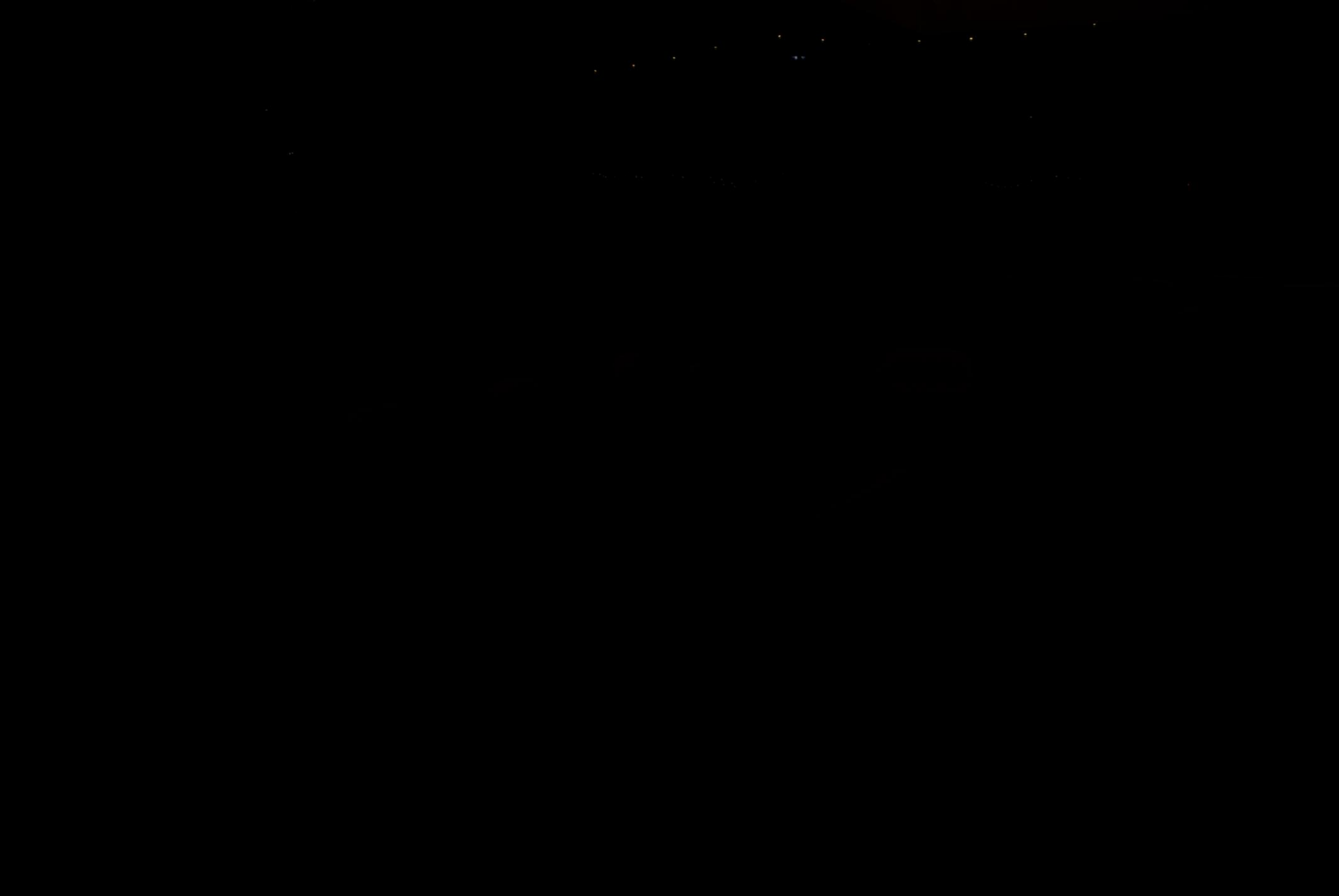}}
 	\centerline{Input}
 	\centerline{\includegraphics[width=1\textwidth]{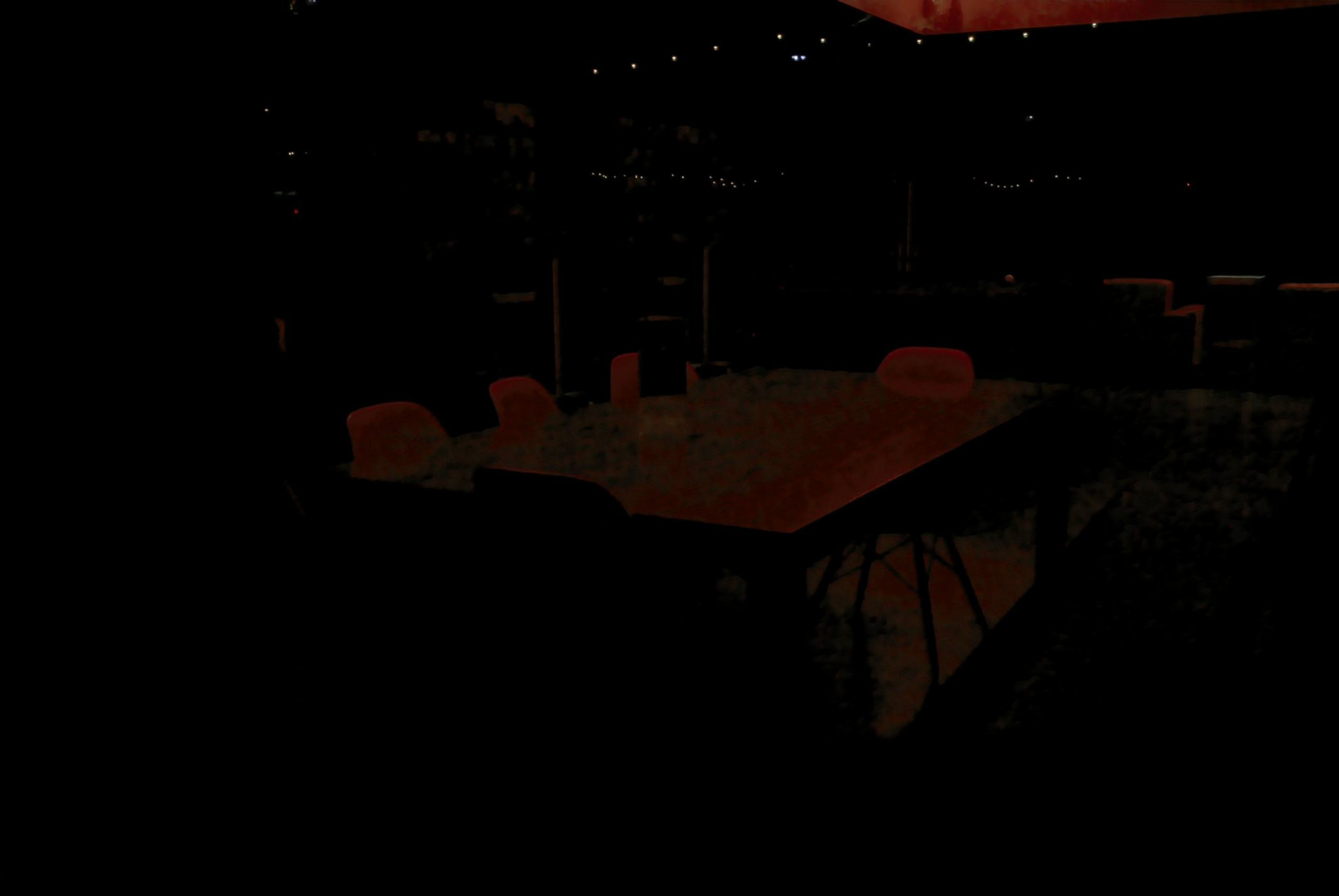}}
 	\centerline{URetinexNet}
        \vspace{3pt}
\end{minipage}
\begin{minipage}{0.245\linewidth}
        \vspace{3pt}
 	\centerline{\includegraphics[width=1\textwidth]{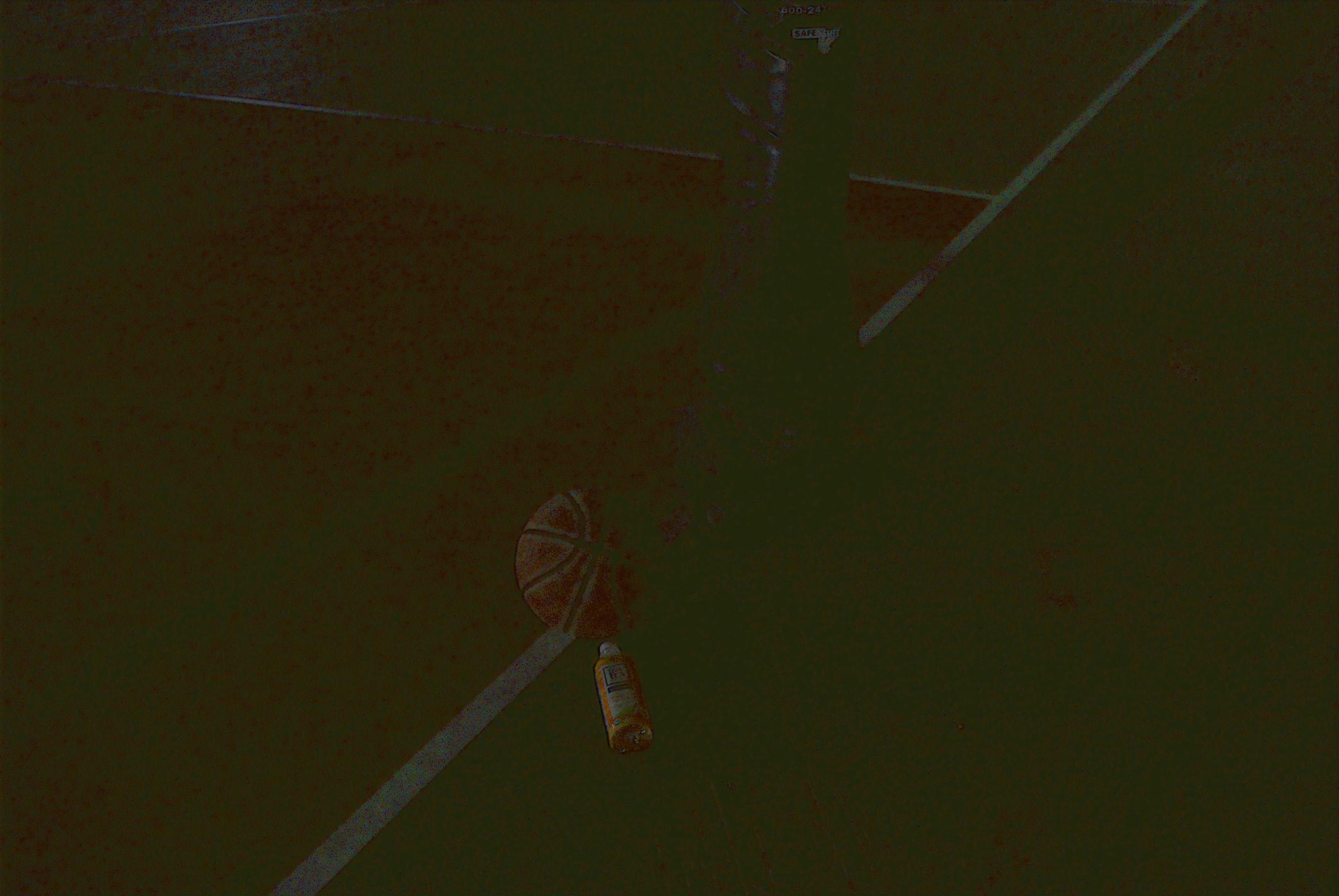}}
 	\centerline{RetinexNet}
        \vspace{2pt}
 	\centerline{\includegraphics[width=1\textwidth]{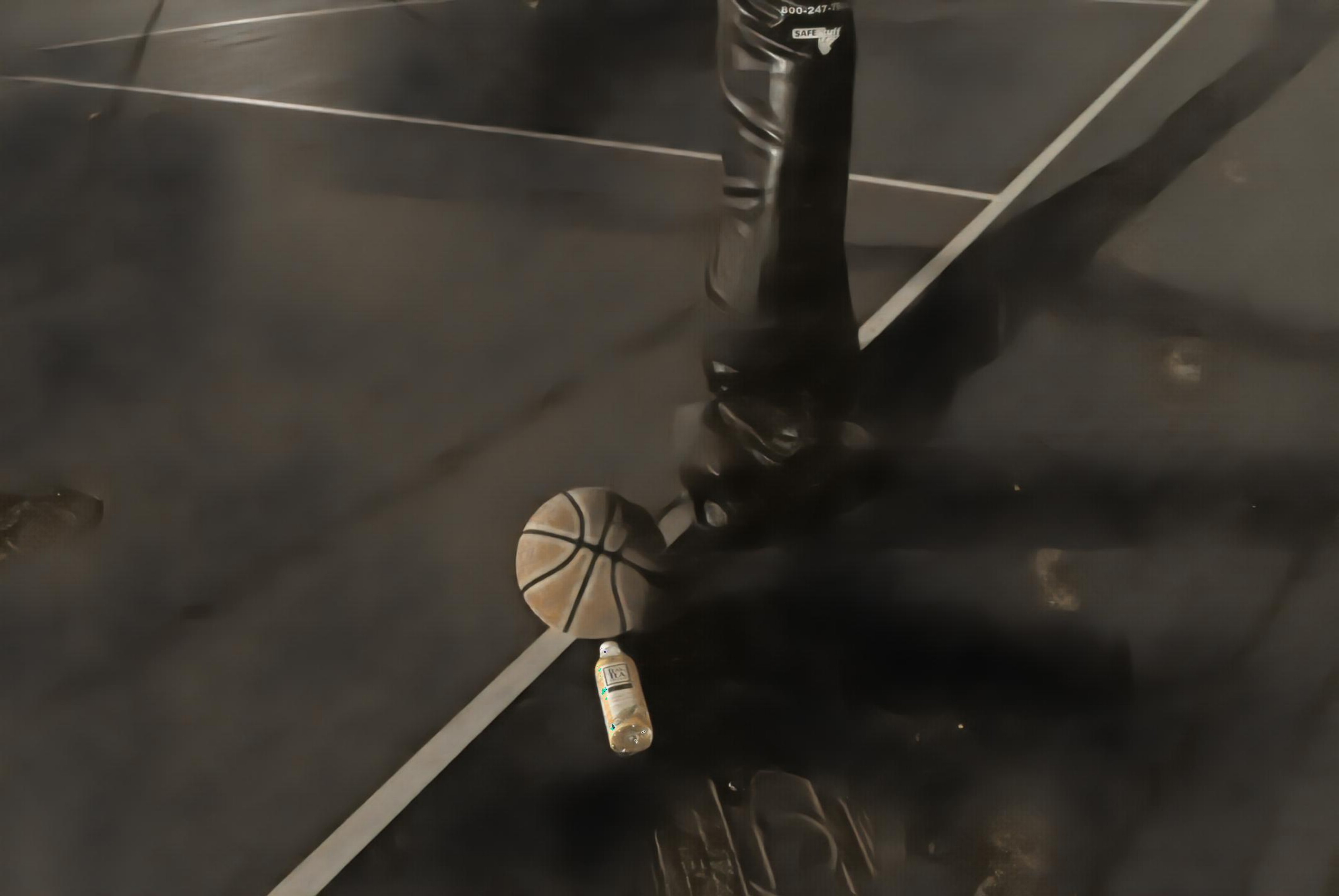}}
 	\centerline{LEDNet}
        \vspace{3pt}
 	\centerline{\includegraphics[width=1\textwidth]{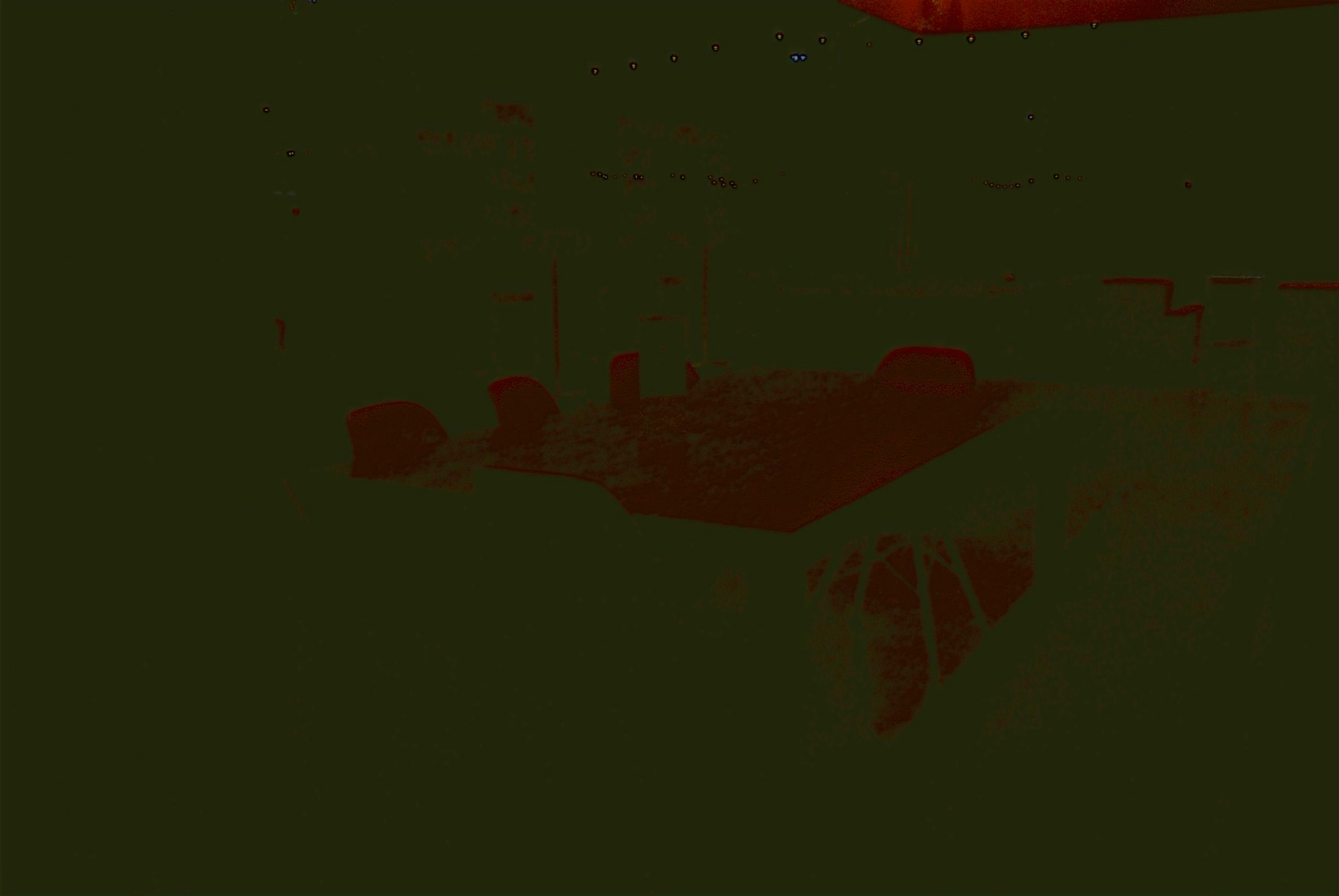}}
 	\centerline{RetinexNet}
        \vspace{2pt}
 	\centerline{\includegraphics[width=1\textwidth]{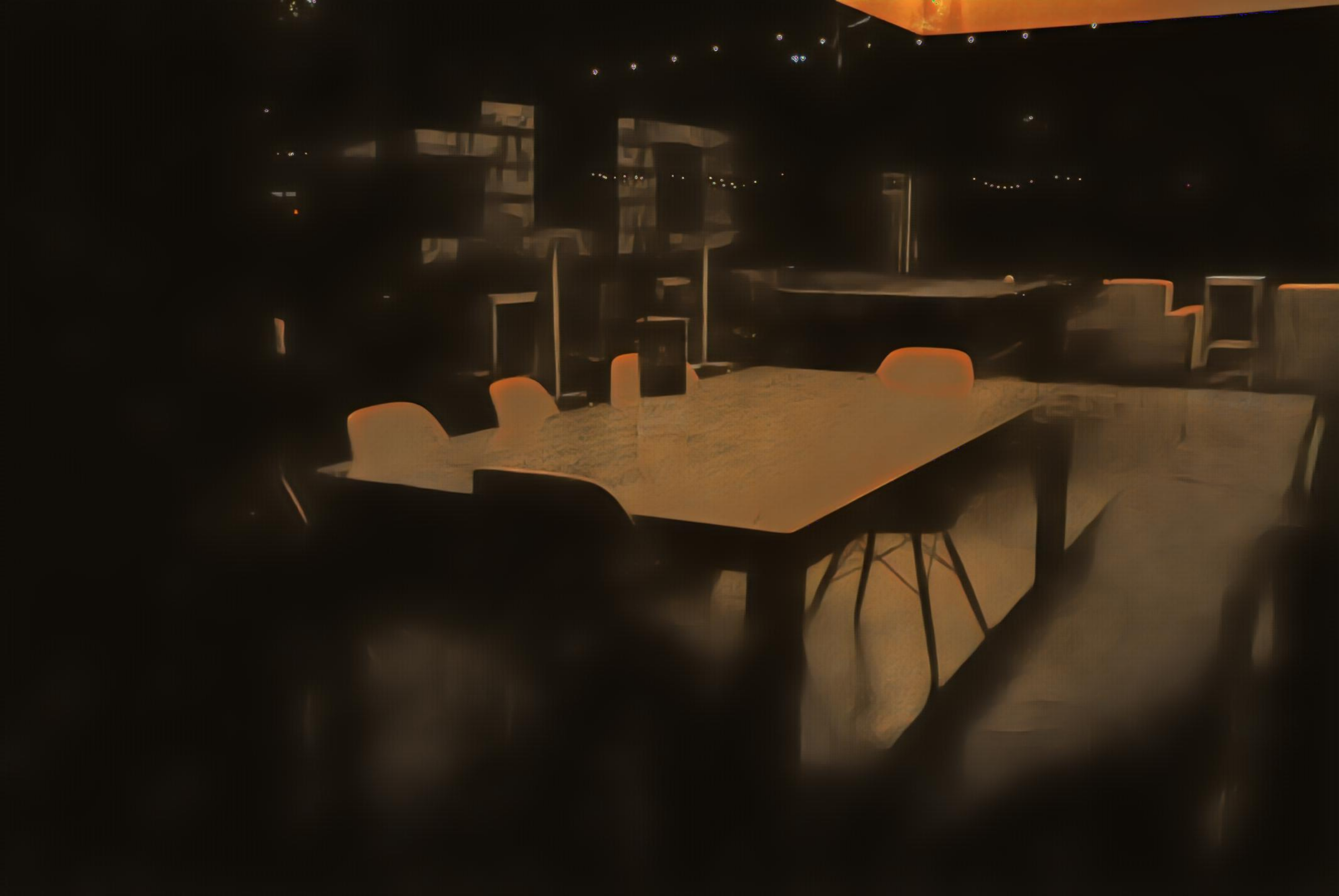}}
 	\centerline{LEDNet}
        \vspace{3pt}
 \end{minipage}
\begin{minipage}{0.245\linewidth}
 	\vspace{3pt}
 	\centerline{\includegraphics[width=1\textwidth]{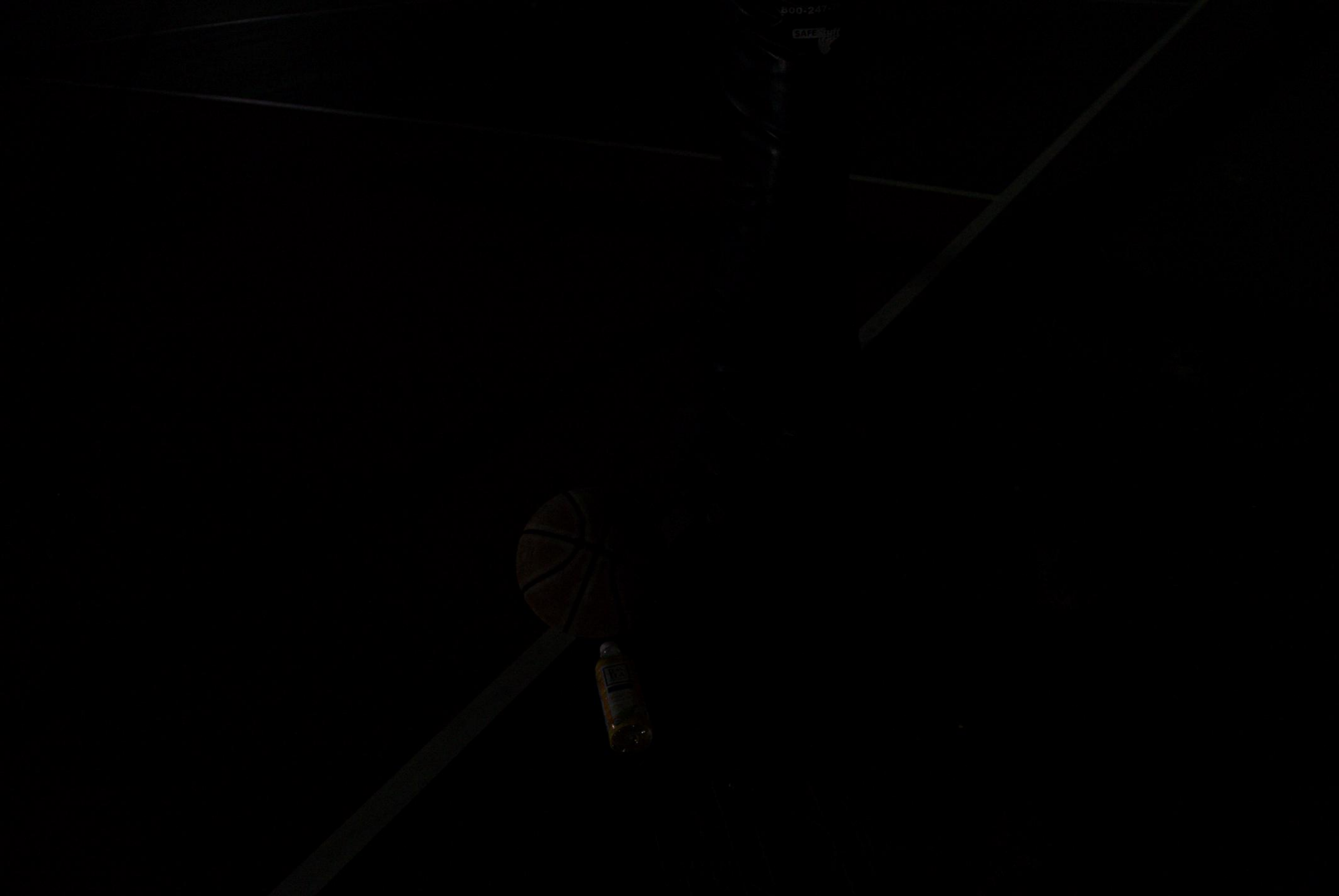}}
 	\centerline{RUAS}
 	\vspace{2pt}
 	\centerline{\includegraphics[width=1\textwidth]{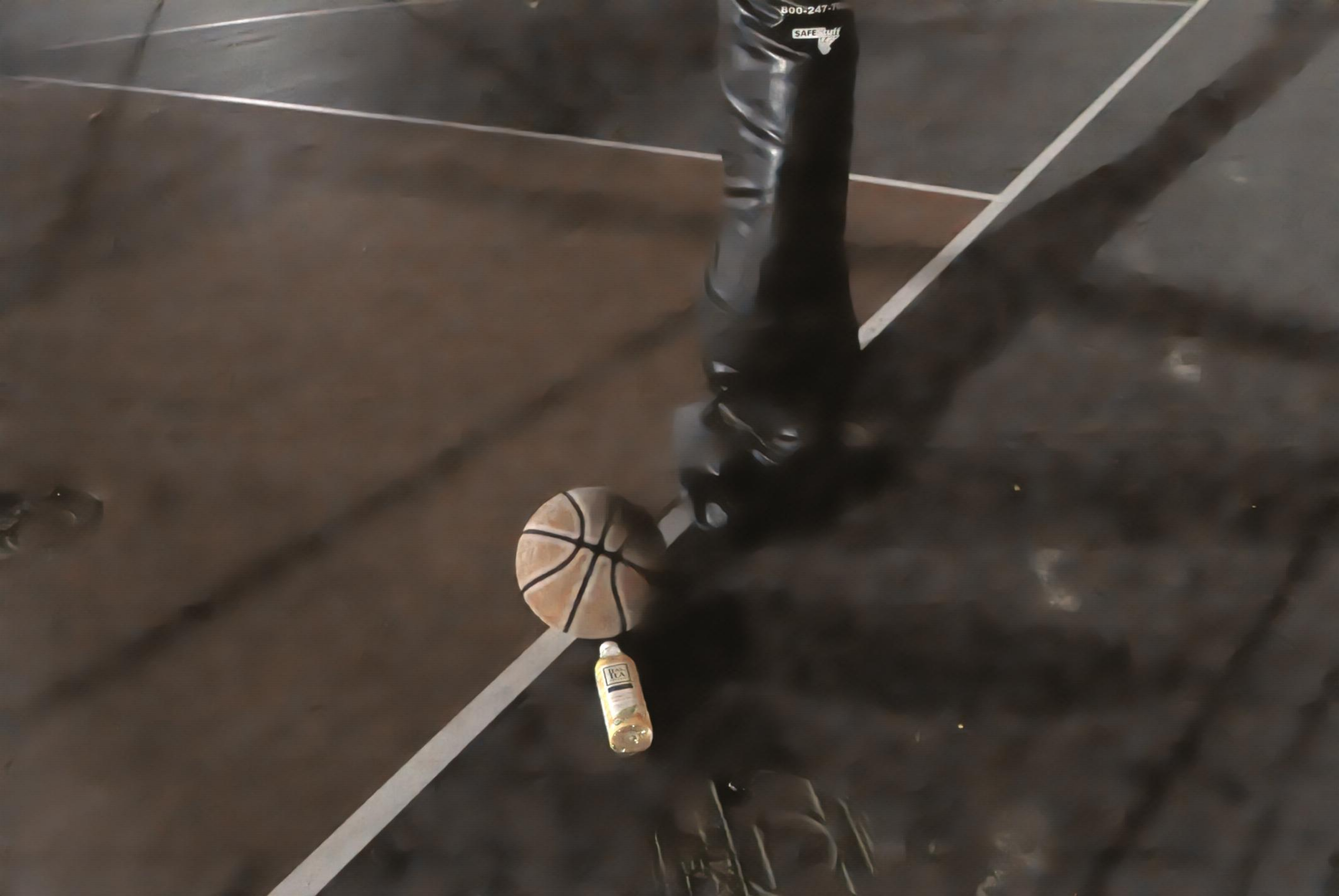}}
 	\centerline{RetinexFormer}
 	\vspace{3pt}
 	\centerline{\includegraphics[width=1\textwidth]{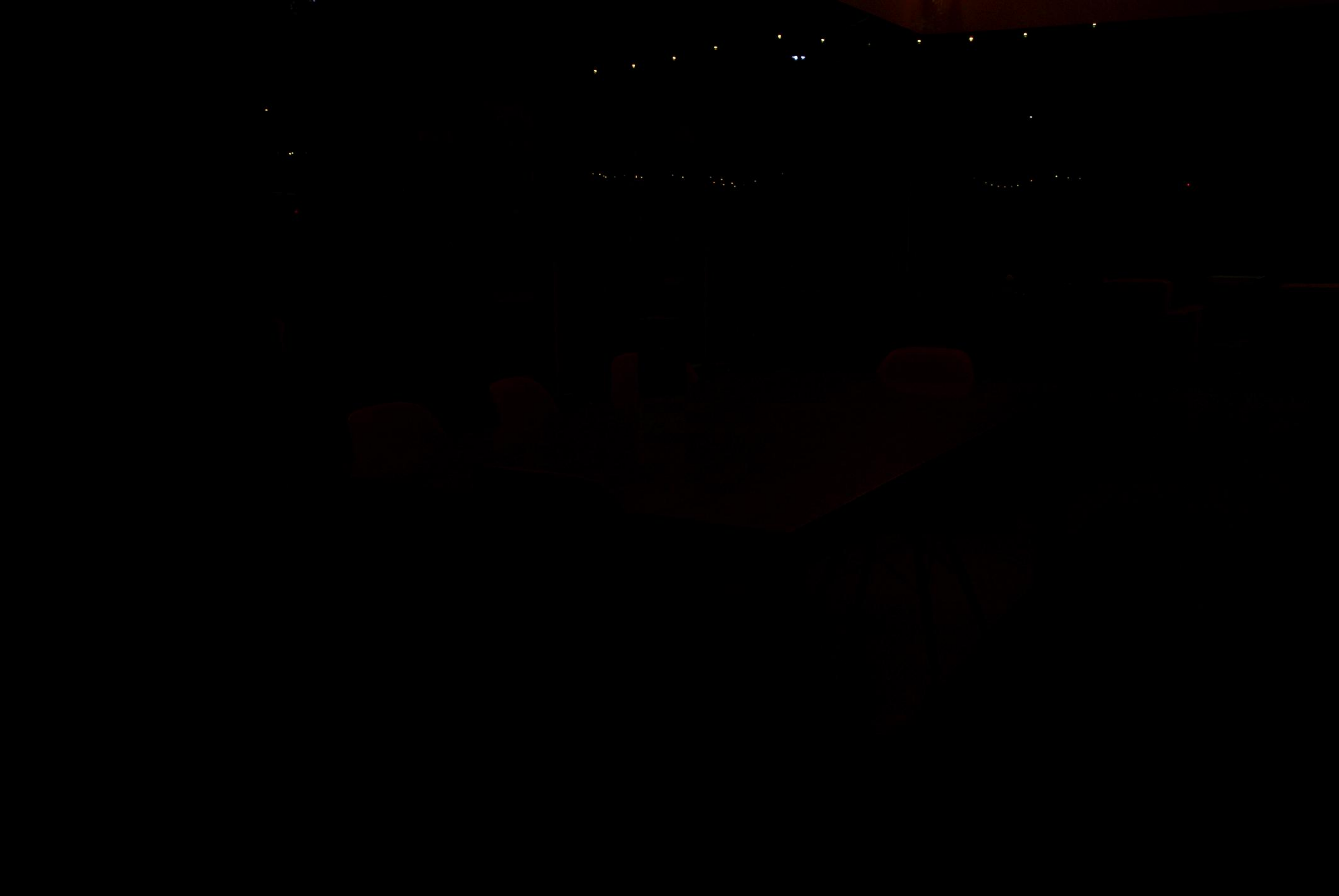}}
 	\centerline{RUAS}
 	\vspace{2pt}
 	\centerline{\includegraphics[width=1\textwidth]{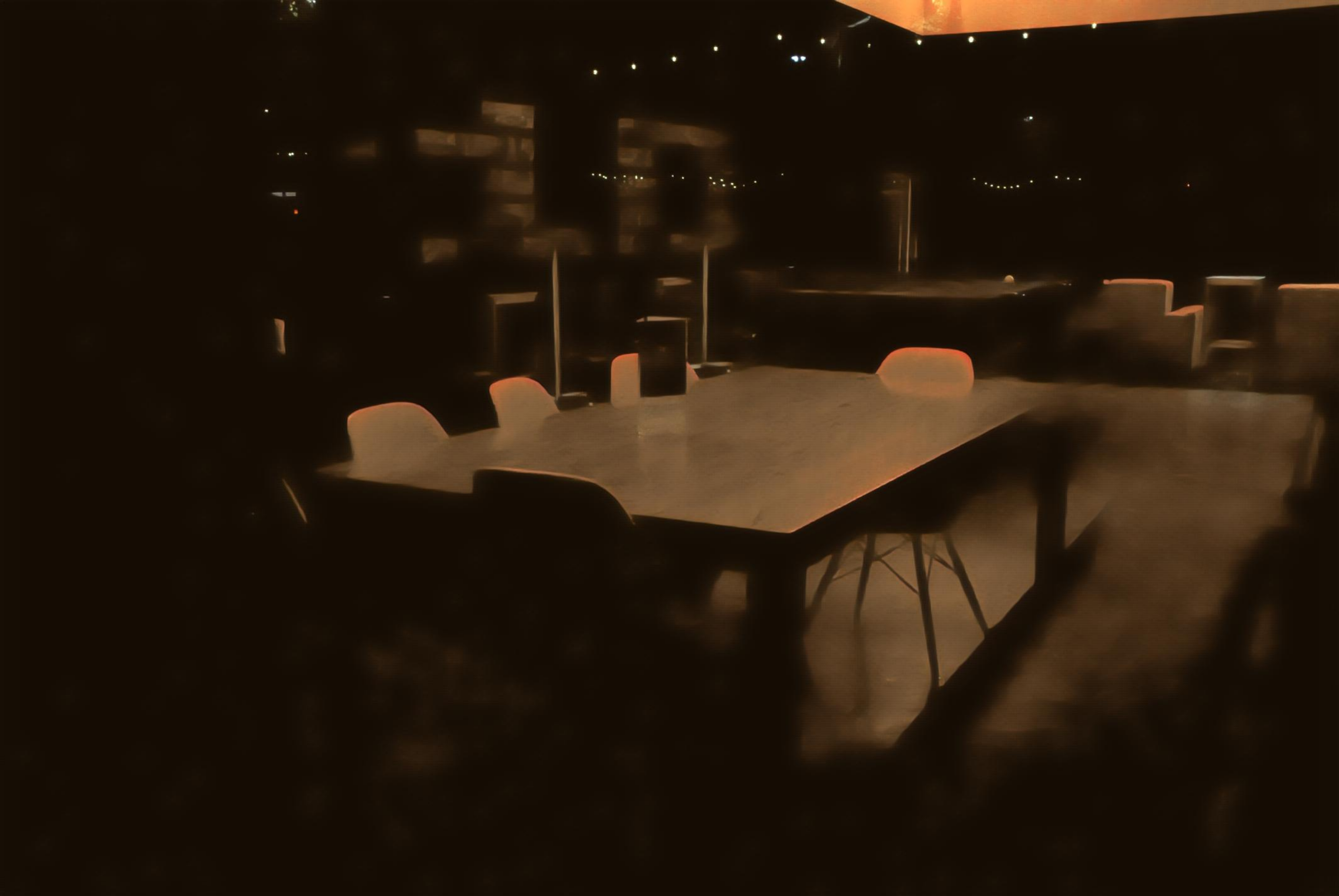}}
 	\centerline{RetinexFormer}
 	\vspace{3pt}
 \end{minipage}
 \begin{minipage}{0.245\linewidth}
 	\vspace{3pt}
 	\centerline{\includegraphics[width=1\textwidth]{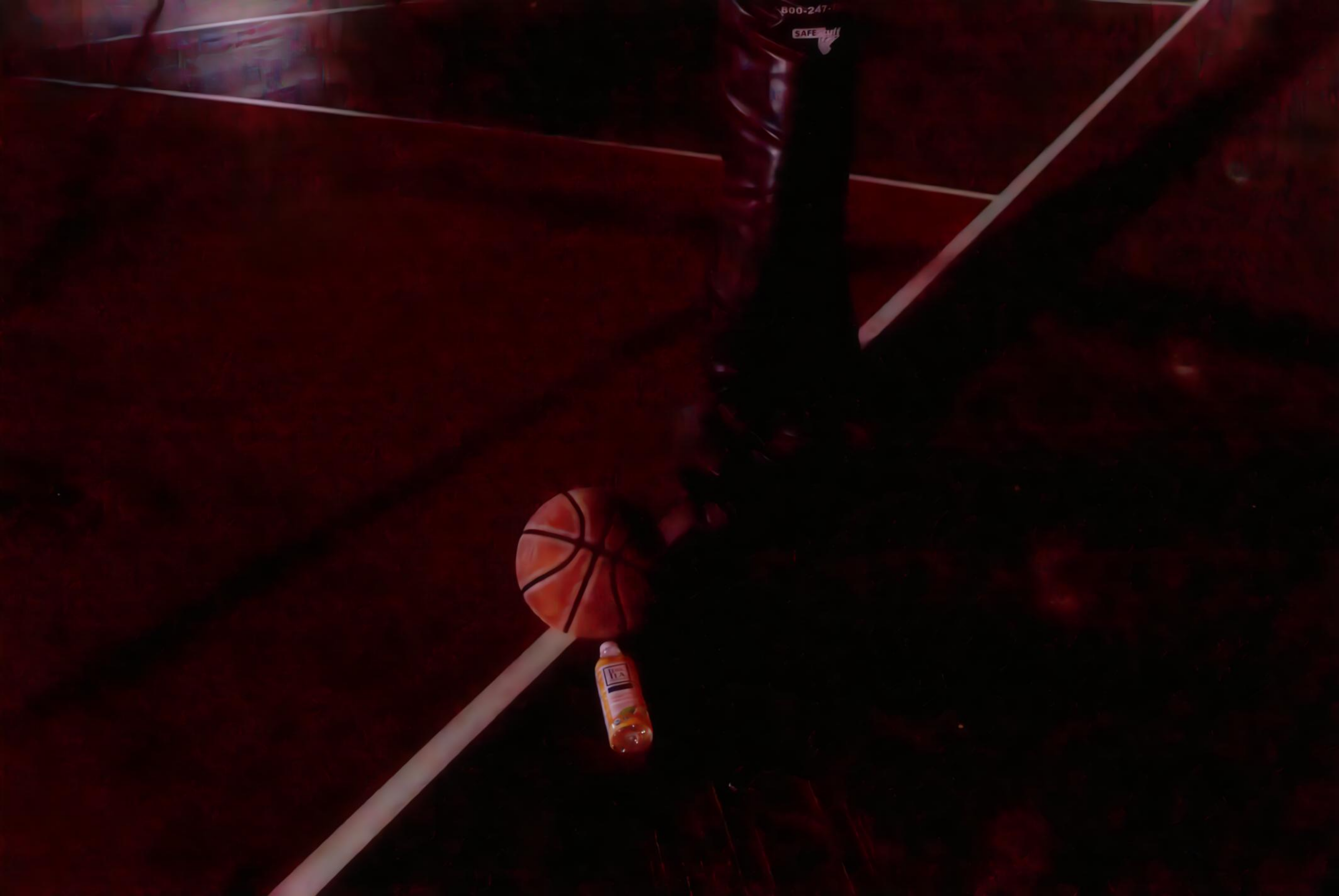}}
 	\centerline{LLFlow}
 	\vspace{2pt}
 	\centerline{\includegraphics[width=1\textwidth]{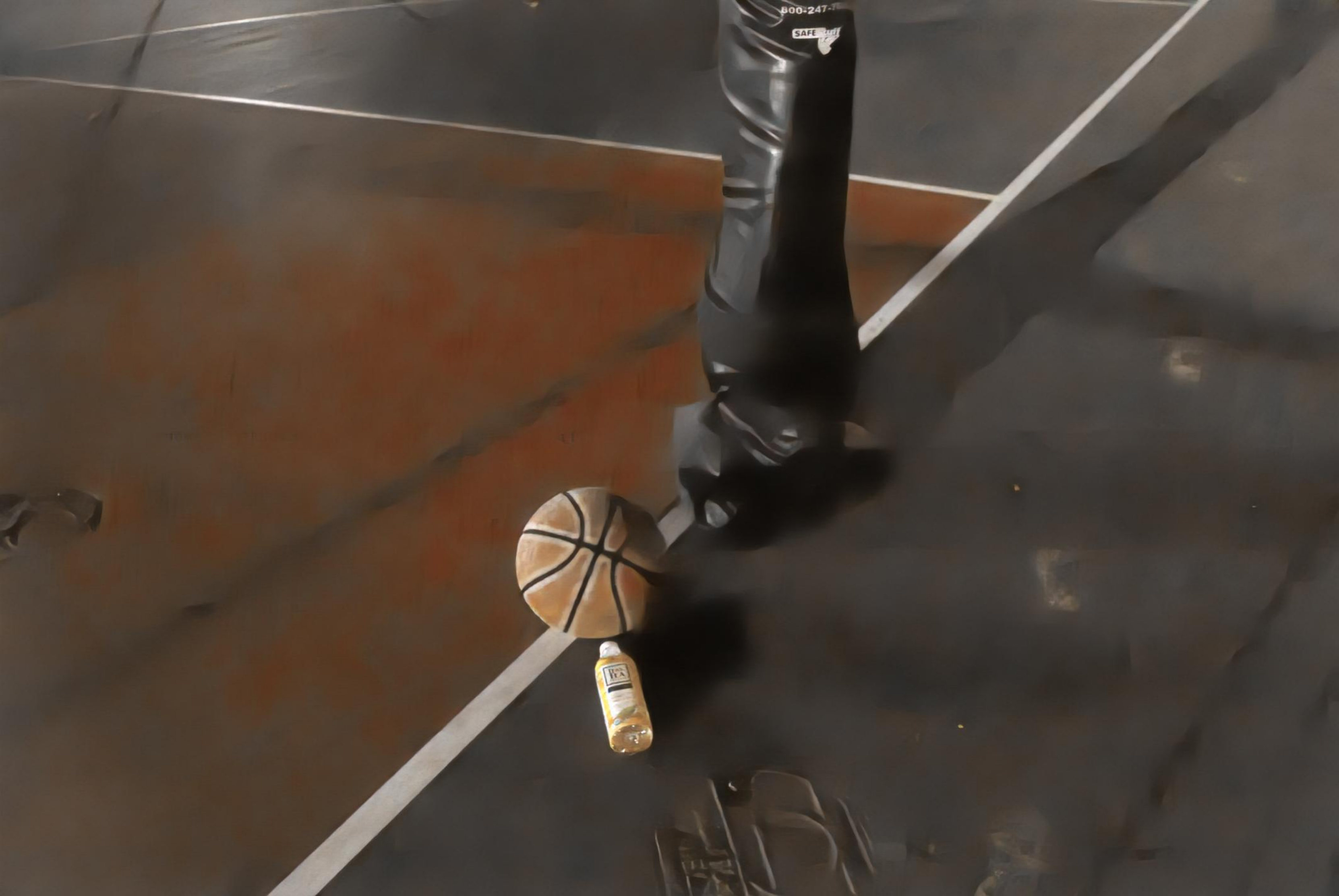}}
 	\centerline{CIDNet}
 	\vspace{3pt}
 	\centerline{\includegraphics[width=1\textwidth]{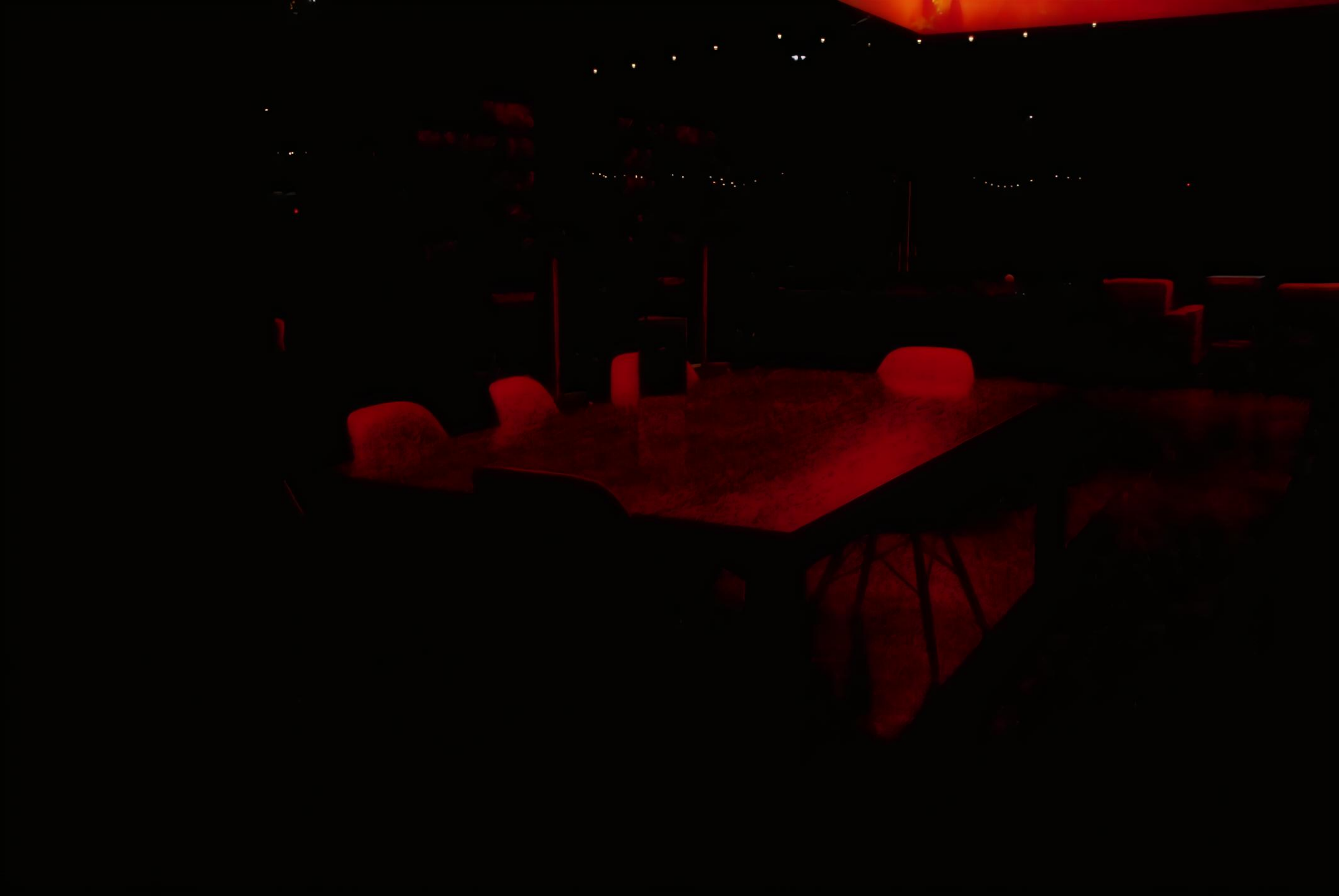}}
 	\centerline{LLFlow}
 	\vspace{2pt}
 	\centerline{\includegraphics[width=1\textwidth]{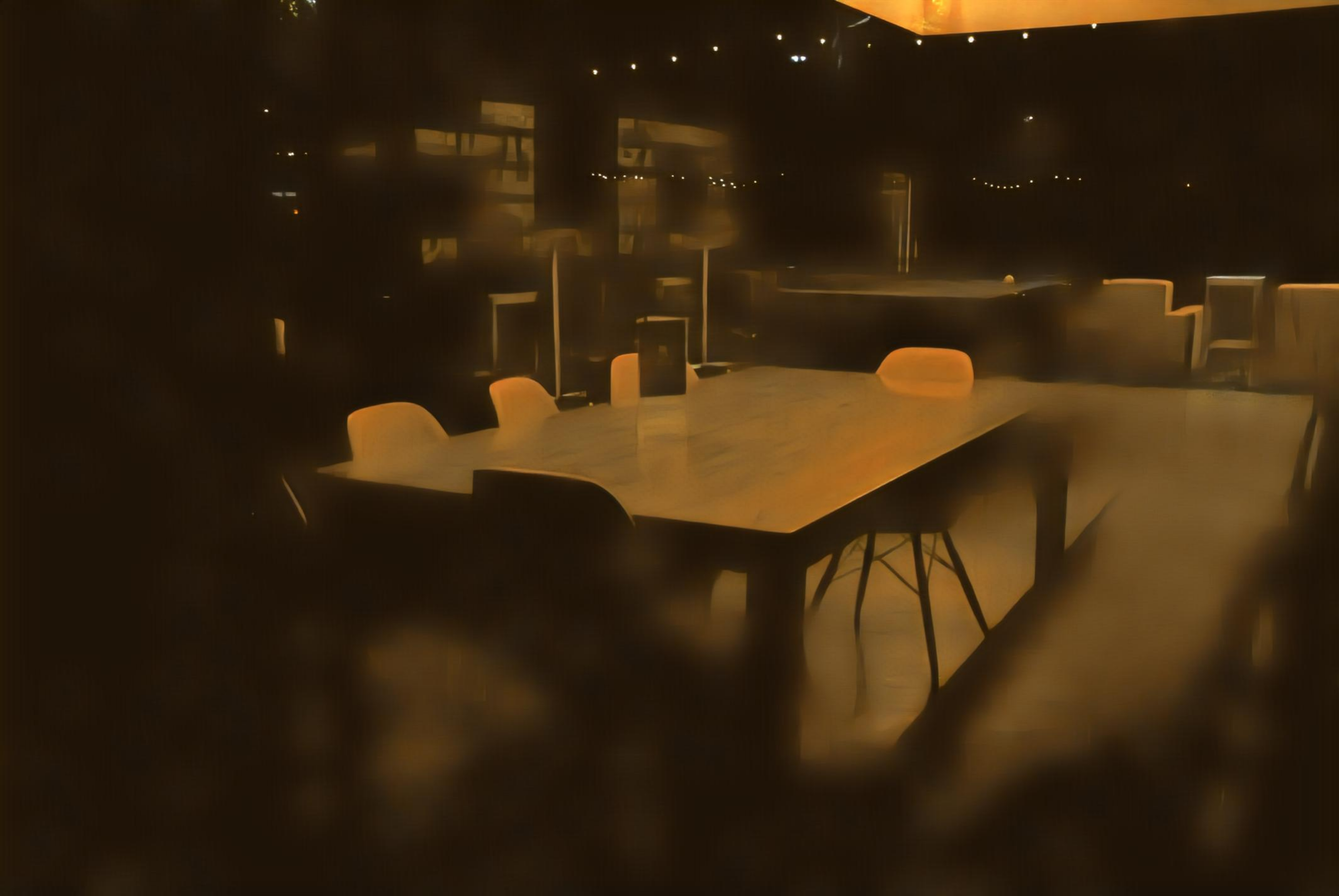}}
 	\centerline{CIDNet}
 	\vspace{3pt}
 \end{minipage}
 \caption{Visual examples for extremely low-light image enhancement on Sony-Total-Dark dataset \cite{SID}. As shown, almost no low-light features were extracted by RetinexNet \cite{RetinexNet}, RUAS \cite{RUAS} and LLFlow \cite{LLFlow}. Although URetinexNet \cite{URetinexNet}, LEDNet \cite{LEDNet} and RetinexFormer \cite{RetinexFormer} brightens the image, the results in more artifacts and color bias. Only our CIDNet achieves a reasonable and more realistic enhancing effect.}
 \label{fig:Sony}
\end{figure*}

\begin{figure*}[h]
\centering
 \begin{minipage}{0.32\linewidth}
 \centering
        \vspace{3pt}
 	\centerline{\includegraphics[width=1\textwidth]{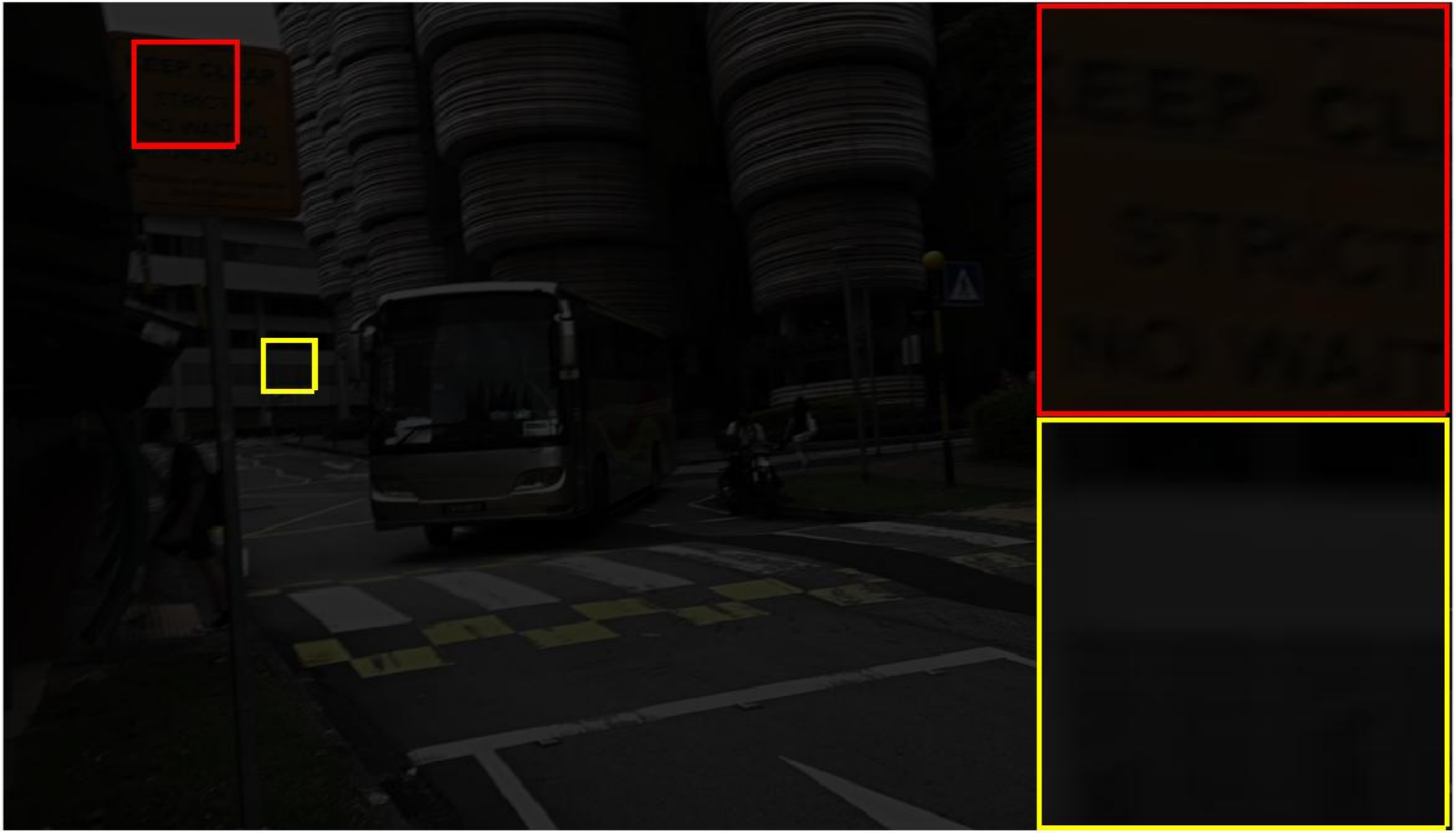}}
 	\centerline{Input}
 	\centerline{\includegraphics[width=1\textwidth]{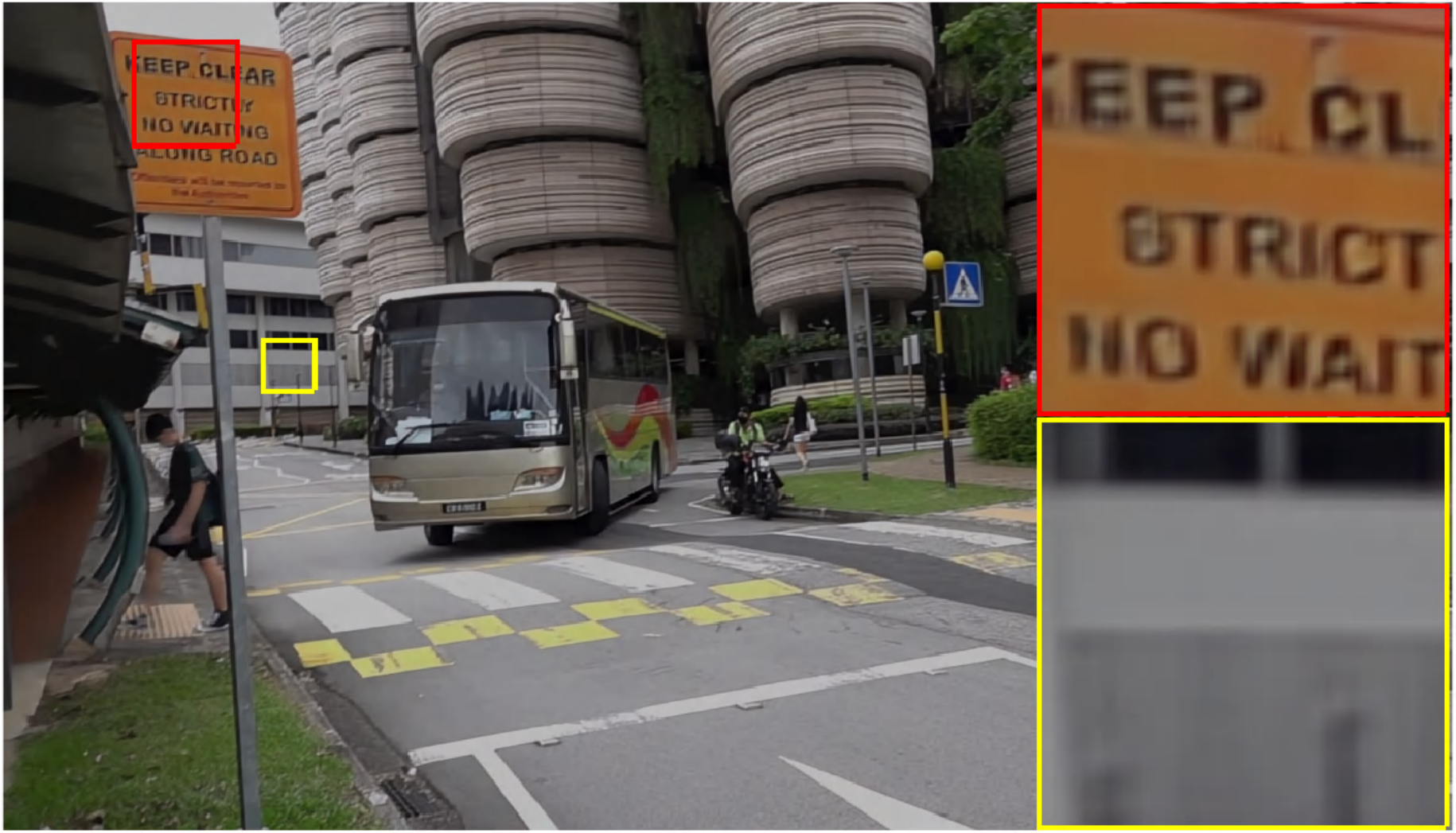}}
 	\centerline{LEDNet}
        \vspace{3pt}
 	\centerline{\includegraphics[width=1\textwidth]{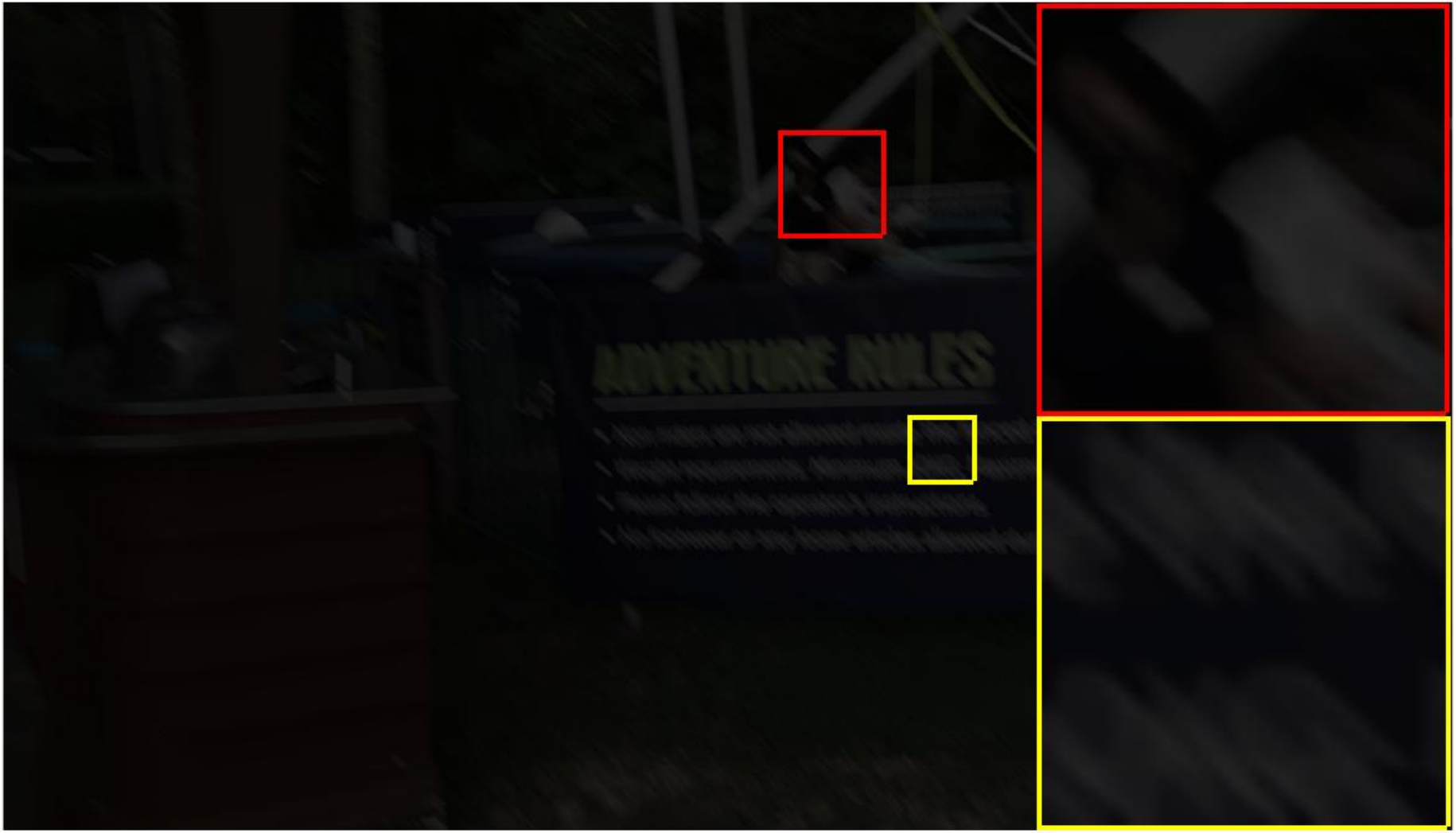}}
 	\centerline{Input}
 	\centerline{\includegraphics[width=1\textwidth]{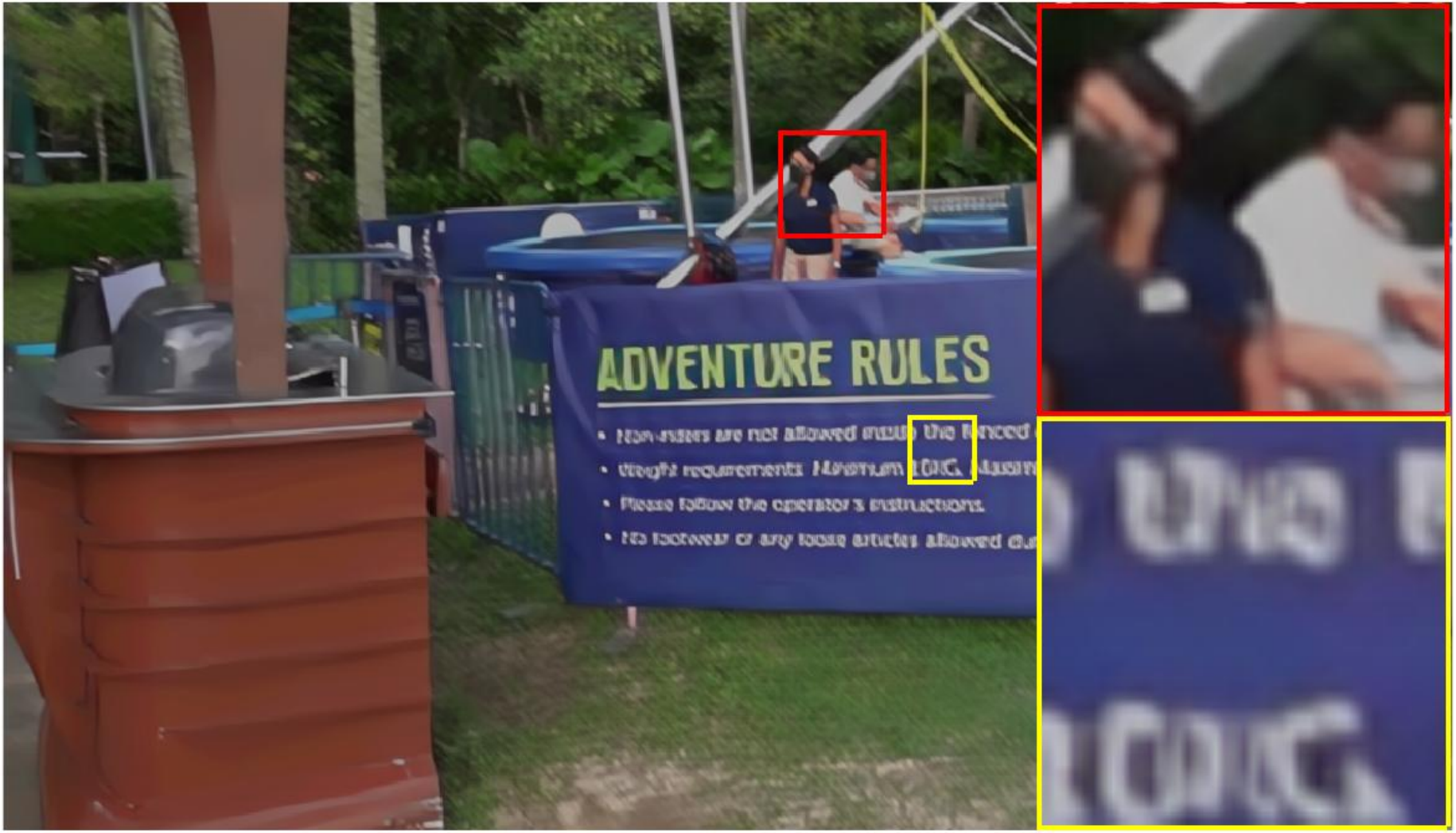}}
 	\centerline{LEDNet}
        \vspace{3pt}
\end{minipage}
\begin{minipage}{0.32\linewidth}
        \vspace{3pt}
 	\centerline{\includegraphics[width=1\textwidth]{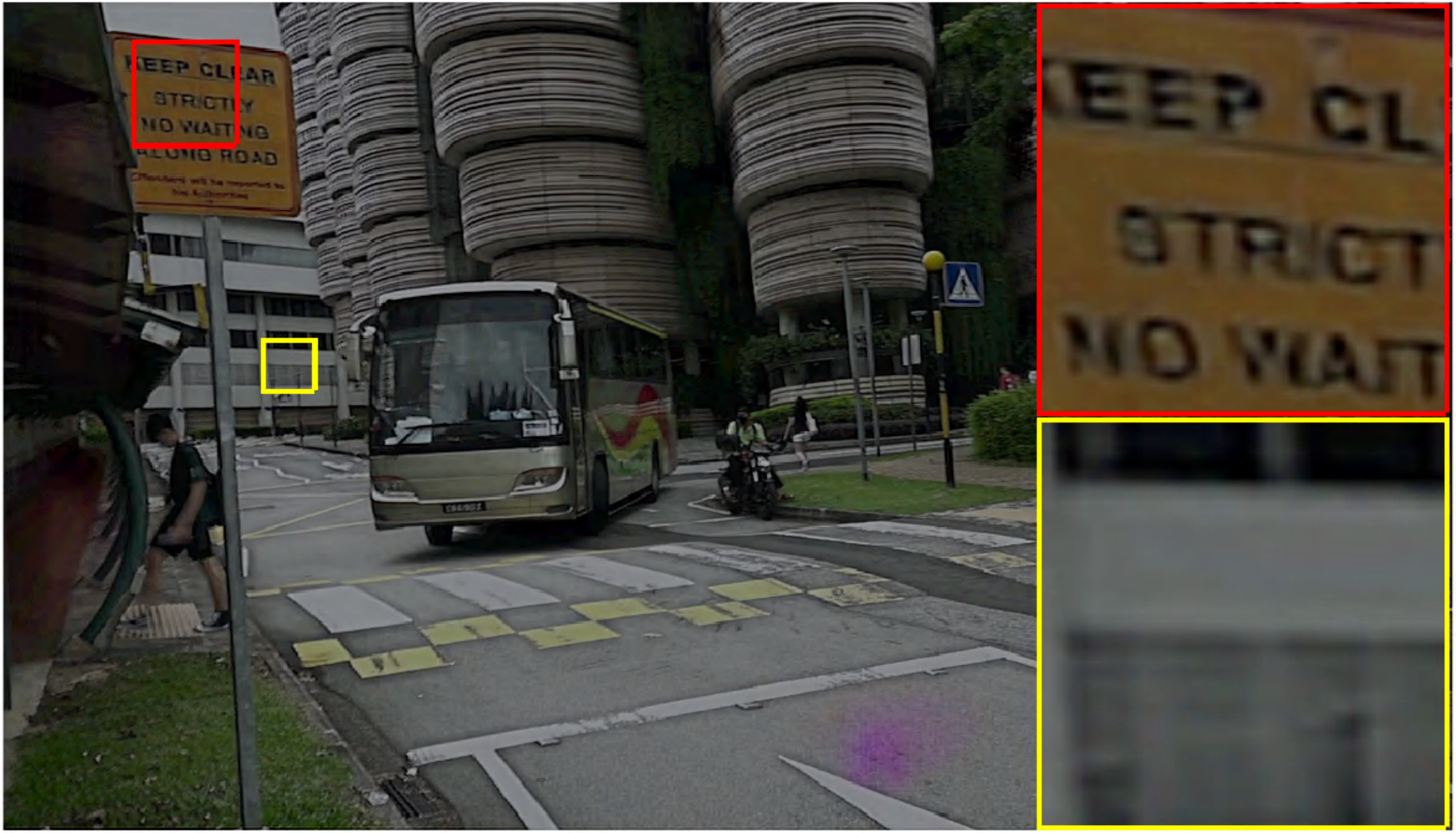}}
 	\centerline{DeblurGANv2 $\rightarrow$ ZeroDCE}
        \vspace{2pt}
 	\centerline{\includegraphics[width=1\textwidth]{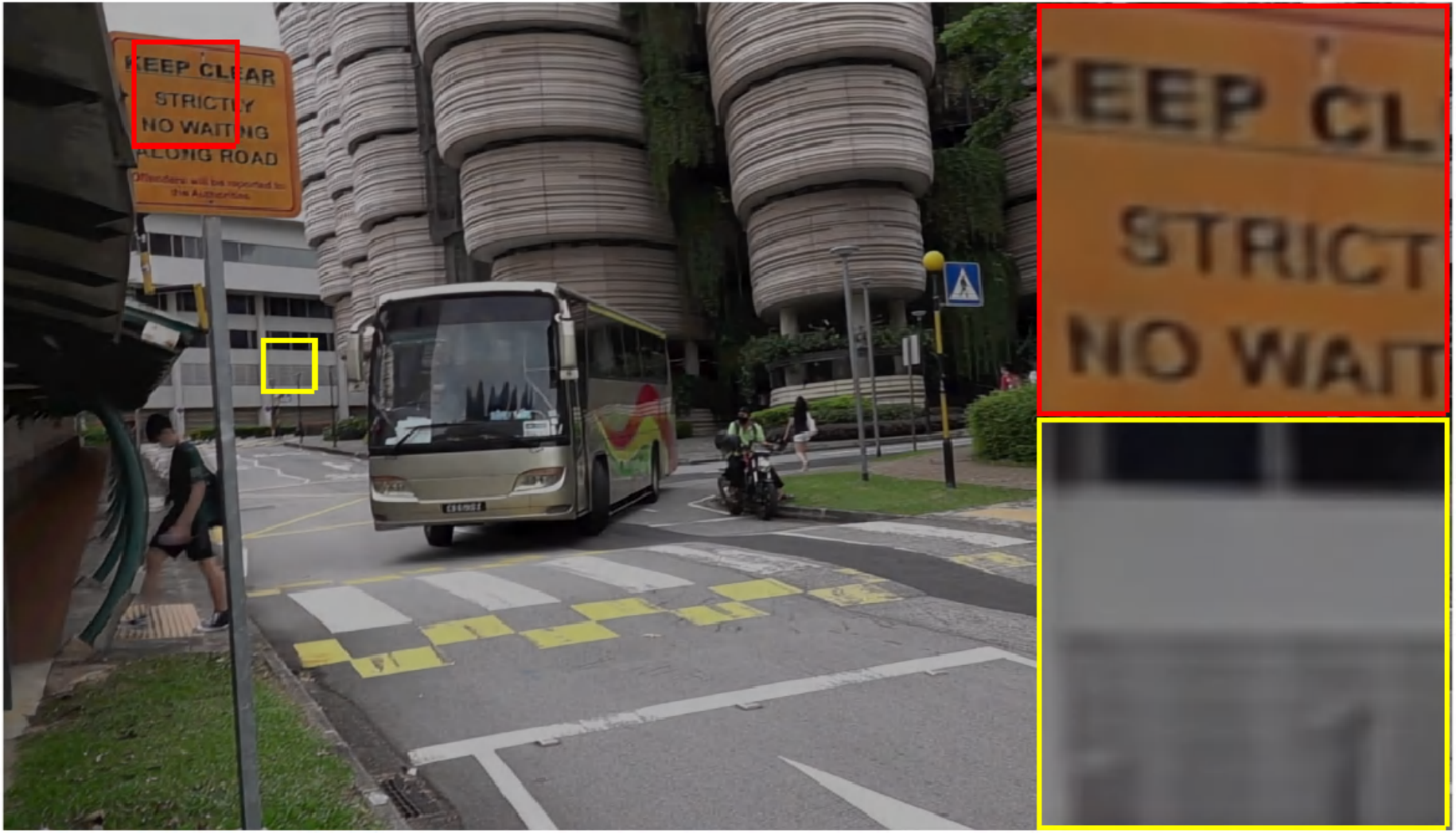}}
 	\centerline{CIDNet}
        \vspace{3pt}
 	\centerline{\includegraphics[width=1\textwidth]{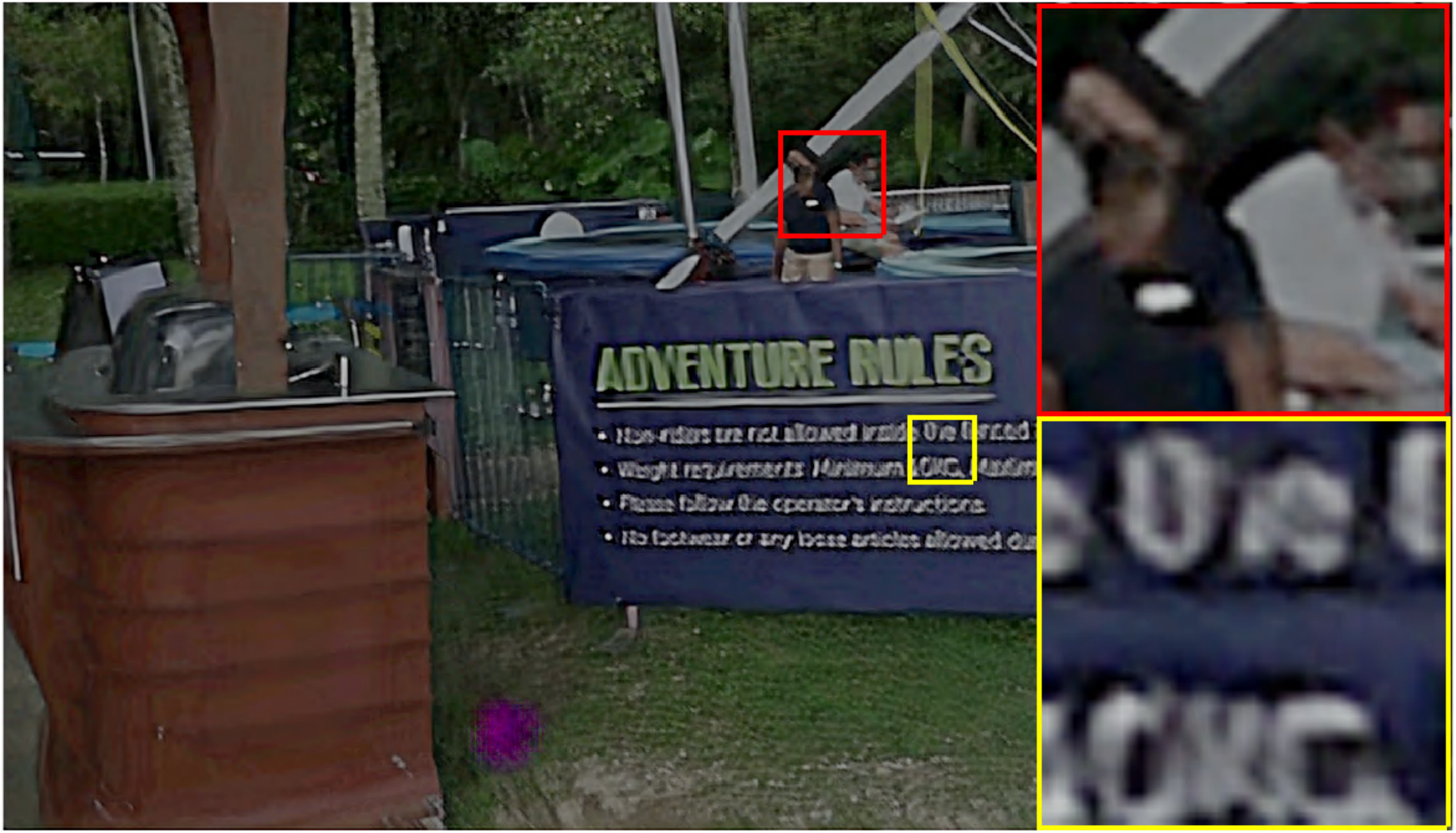}}
 	\centerline{DeblurGANv2 $\rightarrow$ ZeroDCE}
        \vspace{2pt}
 	\centerline{\includegraphics[width=1\textwidth]{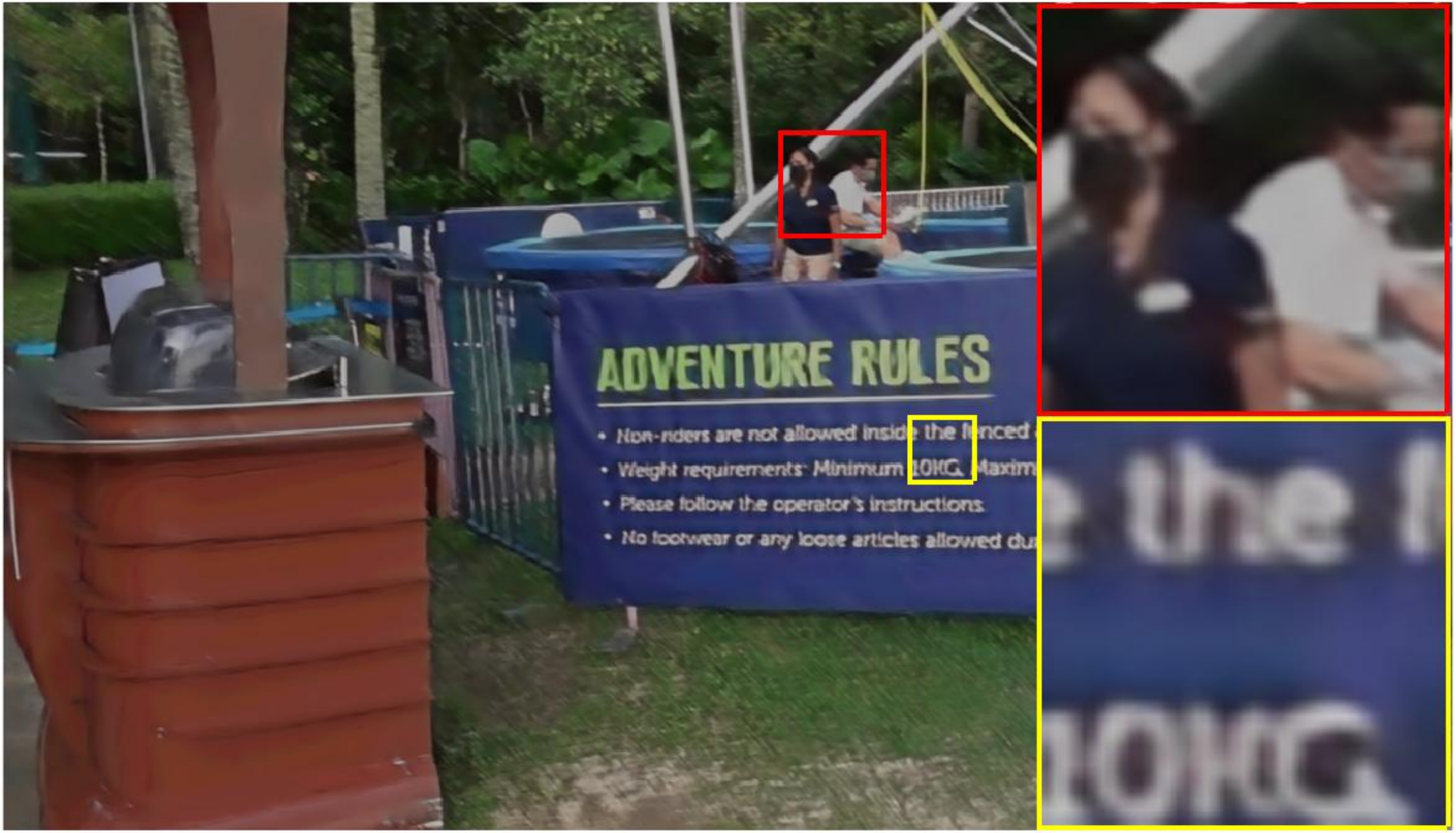}}
 	\centerline{CIDNet}
        \vspace{3pt}
 \end{minipage}
\begin{minipage}{0.32\linewidth}
 	\vspace{3pt}
 	\centerline{\includegraphics[width=1\textwidth]{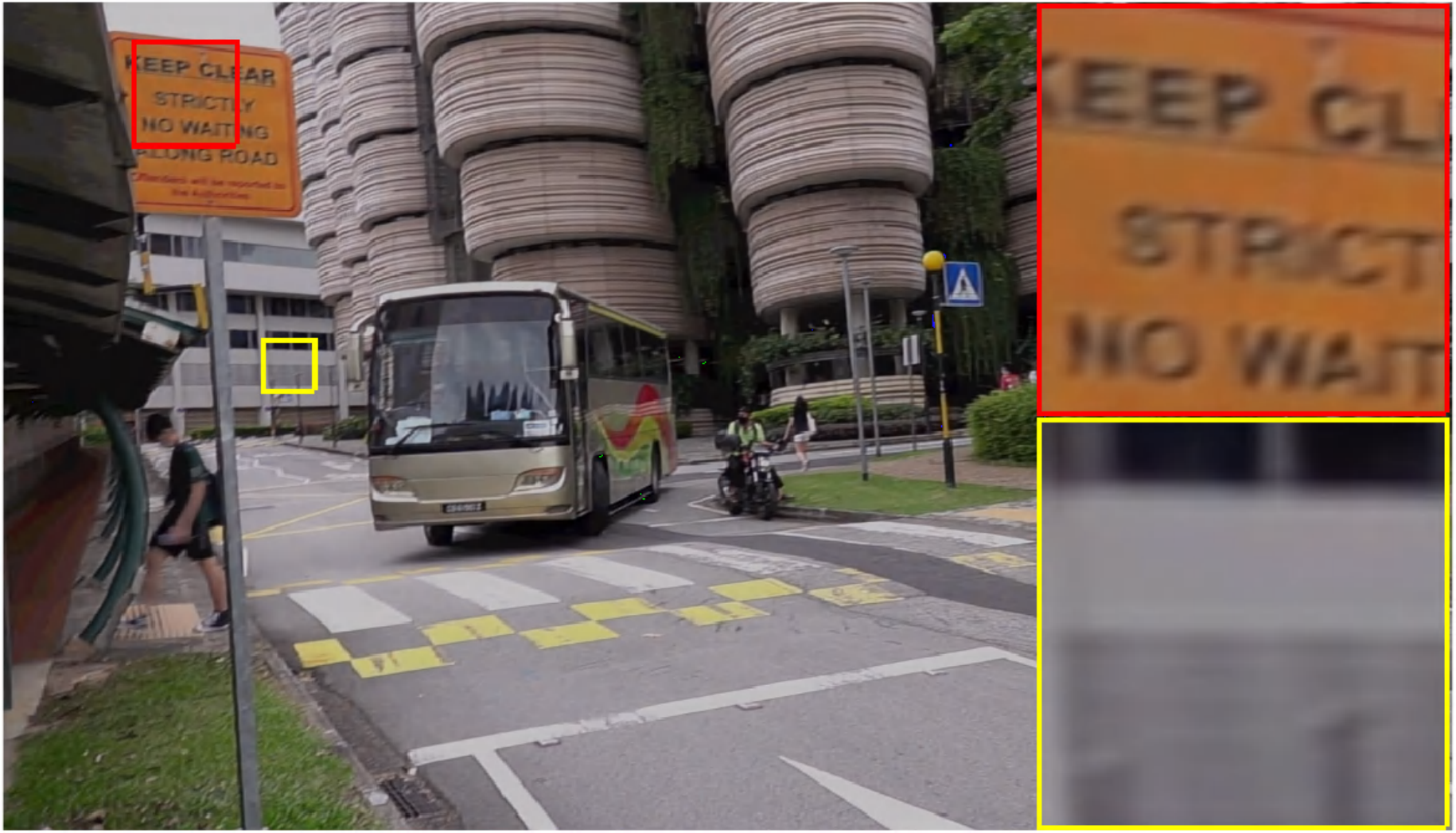}}
 	\centerline{RetinexFormer}
 	\vspace{2pt}
 	\centerline{\includegraphics[width=1\textwidth]{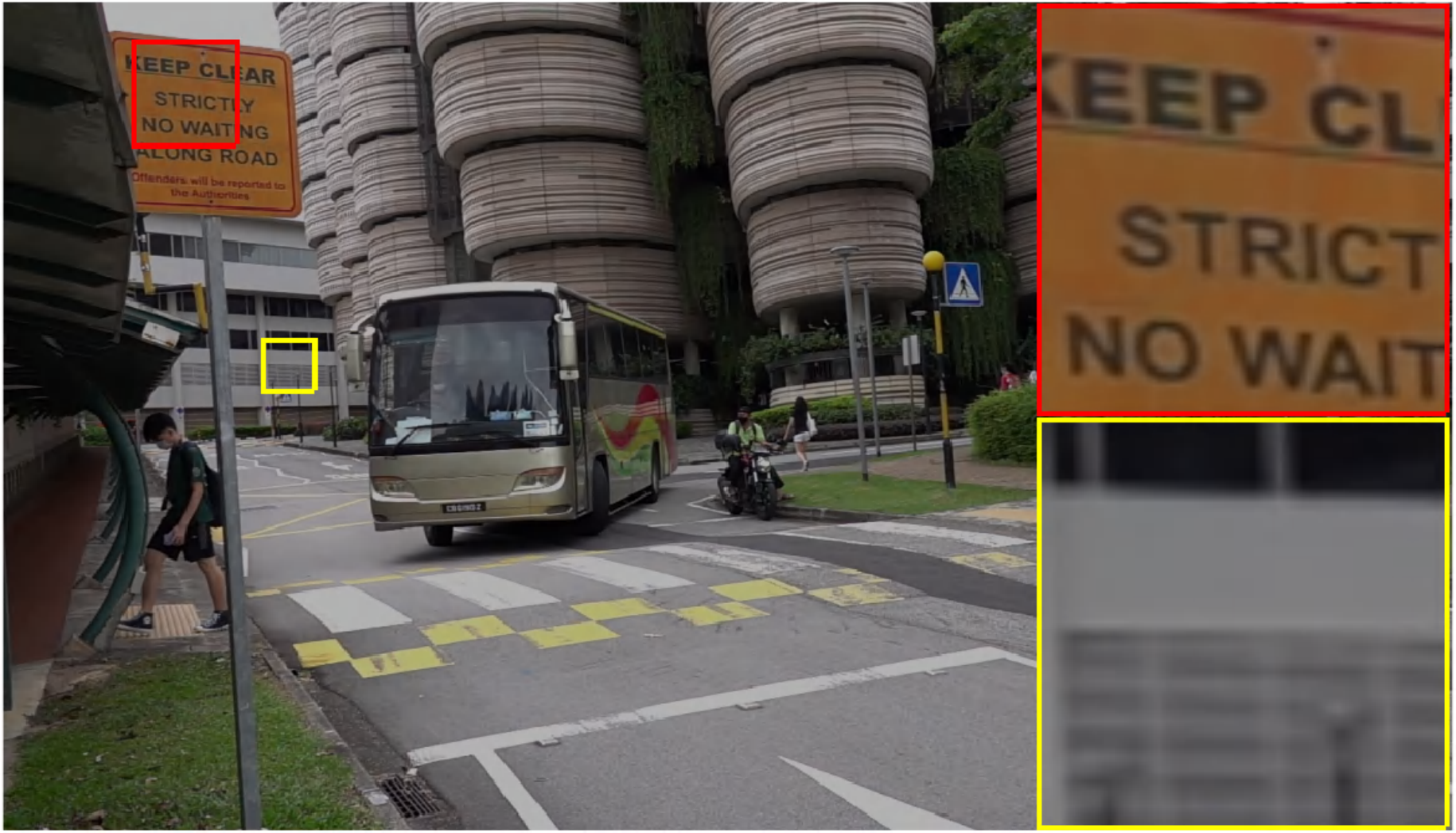}}
 	\centerline{GroundTruth}
 	\vspace{3pt}
 	\centerline{\includegraphics[width=1\textwidth]{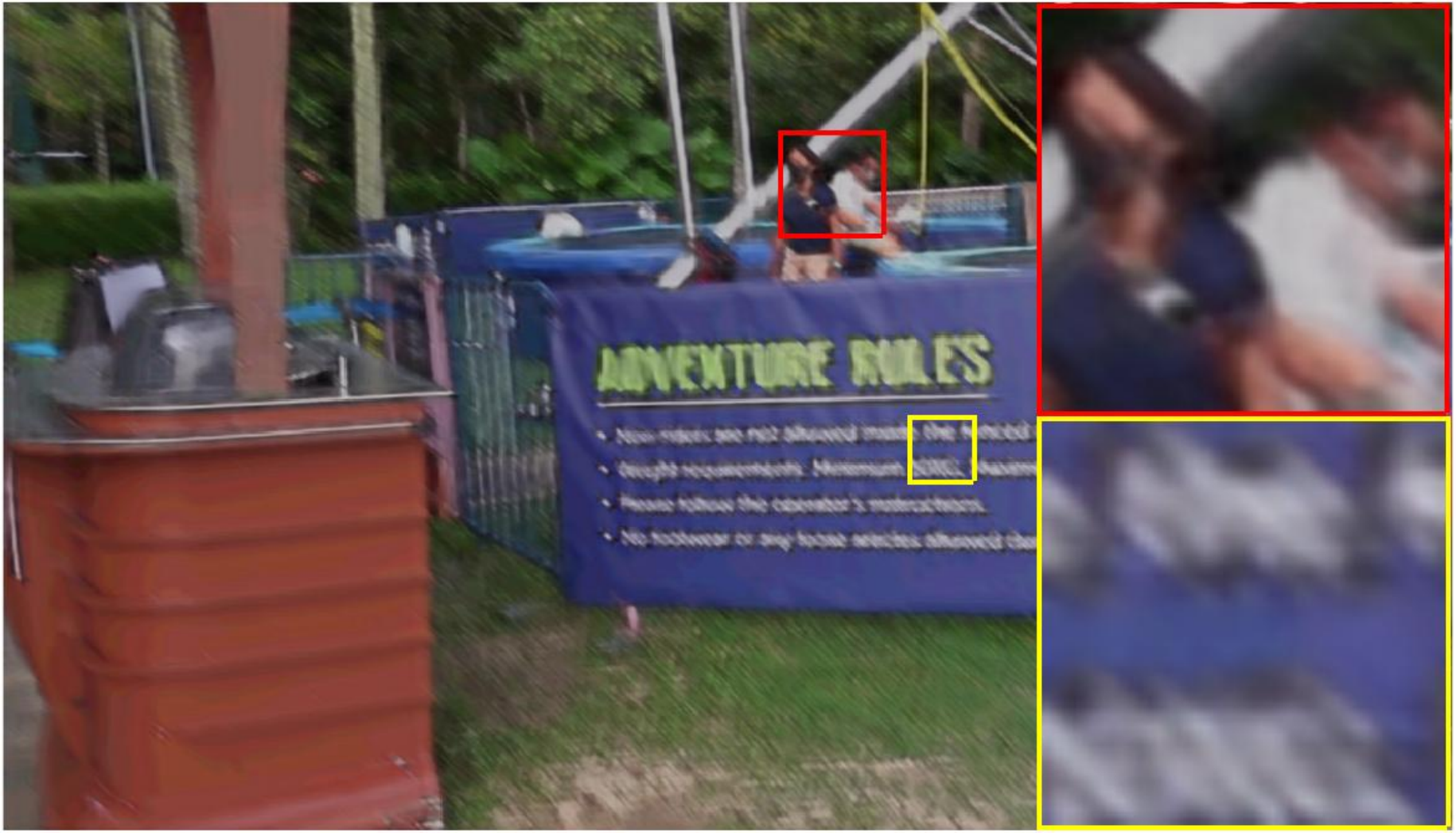}}
 	\centerline{RetinexFormer}
 	\vspace{2pt}
 	\centerline{\includegraphics[width=1\textwidth]{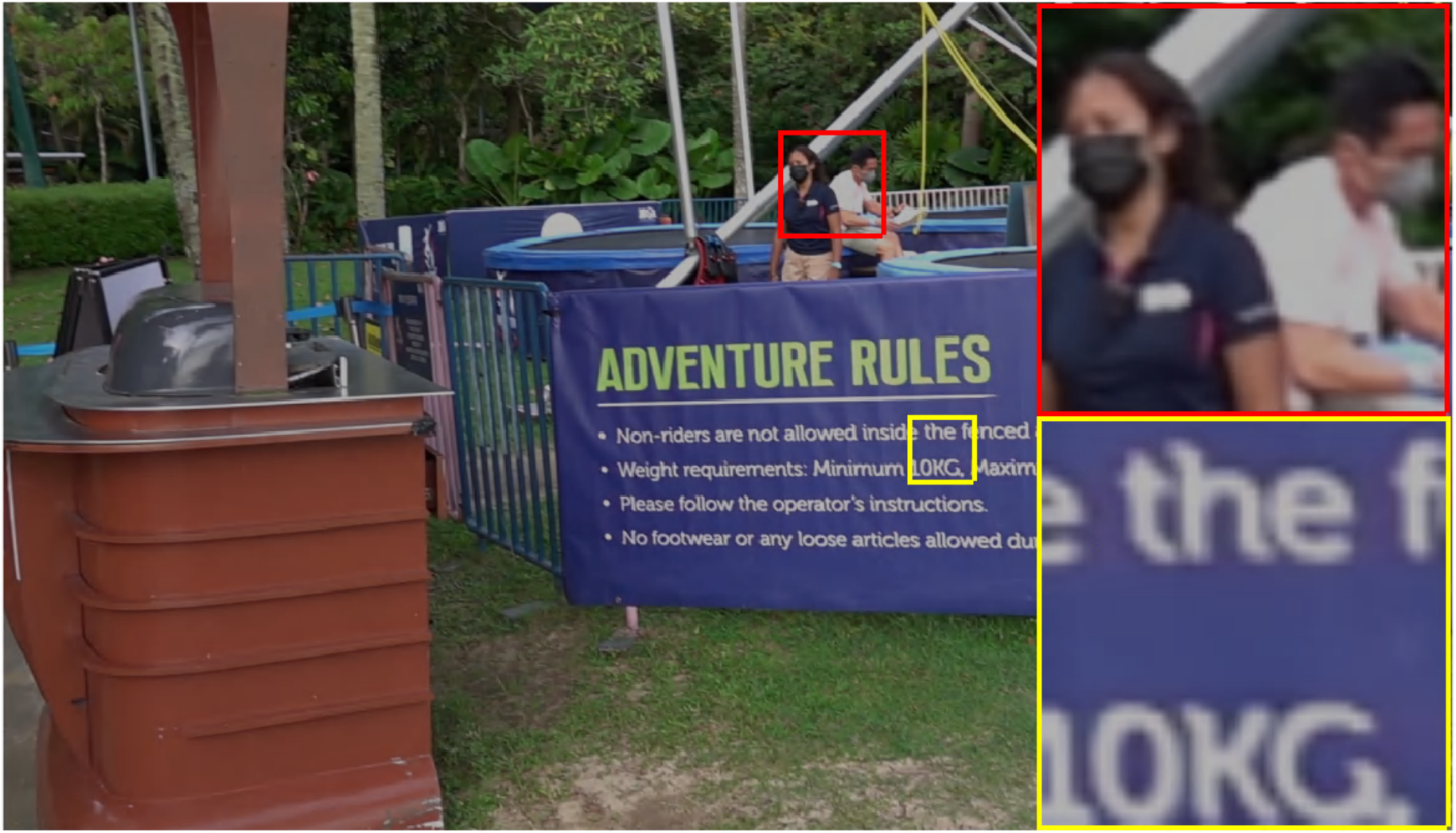}}
 	\centerline{GroundTruth}
 	\vspace{3pt}
 \end{minipage}
 \caption{Visual examples for low-light image enhancement and deblurring on LOL-Blur dataset \cite{LEDNet} among DeblurGANv2 \cite{DeblurGANv2} to ZeroDCE \cite{Zero-DCE}, RetinexFormer \cite{RetinexFormer}, LEDNet \cite{URetinexNet}, and \textbf{proposed CIDNet}. We use DeblurGAN-v2 trained on RealBlur \cite{RealBlur} dataset, and ZeroDCE trained on LOLv1 dataset \cite{RetinexNet}. Compared to other methods, our CIDNet recovers a more recognizable image with realistic colors.}
 \label{fig:LOLBlur}
\end{figure*}

\textbf{Sony-Total-Dark.} The visual results on extream low-light dataset Sony-Total-Dark \cite{SID} is shown in Fig. \ref{fig:Sony}. We did not provide GroundTruth for comparison because all methods had poorer results.

\section{Limitation and Unstudied issues}

\textbf{Intensity Collapse Function $\mathbf{C}_k$ in HVI Color Space.}
In Eq. \ref{eq:2}, we use sine function (as Eq. \ref{eq:f1}) and Eq. \ref{eq:col} to collapse the dark region in color space domain. Further research remains to be done as to whether a more appropriate function than Eq. \ref{eq:f1} exists.

\textbf{The Generalizability of the HVI Color Space in Other Tasks.} When applying the HVI Transformation to the SwinIR method \cite{liang2021swinir} for classic super-resolution $\times2$ task, we also observed an average PSNR improvement of 0.14 dB. Therefore, HVI is expected to have good applicability to other low-level vision tasks. 
(The superior performance remain unexplored.)
Unfortunately, time constraints prevented us from exploring HVI on additional tasks. 
In the future, we will focus on verifying the proposed HVI color space on other vision tasks, \eg, visual recognition tasks.

\textbf{Model training method.} Our model employs traditional supervised learning to validate the effectiveness of HVI-CIDNet. However, we have not explored training under unsupervised, semi-supervised, or zero-shot learning paradigms, which presents an interesting avenue for future research.

\textbf{Will training HVI and CIDNet separately impact performance? } HVI-CIDNet is a novel framework for low-light image enhancement, consisting of HVI transformation and enhancement network. Since we don't have ground truth of HVI, the proposed HVI-CIDNet can not be trained separately. 

\textbf{Other Unstudied Questions.} Would replacing the Transformer module in CIDNet with Mamba \cite{gu2023mamba,xu2024survey,guo2025mambair} yield better performance? Furthermore, can our HVI color space be effectively applied to other large vision models \cite{wang2023review} or adapted for use in other vision tasks \cite{zhang2024vision}?

\end{document}